\newif\ifanonymous \anonymousfalse
\newif\ifdraft \draftfalse
\newif\ifarxiv \arxivtrue
\newif\ifappendix \appendixtrue
\ifarxiv
    \documentclass[acmtog,nonacm]{acmart}
    \setcopyright{none}
\else
    \ifanonymous
    \documentclass[acmtog,anonymous,review]{acmart}
    \else
    \documentclass[acmtog]{acmart}
    \fi
\fi

\usepackage{enumitem}
\AtBeginDocument{%
  }

\setcopyright{none}

\acmDOI{XXXXXXX.XXXXXXX}

\acmSubmissionID{1576}

\ifarxiv\else
  \usepackage{setspace}
\fi

\usepackage{enumitem} %
\usepackage{multirow}
\usepackage{makecell}
\usepackage{xspace}
\usepackage[normalem]{ulem}
\usepackage{hyperref}
\usepackage{cleveref}
\usepackage{amsmath}
\usepackage{algorithm}
\usepackage{algorithmicx}
\usepackage{algpseudocode} %
\usepackage{changepage} %
\usepackage{xr}
\usepackage{subcaption}
\usepackage{lipsum}
\usepackage{graphicx}
\usepackage{tikz}

\usetikzlibrary{calc}
\usepackage[most]{tcolorbox} %
\tcbuselibrary{listings,breakable} %

\crefname{section}{Sec.}{Secs.}
\Crefname{section}{Section}{Sections}
\Crefname{table}{Table}{Tables}
\crefname{table}{Tab.}{Tabs.}
\Crefname{figure}{Figure}{Figures}
\crefname{figure}{Fig.}{Figs.}
\Crefname{algorithm}{Algorithm}{Algorithms}
\crefname{algorithm}{Alg}{Algs}
\crefname{appendix}{Appendix}{Appendices}
\Crefname{appendix}{Appendix}{Appendices}

\newcommand{\algoname}{FoldingAgent\xspace}
\newcommand{\benchname}{PurelandFold\xspace}

\ifdraft
\newcommand{\srcolor}{violet}
\newcommand{\src}[1]{\textcolor{\srcolor}{Sigal: #1}}

\newcommand{\yael}[1]{\textcolor{orange}{Yael: #1}}
\newcommand{\tali}[1]{\textcolor{cyan}{Tali: #1}}
\newcommand{\maya}[1]{\textcolor{green!60!black}{Maya: #1}}
\newcommand{\mmscolor}{magenta}
\newcommand{\mmc}[1]{\textcolor{\mmscolor}\textbf{{MM:} #1}}

\newcommand{\flag}[1]{{\color{red} #1}}
\else
\newcommand{\src}[1]{}

\newcommand{\yael}[1]{}
\newcommand{\tali}[1]{}
\newcommand{\maya}[1]{}
\newcommand{\mmc}[1]{}

\newcommand{\flag}[1]{}

\fi
\makeatletter
\ifundef{\onedot} 
{
\DeclareRobustCommand\onedot{\futurelet\@let@token\@onedot}
\def\@onedot{\ifx\@let@token.\else.\null\fi\xspace}
}
{}
\ifundef{\eg} {\def\eg{\emph{e.g}\onedot}} {} \ifundef{\Eg} {\def\Eg{\emph{E.g}\onedot}} {} 
\ifundef{\ie} {\def\ie{\emph{i.e}\onedot}} {} \ifundef{\Ie} {\def\Ie{\emph{I.e}\onedot}} {} 
\ifundef{\cf} {\def\cf{\emph{cf}\onedot}} {} \ifundef{\Cf} {\def\Cf{\emph{Cf}\onedot}} {} 
\ifundef{\etc} {\def\etc{\emph{etc}\onedot}} {} \ifundef{\vs} {\def\vs{\emph{vs}\onedot}} {} 
\ifundef{\wrt} {\def\wrt{w.r.t\onedot}} {} \ifundef{\dof} {\def\dof{d.o.f\onedot}} {} 
\ifundef{\iid} {\def\iid{i.i.d\onedot}} {}  \ifundef{\wolog} {\def\wolog{w.l.o.g\onedot}} {} 
\ifundef{\etal} {\def\etal{\emph{et al}\onedot}} {} 
\ifappendix
  \newcommand{\supp}[1]{\cref{#1}}
\else
  \newcommand{\supp}[1]{the supplementary}
\fi
\ifappendix
  \newcommand{\suppnoref}{the Appendix\xspace}
\else
  \newcommand{\suppnoref}{the supplementary\xspace}
\fi
\makeatother

\algnewcommand{\Input}[1]{%
  \State \parbox[t]{\dimexpr\linewidth-\algorithmicindent\relax}{%
    \textbf{Input:} #1%
  }%
}
\algnewcommand{\Output}[1]{%
  \State \parbox[t]{\dimexpr\linewidth-\algorithmicindent\relax}{%
    \textbf{Output:} #1%
  }%
}

\makeatletter
\newcommand\myparagraph{\@startsection{paragraph}{4}{\z@}%
  {-.01\baselineskip \@plus -2\p@ \@minus -.2\p@}%
  {-3.5\p@}%
  {\bf\ACM@NRadjust{\@parfont\@adddotafter}}}
\makeatother

\newtcblisting{promptbox}[1][]{%
  colback=gray!5,
  colframe=gray!60,
  fonttitle=\bfseries\normalsize,
  title={#1},
  breakable,
  listing only,
  listing options={
    basicstyle=\ttfamily\footnotesize,
    breaklines=true,
    breakatwhitespace=true,
    breakautoindent=false,
    breakindent=0pt,
    columns=fullflexible,
    keepspaces=true,
    showstringspaces=false,
    inputencoding=utf8,
    extendedchars=false,
    aboveskip=0pt,          %
    belowskip=0pt           %
  },
  left=4pt,
  right=4pt,
  top=2pt,
  bottom=2pt,
  before upper=\par,       %
  after upper=\par,
  before={\par\addvspace{6pt}},
  after={\par\addvspace{6pt}}
}

\makeatletter
\BeforeBeginEnvironment{promptbox}{%
  \begingroup
  \let\@vspace\@vspace@orig
  \let\@vspacer\@vspacer@orig
}
\AfterEndEnvironment{promptbox}{\endgroup}
\makeatother

\begin{document}
\title[\algoname]{\algoname:
Inferring Parametric Origami Procedures from Demonstration Videos}

\author{Maya Moriya}
\affiliation{%
  \institution{Weizmann Institute of Science}
  \city{Rehovot}
  \country{Israel}
}
\author{Sigal Raab}
\affiliation{%
  \institution{Weizmann Institute of Science}
  \city{Rehovot}
  \country{Israel}
}
\author{Yael Vinker}
\affiliation{%
  \institution{MIT}
  \city{Boston}
  \country{USA}
}
\author{Tali Dekel}
\affiliation{%
  \institution{Weizmann Institute of Science}
  \city{Rehovot}
  \country{Israel}
}

\begin{teaserfigure}
    \centering
    \includegraphics[width=1\linewidth]{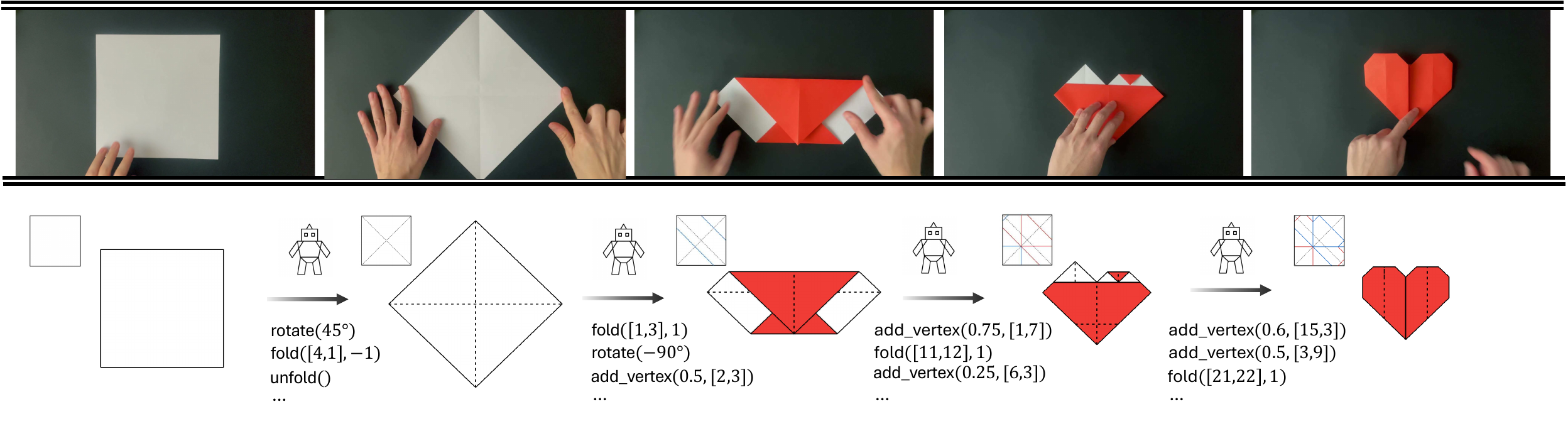}
    \vspace{-0.8cm}
    \caption{
    Given a sequence of keyframes from an instructional origami video (top), our framework, FoldingAgent, infers the full folding procedure as a sequence of executable simulator actions. 
    For each transition, we illustrate the current estimated paper geometry, the predicted simulator actions used to obtain the next paper state, and the accumulated crease pattern.
    The recovered program is parametric and structured, supporting re-rendering, editing, and analysis. 
    }
    \label{fig:teaser}
    \Description[]{}  %
\end{teaserfigure}

\begin{abstract}

We present \emph{FoldingAgent}, an agentic framework for inferring explicit parametric folding programs directly from Origami demonstration videos. Our framework leverages the reasoning power of a pre-trained Vision-Language Model (VLM) equipped with a suite of specialized tools that enable the agent to simulate geometric transitions, verify physical plausibility, retrieve and compare visual content, and evaluate its own predictions. To translate visual content into folding programs, we define a parametric space that consists of the paper's geometry and a set of parametric folding actions.
Unlike models that predict static crease patterns, our agent operates sequentially and possesses the ability to re-plan its actions, effectively mitigating the compounding errors inherent in multi-step folding. Our approach takes a step toward closing the gap between human origami knowledge, which is primarily shared through unstructured visual demonstrations, and computational methods, which typically rely on structured, parametric representations such as a crease pattern or an executable parametric plan. 
We evaluate our approach on \benchname, a newly curated benchmark of diverse Pureland origami videos with ground-truth geometry and action labels. Our results demonstrate that by combining VLM reasoning with a set of specialized tools and physical simulation, we can successfully transform unstructured visual demonstrations into executable, physically plausible folding procedures. 
\ifanonymous
Code and data can be found at \url{https://www.TBD.url}.
\else
Project page: https://maya-moriya.github.io/origami-page.
\fi

\end{abstract}

\maketitle

\section{Introduction} \label{sec:intro}
Origami is the art of transforming a flat sheet of paper into a sculpted form through folding. It combines simple materials with rich expressiveness, making it a widely practiced traditional art form. Beyond art, its unique interplay of geometry, structure, and transformation has made it a subject of study across mathematics~\cite{demaine2007geometric,eppstein2025complexities}, engineering~\cite{filipov2017bar}, design~\cite{lang1996computational,dudte2021additive}, and robotics~\cite{balkcom2008robotic,namiki2021origami}.

Despite its widespread multidisciplinary applications, a profound disconnect remains between the human practice of origami and its computational counterpart.
Origami is fundamentally procedural: a final form emerges from an ordered sequence of folding actions. In practice, this procedure is typically communicated through intuitive visual instructions, such as step-by-step illustrations or video demonstrations.
Conversely, computational approaches typically require structured, explicit forms such as a static crease pattern -- a compact, order-independent representation of the unfolded paper annotated with all mountain and valley creases required to produce the model. This creates a significant semantic gap between the ``language of folding'' used by humans and the structured representations required for machine analysis. Addressing this gap is essential for unlocking the vast repository of human origami knowledge for computational use, for example, by enabling robots to learn complex paper-handling skills directly from the thousands of instructional videos available online or by developing origami generative models.

In this work, we take a step toward this goal by posing a new task: translating an origami demonstration video into procedural folding programs that capture the underlying sequence of human folding actions. Specifically, given a sequence of frames, our goal is to infer a parametric geometric representation of the paper that captures its shape and layered structure, together with the action that transforms each state into the next (see \cref{fig:teaser}).

Inverting video frames into procedural folding programs poses several key challenges. First, the space of folding actions is diverse and highly complex: operations vary in their geometric effect and in the number of layers they manipulate, ranging from simple ``mountain folds'' (a basic crease across the paper) to intricate ``petal folds'' that involve shifting multiple layers into new, complex arrangements. This diversity raises the question of parametrization: how to effectively map the intuitive action taken by humans into computational operations executed on the paper's geometry? Even with a defined action space, inferring these actions from video remains challenging: the folder's hands may often occlude the paper, and as the model progresses, self-occluding layers accumulate and produce a highly complex internal configuration that is not visible from the video alone. Folds also interact through long-range dependencies -- a single action can reposition many layers at once and induce new internal creases far from the visible fold line. Thus, each step must be interpreted in light of the full accumulated state of the paper.

We address these challenges via a novel agentic framework that combines the reasoning capabilities of a pre-trained Visual-Language Model (VLM) with a suite of specialized tools that enable the agent to simulate geometric transitions, verify physical plausibility, retrieve and compare visual content, and self-evaluate its own predictions. Rather than attempting to predict the full procedure at once, the agent operates sequentially across a set of keyframes such that its decisions are made in light of the current paper state produced by previous operations. 

The scope of our task encompasses
Pureland Origami~\cite{smith1980pureland}, a well-studied subset in which every fold is a flat fold along a single crease. Pureland retains substantial expressive range while reducing complexity to a small set of primitive actions: fold, unfold, rotate, and flip. We expose these operations to the agent through a novel parameterized action space, paired with a simulator that executes each operation and maintains the current paper state -- its geometry, layer structure, and crease pattern -- at every step. 
This approach allows the agent to move beyond raw pixels and reason about the underlying physical rules of the paper. A key property of our design is allowing the agent to revert and re-plan a given step, thus 
mitigating the long-range error compounding that follows from the sequential nature of folding. Together, these components turn an unstructured origami video into an executable procedure, producing a sequence of actions, intermediate graphs, and a final crease pattern. Our evaluation over a diverse set of folding sequences demonstrates the power of our agentic approach and the dramatic gain over vanilla VLM inference. 

To summarize, our contributions are as follows:
\begin{itemize}[leftmargin=1.2em]
    \item Introduced the task of inferring procedural folding programs directly from origami demonstration sequences.
    \item Designed a new parametric action space and a simulator for procedural Pureland origami.
    \item Developed a new agentic framework that combines a state-of-the-art VLM with a suite of specialized tools, enabling self-verification and replanning, as well as physical verification. 
    \item Curated \benchname, a dataset containing diverse Pureland origami videos with labeled geometry and folding actions. 
\end{itemize}

\begin{figure*}
    \centering
    \includegraphics[width=1\linewidth]{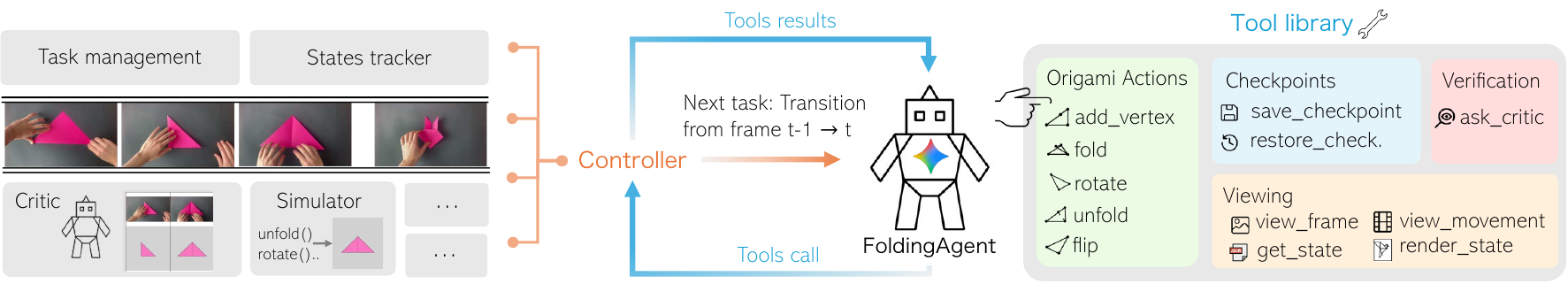}
    \caption{\emph{FoldingAgent Pipeline.} At each step, the folding agent (a pretrained VLM) infers the action transition between two frames. The agent has the flexibility to issue calls to different tools from the Tool Library (right) -- proposing candidate actions, verifying them, revising past decisions, and more. The Controller (orange) holds the agent's action-prediction timeline, the state history, the input video frames, and can call the origami simulator and the critic (a separate VLM). Tool calls are dispatched by the Controller, with results returned to the agent for further reasoning (blue loop). The agent continues until it commits a checkpoint for frame $t$, after which the next transition begins.}
    \label{fig:pipeline} %
    \Description[]{}  %
\end{figure*}
\section{Related Work} \label{sec:rw}

\subsection{Computational Origami} \label{sec:comp_origami}

\paragraph{\textbf{Geometry-based Computation}}
Most computational origami methods operate under the assumption that the relevant geometry is already available in an explicit, structured form, such as a crease pattern or a folded-state graph. 
This assumption underlies works in various areas, including: \begin{itemize}[leftmargin=1.2em]
\item \emph{Analysis \& Simulation:} A vast body of work focuses on flat- and rigid-foldability, determining whether a given crease pattern admits a valid folded configuration~\cite{bern1996complexity, akitaya2020rigid, feng2020rigid, tachi2017selffoldability}. Similarly, simulators and mechanical models operate on these geometric representations to predict physical behavior and stress~\cite{tachi2009simulation, ku2022flatfolder, mitani2007development, filipov2017bar, hu2020folding, yasuda2020data}.
\item \emph{Design \& Synthesis:} 
Design methods typically start from a given geometry and synthesize new origami objects or folding procedures. This includes classical and inverse origami design \cite{lang1996computational,TachiFreeformOrigami2010,dudte2021additive,zhu2022harnessing}, interactive design \cite{ghassaei2018fast}, and folding-sequence generation from a given geometry \cite{akitaya2013generating,huang2026learn2fold,agarwal2026origamibench}. 
\item \emph{Fabrication \& Robotics:} 
In engineering, structured representations support downstream fabrication and deployment in 3D/4D printing \cite{sundaram20173d,mao2015sequential,narumi2023inkjet}, industrial sheet-metal folding \cite{qattawi2014design,ablat2018finite}, robotic folding \cite{balkcom2008robotic}, and self-folding systems \cite{hawkes2010programmable,felton2014method}. 
\end{itemize}

While these works demonstrate the power of explicit geometric models, they do not address the fundamental question of how such representations are obtained in practice. Thus, they cannot ingest the unstructured visual demonstrations through which origami knowledge is naturally shared.

\paragraph{\textbf{Vision-based Reconstruction and Reasoning}}
This body of work focuses on recovering origami representations from visual observations. Early efforts focused on interpreting clean drill-book illustrations~\cite{shimanuki2003recognition} or predicting state classes from controlled camera setups~\cite{shimanuki2012folding, namiki2021origami}. Additional works infer line labels or 3D interpretations from origami line drawings \cite{kanade1980theory,parodi1995complexity,sabbah1985computing}, provide interactive folding guidance from camera/MR observations \cite{chen2025origamisensei,chen2023origamisensei,zhu2010origami}, or estimate physical folding states in robotic/deployable systems \cite{namiki2021origami,lal2023unsupervised,ray2024origami}. More recently, \citet{kato2025origami} have predicted crease lines from before/after image pairs. However, their approach is limited to isolated action vocabulary in a controlled setting, and does not support the broader Pureland action vocabulary used in our task (e.g., unfold, rotate, or flip).
Unlike existing works, which primarily focus on state classification or single-step crease prediction, our method is the first to translate unstructured video keyframes into executable procedural programs. By providing an automated, closed-loop system to convert human-oriented demonstrations into formal representations, we take a step toward bridging the gap between geometric modeling and real-world origami applications.

\subsection{VLM-reasoning for Inverse Vision Tasks} As the reasoning capabilities of VLMs become more powerful, inverse vision is increasingly recast as a language-mediated, agentic process. IG-LLM~\shortcite{kulits2024rethinking} demonstrated that VLMs can decode visual embeddings directly into structured 3D programs via spatial reasoning. Recent agentic frameworks follow ReAct-style loops \cite{yao2023react} to interleave reasoning with action; for instance, VIGA~\shortcite{yin2026viga} and IR3D-Bench~\shortcite{liu2025ir3dbench} employ VLM agents to iteratively write, render, and revise graphics programs based on visual feedback.

This reasoning-centric approach allows VLMs and LLMs to produce editable visual artifacts through executable scripts, \eg by using Blender \cite{lu2025ll3m}, SVG \cite{vinker2025sketchagent, cai2024delving},
Python~\cite{Han2023ChartLlamaAM}, Processing~\cite{sharma2024vision}, or TikZ~\cite{bubeck2023sparks}.
Finally, VLMs serve as powerful critics and reward models for 3D generation. \citet{bai2025vlm3drewards} and \citet{chen2026know3d} utilize VLM spatial-relation scores to guide text-to-3D optimization. Furthermore, GPT-4V \shortcite{wu2024gpt4v} and Gen3DEval \shortcite{maiti2025gen3deval} assess generated assets using human-aligned VLM evaluators, providing a scalable alternative to manual benchmarking.

To the best of our knowledge, Learn2Fold \cite{huang2026learn2fold} is the only prior work to apply LLMs to an inverse origami task. 
However, it targets a fundamentally different setting: generating folding sequences from a given final crease pattern, which is a structured representation rarely available in real-world origami settings. In contrast, our framework operates directly on instructional origami videos, without requiring additional parametric information, tackling the reconstruction task in a zero-shot agentic fashion.

\section{Preliminaries}
\label{sec:prelim}

\subsection{FOLD Representation}
Flexible Origami List Datastructure (FOLD) is a JSON-based format designed for computational origami models \cite{FOLD_CGW2016}. FOLD has emerged as a standard interchange format, widely adopted across various origami design tools and simulators \cite{ghassaei2018fast,TachiFreeformOrigami2010,RabbitEar,mitani2007development}. Like standard mesh formats (OBJ, DXF, etc.), FOLD describes a surface as a set of vertices, edges, and faces, and a per-edge label specifying its role in the fold --- mountain, valley, boundary, or flat. While standard mesh representations are sufficient for general 3D geometry, they are inadequate for flat-folded origami, where overlapping faces stack in a specific order that is part of the folded state. FOLD addresses this with a dedicated field (\texttt{faceOrders}) encoding pairwise above/below relationships between faces, allowing it to fully specify flat-folded configurations. In our work, we adopt and extend the FOLD representation (see \cref{sec:representation}).

\section{Method} \label{sec:method}
Given an instructional origami video, we first manually extract a sequence of keyframes $\{I_1, \dots, I_K\}$, where each $I_t$ depicts a meaningful interaction between the demonstrator's hands and the paper sheet. 
Our goal is to recover a parametric, executable representation of the demonstrated folding procedure. We formulate the folding procedure as a set of states $\{S_1, \dots, S_K\}$, representing the paper's geometry at matching keyframes (\cref{sec:representation}), and corresponding transition actions $\{a_2, \dots, a_K\}$ matching the Pureland action space (\cref{sec:actions}).
We define a novel simulator, detailed in \supp{sec:simulator}, that executes an action (or a short sequence of actions) to transform one state into another:
\begin{equation}
    \texttt{Origami-Sim}(S_{t-1}, a_{t-1}) = S_{t}
\end{equation}

This formulation enables us to build an automated framework in which a pretrained vision-language model (the \emph{agent}) interacts with the simulator through a dedicated tool library. The agent can propose folding actions, use verification tools to test hypotheses, request additional visual information, and revert earlier decisions, until it reaches a consistent reconstruction. The process is performed sequentially, with rollback available throughout to prevent cumulative errors. A high-level illustration is shown in \cref{fig:pipeline}.

\begin{figure}[t]
    \centering
    \includegraphics
    [width=\columnwidth]
    {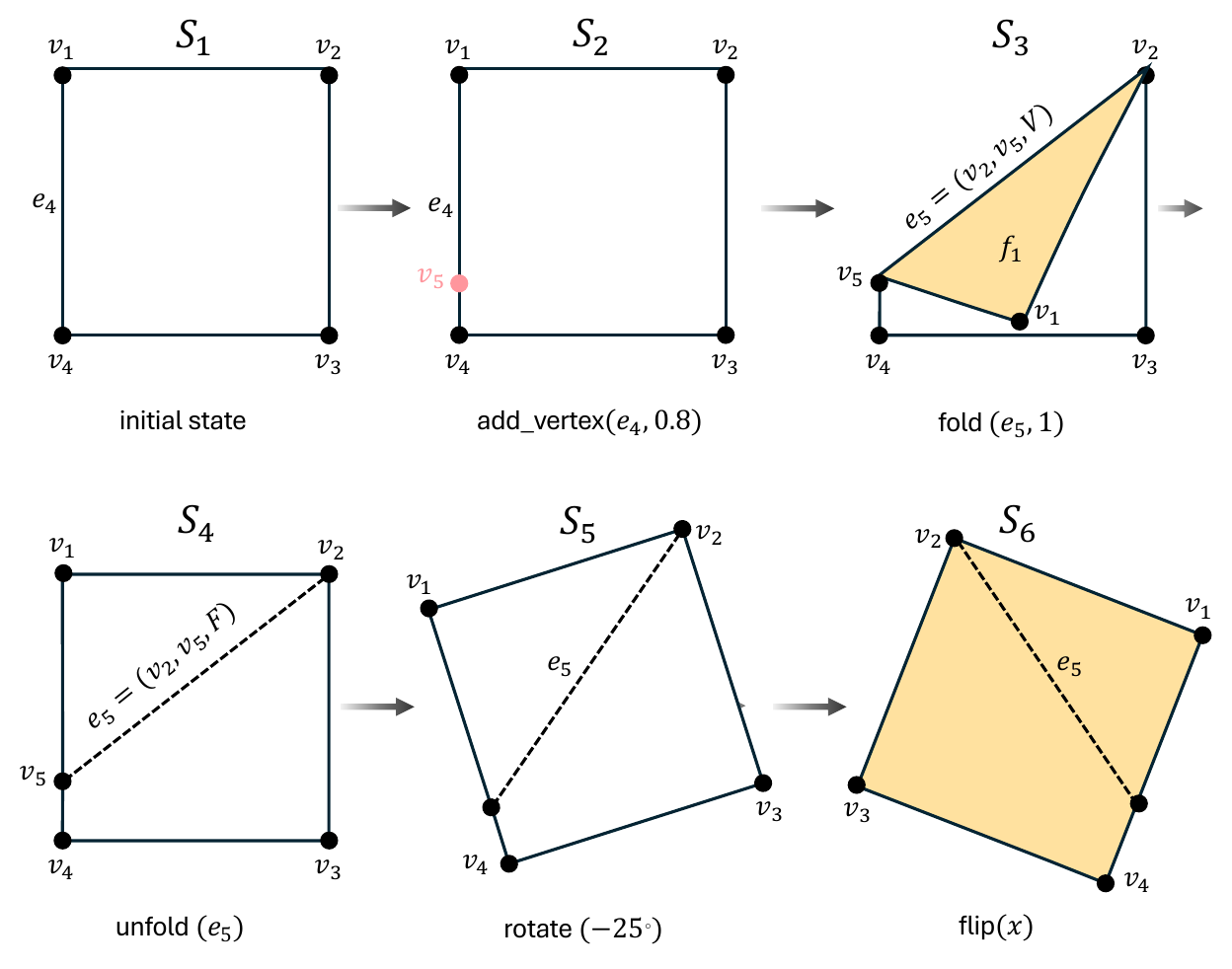}
    \caption{\emph{State representation and action space.} Starting from a flat square $S_1$, we apply the five primitives in sequence (\texttt{add\_vertex}, \texttt{fold}, \texttt{unfold}, \texttt{rotate}, \texttt{flip}) and show the states they produce. New vertices, edges, and faces are introduced as the graph evolves; yellow faces are back-side up. Active folds are shown solid, flat creases dashed.}    
    \label{fig:action_space} %
    \Description[]{}  %
\end{figure}

\subsection{Origami State Representation}
\label{sec:representation}
 The scope of our task encompasses 2D Pureland origami, which provides a compact and tractable representation space while remaining sufficiently expressive to cover a wide range of real-world instructional folding sequences. Specifically, it restricts the allowed folds to simple mountain and valley folds along straight creases, together with the basic
operations unfold, rotate, and flip. Pureland models are accessible to beginners and ensure that every intermediate folded state is a flat-folded layout of polygonal faces, which simplifies both representation and simulation. This setting has been studied in the context of origami notation and robotics \cite{konjevod2009notation,balkcom2004robotic}, and adopted in instructional resources \cite{smith1980pureland,smith1989pureland2,smith1993pureland3}. 

Each folded state $S$ is defined as a planar graph as follows:
\begin{equation}
    S = (V, F, E, O, L),
\end{equation}
where $V$, $F$, and $E$ are the standard FOLD vertices, faces, and edges, respectively (\cref{sec:prelim}). Vertices are defined by 2D coordinates, while edges are defined by their endpoint vertices and carry labels in $\{B, F, M, V\}$ (boundary, flat, mountain, and valley). Boundary edges define the outer paper contour, and mountain and valley edges define active folds -- creasing the paper so the ridge points up or sinks down, respectively.
Finally, flat edges connect coplanar faces without a visible crease. Faces are defined by their boundary vertices, and the complete representation is illustrated in \cref{fig:action_space}.

$O$ and $L$ are two extensions required for our visual task.
$O = \{f \to o_f\}$ is the per-face orientation, where $o_f \in \{0, 1\}$ indicates whether, at face $f$, the front (0) or back (1) side of the paper is facing up. Differentiating the paper's front and back sides improves the VLM's visual state tracking.
$L = [L_1, \dots, L_M]$ is an explicit layer ordering over coplanar faces, listed bottom to top. Each $L_i$ contains several planes, all located at the same topological depth in the folded model; each plane comprises a maximal set of faces connected by flat edges. The faces within each plane move together under any subsequent fold. Operating at the granularity of planes makes layer recomputation tractable after a fold and plays the role of the \texttt{faceOrders} field in FOLD. 
Together, these fields form a compact representation of a flat-folded pureland origami state, shared by our simulator and agent. Note that our geometric representation can be straightforwardly converted 
to the crease pattern (CP) format.

\subsection{Folding Action Space}
\label{sec:actions}
We define a compact, composable action space to express the pureland folding space. An action $a_t$ is drawn from five primitives:
\begin{description}
    \item[\texttt{add\_vertex}$(e, p)$] subdivides edge $e$ at fractional position $p \in [0, 1]$, returning a new vertex. This enables folds along creases that do not pass through existing vertices.
    \item[\texttt{fold}$(e, d)$] folds the paper along edge $e$, with direction $d \in \{+1, -1\}$ specifying which side of the crease moves toward the viewer.
    \item[\texttt{unfold}$(e)$] reverses a previously applied fold along $e$, leaving a flat edge as a crease mark.
    \item[\texttt{rotate}$(\theta)$] rotates the entire model by $\theta$ degrees in the plane.
    \item[\texttt{flip}$(\alpha)$] reflects the model about an axis $\alpha \in \{x,\, y,\, y{=}x,\, y{=}{-}x\}$.
\end{description}

\cref{fig:action_space} illustrates the actions and their cumulative effect on the rendered states.
Note that a single transition $I_{t-1} \to I_{t}$ may require composing several primitives: for example, folding along a non-vertex crease entails an \texttt{add\_vertex} followed by a \texttt{fold}, and reorienting the paper before a fold may require a \texttt{rotate} or \texttt{flip}. By keeping the primitives minimal and composable, the agent can express a wide range of folds with a small vocabulary, while each individual action has an easy-to-verify effect on the state.

\begin{figure}
    \centering
    \includegraphics[width=1\linewidth]{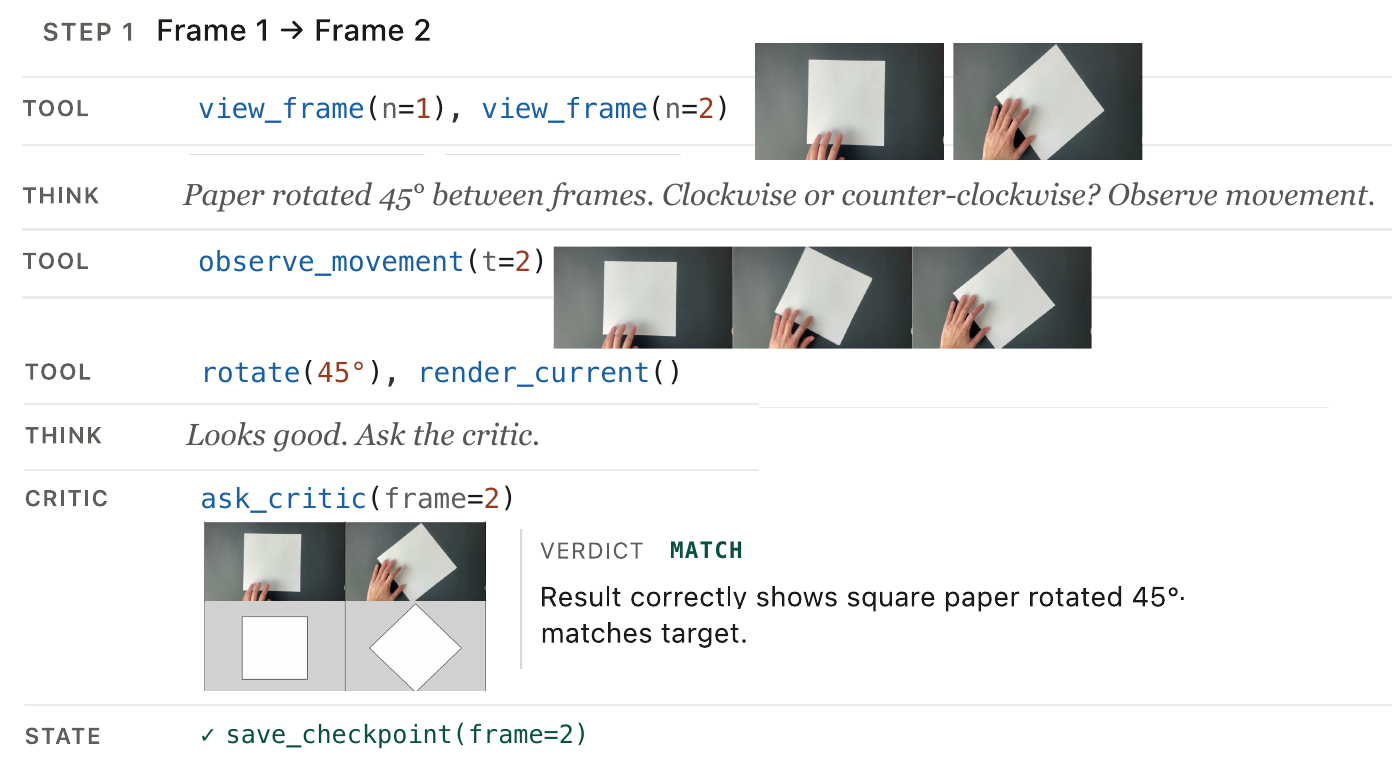}
    \\[-6pt]
    \Description{An example agent trajectory showing interleaved reasoning, tool calls, and verification for paper rotation.}
    \caption{\emph{Example agent trajectory for a single transition.} A real log showing the agent interleaving reasoning (italic), tool calls, and verification. Here the agent rotates the paper by $45^\circ$ and confirms the result via the critic, which receives source/target photos (A, B) and rendered states (C, D).} %
    \label{fig:one_step}
\end{figure}

\subsection{Agentic Framework}
\label{sec:agentic}
We use a pretrained vision-language model as our \emph{agent}, in a zero-shot manner without any fine-tuning. Starting from a square paper $S_1$, the agent's task is to infer $a_t$ for each transition $I_{t-1} \to I_{t}$. A representative single-transition trajectory is shown in \cref{fig:one_step}.

Our system supports the agent through three main components, as illustrated in \cref{fig:pipeline}:

\paragraph{\bf Tool Library}
\label{sec:tools}
We design four categories of specialized tools:
\begin{itemize}
    \item \textbf{Origami Actions.} The folding primitives in~\cref{sec:actions} (\texttt{add\_vertex}, \texttt{fold}, \texttt{unfold}, \texttt{rotate}, \texttt{flip}) are exposed as callable tools. The simulator applies each tool to the input state and returns the updated state.

    \item \textbf{Viewing.}
    The agent can inspect both the input sequence and its own progress. \texttt{view\_frame}$(t)$ retrieves frame $I_t$ and feeds it to the agent; \texttt{observe\_movement}$(t)$ returns a filmstrip of the motion between $I_{t-1}$ and $I_t$, used when a single keyframe is insufficient to disambiguate the action; \path{get_current_state} returns the current state $S$ as JSON; and \texttt{render\_current} returns a 2D diagram of $S$ that visually mirrors the input frames (front/back-side coloring, solid edges for active folds, dashed edges for crease marks). The diagram is the agent's primary visual feedback signal.

    \item \textbf{Checkpoints.} \texttt{save\_checkpoint}$(t)$ records the current actions and state as the solution for keyframe $I_t$, and \linebreak\texttt{restore\_checkpoint}$(t)$ reverts the simulator to the state saved at $t$. Checkpoints serve both as the final output of the system and as recovery points during exploration.

    \item \textbf{Verification.} \texttt{ask\_critic} invokes a separate visual critic (described below), and \texttt{generate\_checkpoint\_overview} produces a side-by-side panel of all saved checkpoints against their target photos, to identify where reconstruction drifted.

\end{itemize}

\paragraph{\textbf{Controller}}
The controller (marked in orange in \cref{fig:pipeline}) is the deterministic execution layer between the agent and the rest of the system. It parses each tool call emitted by the agent, dispatches it to the appropriate backend (simulator for actions, renderer for diagrams, critic for verification), maintains the simulator state across calls, and returns structured results to the agent. The agent does not manipulate states directly; all changes flow through the controller.

\paragraph{\textbf{Visual Critic}} \label{sec:critic}
A central design choice in our framework is to delegate verification to a dedicated vision-language model rather than to the agent itself. When the agent believes a transition is complete, it calls \texttt{ask\_critic}, which assembles a four-image grid: the real photos of the source and target keyframes $(I_{t-1}, I_t)$, and the rendered diagrams of the corresponding simulator states $(S_{t-1}, S_t)$. The critic compares the candidate diagram against the target photo at the level of geometric structure while tolerating the stylistic gap between photo and diagram. 
The critic returns one of three verdicts --- \textsc{Match}, \textsc{Mismatch}, or \textsc{Extreme Divergence} --- together with a written analysis and a list of discrepancies. Separating the proposer from the verifier in this way reduces the risk of the agent rationalizing its own mistakes.

\paragraph{\textbf{Reconstruction loop}} With these components in hand, the agent operates sequentially. For a transition $S_{t-1} \to S_t$, the agent is guided via our system prompt to view $I_{t-1}$ and $I_t$ (and optionally view filmstrips, or future/past frames), proposes one or more simulation actions, inspects the resulting diagram, and calls the critic. The agent has the flexibility to make other tool calls and generate several predictions before deciding to call the critic. The agent is instructed, but not required, to follow this protocol: on a \textsc{Match} verdict, it should save $S_t$ as a checkpoint, after which the controller advances to the next step; on a \textsc{Mismatch}, it should attempt the transition again, possibly using a different sequence of actions and with the option to roll back to previous steps. If the agent fails to obtain a \textsc{Match} after three attempts on the same transition, it should invoke \texttt{generate\_checkpoint\_overview} to inspect the full reconstruction so far, identify the earliest checkpoint that diverges from its target photo, and roll back to a confident earlier state before attempting again under different assumptions. This rollback protocol is essential: errors in early transitions propagate, and providing the agent with an explicit mechanism to detect and recover from them is what makes long sequences tractable. See \suppnoref for our full protocol.

\paragraph{\textbf{Bounded exploration}}
To prevent excessive rollbacks, we impose a budget of $C$ transition attempts per keyframe, where an attempt is any simulator action applied to the input state, regardless of whether the resulting state is kept (and later revisited) or discarded. Throughout the reconstruction process, the controller records each explored keyframe-transition trajectory as a sequence of actions and resulting states.
If the agent consumes its budget for a keyframe without obtaining a Match verdict, the controller renders all explored states for that keyframe and invokes a lightweight selector agent (see \supp{sec:selector}), which chooses the state that best matches the target. The selected state, as well as all ancestor states along the selected path, are saved as checkpoints.
Importantly, this commitment is final: once the selector agent is invoked for keyframe $t$, \texttt{restore\_checkpoint} to keyframe $t$ or any preceding keyframe is disabled, preventing the agent from revisiting the resolved portion of the sequence. This mechanism trades a small amount of local optimality for the guarantee that exploration remains tractable.
We use $C=5$ in our experiments.

\section{Results} \label{sec:exp}
We evaluate our framework on a collection of instructional pureland origami videos spanning a range of common folding procedures (described in \cref{sec:data}).
We use Gemini 3.1 Pro Preview for our agent and critic models. 
Full implementation details, including our system prompts,
are included in \suppnoref,
along with extensive visualizations of our results, metric details, decisions, tool calls, and reasoning across full sequences.

On average, per sequence, FoldingAgent issues 140 queries, consuming 4M input and 68K output tokens, costing \$8.9. This reflects the cumulative cost of iterative tool calling and self-correction across a full folding sequence.

\ifarxiv
     \begin{figure*}[t]
\centering

\newlength{\imgwfourteen}
\setlength{\imgwfourteen}{0.12\textwidth}

\newlength{\xgapfourteen}
\setlength{\xgapfourteen}{0.126\textwidth}

\newlength{\ygapfourteen}
\setlength{\ygapfourteen}{-0.085\textwidth}

\newcommand{\secondrowfigfourteen}[1]{%
  \includegraphics[
    width=\imgwfourteen,
    trim={0pt 5cm 0pt 5cm},
    clip
  ]{#1}%
}

\begin{tikzpicture}[
    img/.style={
        inner sep=0pt,
        outer sep=0pt
    },
    cmd/.style={
        fill=white,
        fill opacity=0.90,
        text opacity=1,
        draw=black!30,
        rounded corners=1.2pt,
        inner sep=1pt,
        font=\tiny\ttfamily,
        align=left
    },
    frameid/.style={
        fill=white,
        fill opacity=0,
        text opacity=1,
        text=white,
        draw=black!25,
        rounded corners=0pt,
        inner sep=1pt,
        font=\tiny\ttfamily,
        anchor=north west
    }
]

\node[img] (a1)  at (0\xgapfourteen, 0)  {\includegraphics[width=\imgwfourteen]{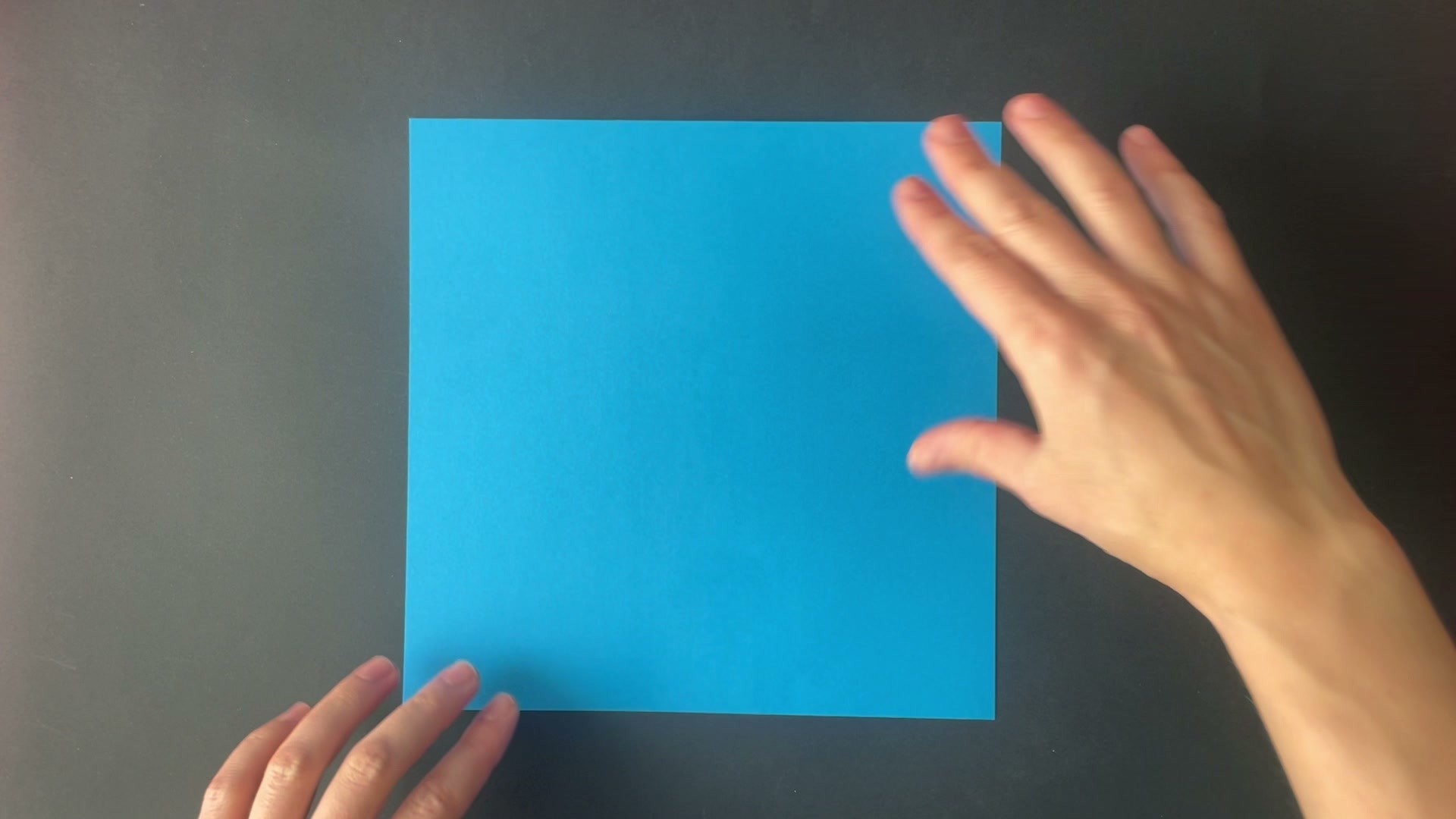}};
\node[img] (a2)  at (1\xgapfourteen, 0)  {\includegraphics[width=\imgwfourteen]{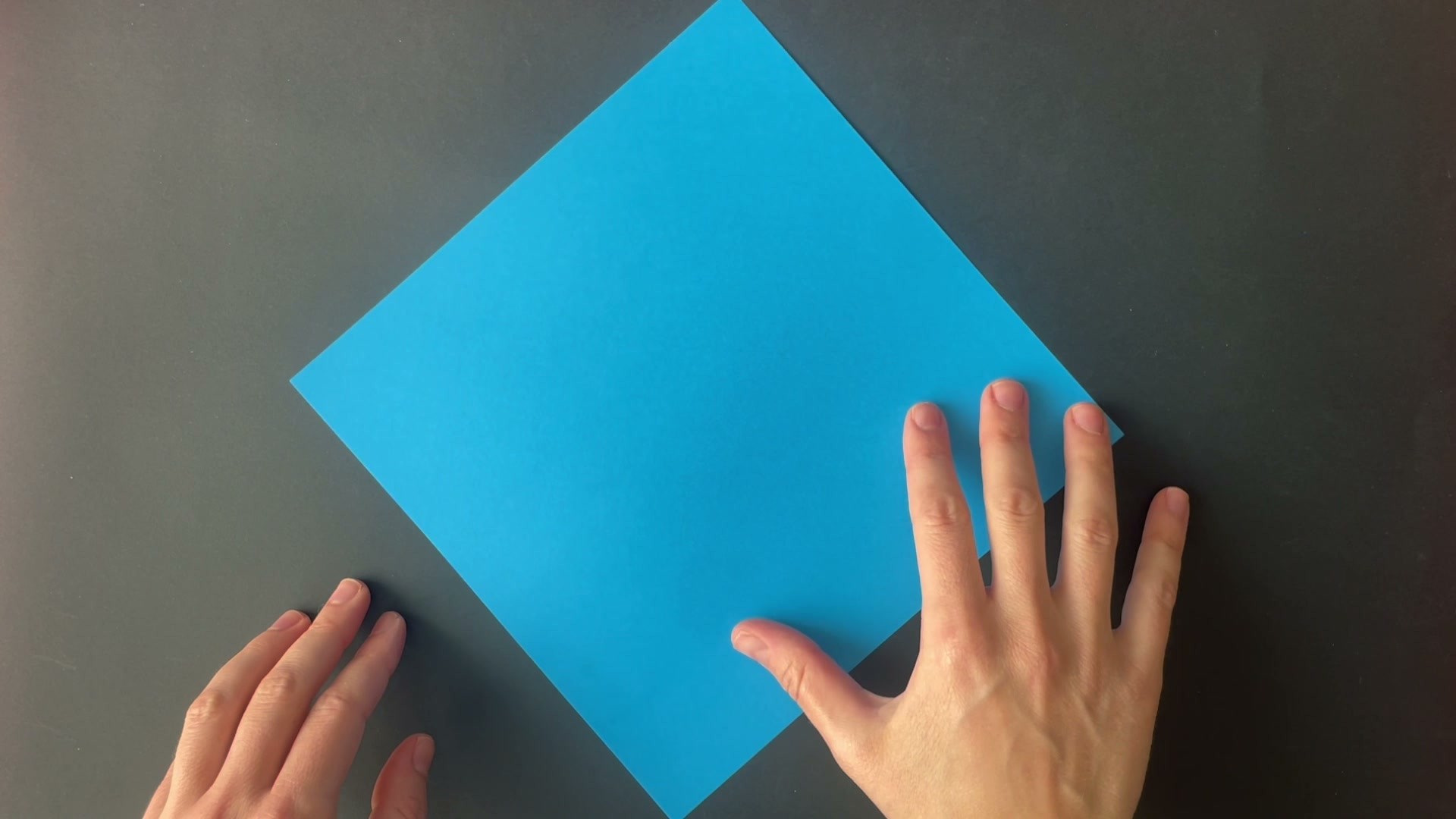}};
\node[img] (a3)  at (2\xgapfourteen, 0)  {\includegraphics[width=\imgwfourteen]{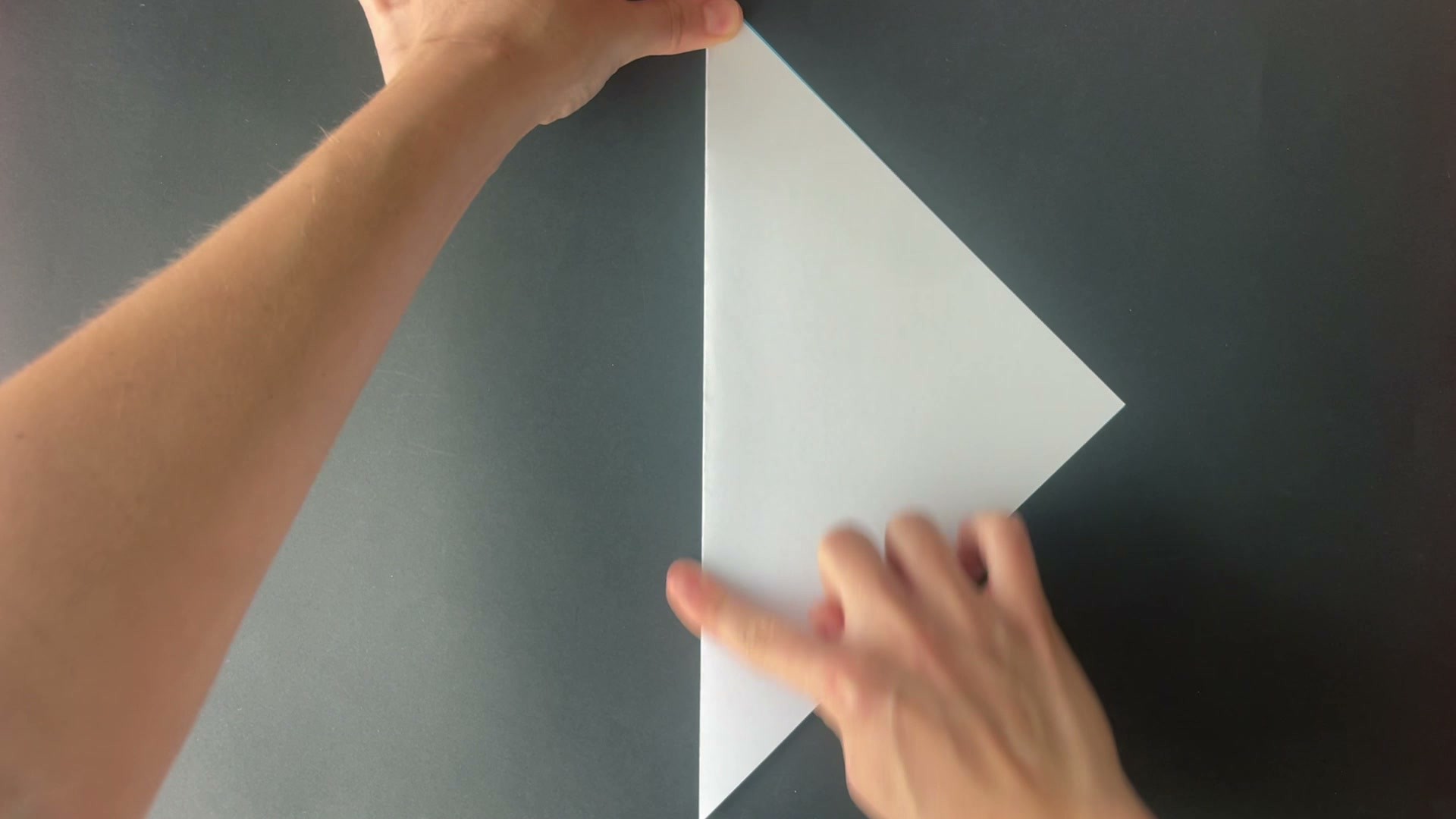}};
\node[img] (a4)  at (3\xgapfourteen, 0)  {\includegraphics[width=\imgwfourteen]{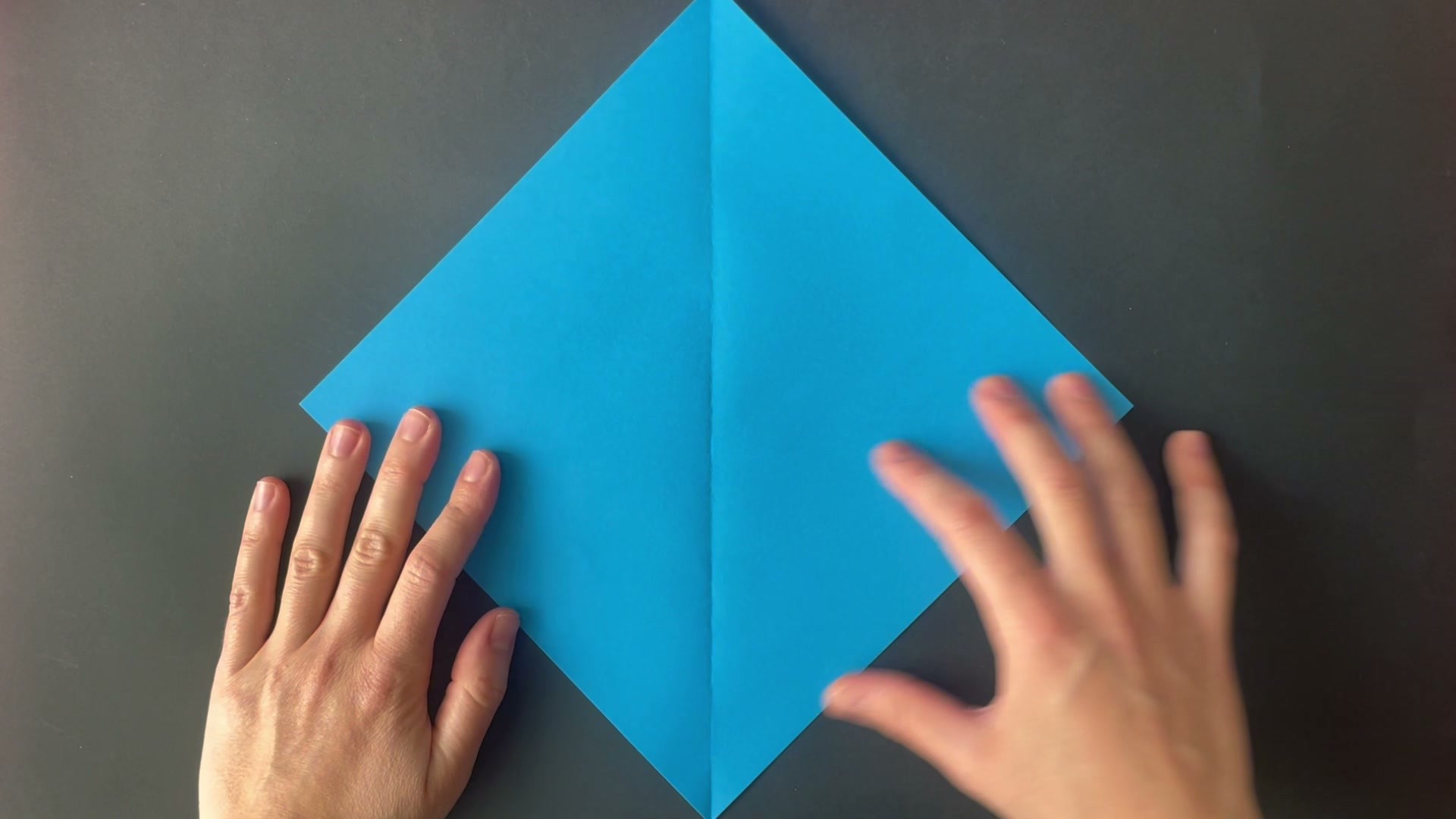}};
\node[img] (a5)  at (4\xgapfourteen, 0)  {\includegraphics[width=\imgwfourteen]{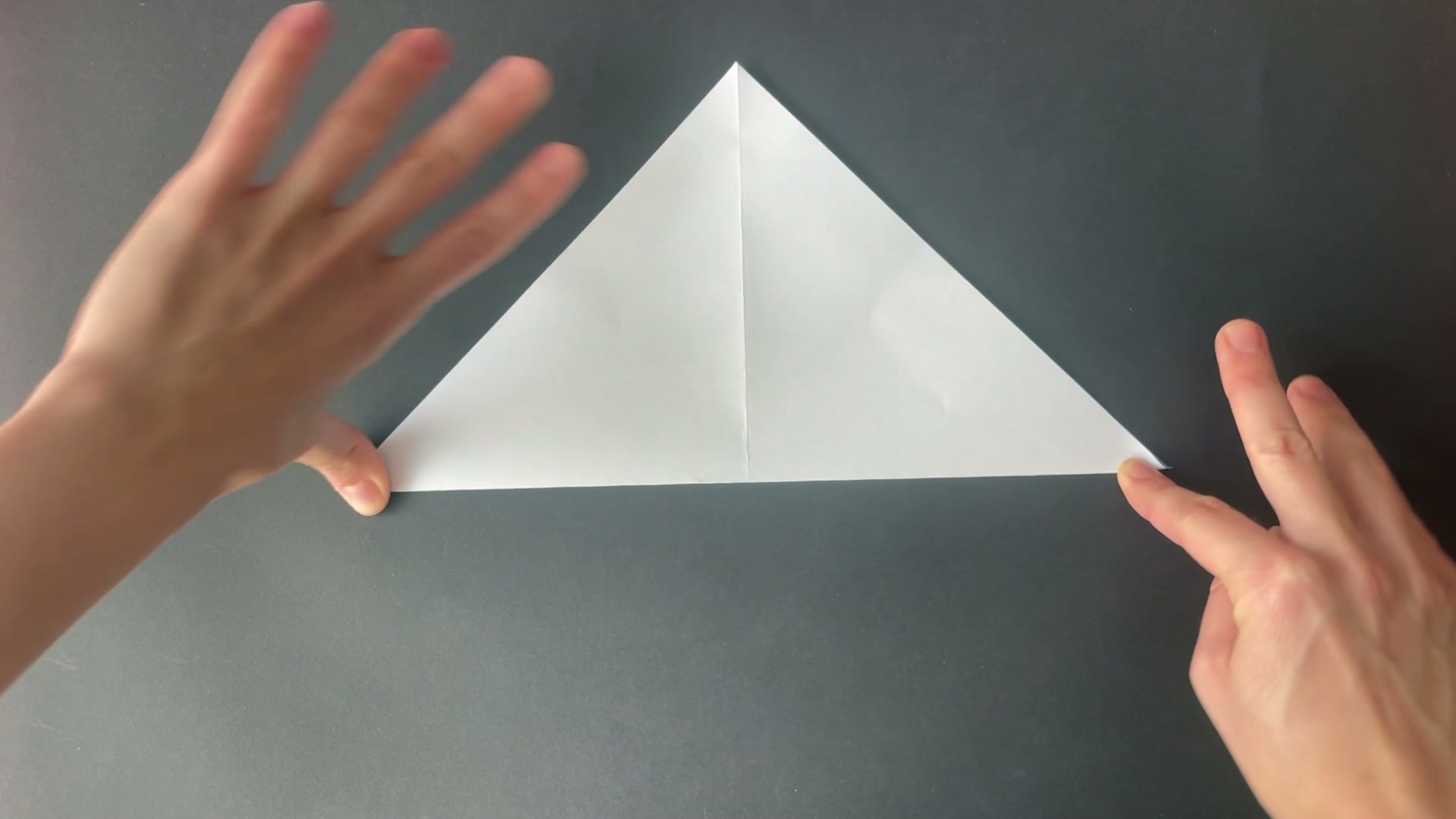}};
\node[img] (a6)  at (5\xgapfourteen, 0)  {\includegraphics[width=\imgwfourteen]{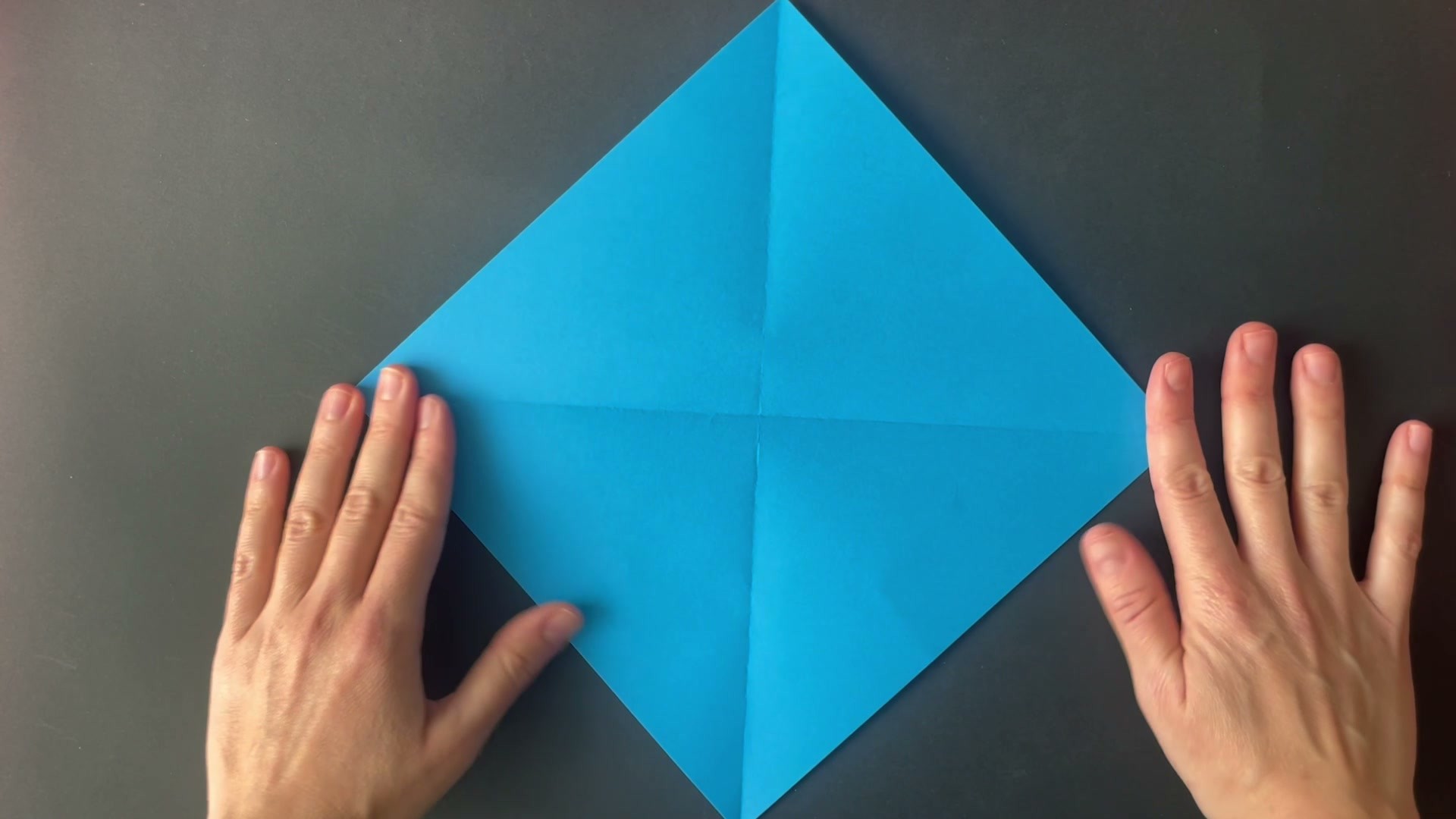}};
\node[img] (a7)  at (6\xgapfourteen, 0)  {\includegraphics[width=\imgwfourteen]{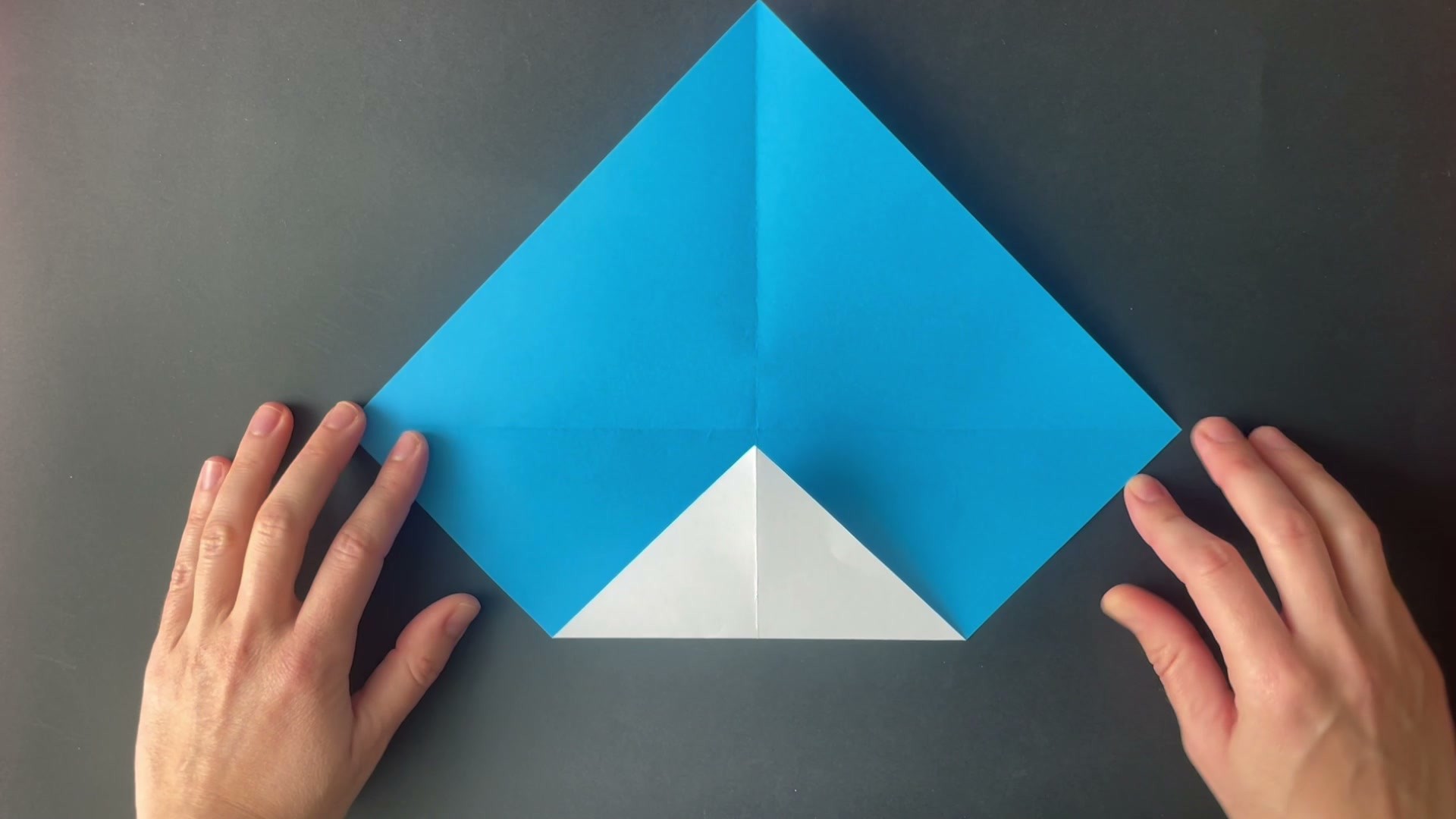}};

\node[frameid] at ([xshift=1pt,yshift=-1pt]a1.north west) {1};
\node[frameid] at ([xshift=1pt,yshift=-1pt]a2.north west) {2};
\node[frameid] at ([xshift=1pt,yshift=-1pt]a3.north west) {3};
\node[frameid] at ([xshift=1pt,yshift=-1pt]a4.north west) {4};
\node[frameid] at ([xshift=1pt,yshift=-1pt]a5.north west) {5};
\node[frameid] at ([xshift=1pt,yshift=-1pt]a6.north west) {6};
\node[frameid] at ([xshift=1pt,yshift=-1pt]a7.north west) {7};

\node[img] (b1)  at (0\xgapfourteen, \ygapfourteen)  {\secondrowfigfourteen{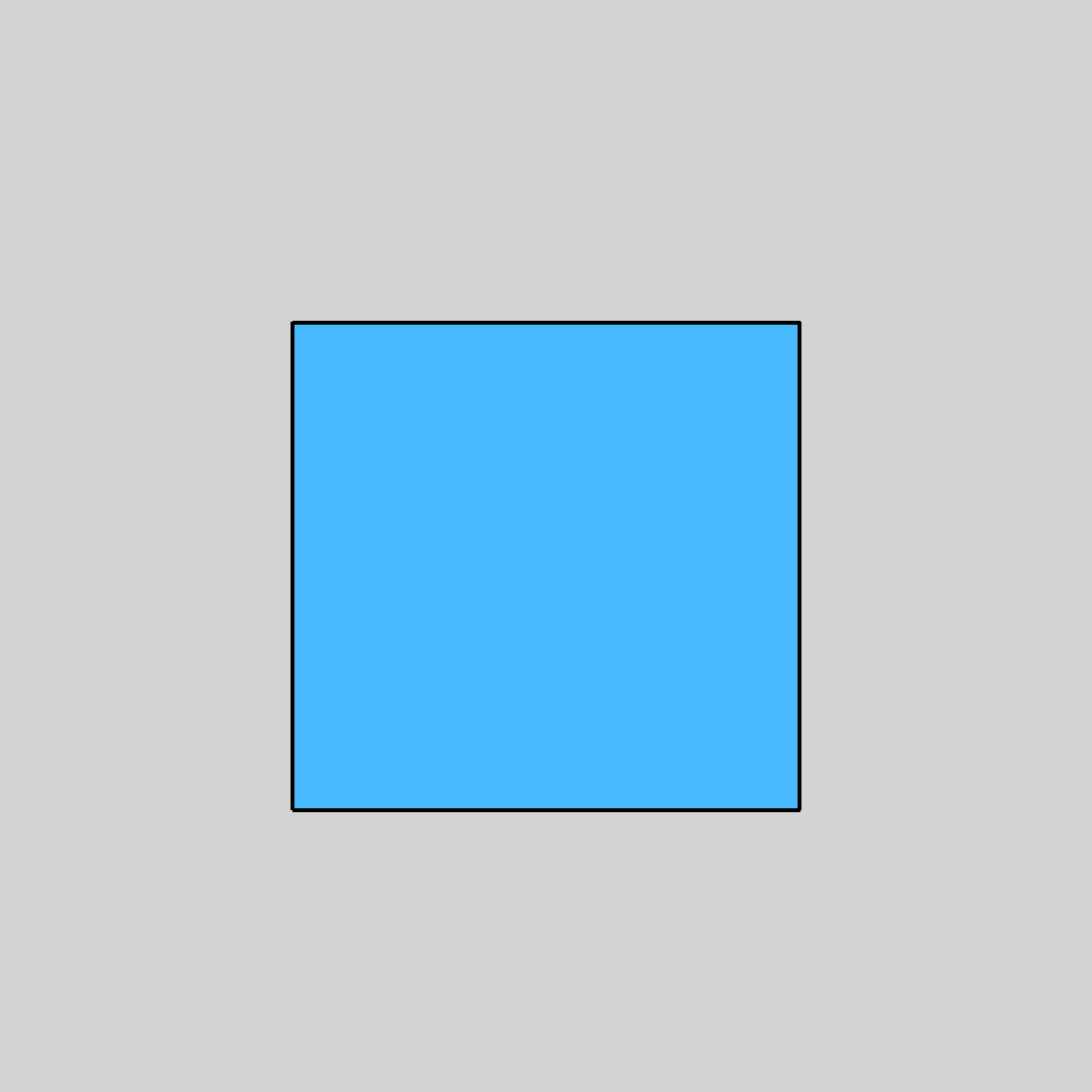}};
\node[img] (b2)  at (1\xgapfourteen, \ygapfourteen)  {\secondrowfigfourteen{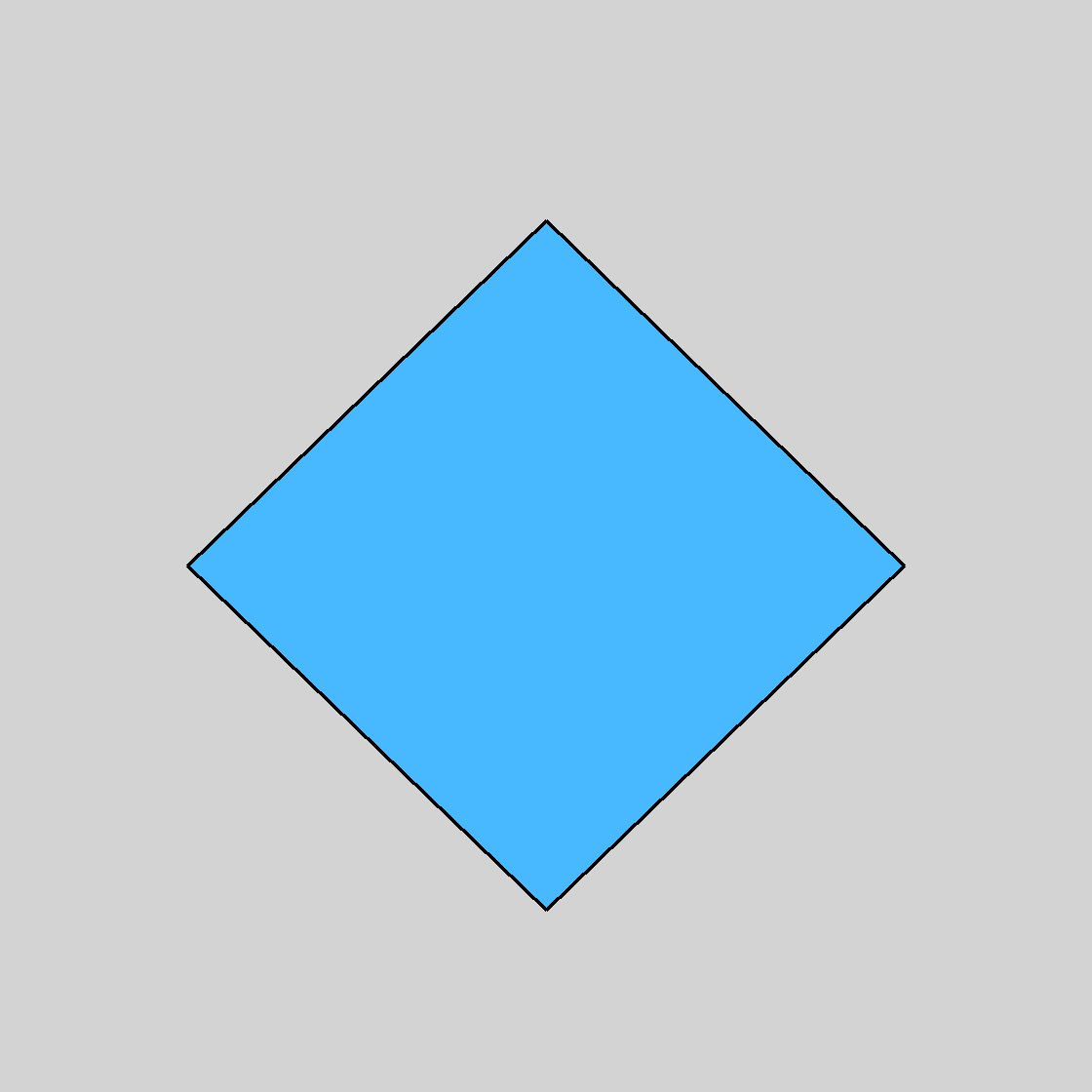}};
\node[img] (b3)  at (2\xgapfourteen, \ygapfourteen)  {\secondrowfigfourteen{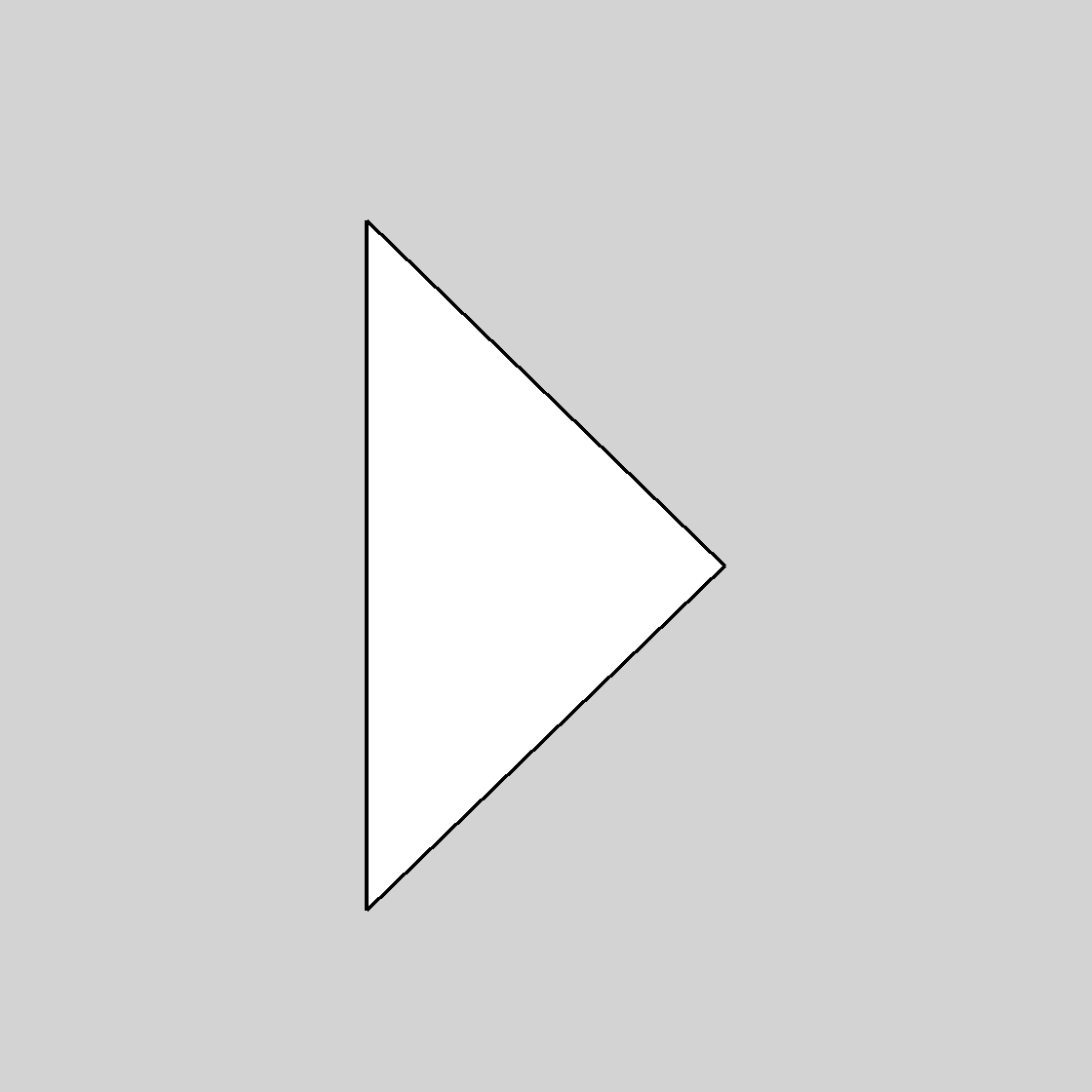}};
\node[img] (b4)  at (3\xgapfourteen, \ygapfourteen)  {\secondrowfigfourteen{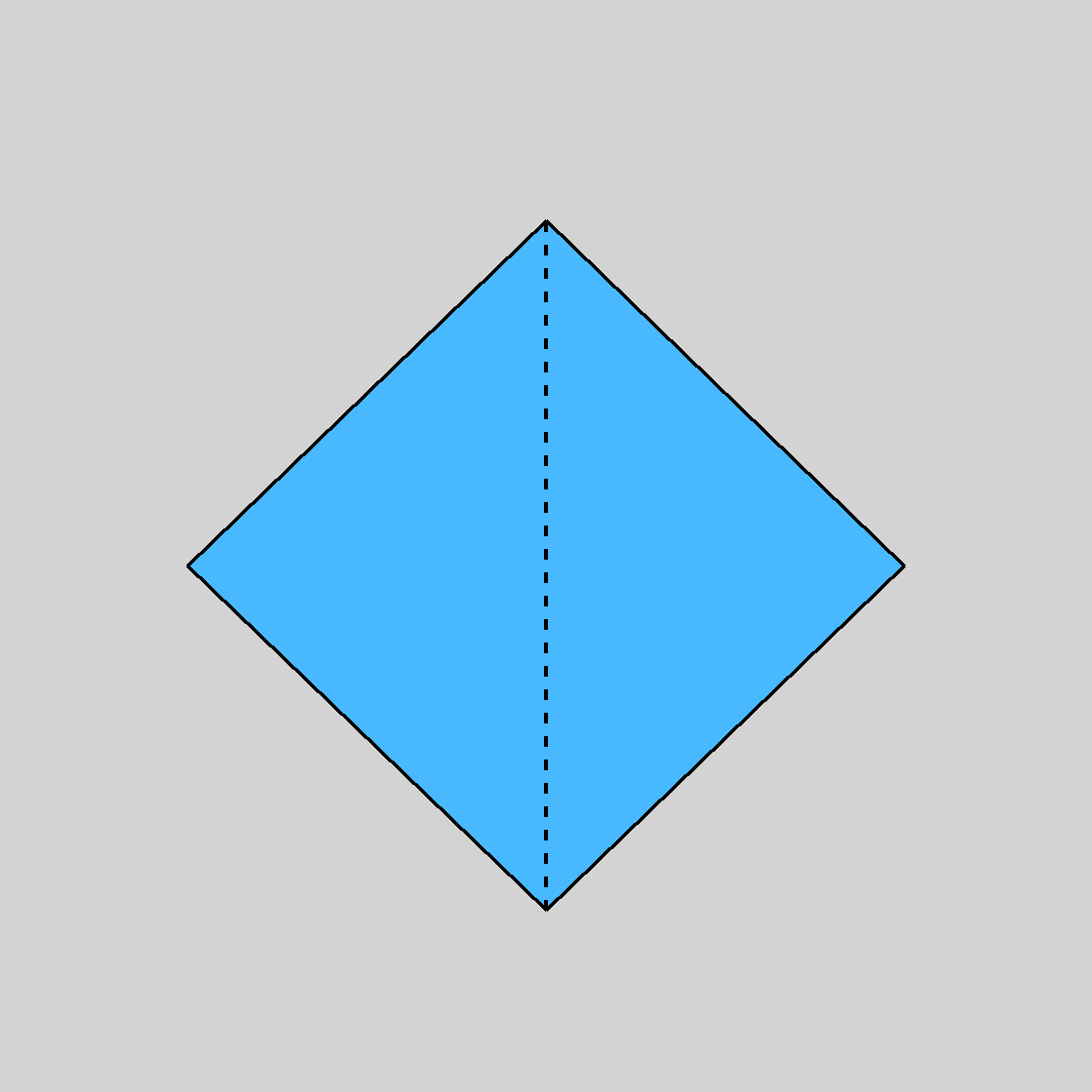}};
\node[img] (b5)  at (4\xgapfourteen, \ygapfourteen)  {\secondrowfigfourteen{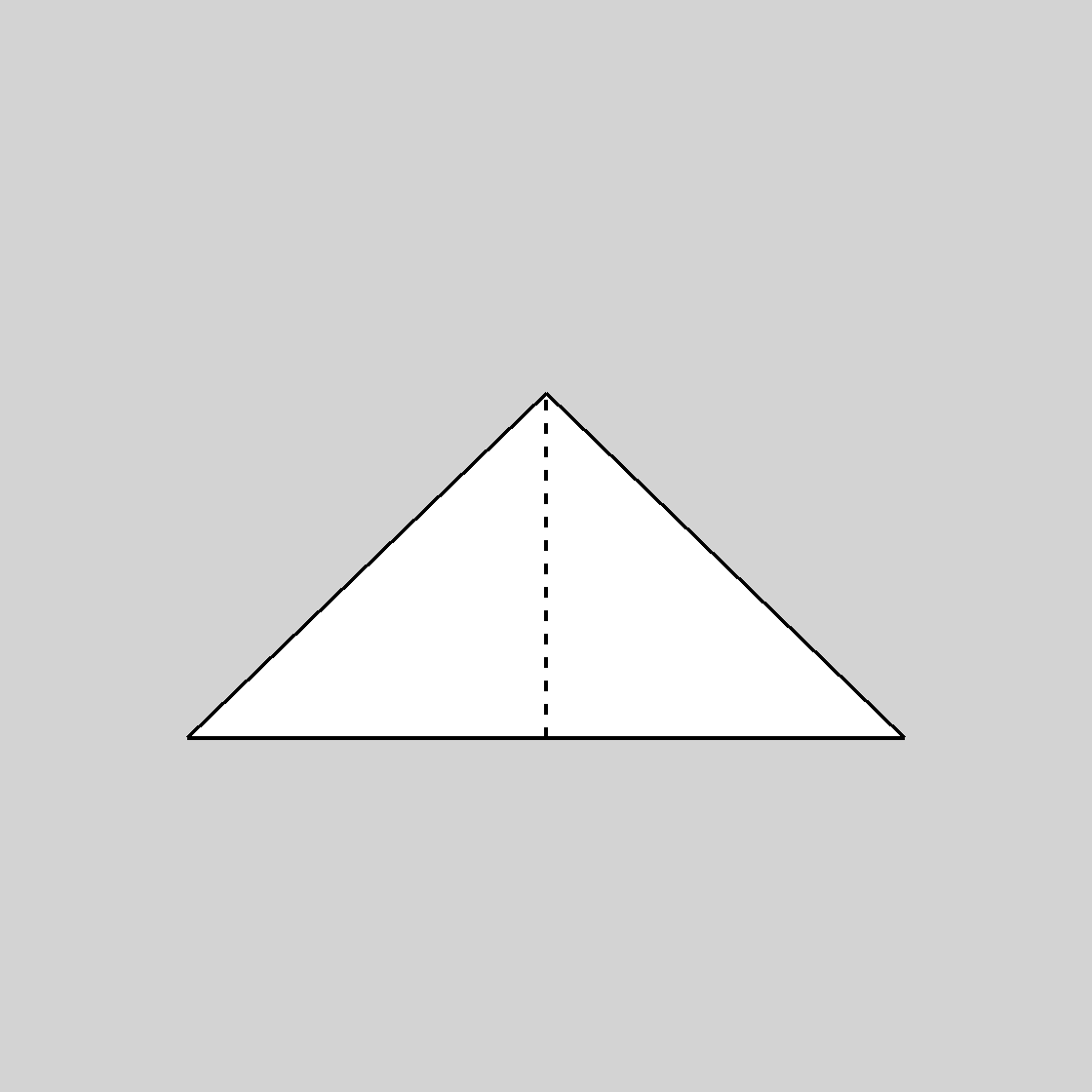}};
\node[img] (b6)  at (5\xgapfourteen, \ygapfourteen)  {\secondrowfigfourteen{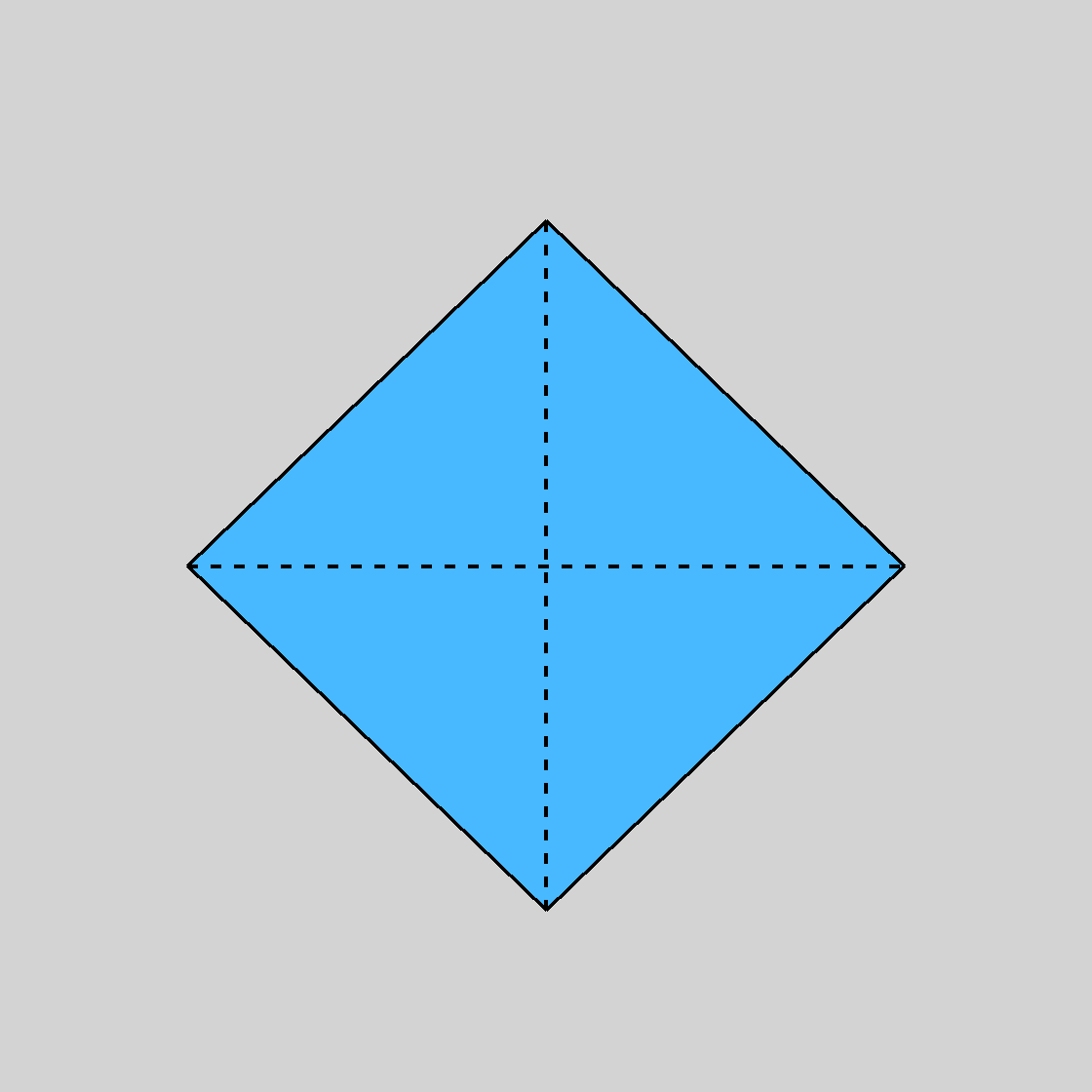}};
\node[img] (b7)  at (6\xgapfourteen, \ygapfourteen)  {\secondrowfigfourteen{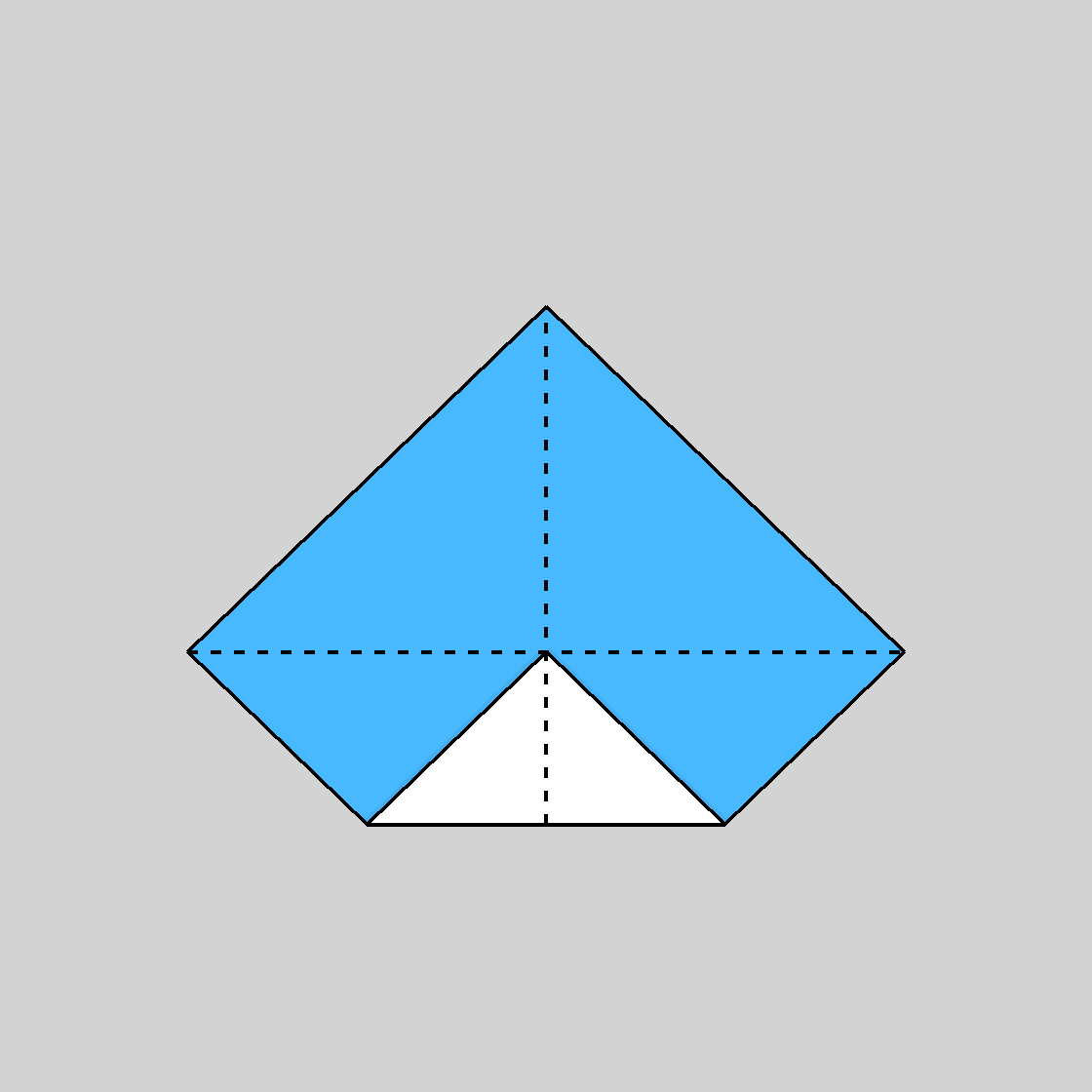}};

\node[cmd] at ($(b1)!0.5!(b2) + (0,-0.77)$) {
rotate(-45$^\circ$)
};

\node[cmd] at ($(b2)!0.5!(b3) + (0,-0.77)$) {
fold([2, 4],-1)
};

\node[cmd] at ($(b3)!0.5!(b4) + (0,-0.77)$) {
unfold()
};

\node[cmd] at ($(b4)!0.5!(b5) + (0,-0.77)$) {
fold([1,3],-1)
};

\node[cmd] at ($(b5)!0.5!(b6) + (0,-0.77)$) {
unfold()
};

\node[cmd] at ($(b6)!0.5!(b7) + (0,-0.77)$) {
add\_v([4,1],0.5)\\
add\_v([3,4],0.5)\\
fold([6,7],-1)
};

\node[cmd] at ($(b6)!0.5!(b7) + (2.1,-0.77)$) {
flip(x)
};

\end{tikzpicture}

\vspace{10pt}

\begin{tikzpicture}[
    img/.style={
        inner sep=0pt,
        outer sep=0pt
    },
    cmd/.style={
        fill=white,
        fill opacity=0.90,
        text opacity=1,
        draw=black!30,
        rounded corners=1.2pt,
        inner sep=1pt,
        font=\tiny\ttfamily,
        align=left
    },
    frameid/.style={
        fill=white,
        fill opacity=0,
        text opacity=1,
        text=white,
        draw=black!25,
        rounded corners=0pt,
        inner sep=1pt,
        font=\tiny\ttfamily,
        anchor=north west
    }
]

\node[img] (a1)  at (0\xgapfourteen, 0)  {\includegraphics[width=\imgwfourteen]{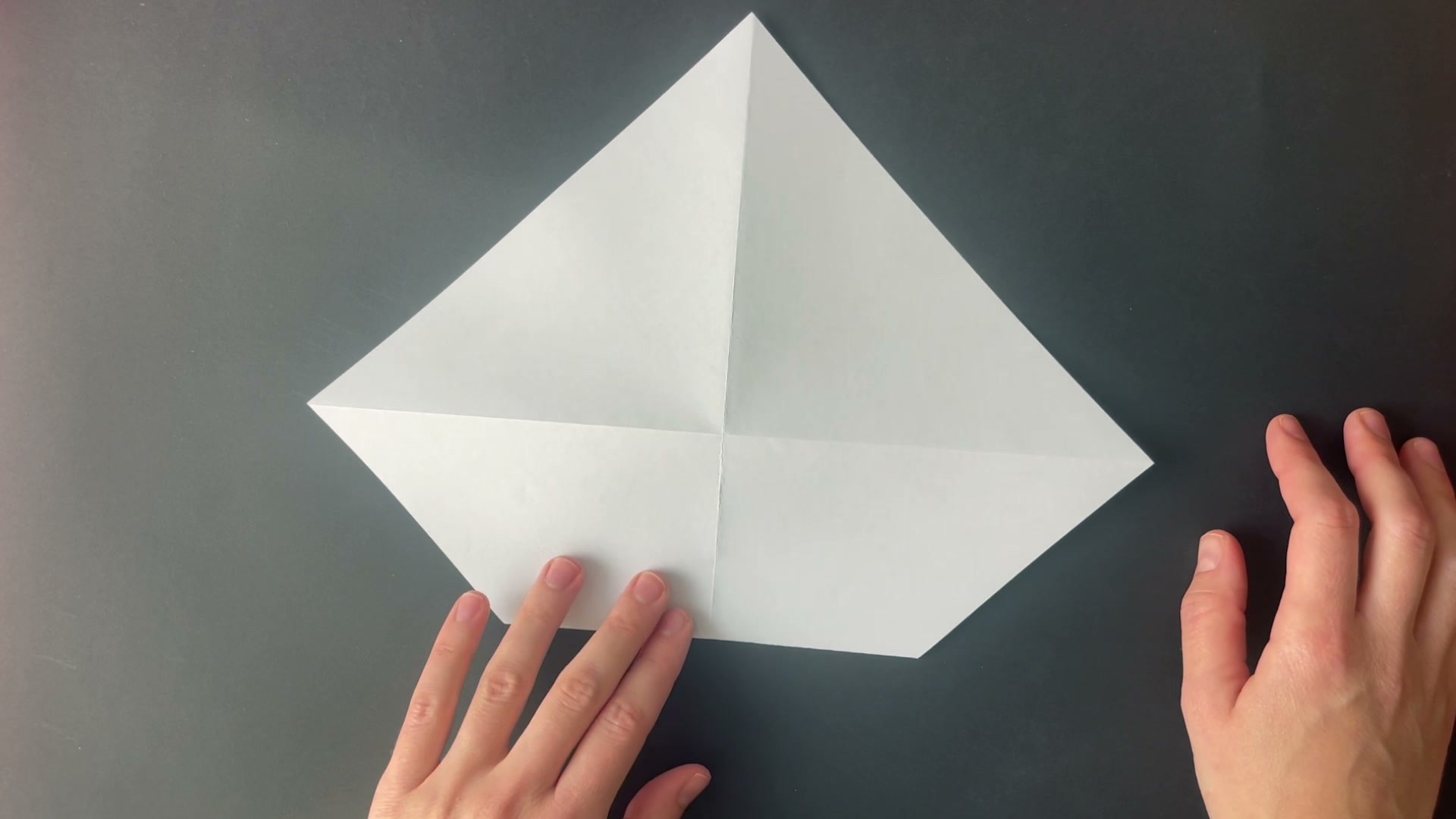}};
\node[img] (a2)  at (1\xgapfourteen, 0)  {\includegraphics[width=\imgwfourteen]{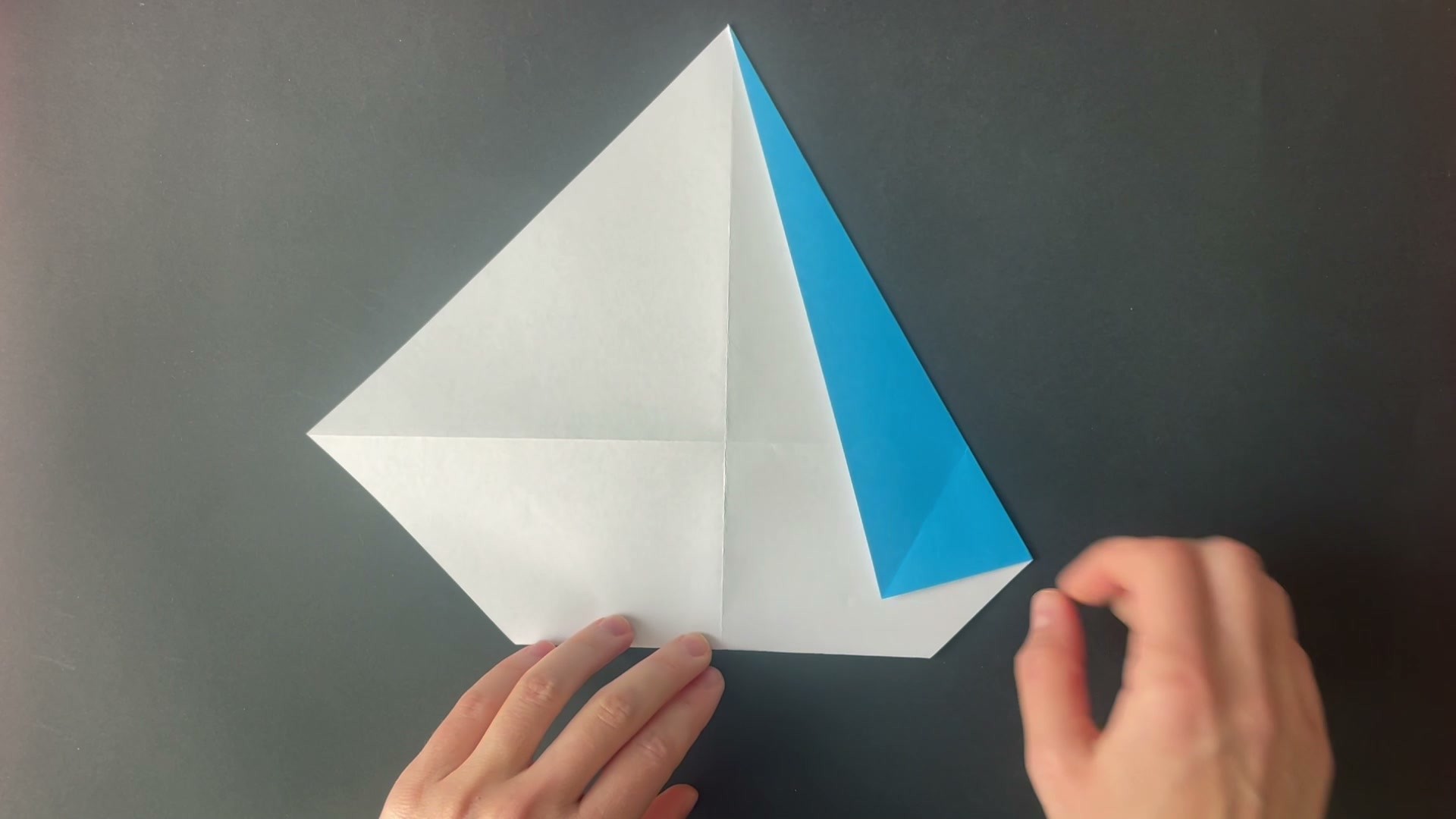}};
\node[img] (a3)  at (2\xgapfourteen, 0)  {\includegraphics[width=\imgwfourteen]{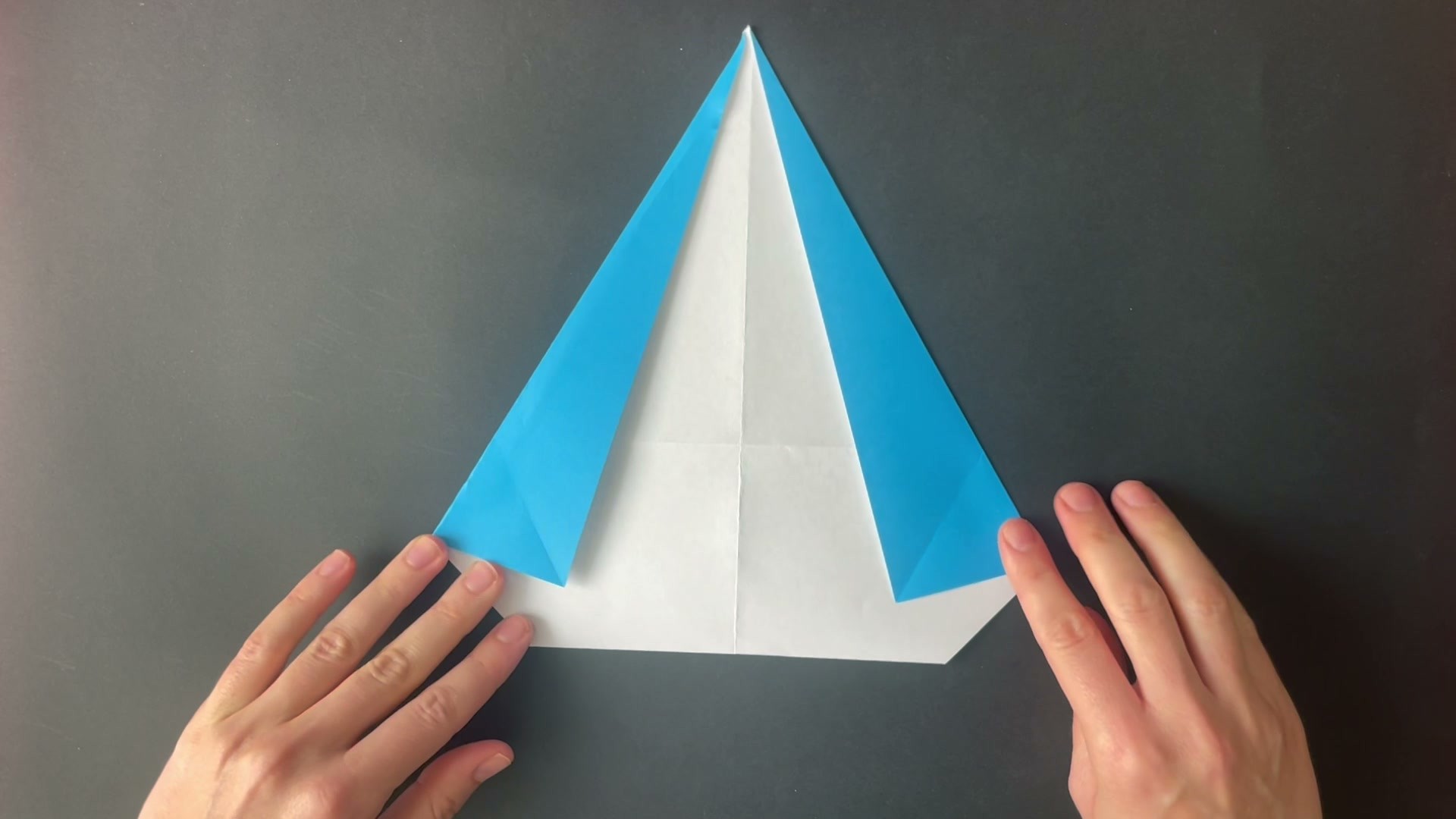}};
\node[img] (a4)  at (3\xgapfourteen, 0)  {\includegraphics[width=\imgwfourteen]{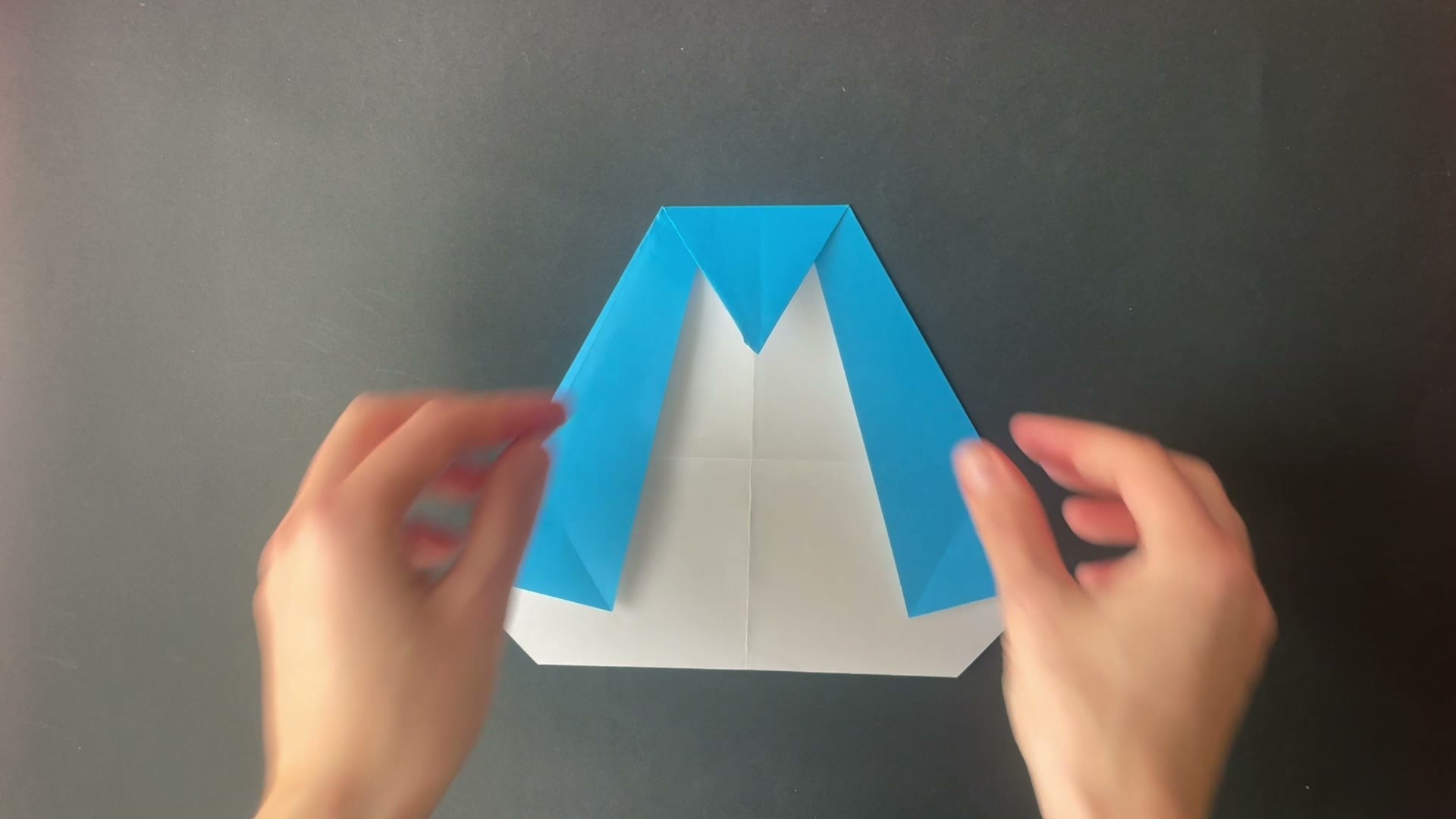}};
\node[img] (a5)  at (4\xgapfourteen, 0)  {\includegraphics[width=\imgwfourteen]{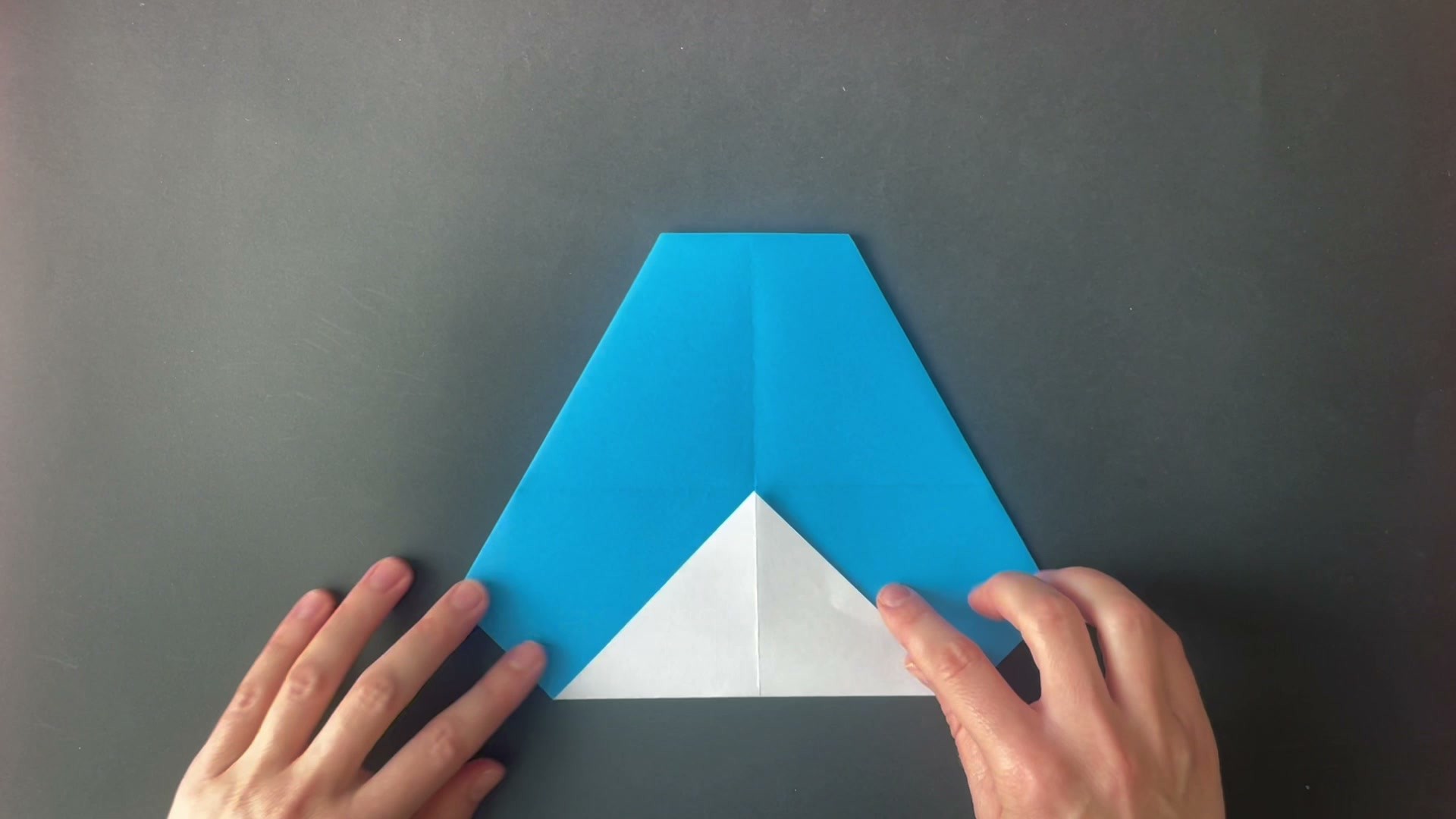}};
\node[img] (a6)  at (5\xgapfourteen, 0)  {\includegraphics[width=\imgwfourteen]{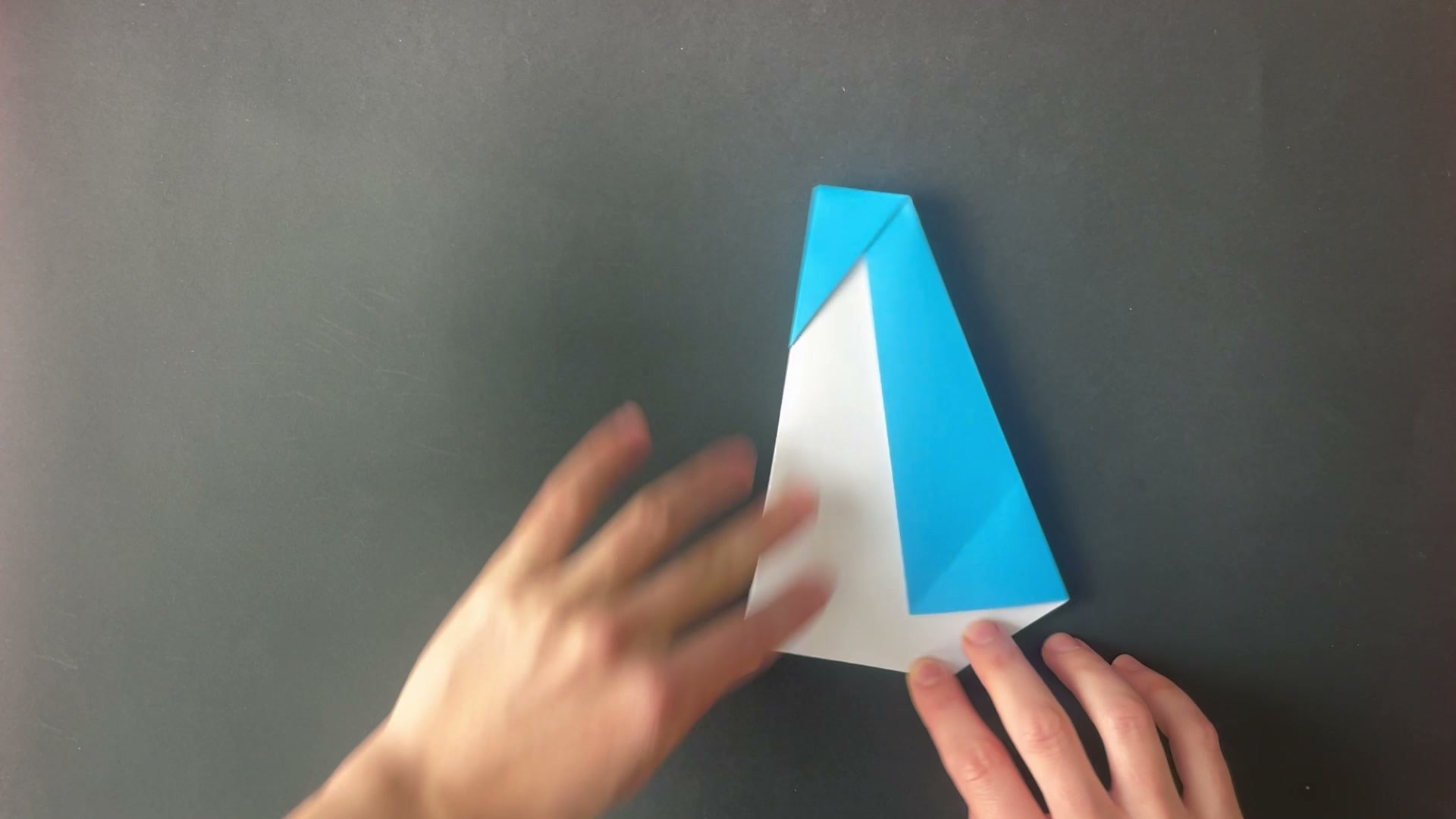}};

\node[frameid] at ([xshift=1pt,yshift=-1pt]a1.north west) {8};
\node[frameid] at ([xshift=1pt,yshift=-1pt]a2.north west) {9};
\node[frameid] at ([xshift=1pt,yshift=-1pt]a3.north west) {10};
\node[frameid] at ([xshift=1pt,yshift=-1pt]a4.north west) {11};
\node[frameid] at ([xshift=1pt,yshift=-1pt]a5.north west) {12};
\node[frameid] at ([xshift=1pt,yshift=-1pt]a6.north west) {13};

\node[img] (b1)  at (0\xgapfourteen, \ygapfourteen)  {\secondrowfigfourteen{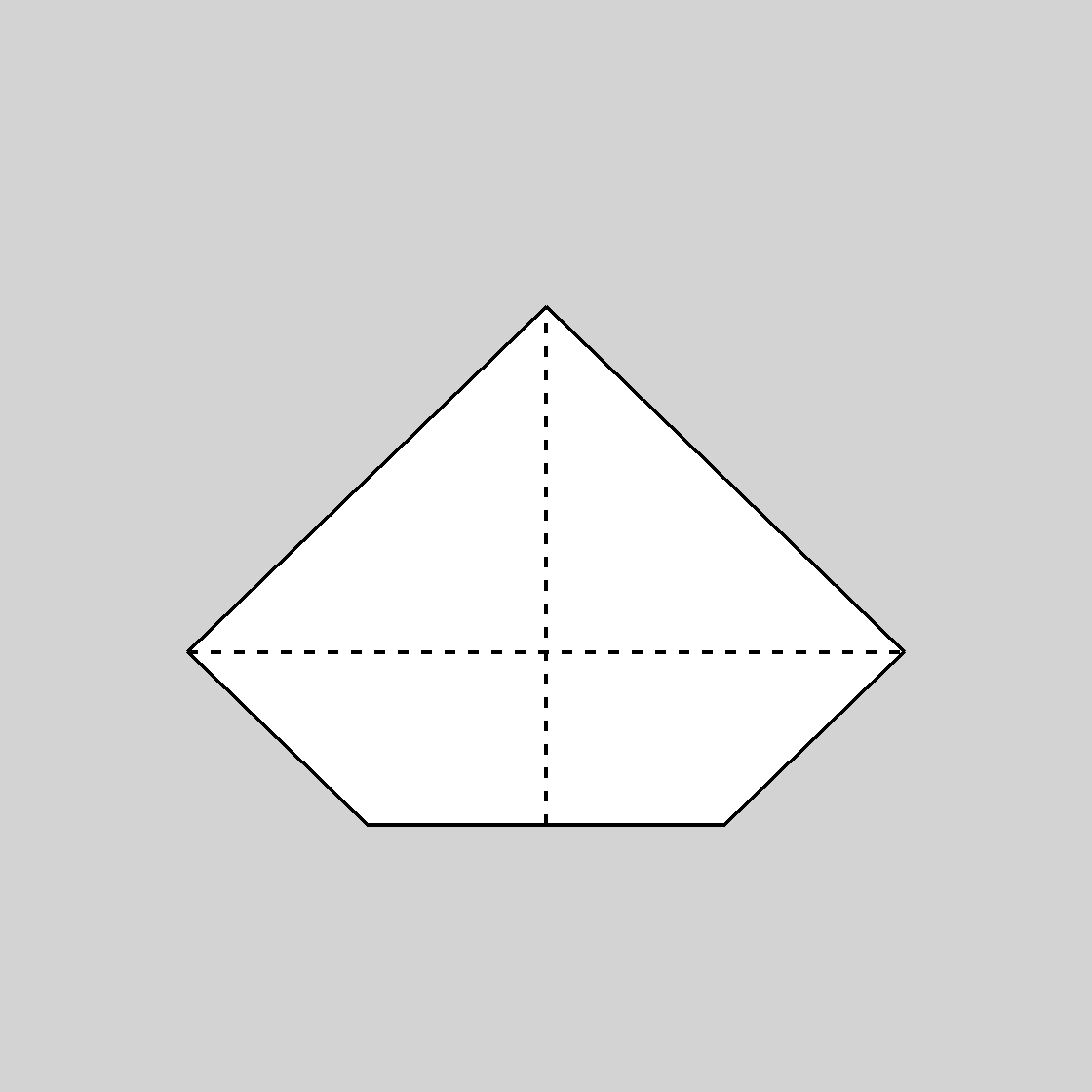}};
\node[img] (b2)  at (1\xgapfourteen, \ygapfourteen)  {\secondrowfigfourteen{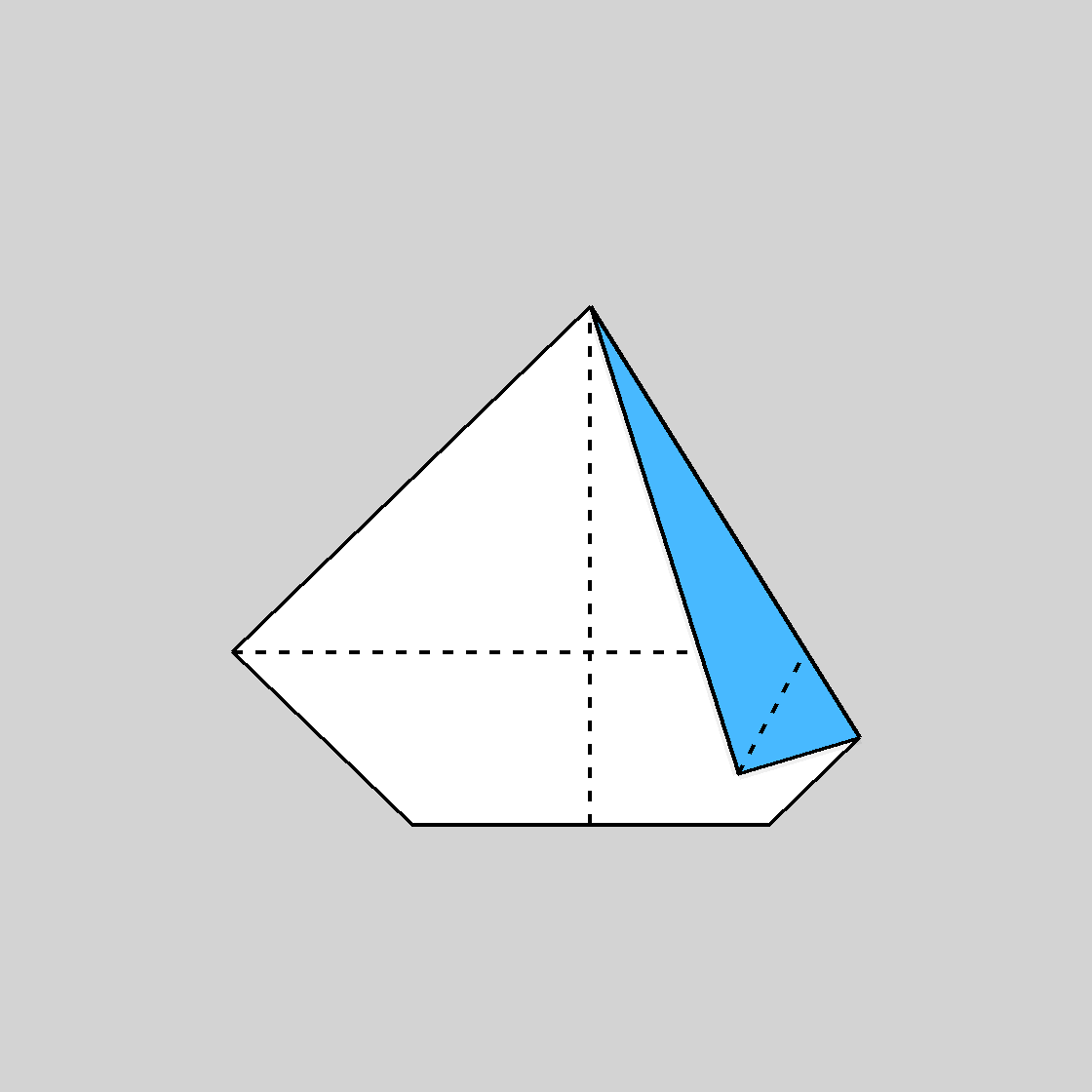}};
\node[img] (b3)  at (2\xgapfourteen, \ygapfourteen)  {\secondrowfigfourteen{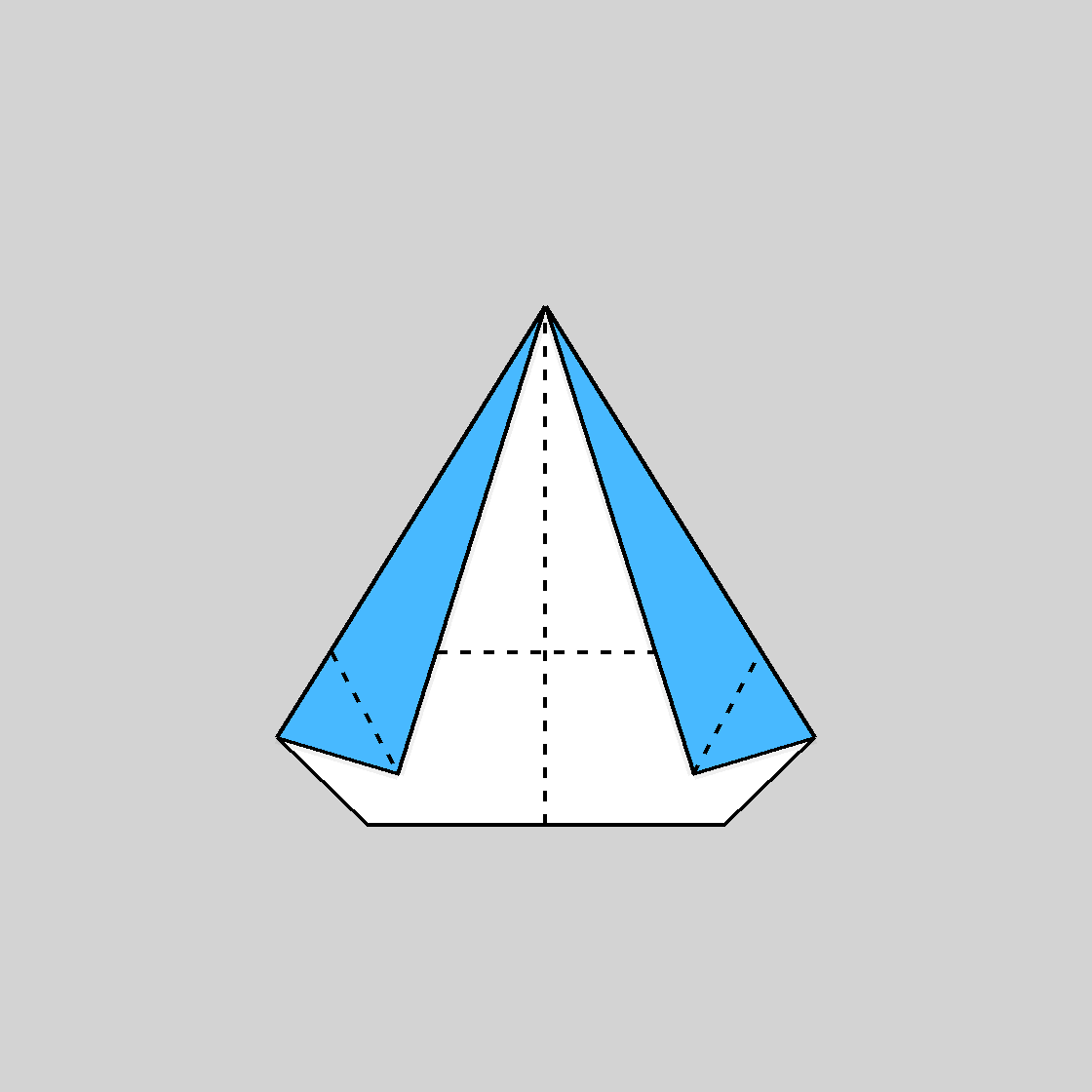}};
\node[img] (b4)  at (3\xgapfourteen, \ygapfourteen)  {\secondrowfigfourteen{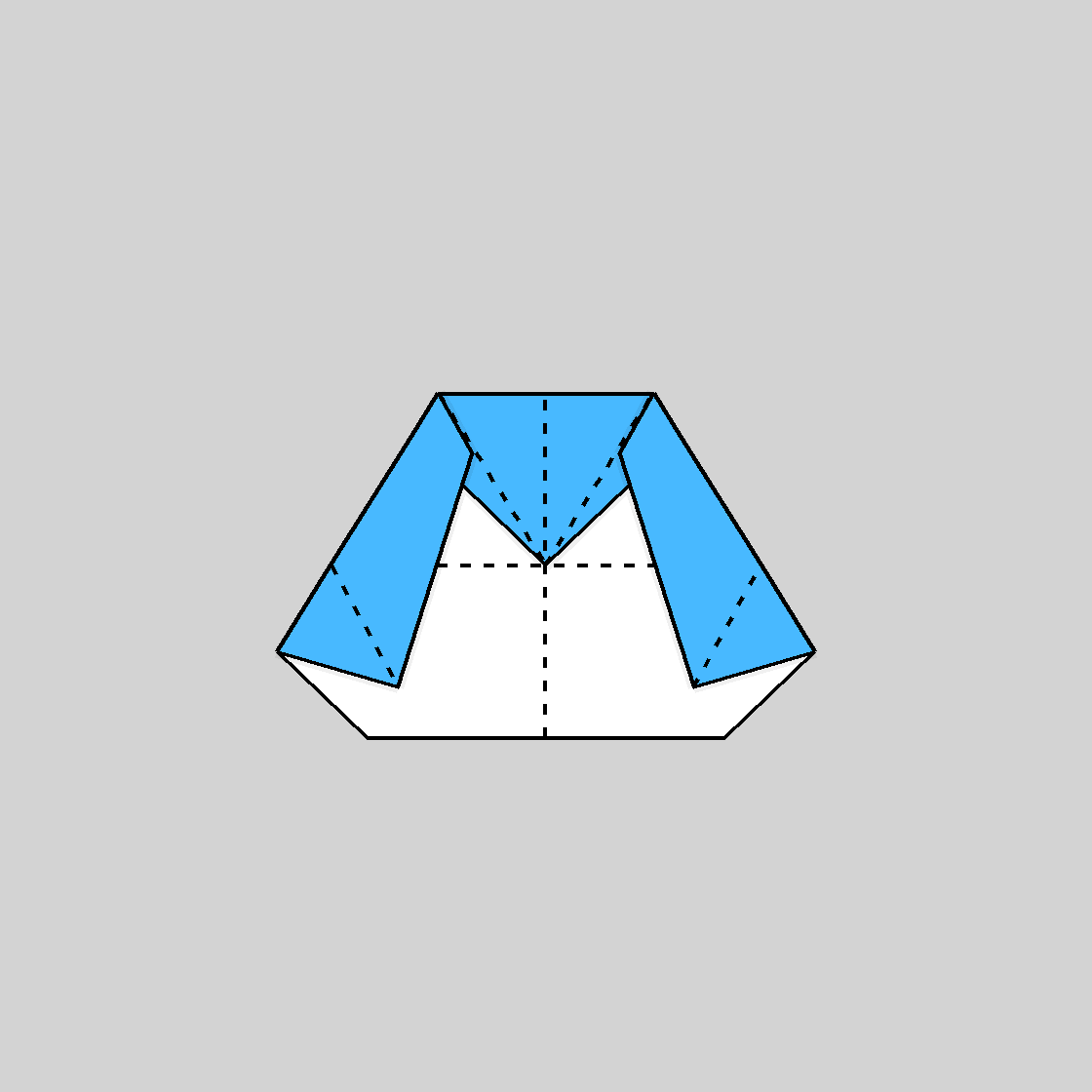}};
\node[img] (b5)  at (4\xgapfourteen, \ygapfourteen)  {\secondrowfigfourteen{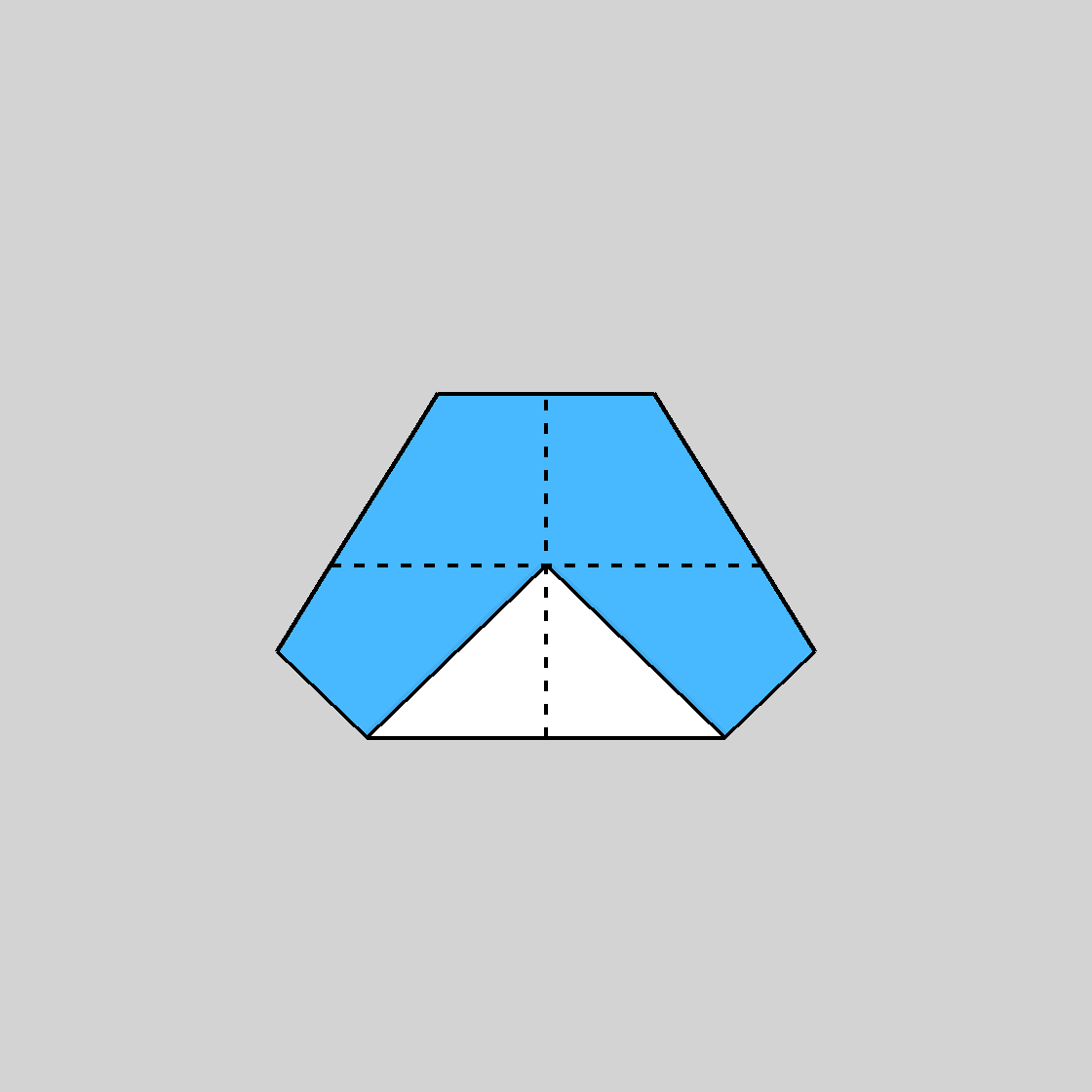}};
\node[img] (b6)  at (5\xgapfourteen, \ygapfourteen)  {\secondrowfigfourteen{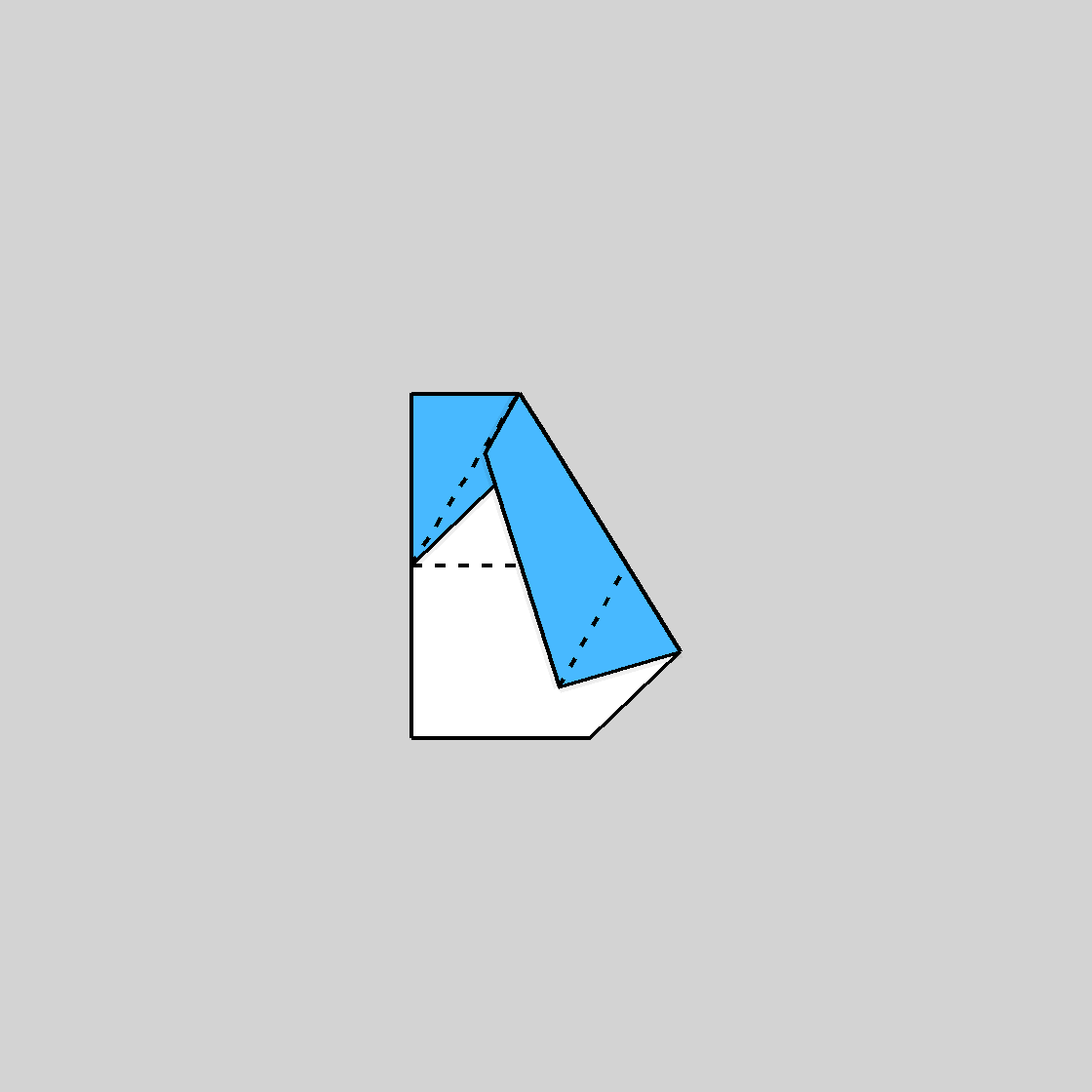}};

\node[cmd] at ($(b1)!0.5!(b2) + (0,-0.77)$) {
add\_v([6,1],0.5)\\
fold([2,9],1)
};

\node[cmd] at ($(b2)!0.5!(b3) + (0,-0.77)$) {
add\_v([3,7],0.5)\\
fold([2,11],1)
};

\node[cmd] at ($(b3)!0.5!(b4) + (0,-0.77)$) {
add\_v([2,3],0.5)\\
add\_v([1,2],0.5)\\
fold([13,14],1)
};

\node[cmd] at ($(b4)!0.5!(b5) + (0,-0.77)$) {
flip(x)
};

\node[cmd] at ($(b5)!0.5!(b6) + (0,-0.77)$) {
fold([5,8],-1)
};

\end{tikzpicture}

\vspace{10pt}

\centering

\setlength{\imgwfourteen}{0.12\textwidth}

\setlength{\xgapfourteen}{0.126\textwidth}

\setlength{\ygapfourteen}{-0.085\textwidth}

\begin{tikzpicture}[
    img/.style={
        inner sep=0pt,
        outer sep=0pt
    },
    cmd/.style={
        fill=white,
        fill opacity=0.90,
        text opacity=1,
        draw=black!30,
        rounded corners=1.2pt,
        inner sep=1pt,
        font=\tiny\ttfamily,
        align=left
    },
    frameid/.style={
        fill=white,
        fill opacity=0,
        text opacity=1,
        text=white,
        draw=black!25,
        rounded corners=0pt,
        inner sep=1pt,
        font=\tiny\ttfamily,
        anchor=north west
    }
]

\node[img] (a1)  at (0\xgapfourteen, 0)  {\includegraphics[width=\imgwfourteen]{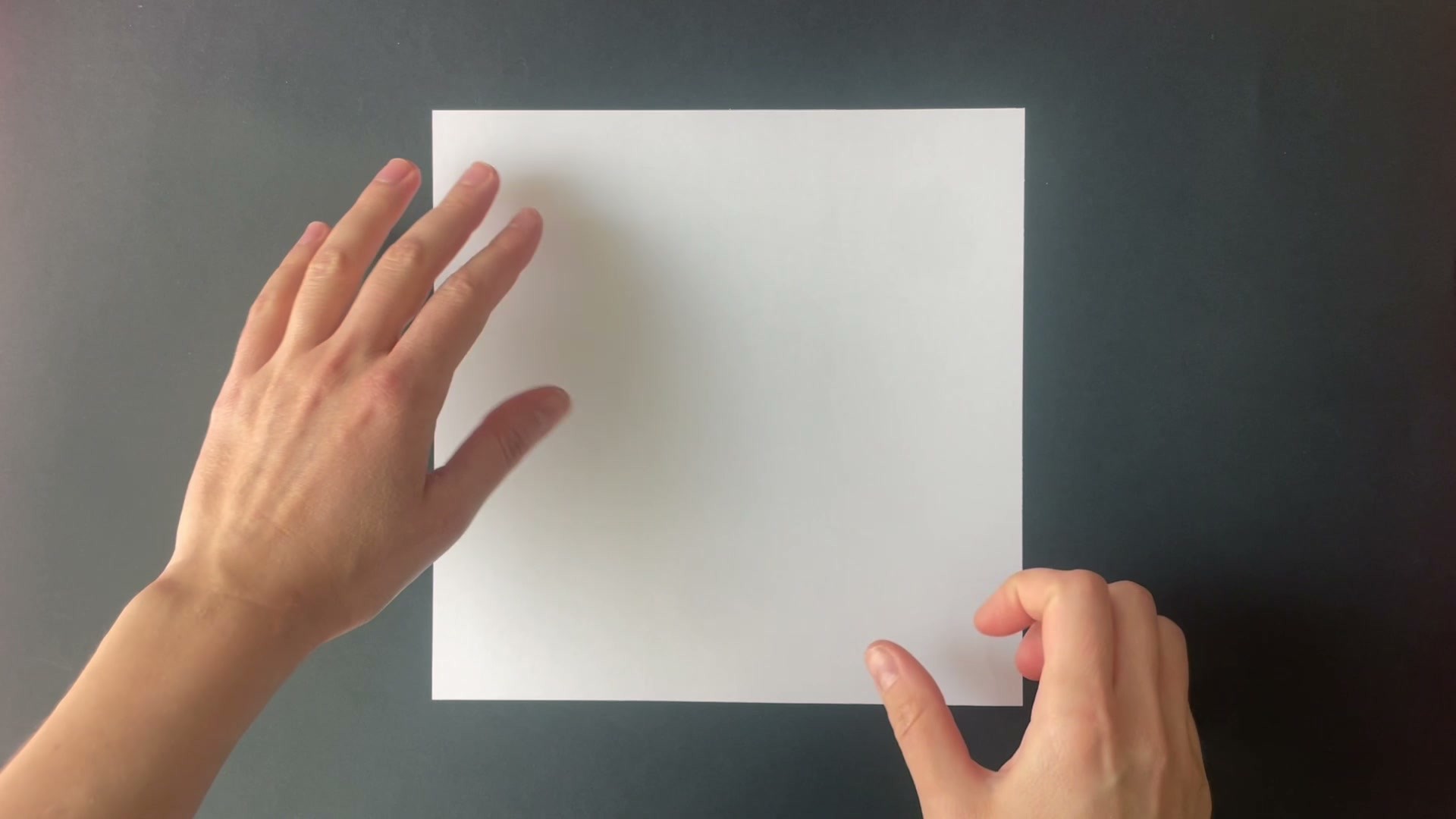}};
\node[img] (a2)  at (1\xgapfourteen, 0)  {\includegraphics[width=\imgwfourteen]{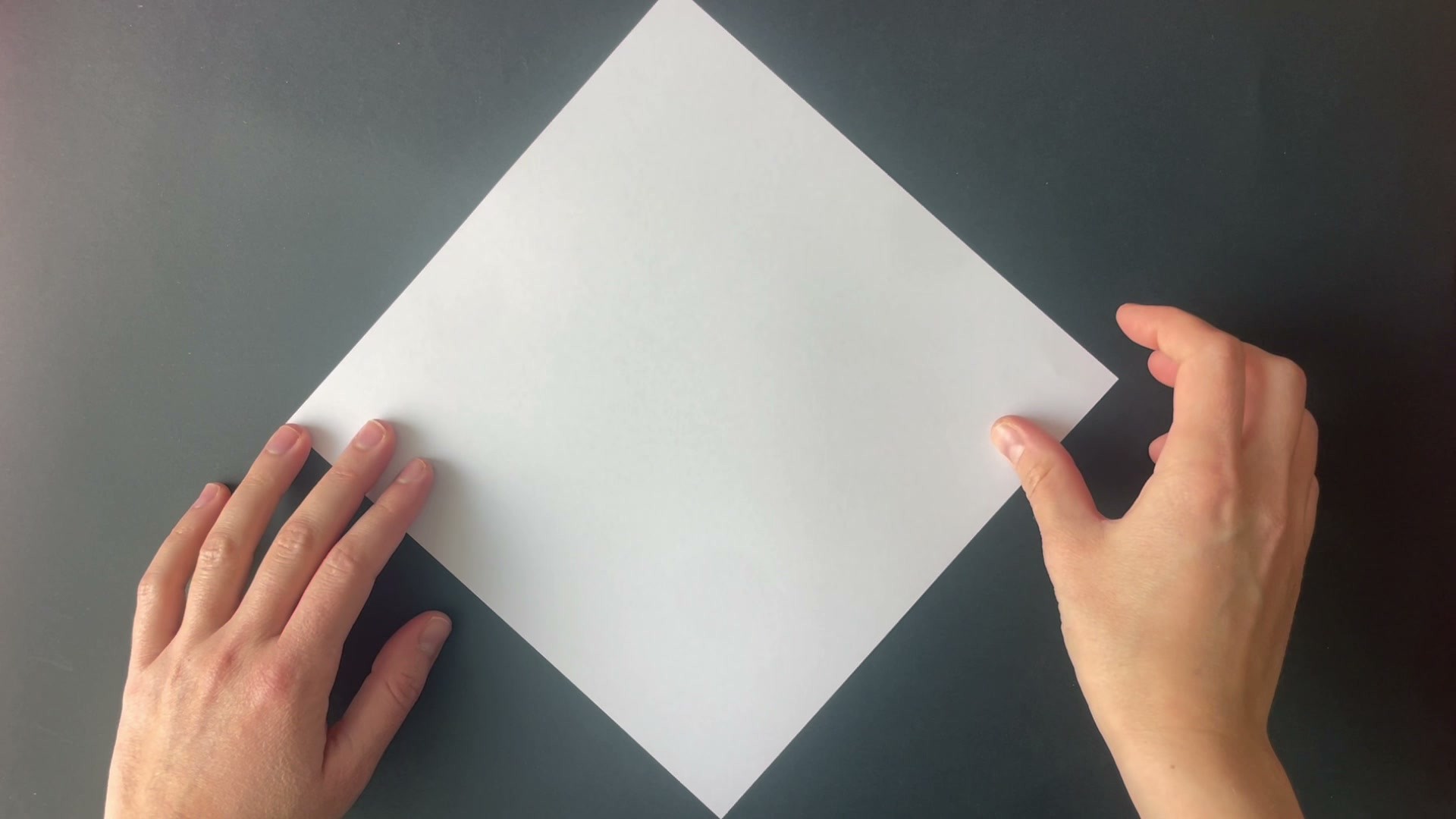}};
\node[img] (a3)  at (2\xgapfourteen, 0)  {\includegraphics[width=\imgwfourteen]{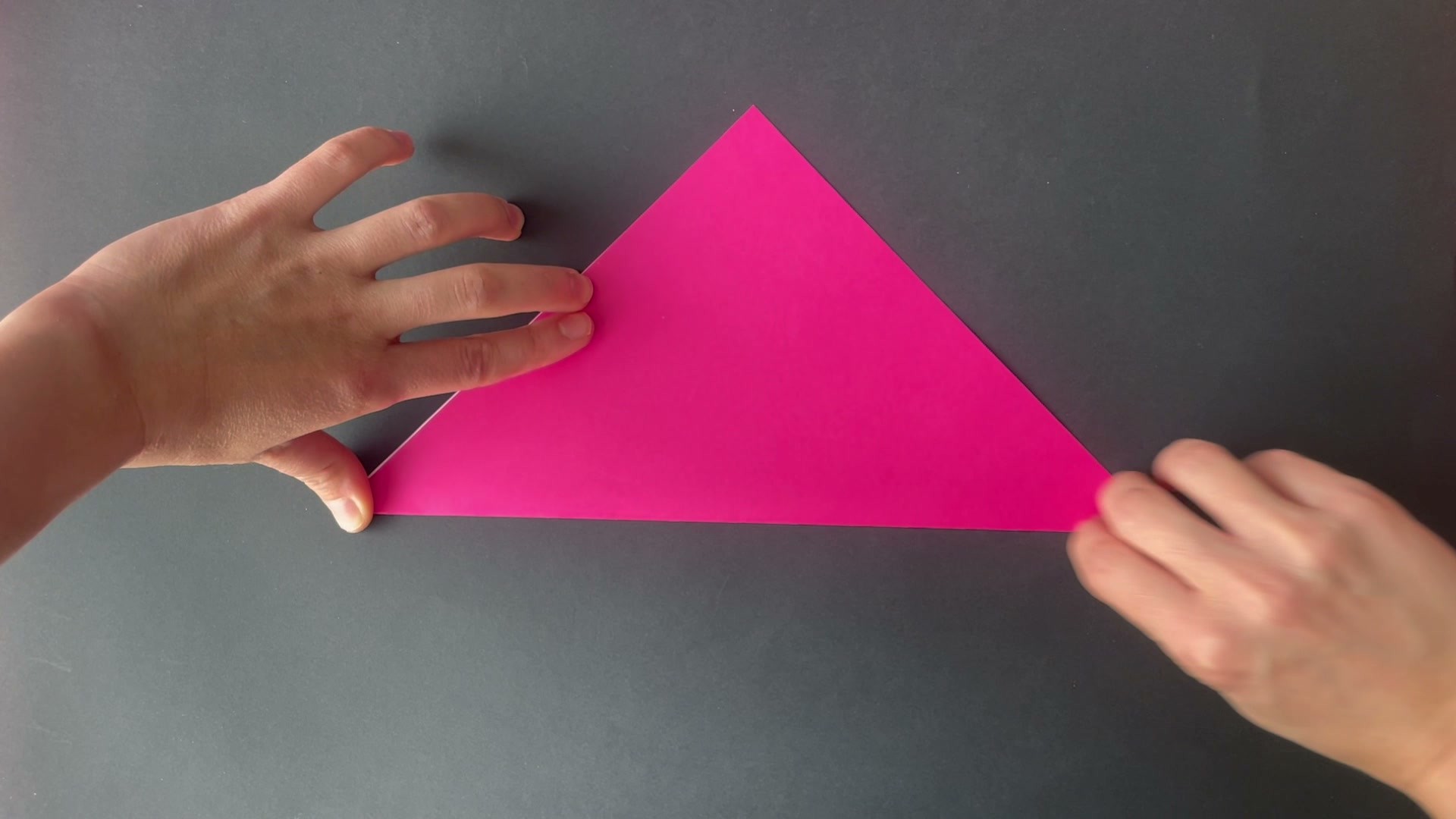}};
\node[img] (a4)  at (3\xgapfourteen, 0)  {\includegraphics[width=\imgwfourteen]{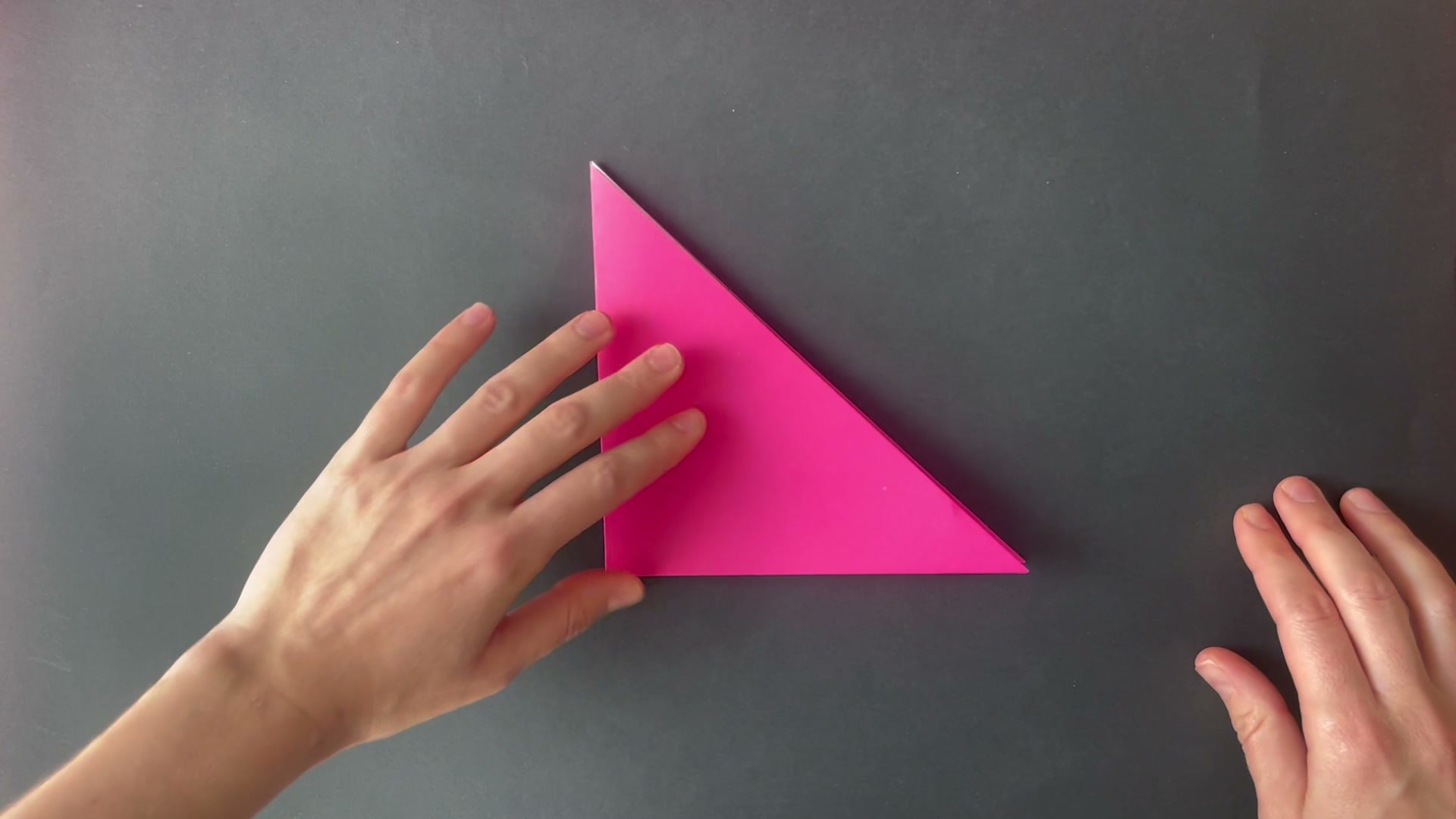}};
\node[img] (a5)  at (4\xgapfourteen, 0)  {\includegraphics[width=\imgwfourteen]{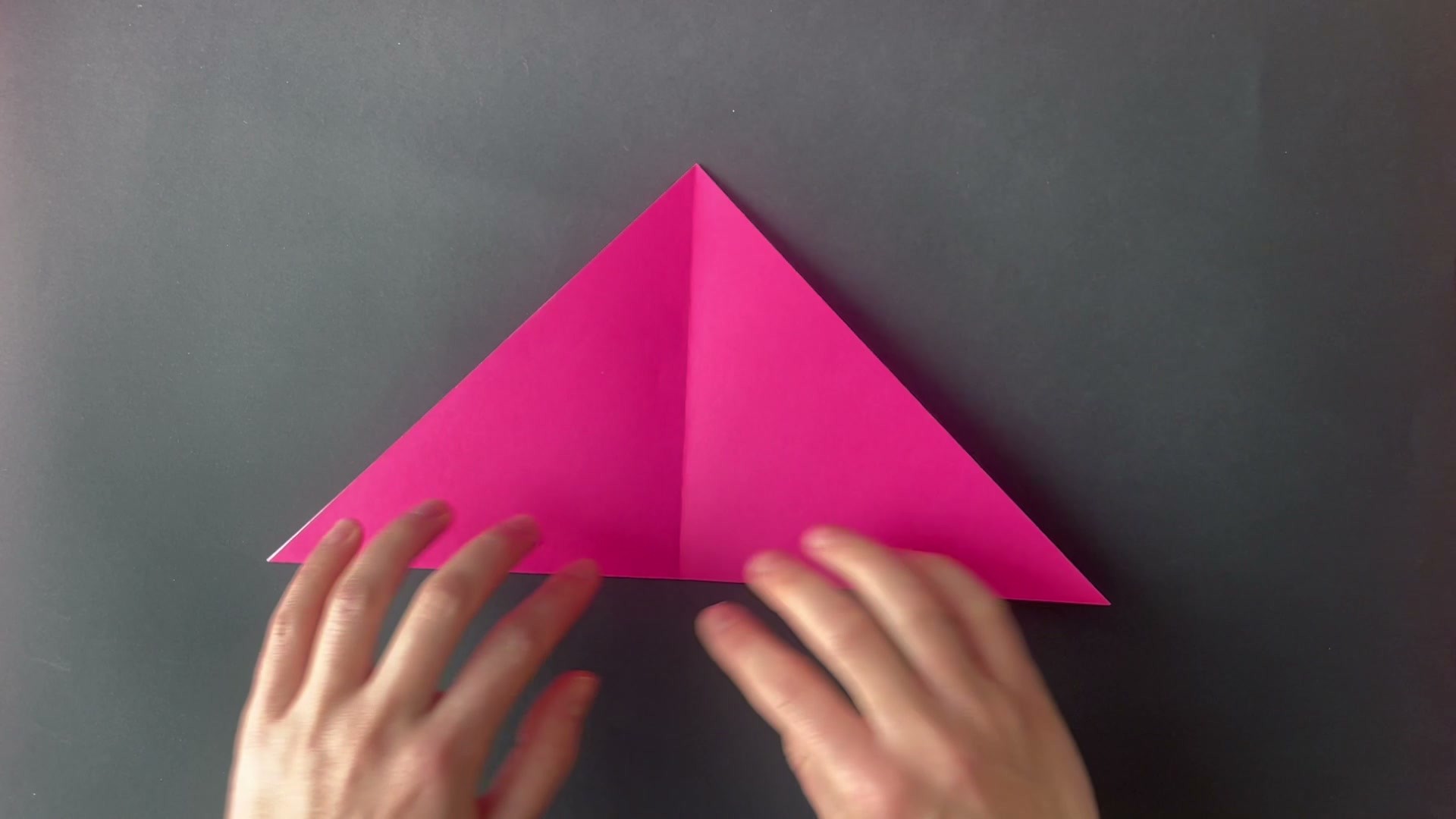}};
\node[img] (a6)  at (5\xgapfourteen, 0)  {\includegraphics[width=\imgwfourteen]{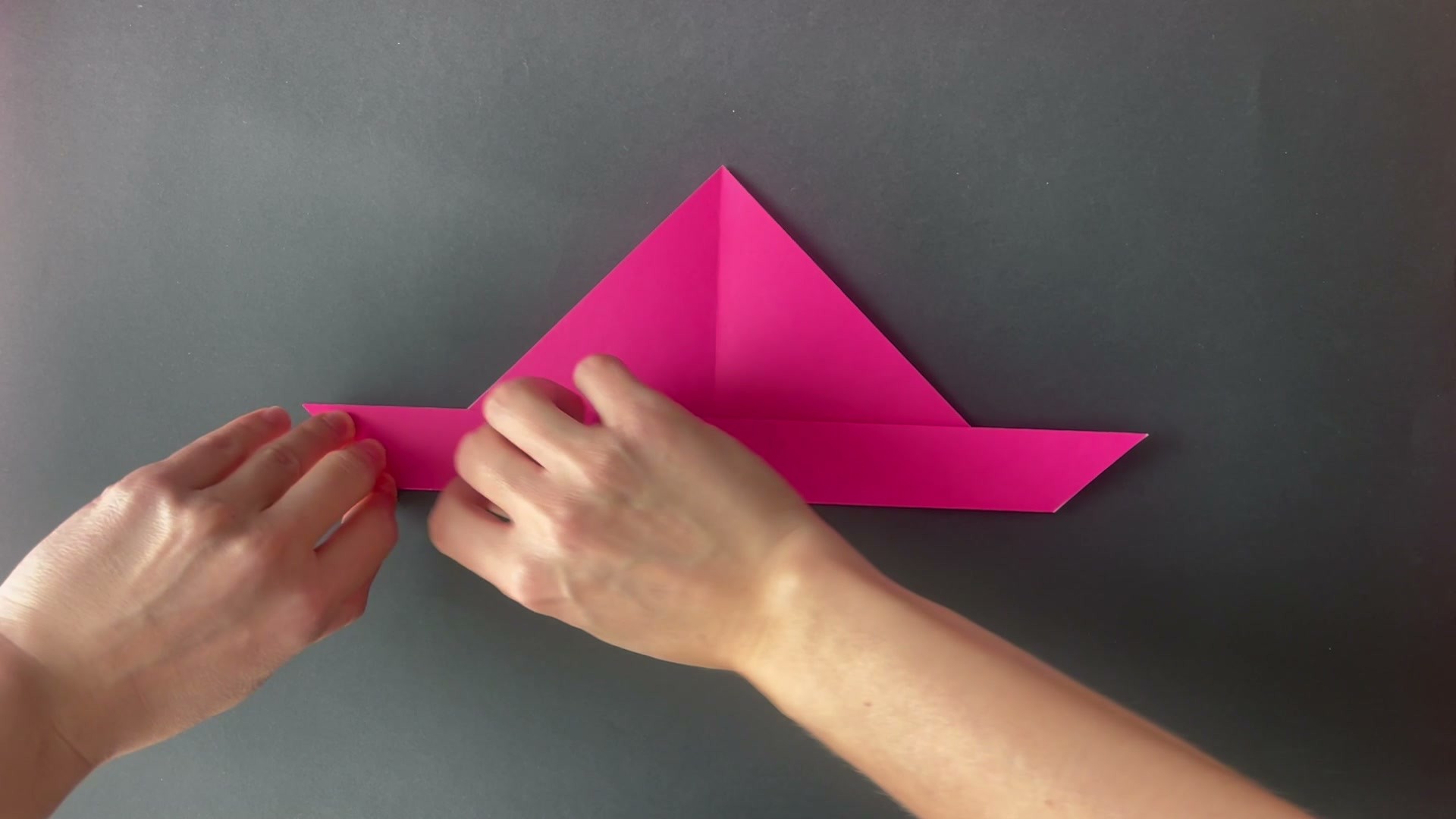}};
\node[img] (a7)  at (6\xgapfourteen, 0)  {\includegraphics[width=\imgwfourteen]{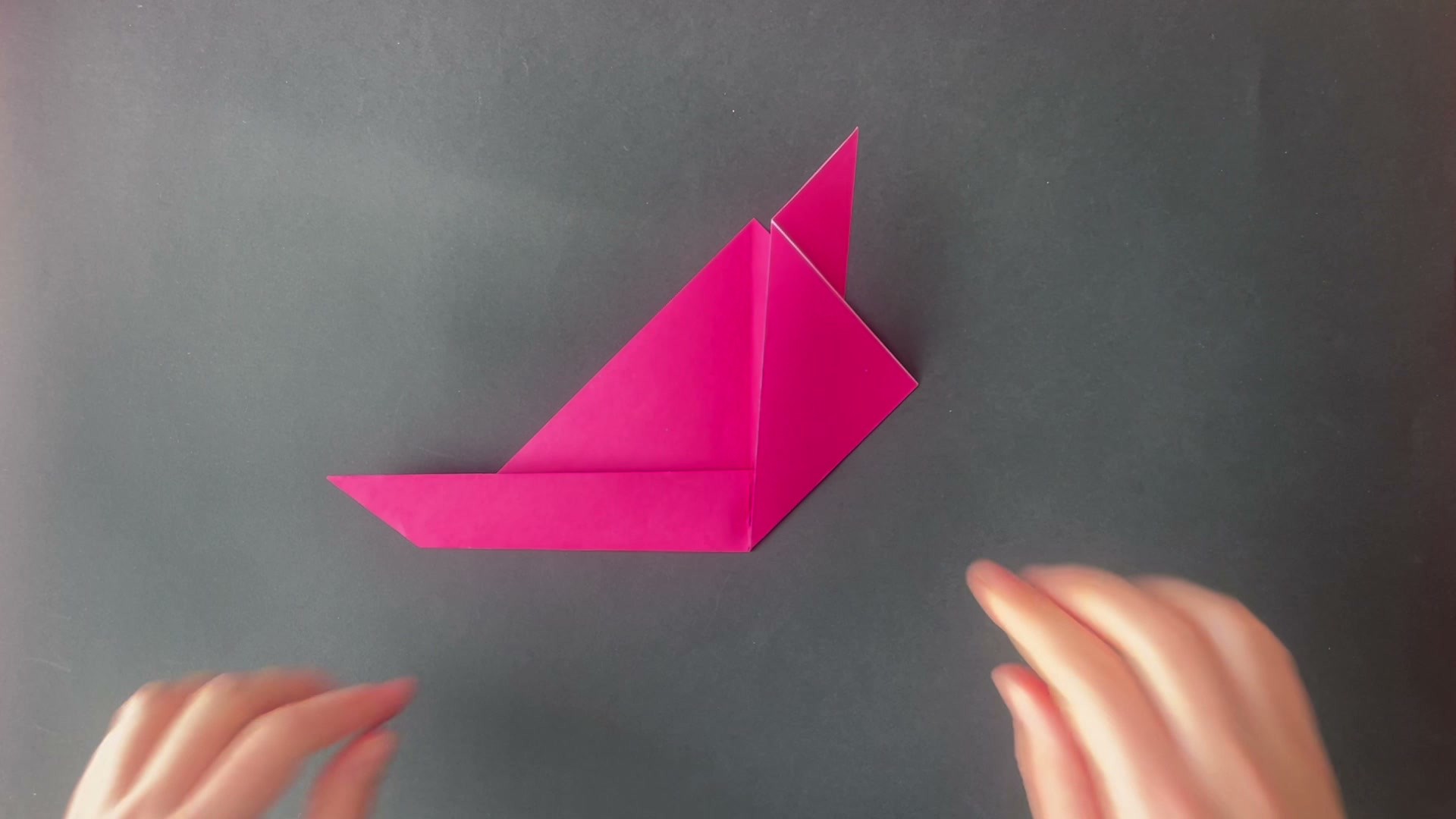}};
\node[img] (a8)  at (6\xgapfourteen, 0)  {};

\node[frameid] at ([xshift=1pt,yshift=-1pt]a1.north west) {1};
\node[frameid] at ([xshift=1pt,yshift=-1pt]a2.north west) {2};
\node[frameid] at ([xshift=1pt,yshift=-1pt]a3.north west) {3};
\node[frameid] at ([xshift=1pt,yshift=-1pt]a4.north west) {4};
\node[frameid] at ([xshift=1pt,yshift=-1pt]a5.north west) {5};
\node[frameid] at ([xshift=1pt,yshift=-1pt]a6.north west) {6};
\node[frameid] at ([xshift=1pt,yshift=-1pt]a7.north west) {7};

\node[img] (b1)  at (0\xgapfourteen, \ygapfourteen)  {\secondrowfigfourteen{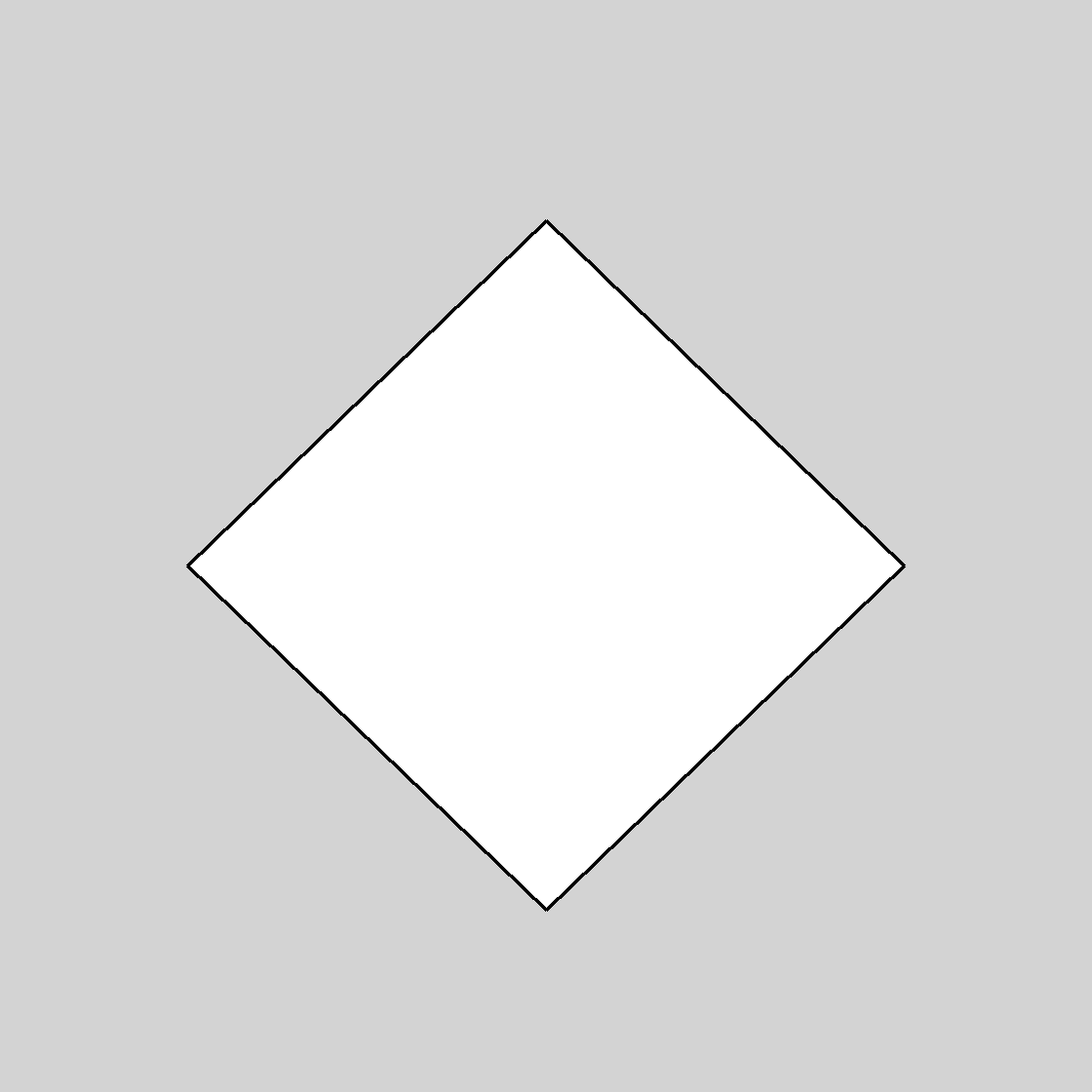}};
\node[img] (b2)  at (1\xgapfourteen, \ygapfourteen)  {\secondrowfigfourteen{pics/rabbit_output_result_diagram_0002.png}};
\node[img] (b3)  at (2\xgapfourteen, \ygapfourteen)  {\secondrowfigfourteen{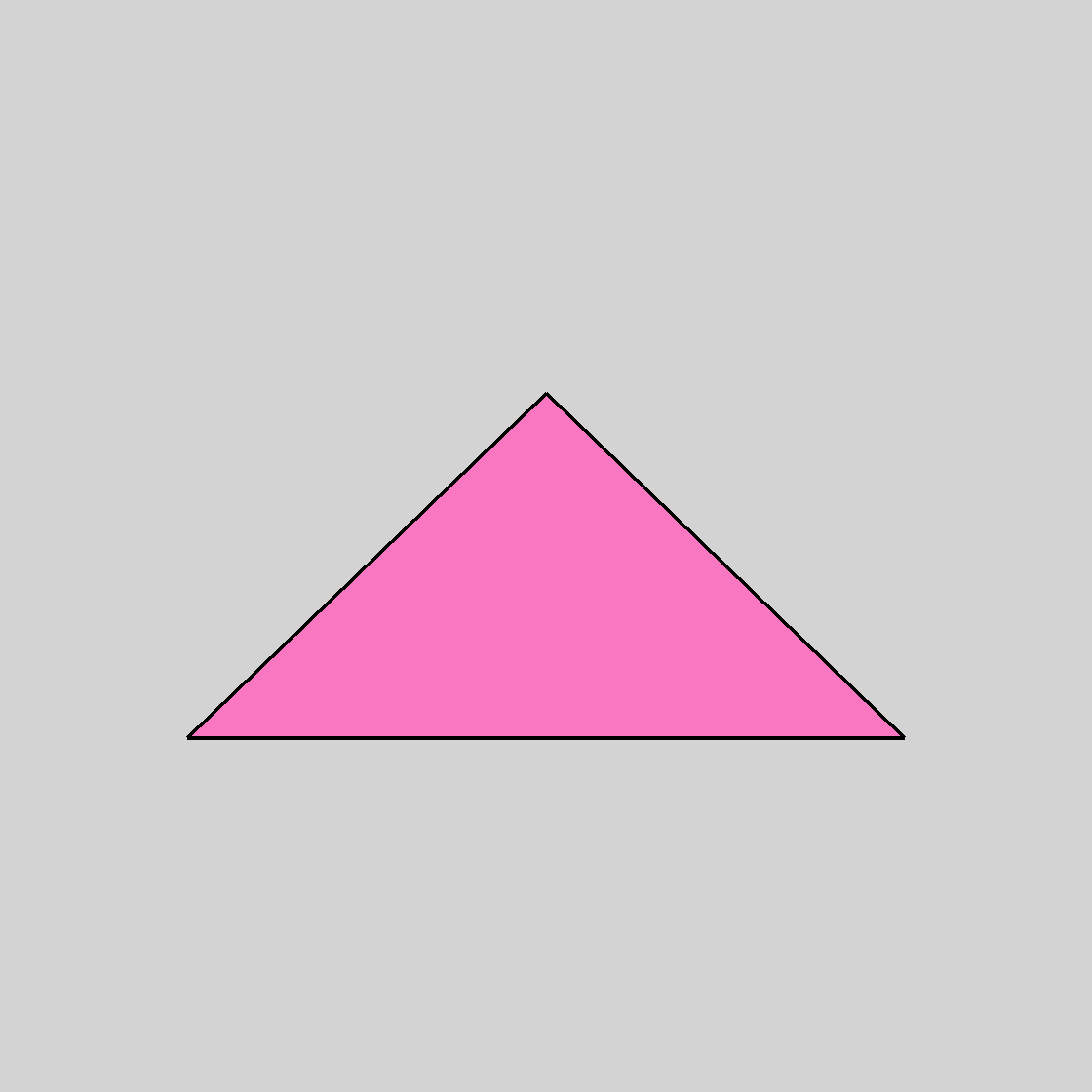}};
\node[img] (b4)  at (3\xgapfourteen, \ygapfourteen)  {\secondrowfigfourteen{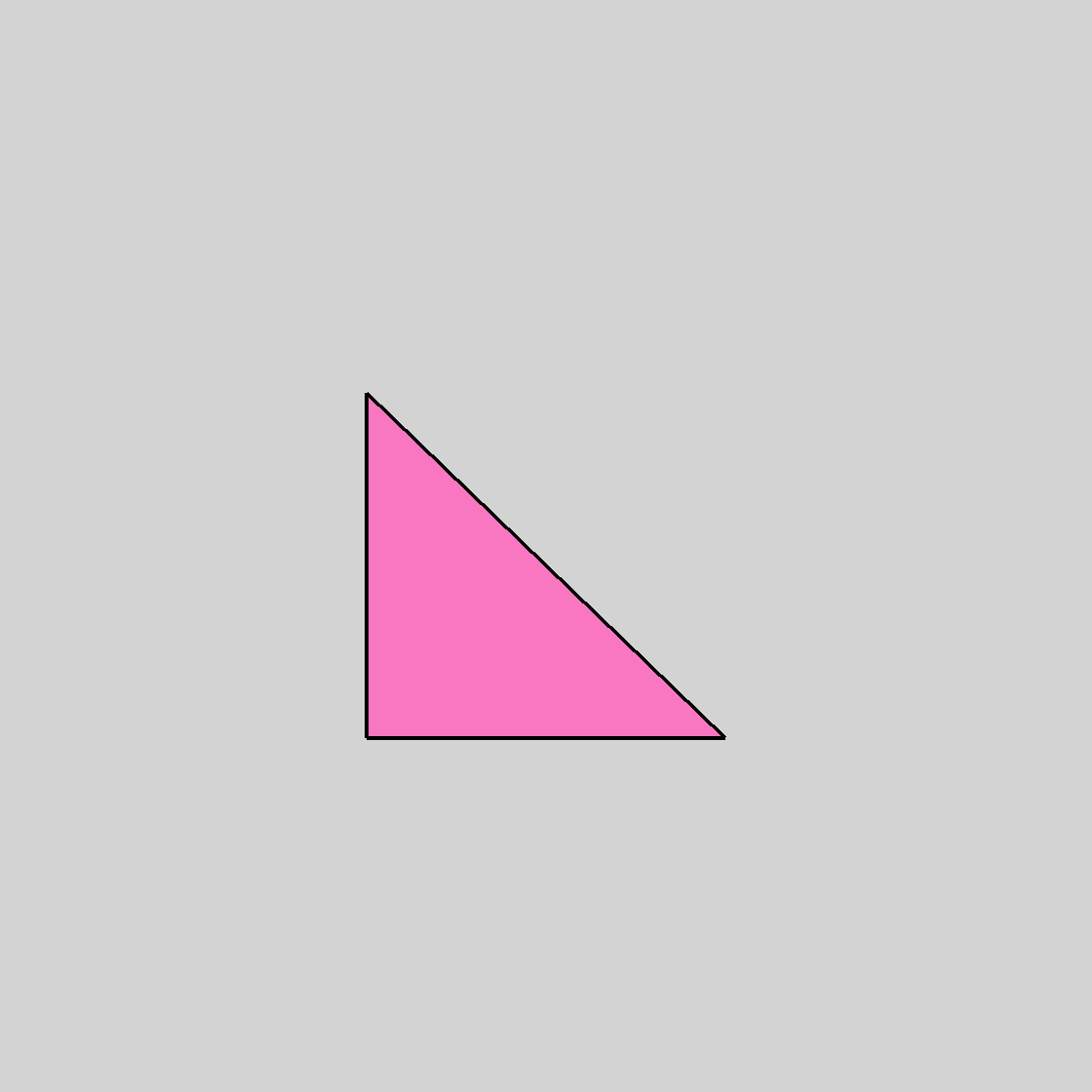}};
\node[img] (b5)  at (4\xgapfourteen, \ygapfourteen)  {\secondrowfigfourteen{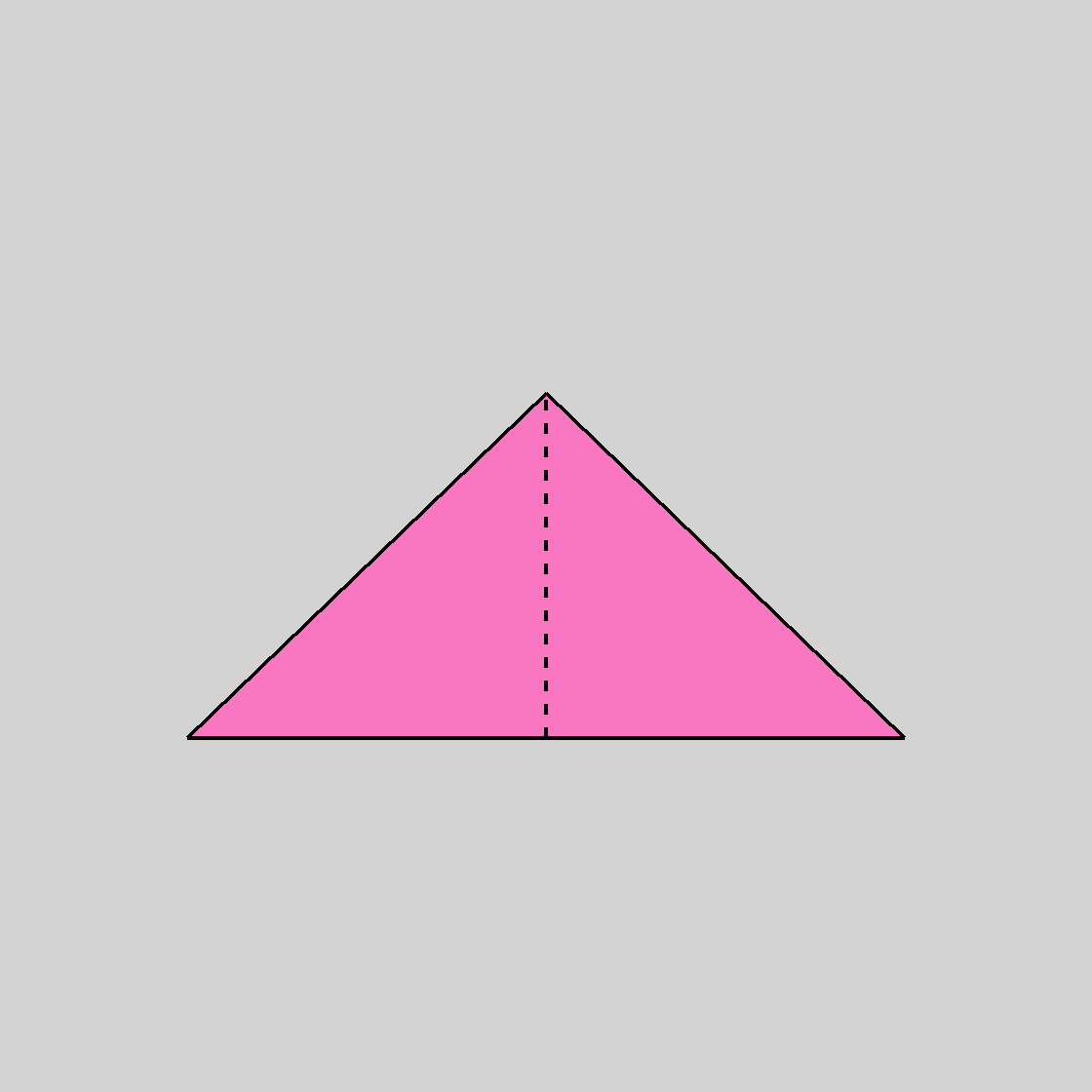}};
\node[img] (b6)  at (5\xgapfourteen, \ygapfourteen)  {\secondrowfigfourteen{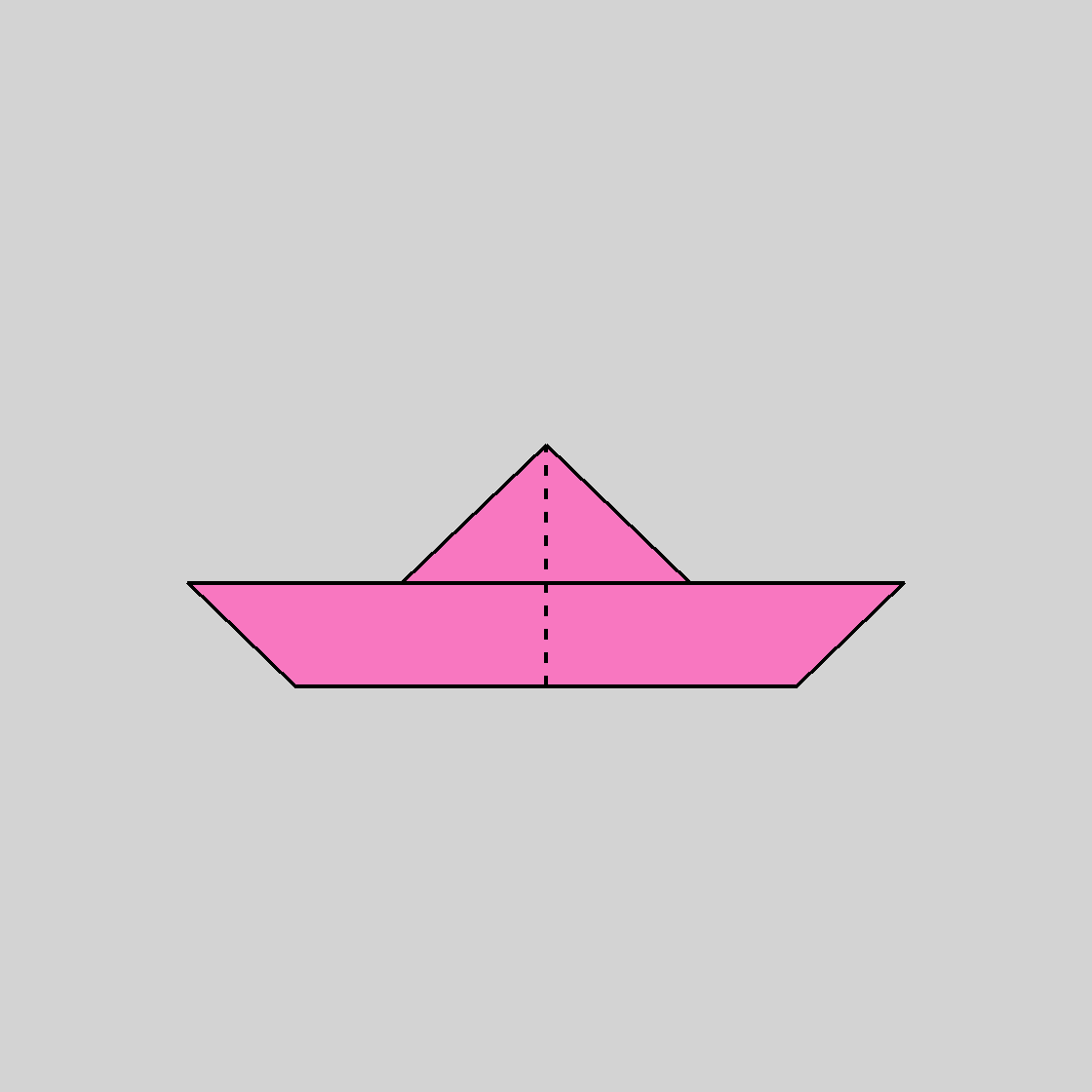}};
\node[img] (b7)  at (6\xgapfourteen, \ygapfourteen)  {\secondrowfigfourteen{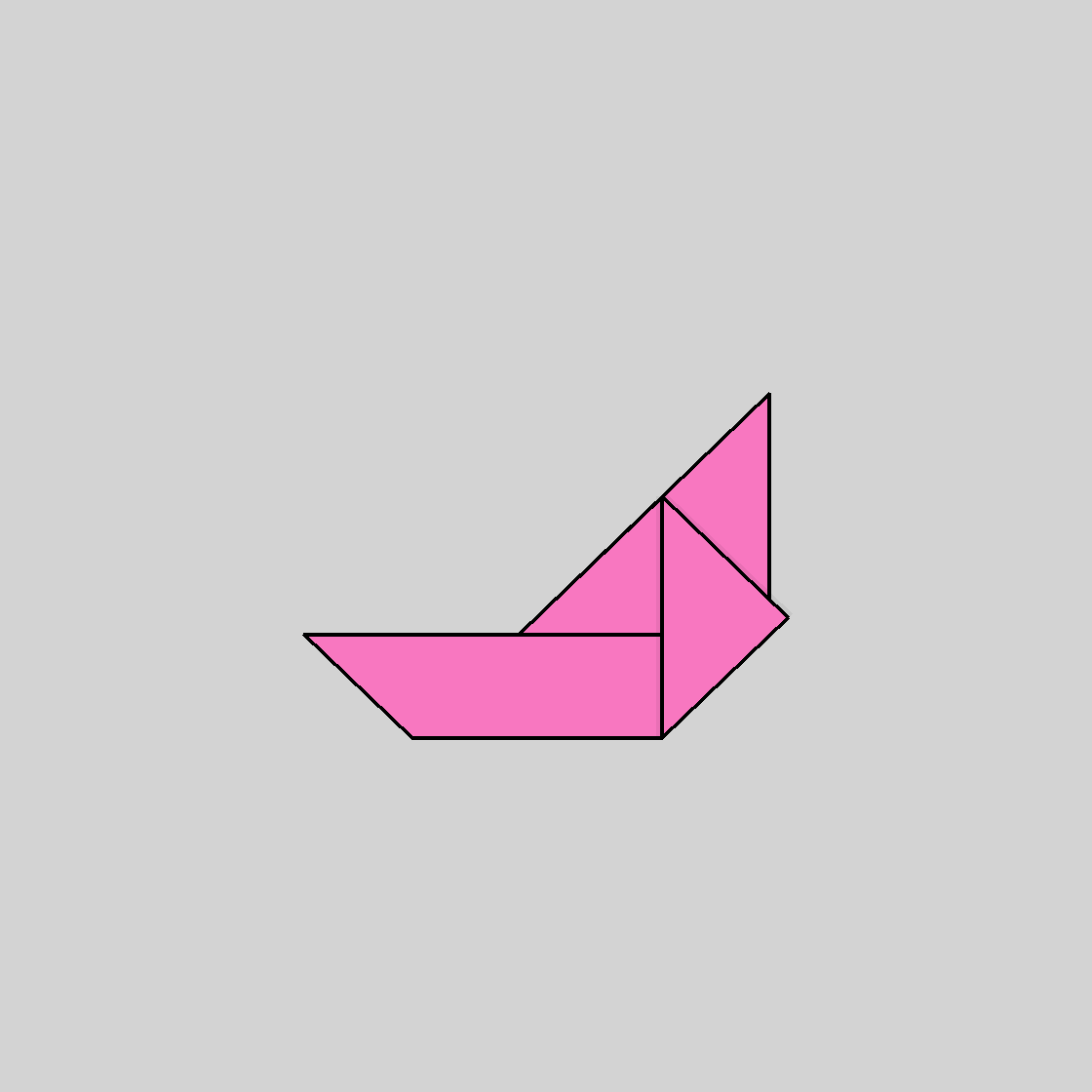}};
\node[img] (b8)  at (6\xgapfourteen, \ygapfourteen)  {};

\node[cmd] at ($(b1)!0.5!(b2) + (0,-0.77)$) {
rotate(45$^\circ$)
};

\node[cmd] at ($(b2)!0.5!(b3) + (0,-0.77)$) {
fold([4,2],-1)
};

\node[cmd] at ($(b3)!0.5!(b4) + (0,-0.67)$) {
add\_v([4,2],0.5)\\
fold([1,5],-1)
};

\node[cmd] at ($(b4)!0.5!(b5) + (0,-0.77)$) {
unfold()
};

\node[cmd] at ($(b5)!0.5!(b6) + (0,-0.6)$) {
add\_v([4,1],0.3)\\
add\_v([1,2],0.7) \\
fold([6,7],-1)
};

\node[cmd] at ($(b6)!0.5!(b7) + (0,-0.67)$) {
add\_v([1,7],0.5)\\
fold([8,12],-1)
};

\node[cmd] at ($(b7)!0.5!(b8) + (1,-0.77)$) {
fold([8,15],-1)
};

\end{tikzpicture}

\vspace{10pt}

\begin{tikzpicture}[
    img/.style={
        inner sep=0pt,
        outer sep=0pt
    },
    cmd/.style={
        fill=white,
        fill opacity=0.90,
        text opacity=1,
        draw=black!30,
        rounded corners=1.2pt,
        inner sep=1pt,
        font=\tiny\ttfamily,
        align=left
    },
    frameid/.style={
        fill=white,
        fill opacity=0,
        text opacity=1,
        text=white,
        draw=black!25,
        rounded corners=0pt,
        inner sep=1pt,
        font=\tiny\ttfamily,
        anchor=north west
    }
]

\node[img] (a1)  at (0\xgapfourteen, 0)  {\includegraphics[width=\imgwfourteen]{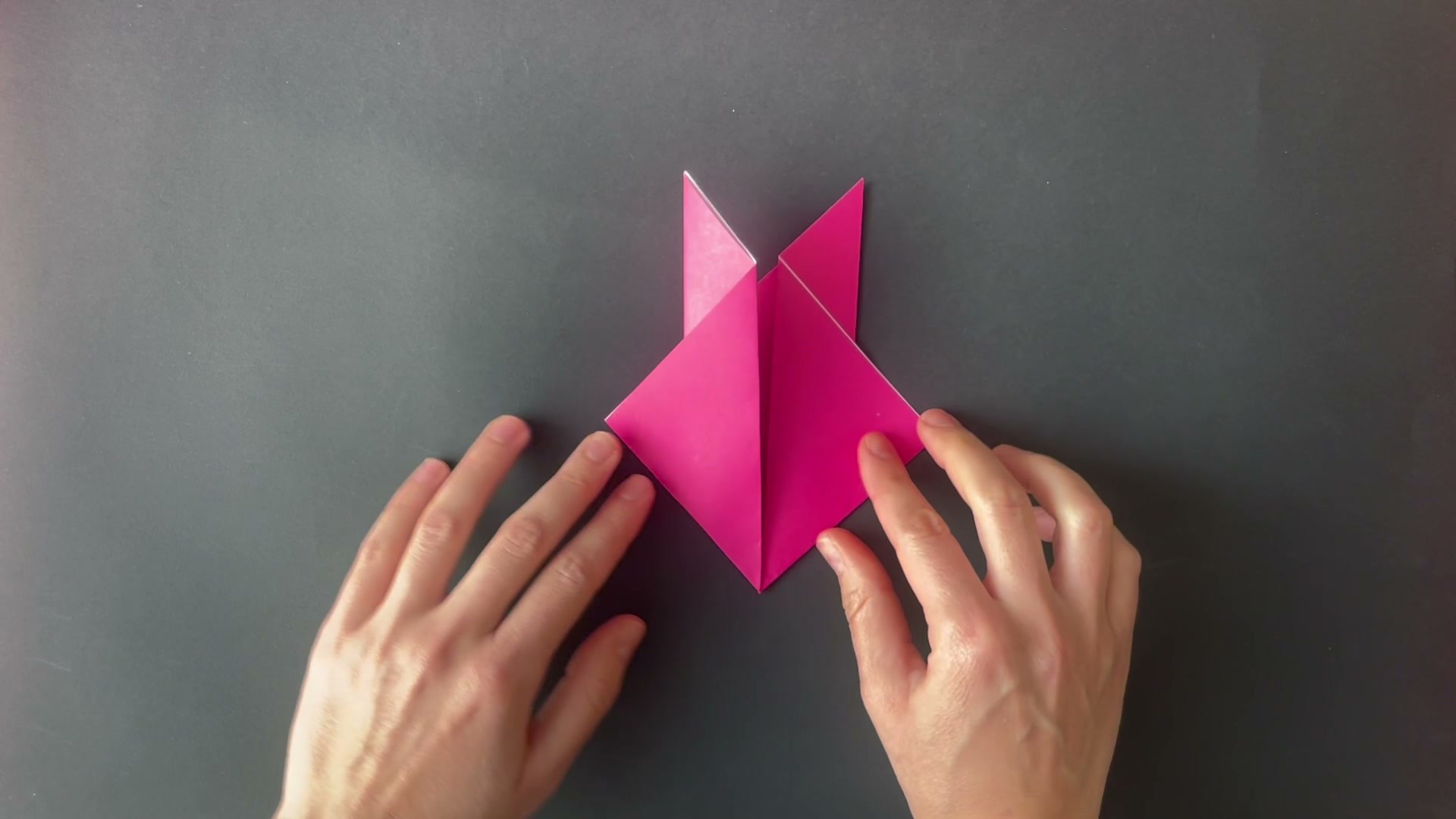}};
\node[img] (a2)  at (1\xgapfourteen, 0)  {\includegraphics[width=\imgwfourteen]{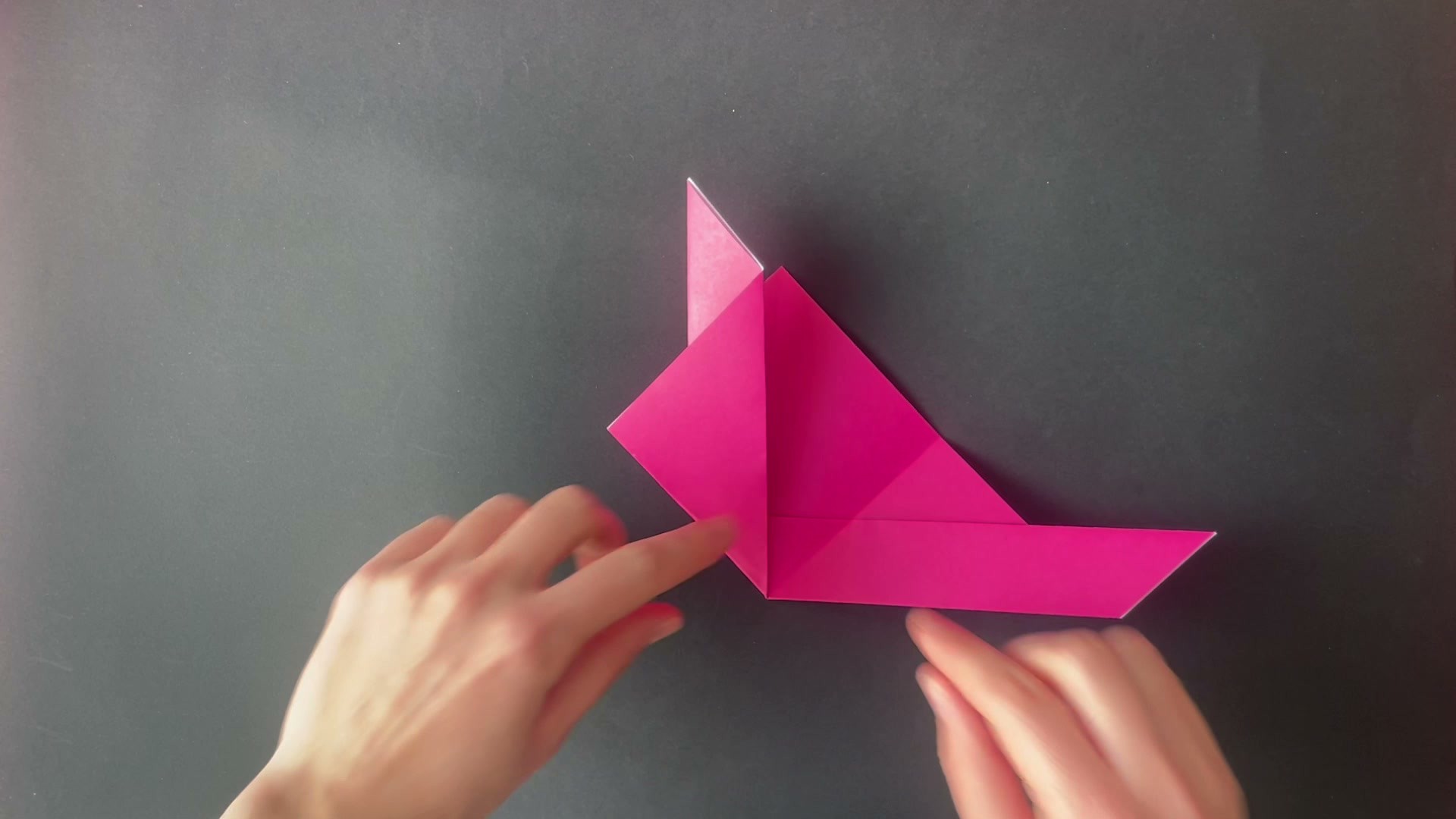}};
\node[img] (a3)  at (2\xgapfourteen, 0)  {\includegraphics[width=\imgwfourteen]{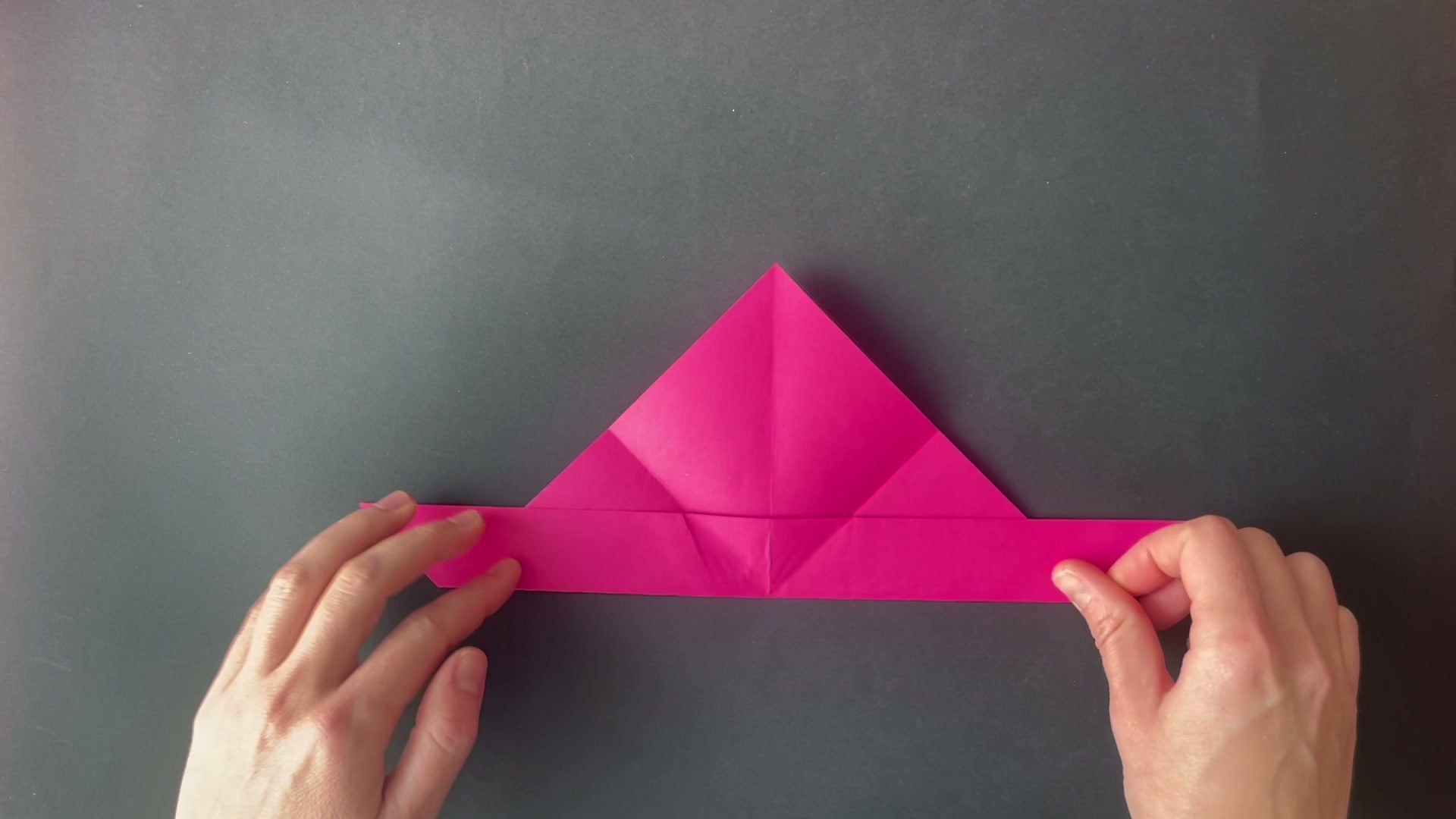}};
\node[img] (a4)  at (3\xgapfourteen, 0)  {\includegraphics[width=\imgwfourteen]{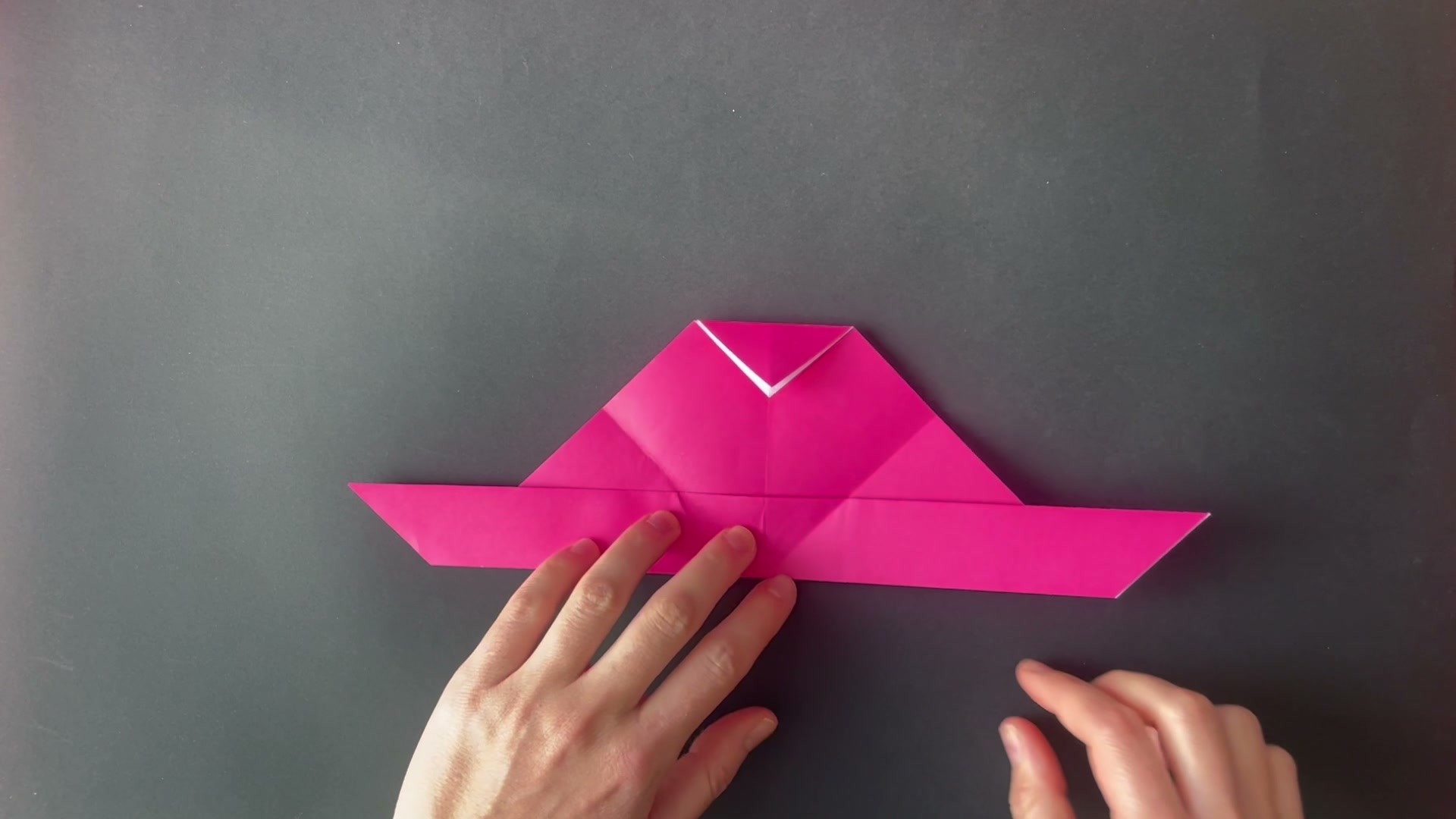}};
\node[img] (a5)  at (4\xgapfourteen, 0)  {\includegraphics[width=\imgwfourteen]{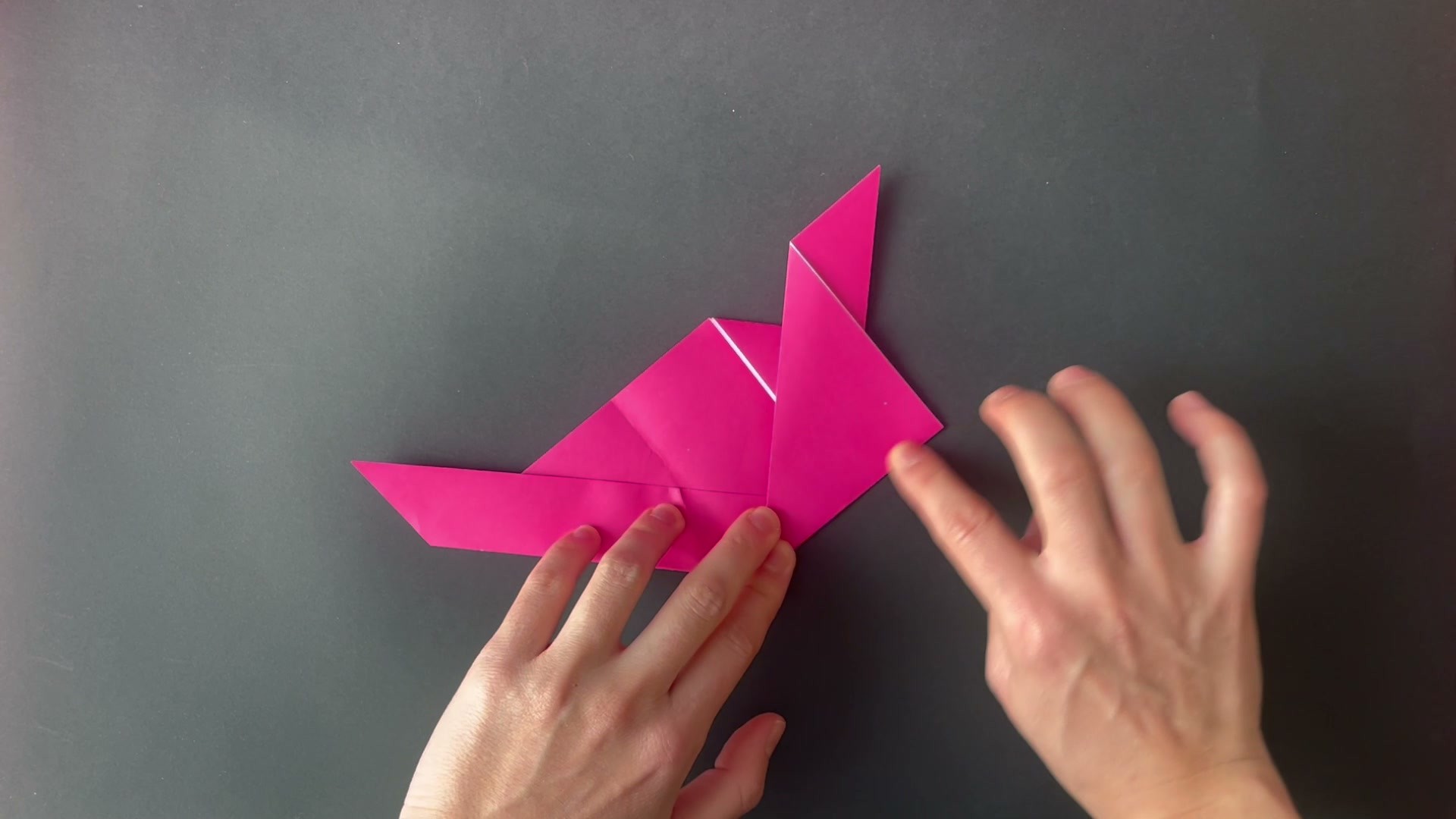}};
\node[img] (a6)  at (5\xgapfourteen, 0)  {\includegraphics[width=\imgwfourteen]{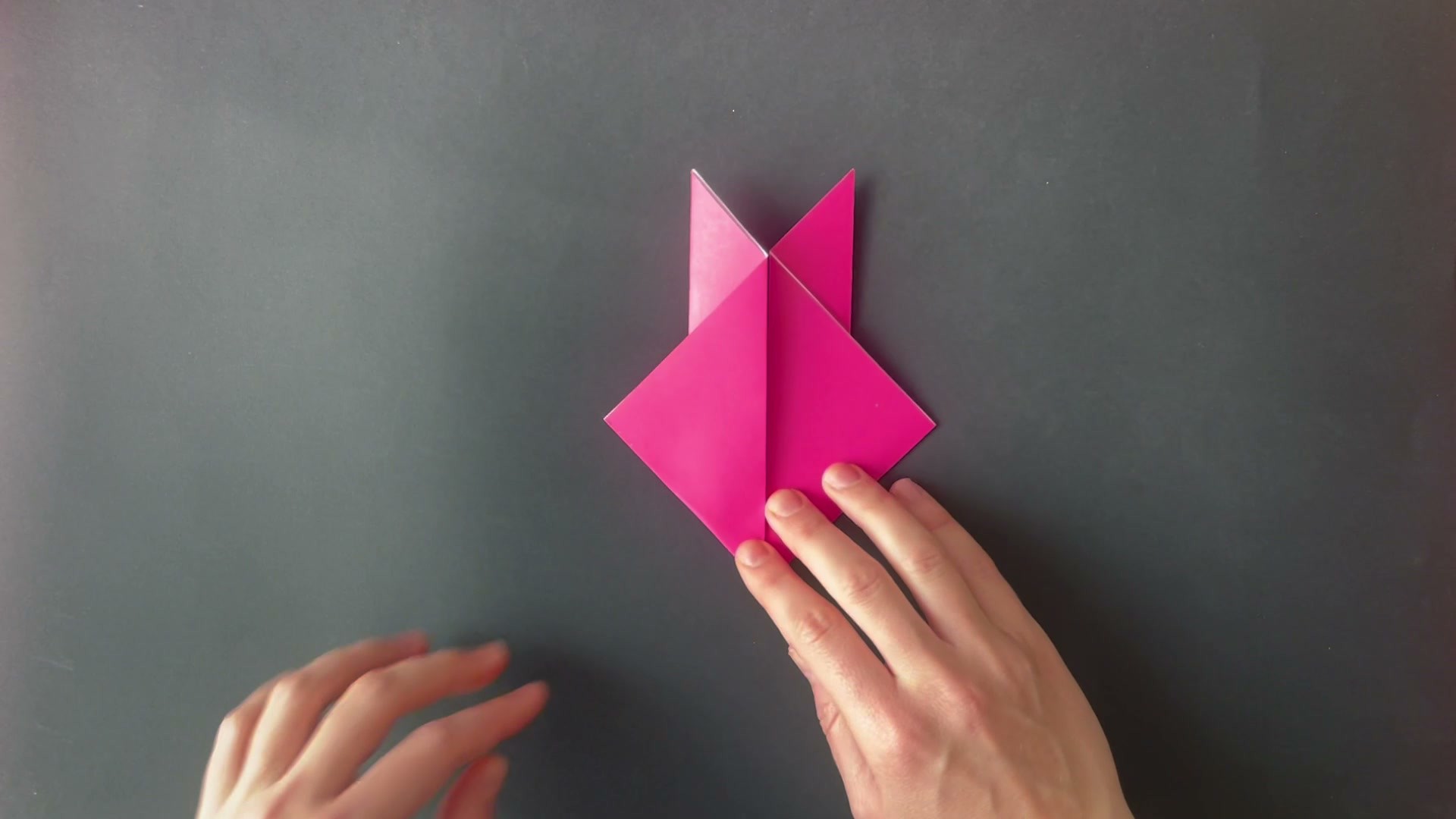}};
\node[img] (a7)  at (6\xgapfourteen, 0)  {\includegraphics[width=\imgwfourteen]{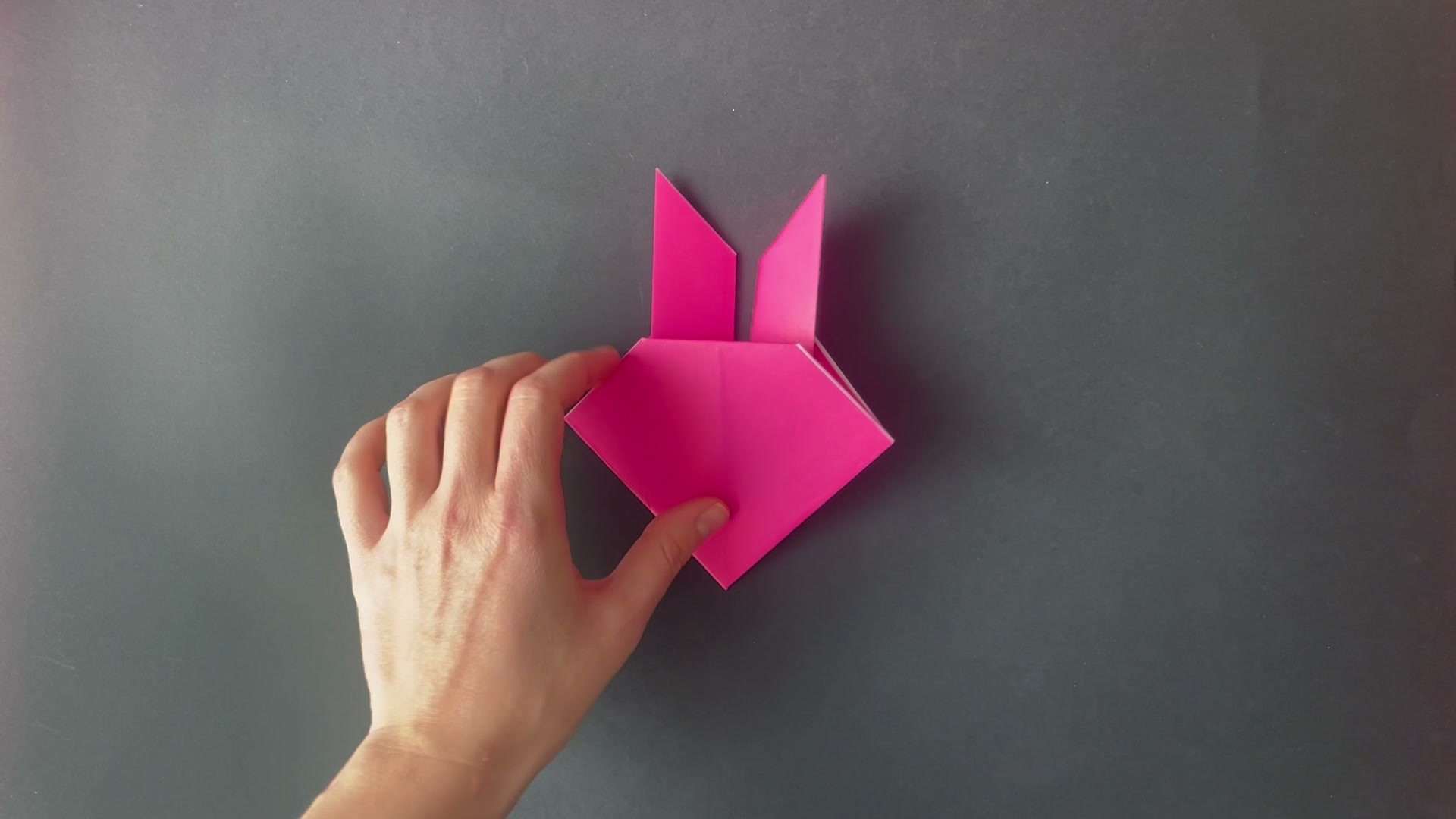}};

\node[frameid] at ([xshift=1pt,yshift=-1pt]a1.north west) {8};
\node[frameid] at ([xshift=1pt,yshift=-1pt]a2.north west) {9};
\node[frameid] at ([xshift=1pt,yshift=-1pt]a3.north west) {10};
\node[frameid] at ([xshift=1pt,yshift=-1pt]a4.north west) {11};
\node[frameid] at ([xshift=1pt,yshift=-1pt]a5.north west) {12};
\node[frameid] at ([xshift=1pt,yshift=-1pt]a6.north west) {13};
\node[frameid] at ([xshift=1pt,yshift=-1pt]a7.north west) {14};

\node[img] (b1)  at (0\xgapfourteen, \ygapfourteen)  {\secondrowfigfourteen{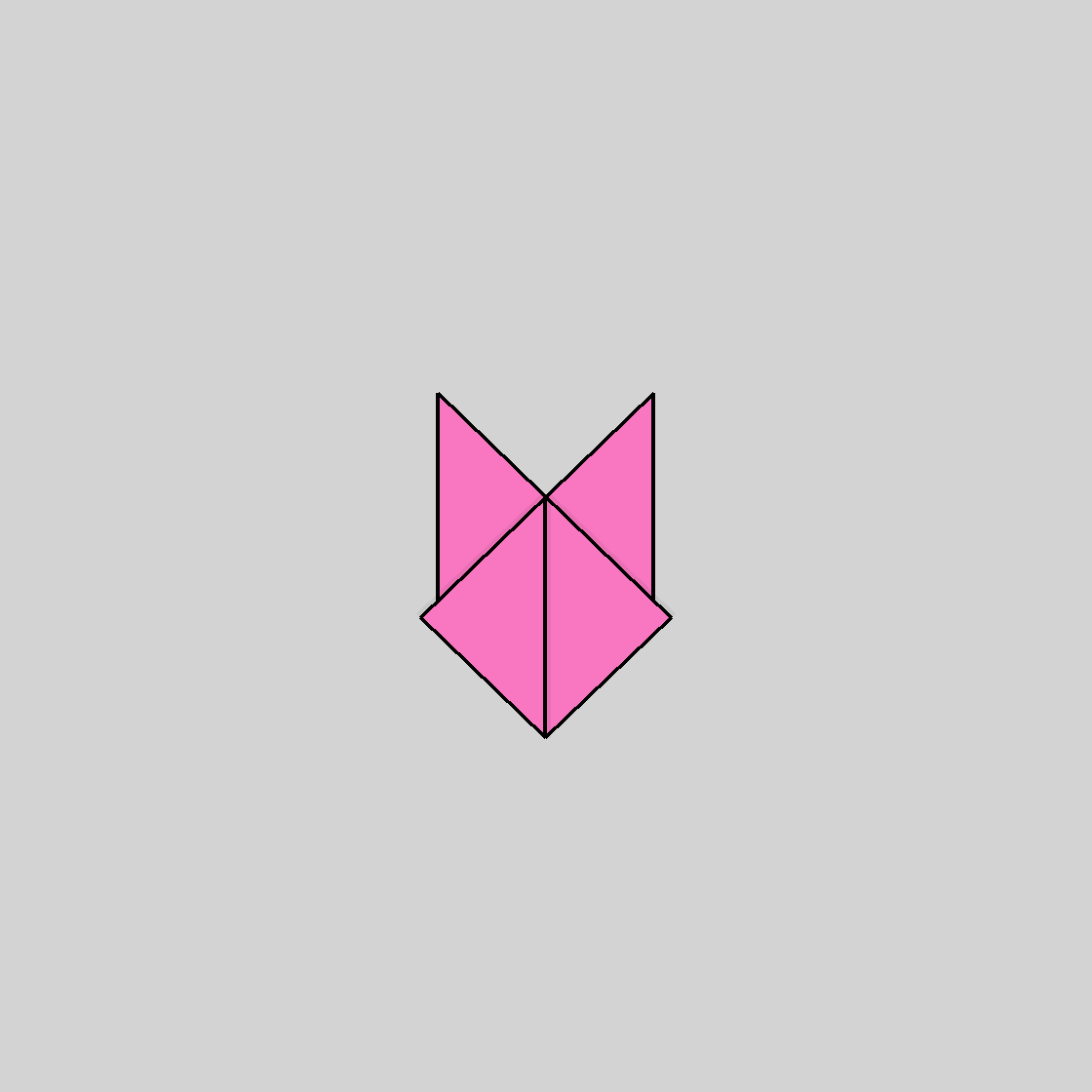}};
\node[img] (b2)  at (1\xgapfourteen, \ygapfourteen)  {\secondrowfigfourteen{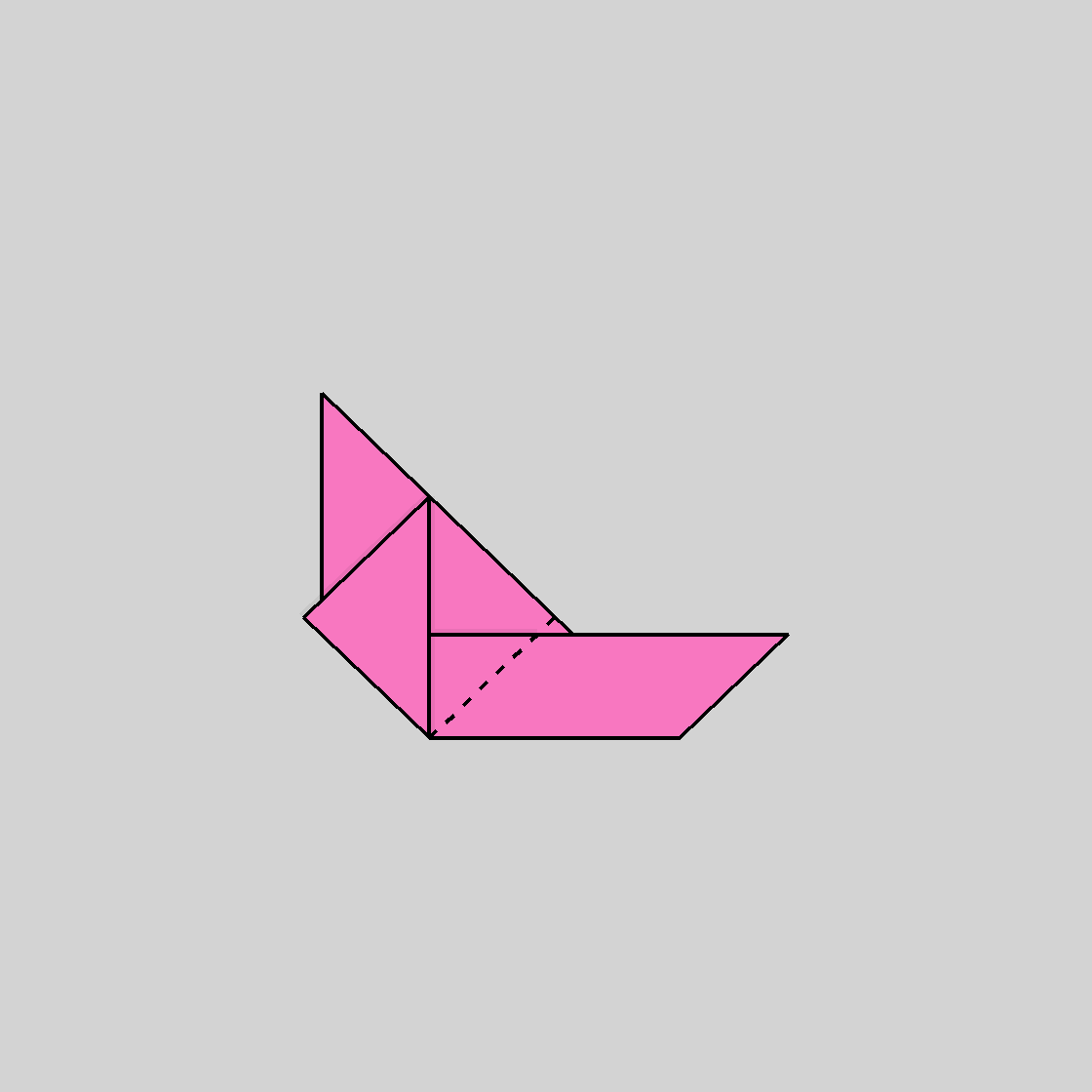}};
\node[img] (b3)  at (2\xgapfourteen, \ygapfourteen)  {\secondrowfigfourteen{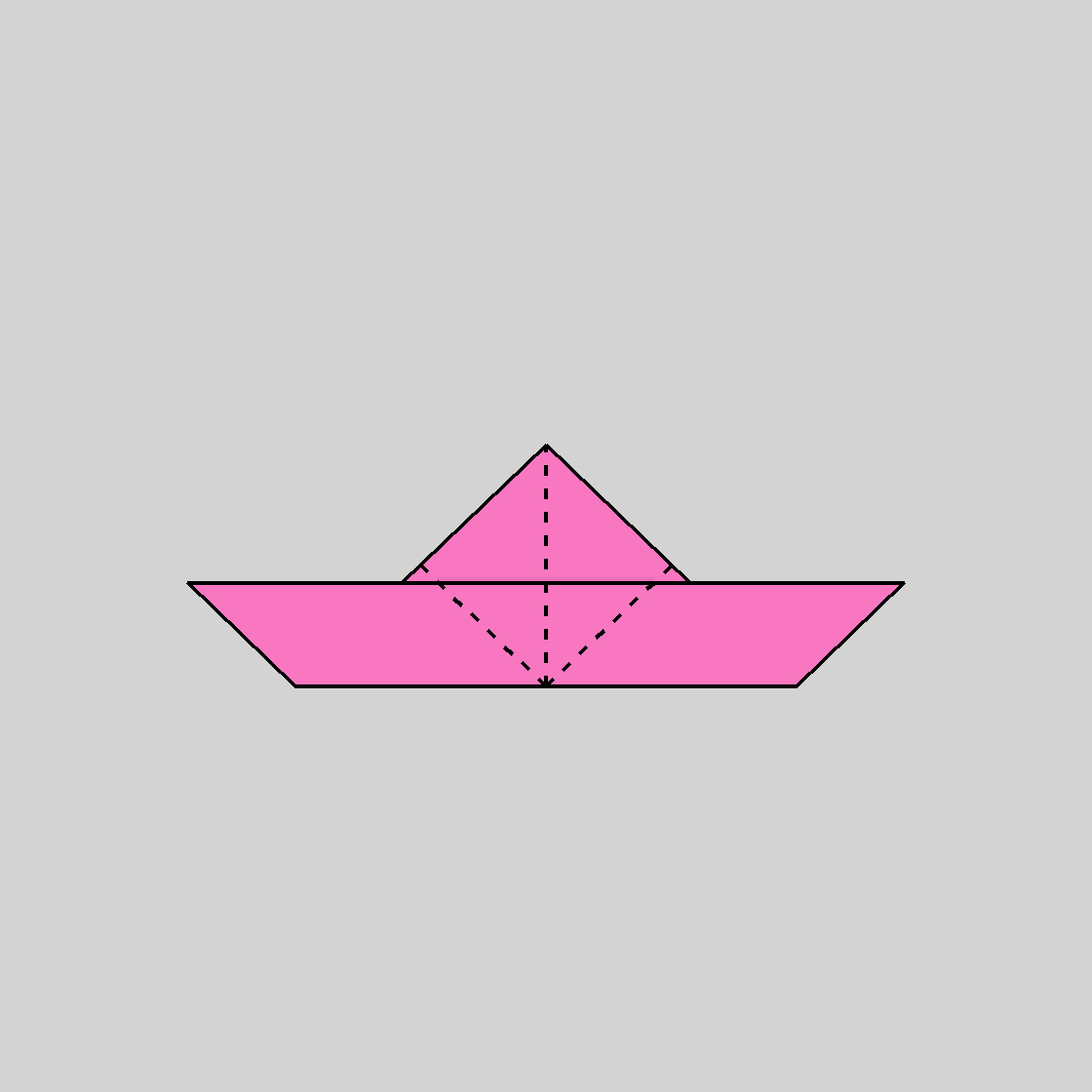}};
\node[img] (b4)  at (3\xgapfourteen, \ygapfourteen)  {\secondrowfigfourteen{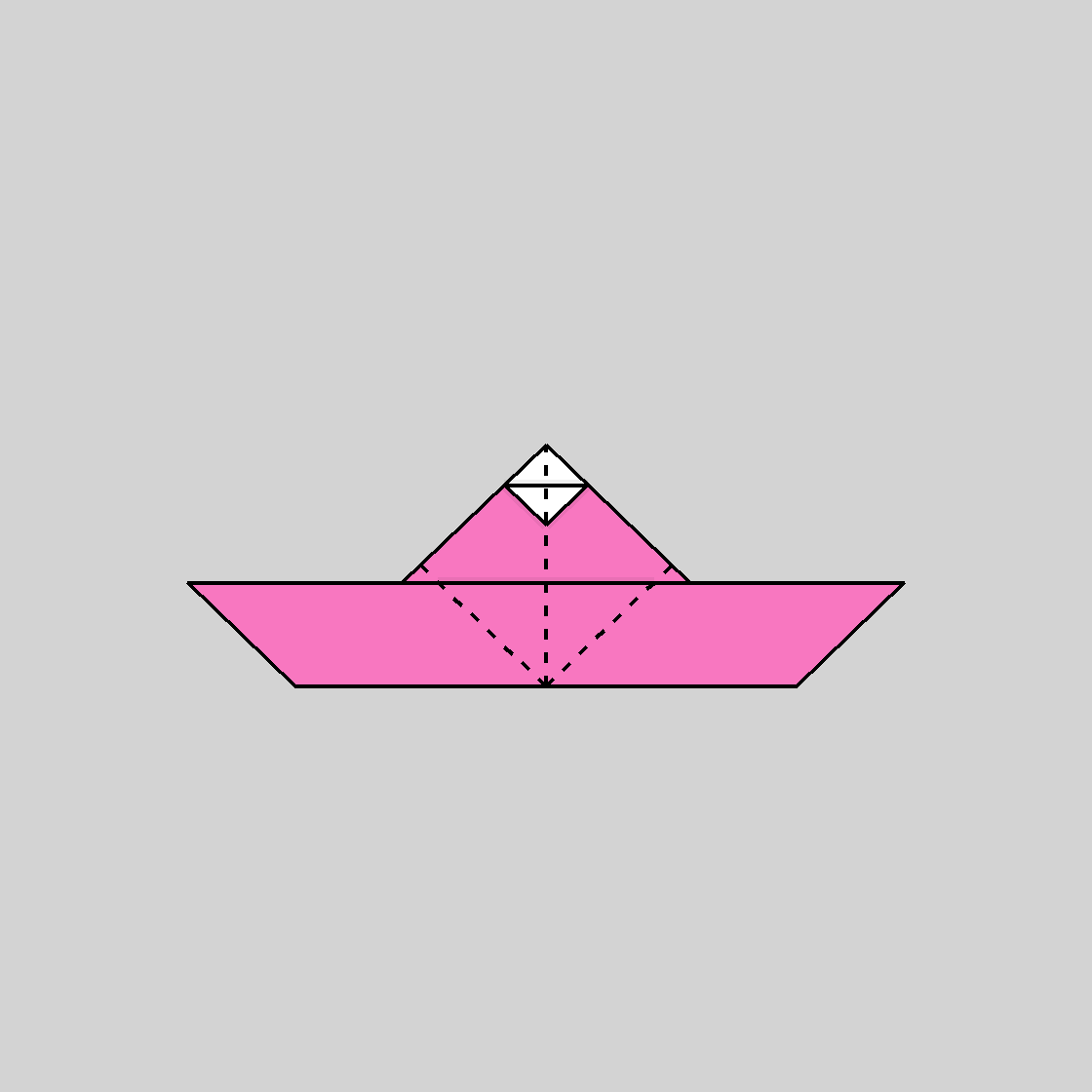}};
\node[img] (b5)  at (4\xgapfourteen, \ygapfourteen)  {\secondrowfigfourteen{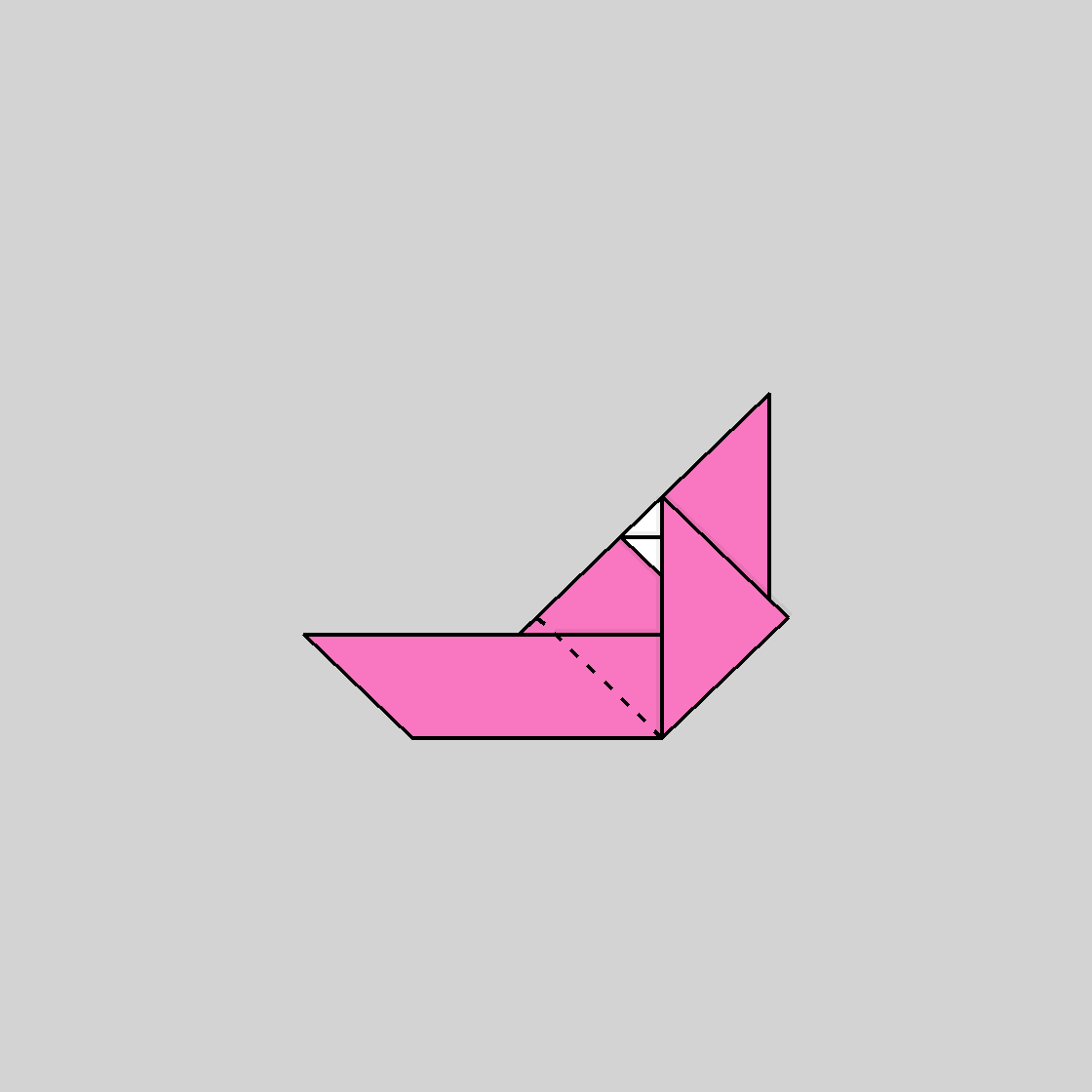}};
\node[img] (b6)  at (5\xgapfourteen, \ygapfourteen)  {\secondrowfigfourteen{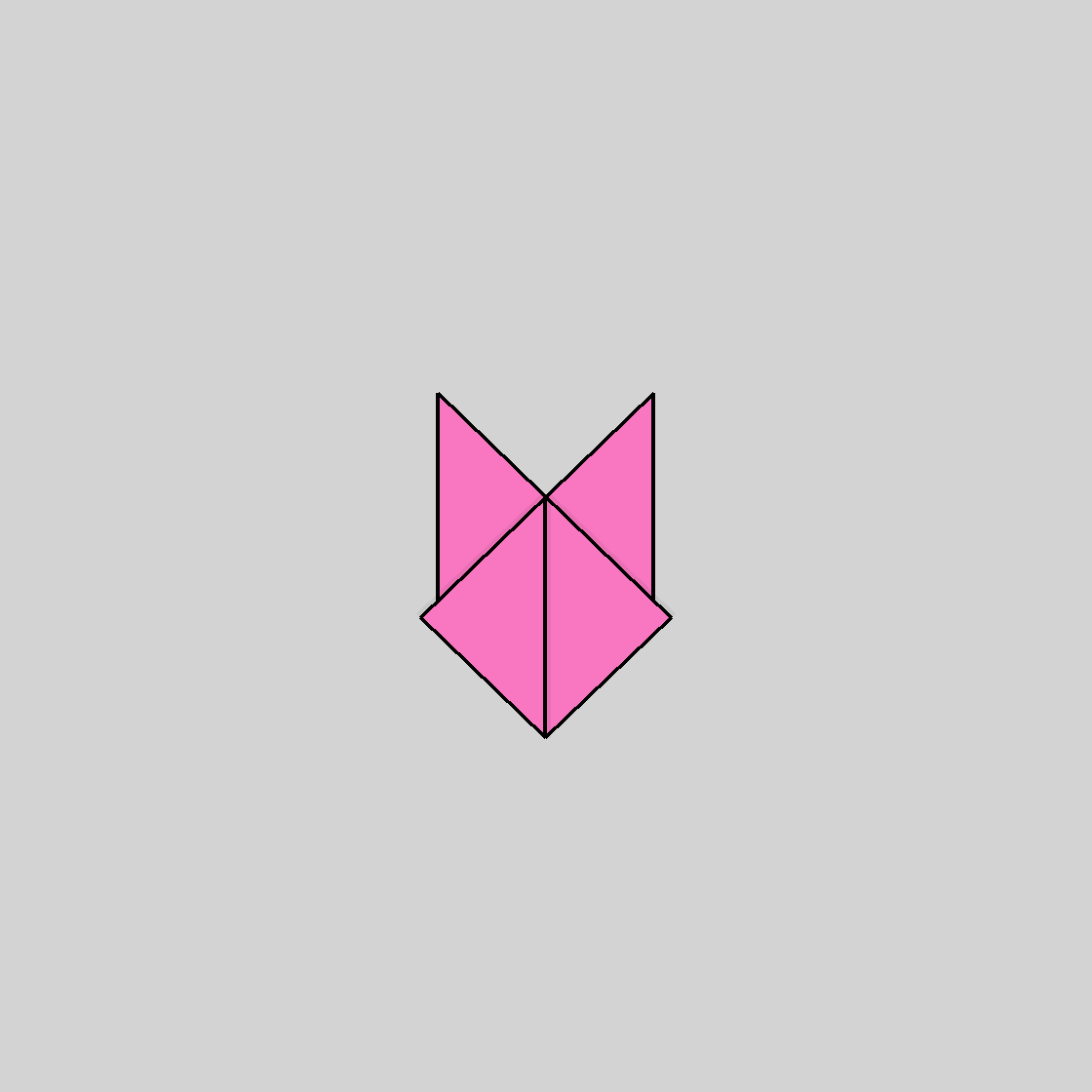}};
\node[img] (b7)  at (6\xgapfourteen, \ygapfourteen)  {\secondrowfigfourteen{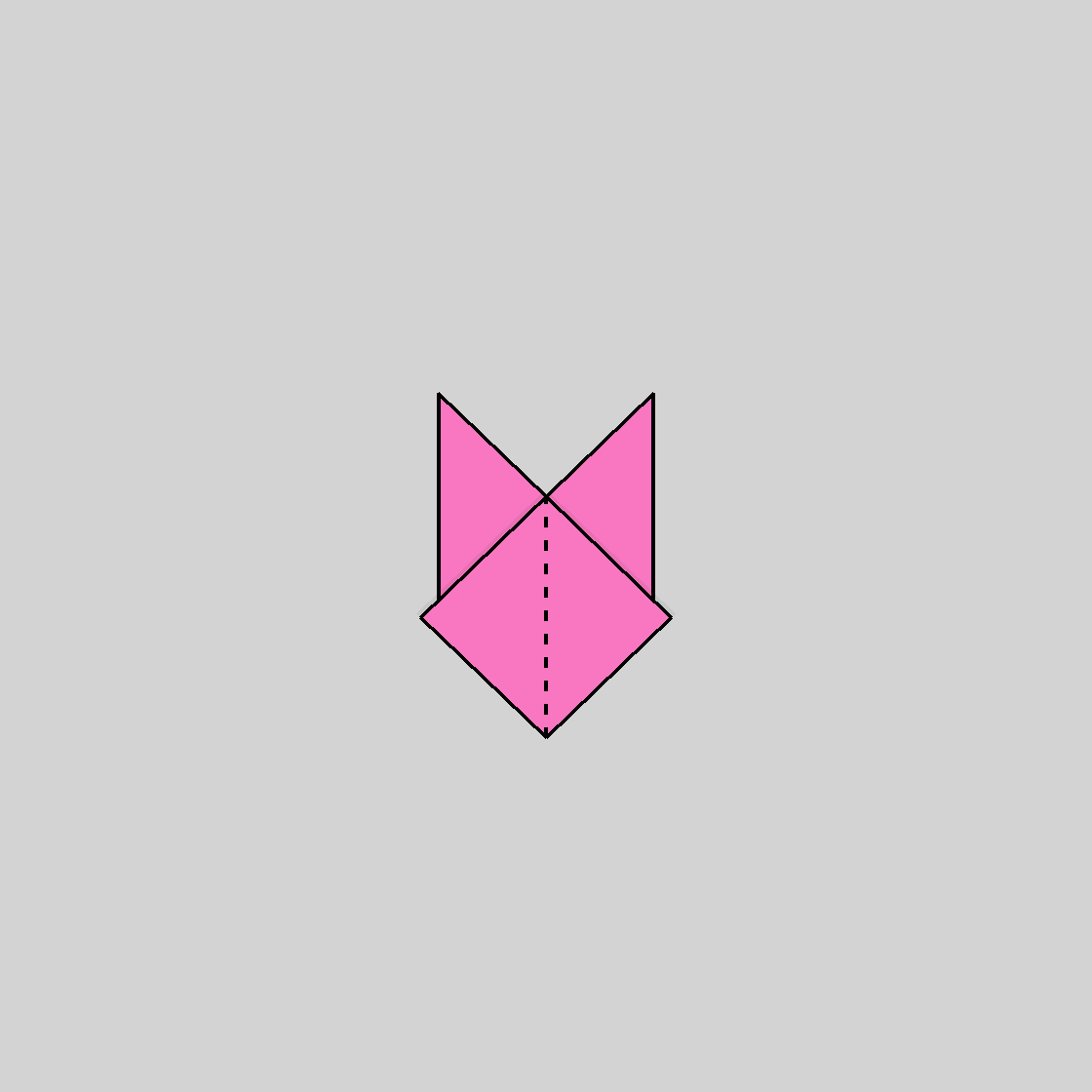}};

\node[cmd] at ($(b1)!0.5!(b2) + (0,-0.77)$) {
unfold([8,12])
};

\node[cmd] at ($(b2)!0.5!(b3) + (0,-0.77)$) {
unfold([8,15])
};

\node[cmd] at ($(b3)!0.5!(b4) + (0,-0.77)$) {
add\_v([3,14],0.33)\\
add\_v([3,16],0.33)\\
fold([19,18],1)
};

\node[cmd] at ($(b4)!0.5!(b5) + (0,-0.77)$) {
fold([8,12],-1)
};

\node[cmd] at ($(b5)!0.5!(b6) + (0,-0.77)$) {
fold([8,15],-1)
};

\node[cmd] at ($(b6)!0.5!(b7) + (0,-0.77)$) {
flip(x)
};

\end{tikzpicture}

\Description{Long-sequence qualitative result.}
\caption{\emph{Long-sequence qualitative result.} For each example, we show the input keyframes (top), rendering of folded states reconstructed by our method (bottom), and the inferred actions between consecutive keyframes. These examples illustrate our method's ability to recover a long folding procedure involving rotations, folds, unfolds, flips, and precise vertex definitions. See \supp{sec:full_results} for the full set of results.}
\label{fig:fourteen_images2}
\end{figure*}
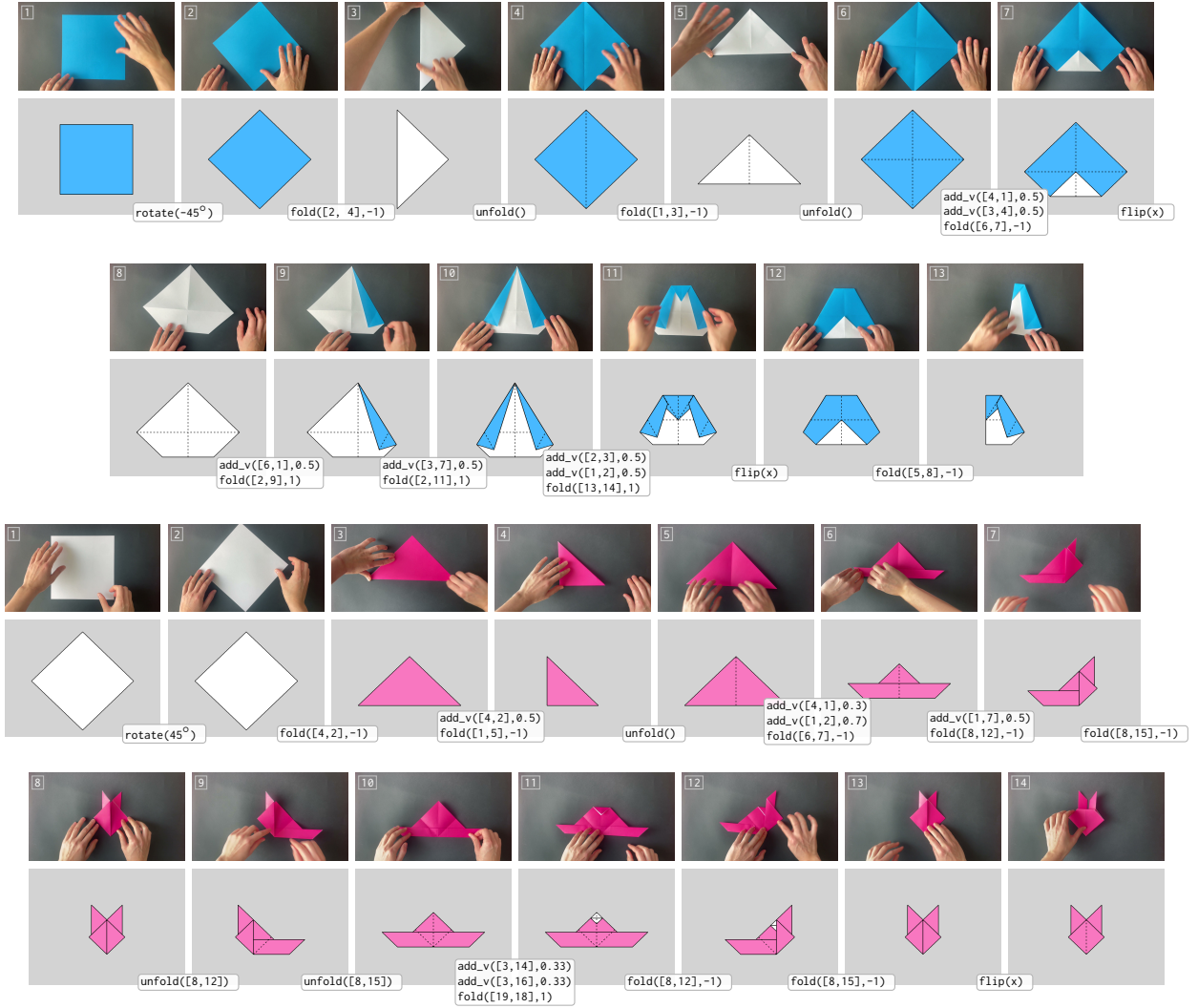

\fi
\begin{figure*}[t]
\centering

\newlength{\imgw}
\setlength{\imgw}{0.12\textwidth}

\newlength{\xgap}
\setlength{\xgap}{0.126\textwidth}

\newlength{\ygap}
\setlength{\ygap}{-0.085\textwidth}

\newcommand{\secondrowfig}[1]{%
  \includegraphics[
    width=\imgw,
    trim={0pt 5cm 0pt 5cm},
    clip
  ]{#1}%
}

\begin{tikzpicture}[
    img/.style={
        inner sep=0pt,
        outer sep=0pt
    },
    cmd/.style={
        fill=white,
        fill opacity=0.90,
        text opacity=1,
        draw=black!30,
        rounded corners=1.5pt,
        inner sep=2pt,
        font=\tiny\ttfamily,
        align=left
    },
    frameid/.style={
        fill=white,
        fill opacity=0,
        text opacity=1,
        text=white,
        draw=black!25,
        rounded corners=0pt,
        inner sep=1pt,
        font=\tiny\ttfamily,
        anchor=north west
    }
]

\node[img] (a1) at (0\xgap, 0)
{\includegraphics[width=\imgw]{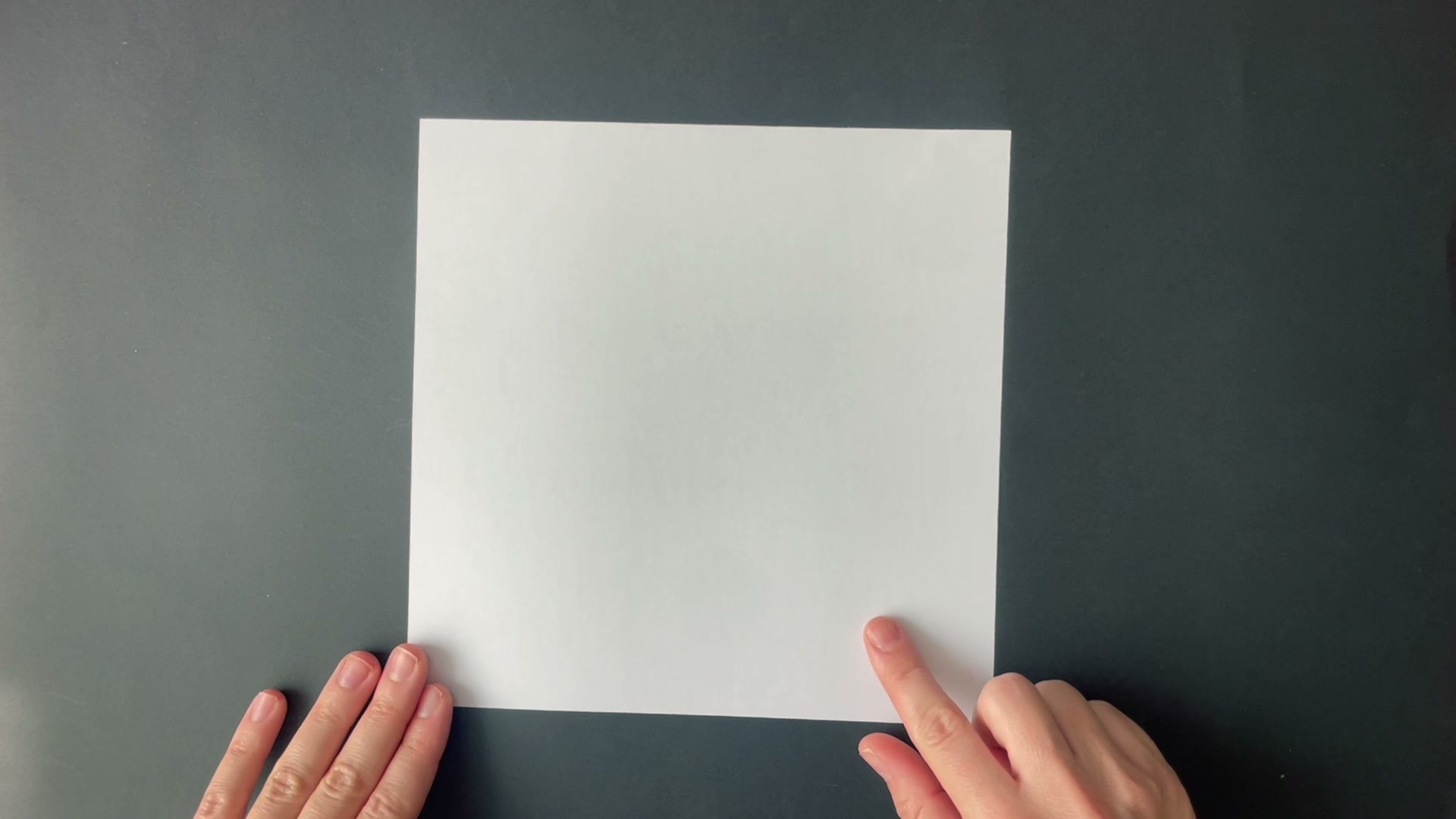}};

\node[img] (a2) at (1\xgap, 0)
{\includegraphics[width=\imgw]{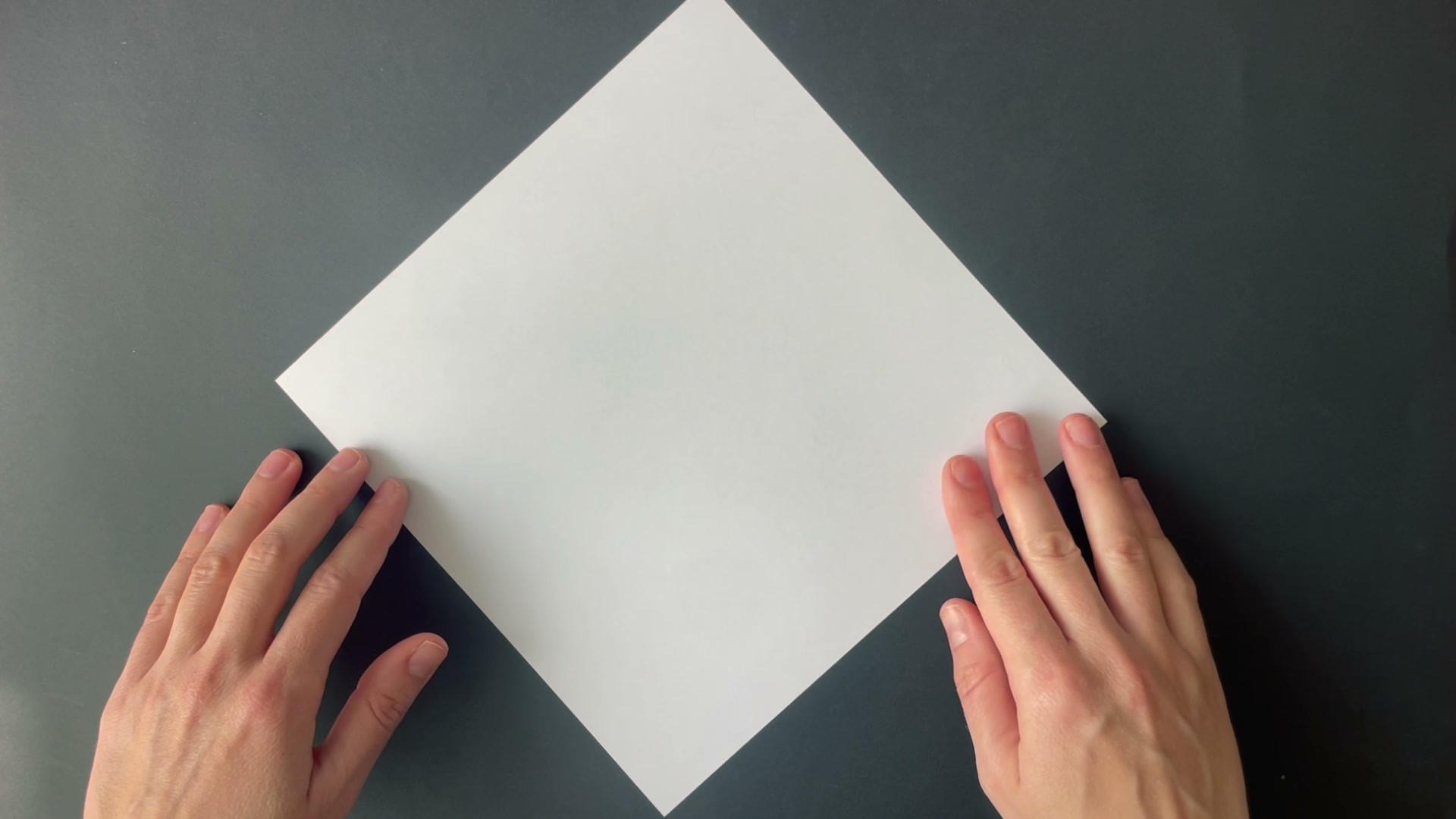}};

\node[img] (a3) at (2\xgap, 0)
{\includegraphics[width=\imgw]{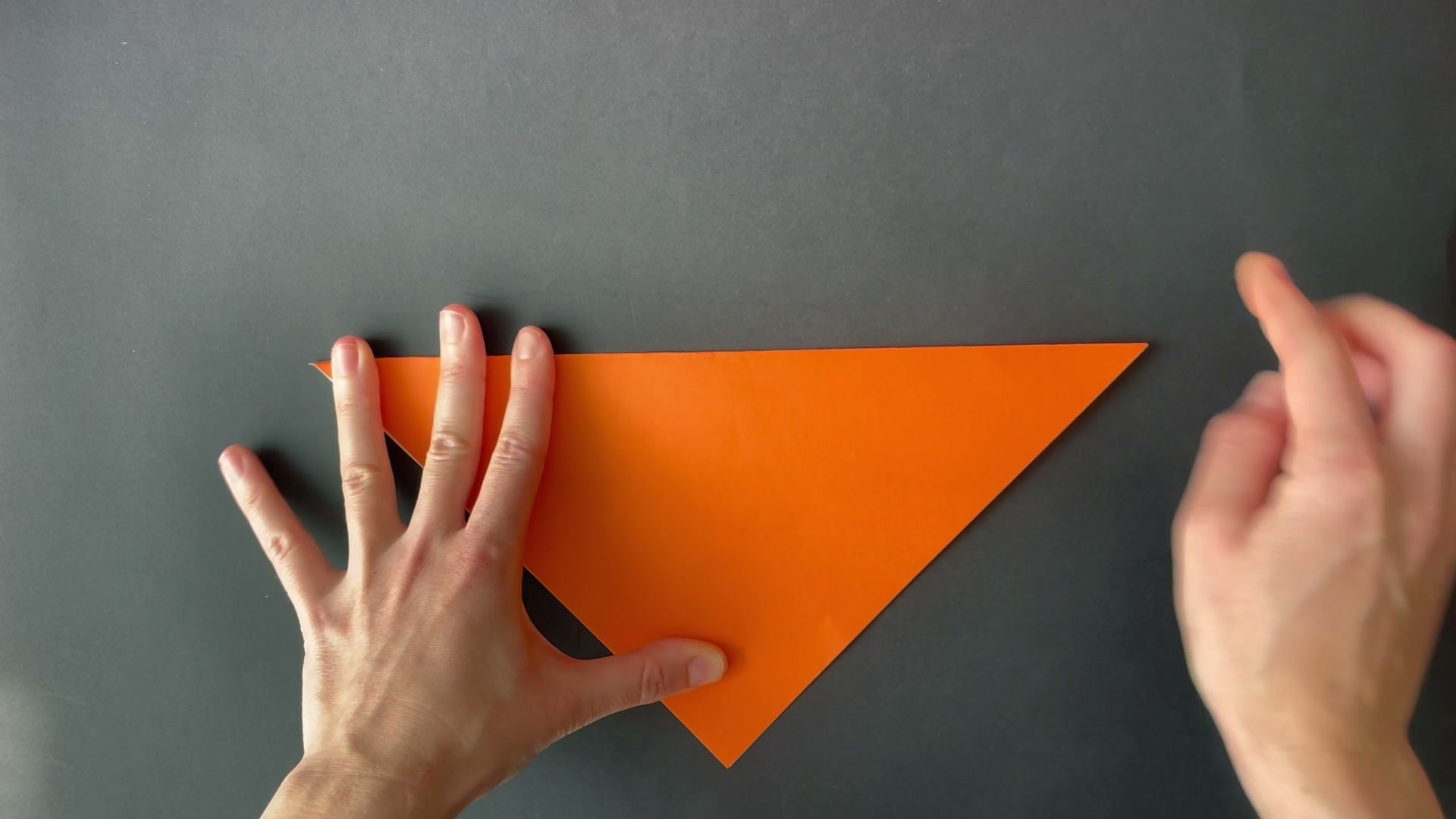}};

\node[img] (a4) at (3\xgap, 0)
{\includegraphics[width=\imgw]{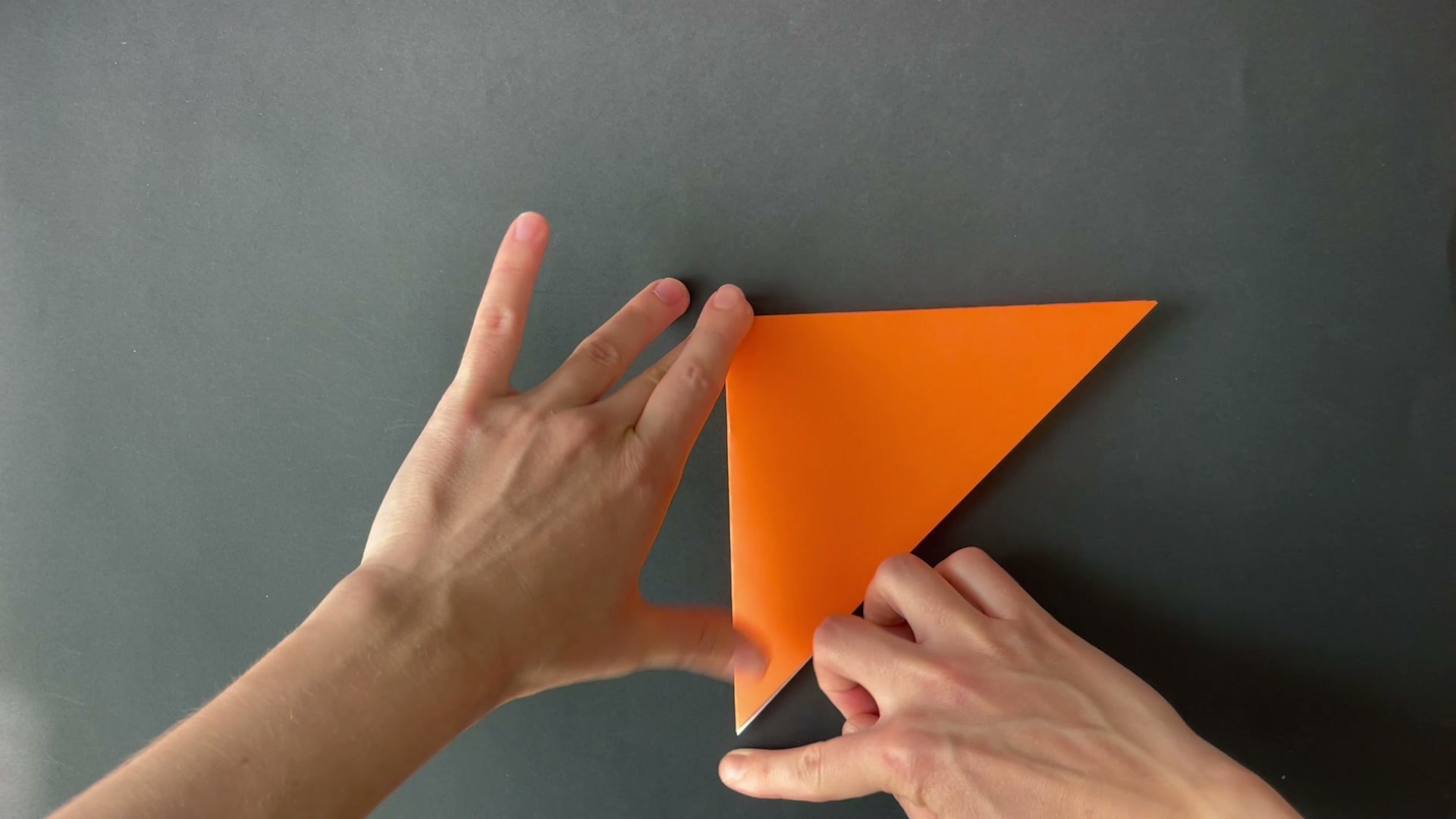}};

\node[img] (a5) at (4\xgap, 0)
{\includegraphics[width=\imgw]{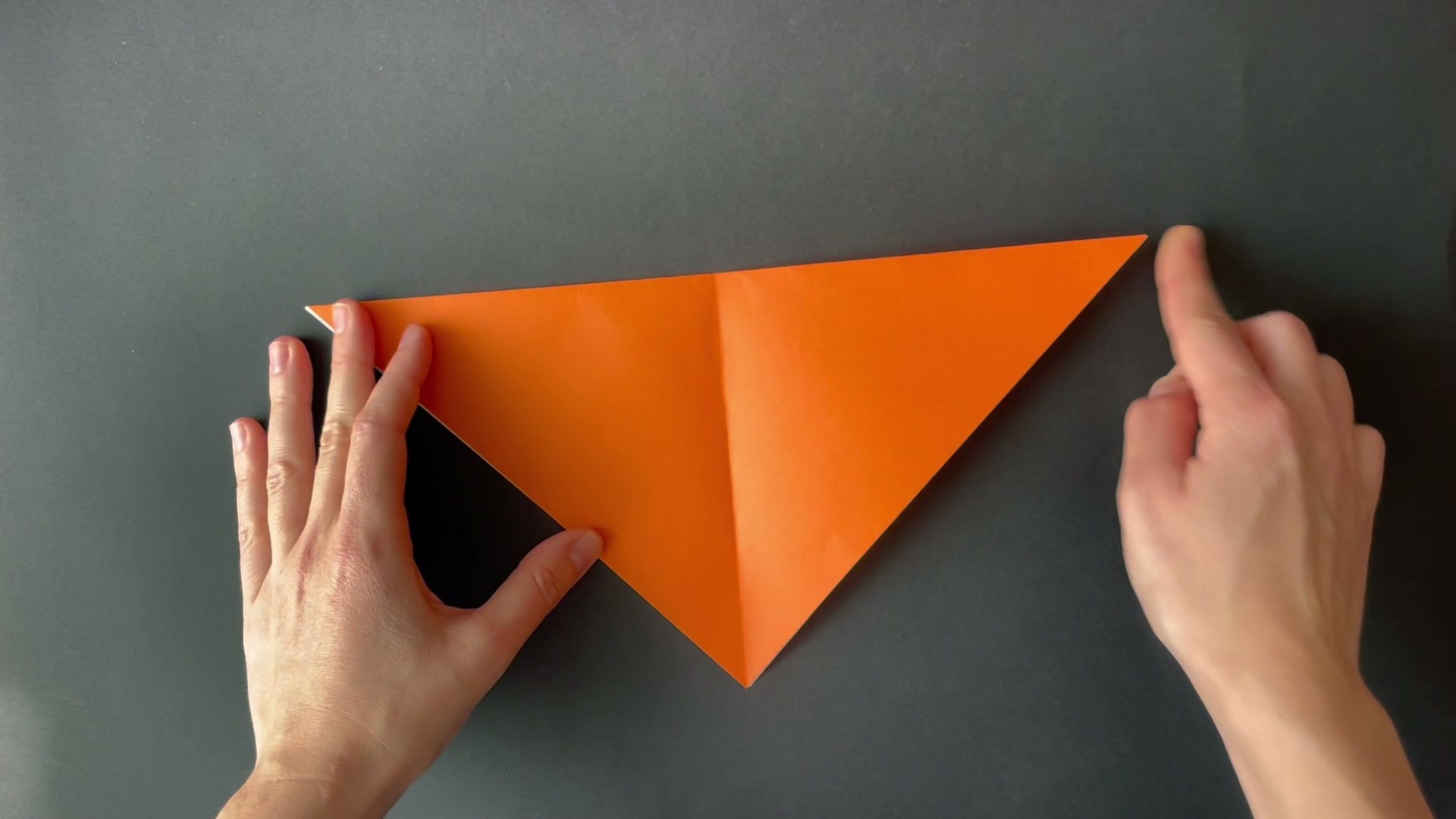}};

\node[img] (a6) at (5\xgap, 0)
{\includegraphics[width=\imgw]{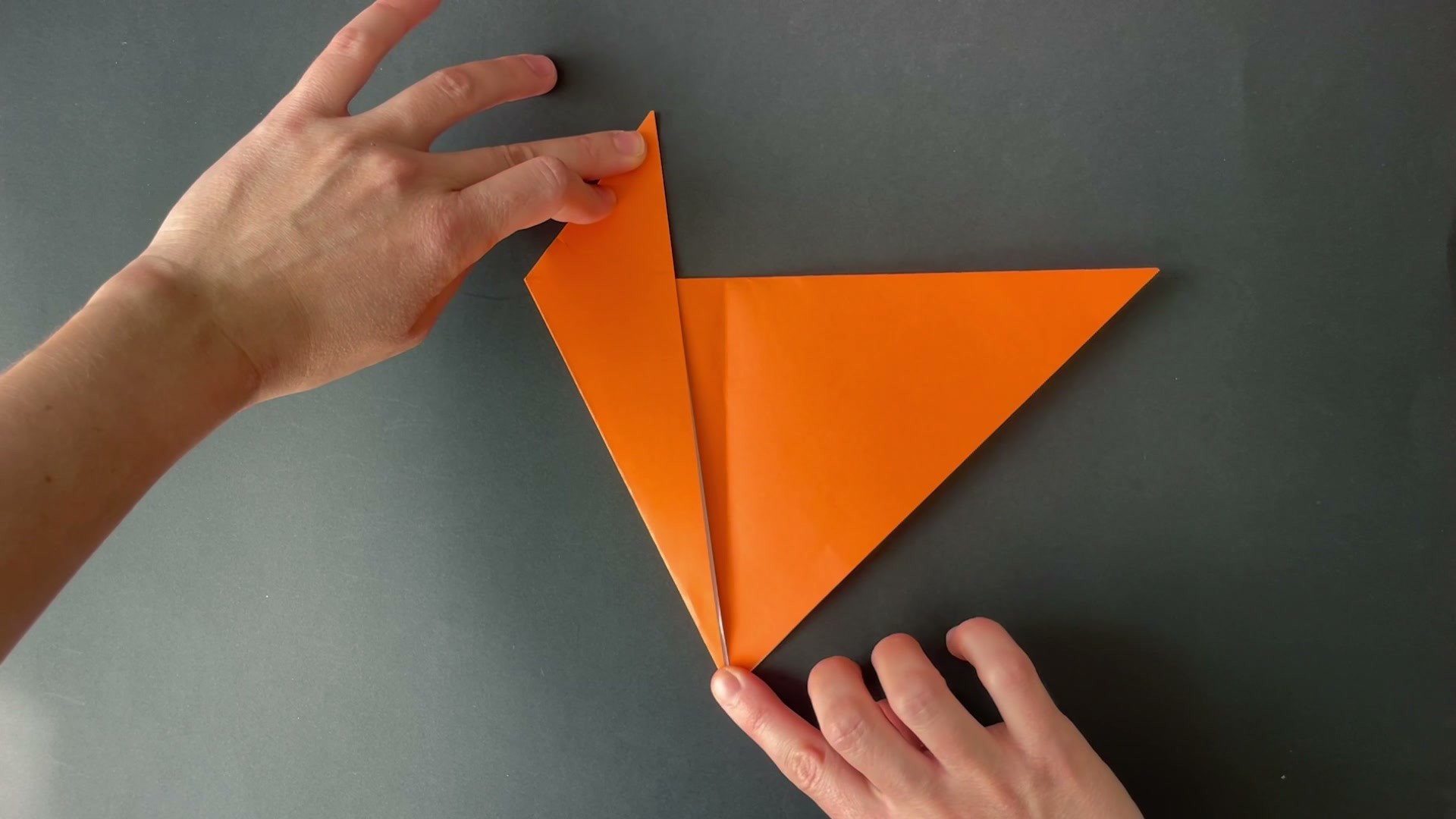}};

\node[img] (a7) at (6\xgap, 0)
{\includegraphics[width=\imgw]{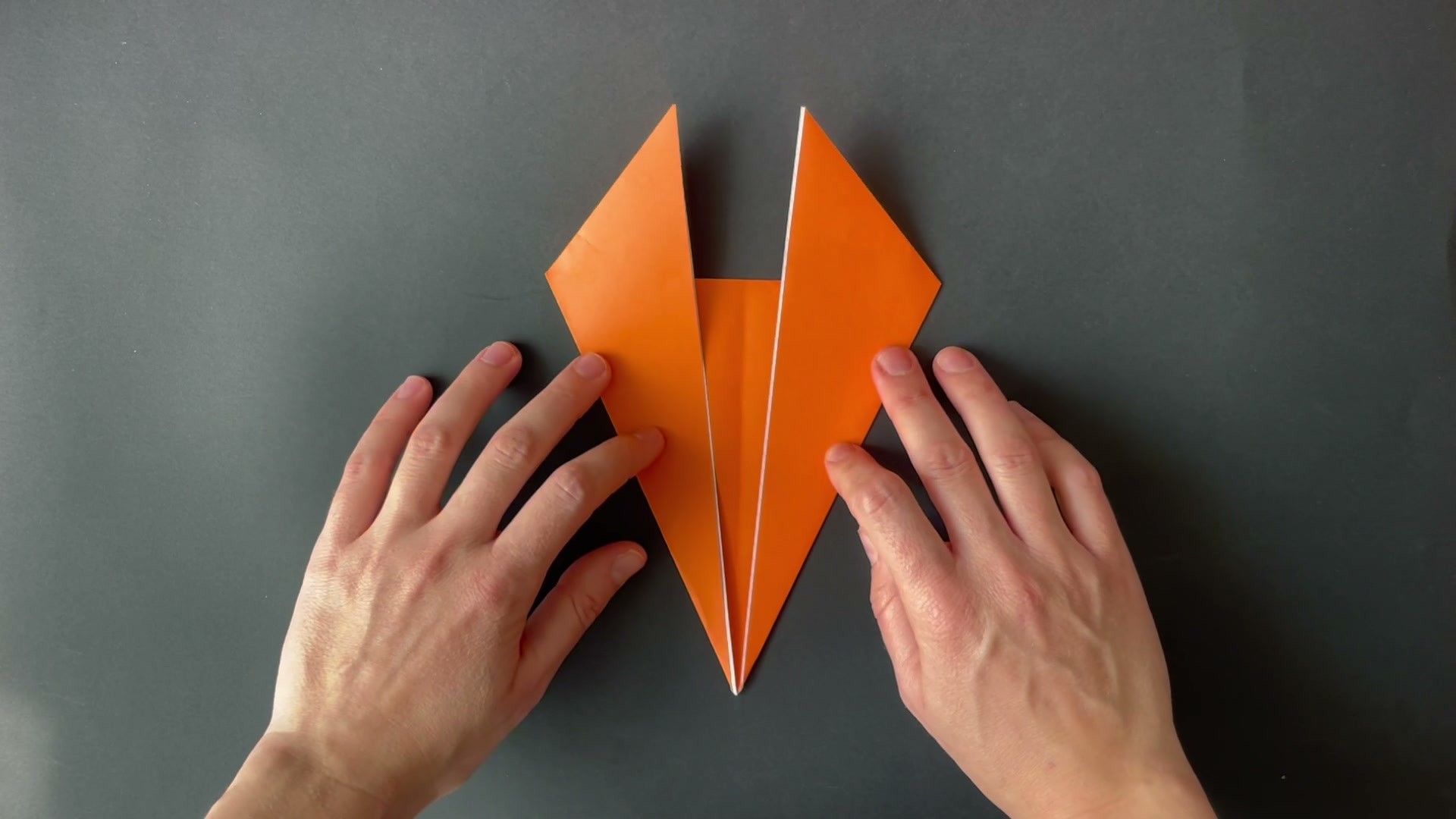}};

\node[img] (a8) at (7\xgap, 0)
{\includegraphics[width=\imgw]{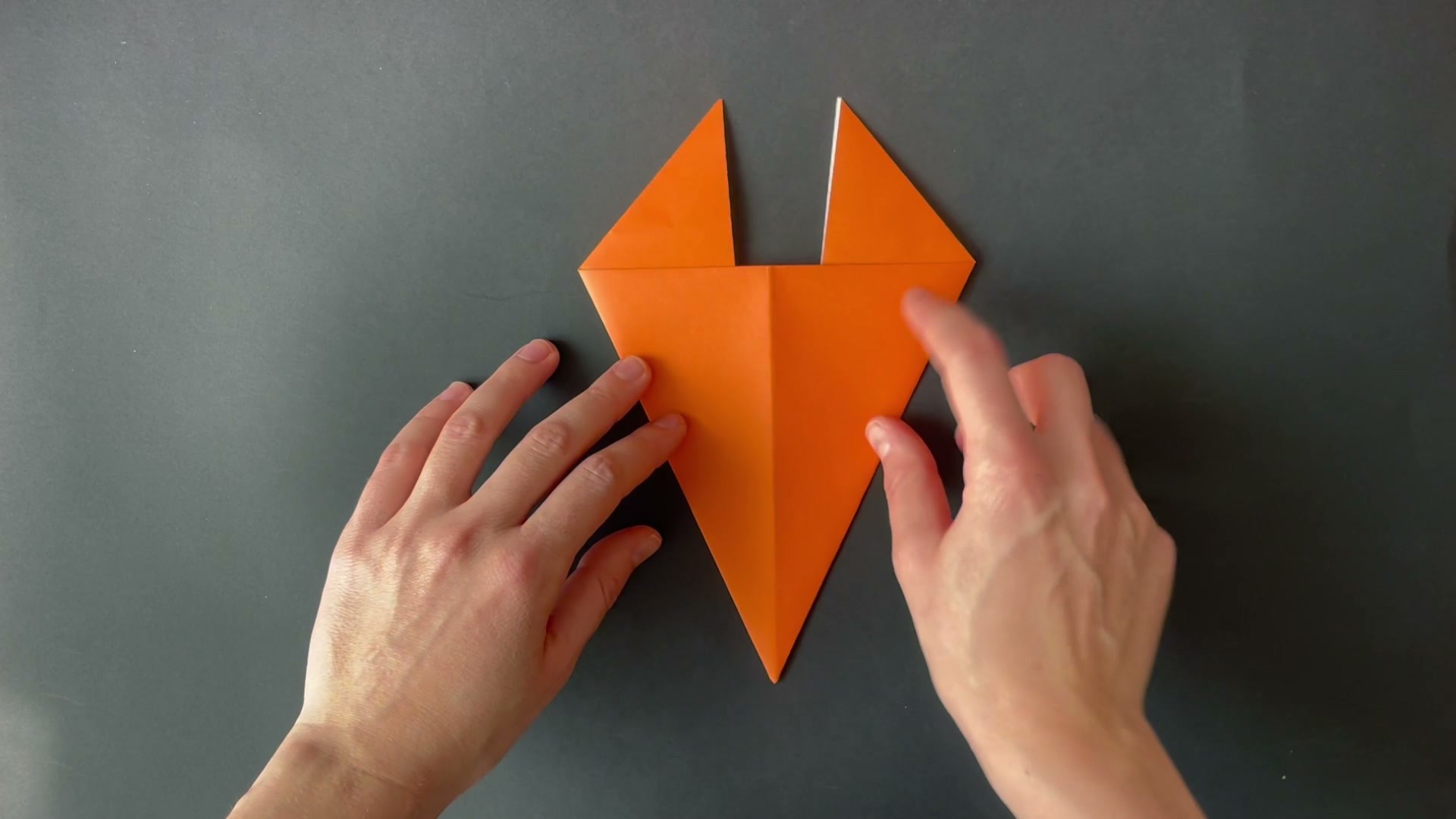}};

\node[frameid] at ([xshift=1pt,yshift=-1pt]a1.north west) {1};
\node[frameid] at ([xshift=1pt,yshift=-1pt]a2.north west) {2};
\node[frameid] at ([xshift=1pt,yshift=-1pt]a3.north west) {3};
\node[frameid] at ([xshift=1pt,yshift=-1pt]a4.north west) {4};
\node[frameid] at ([xshift=1pt,yshift=-1pt]a5.north west) {5};
\node[frameid] at ([xshift=1pt,yshift=-1pt]a6.north west) {6};
\node[frameid] at ([xshift=1pt,yshift=-1pt]a7.north west) {7};
\node[frameid] at ([xshift=1pt,yshift=-1pt]a8.north west) {8};
\node[img] (b1) at (0\xgap, \ygap)
{\secondrowfig{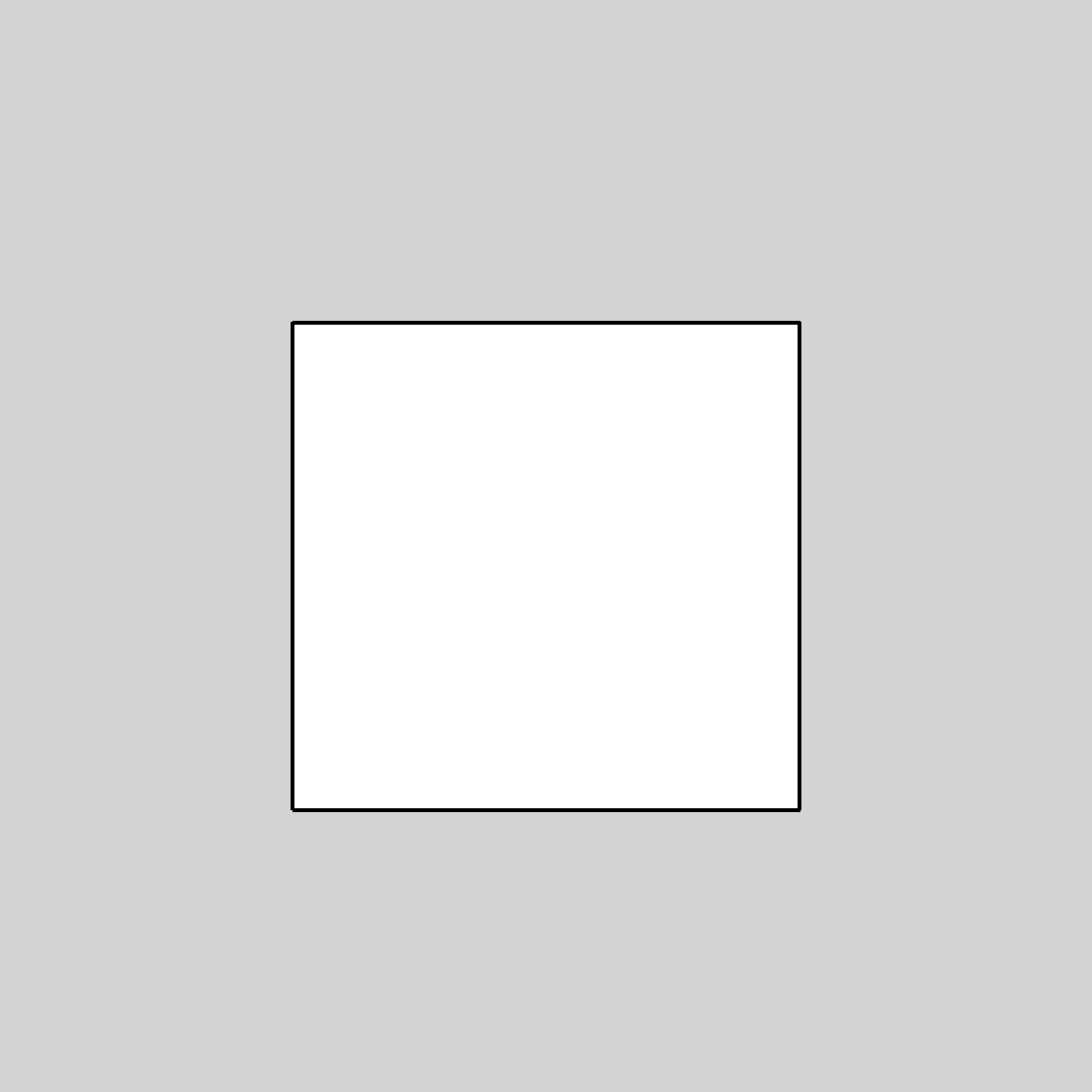}};

\node[img] (b2) at (1\xgap, \ygap)
{\secondrowfig{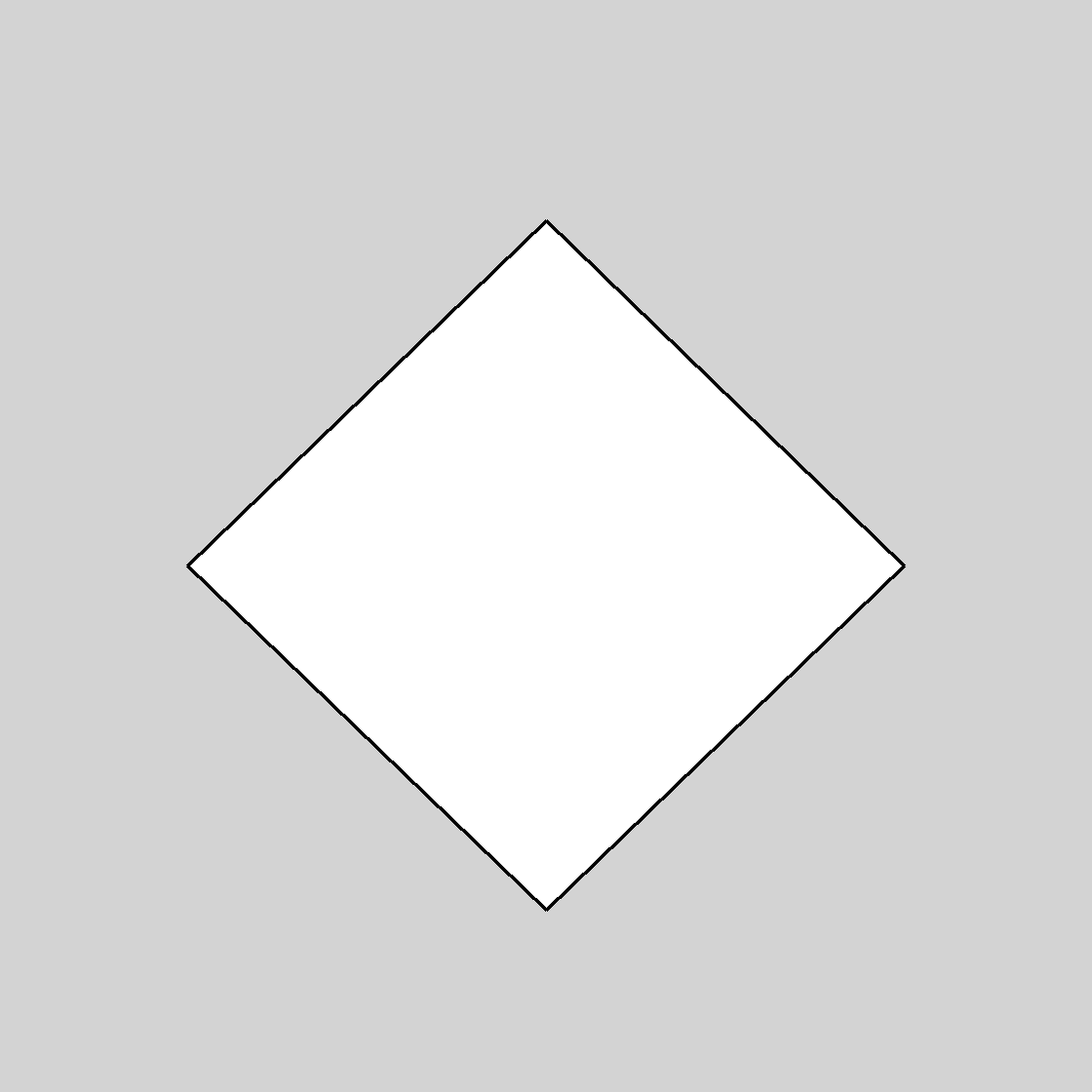}};

\node[img] (b3) at (2\xgap, \ygap)
{\secondrowfig{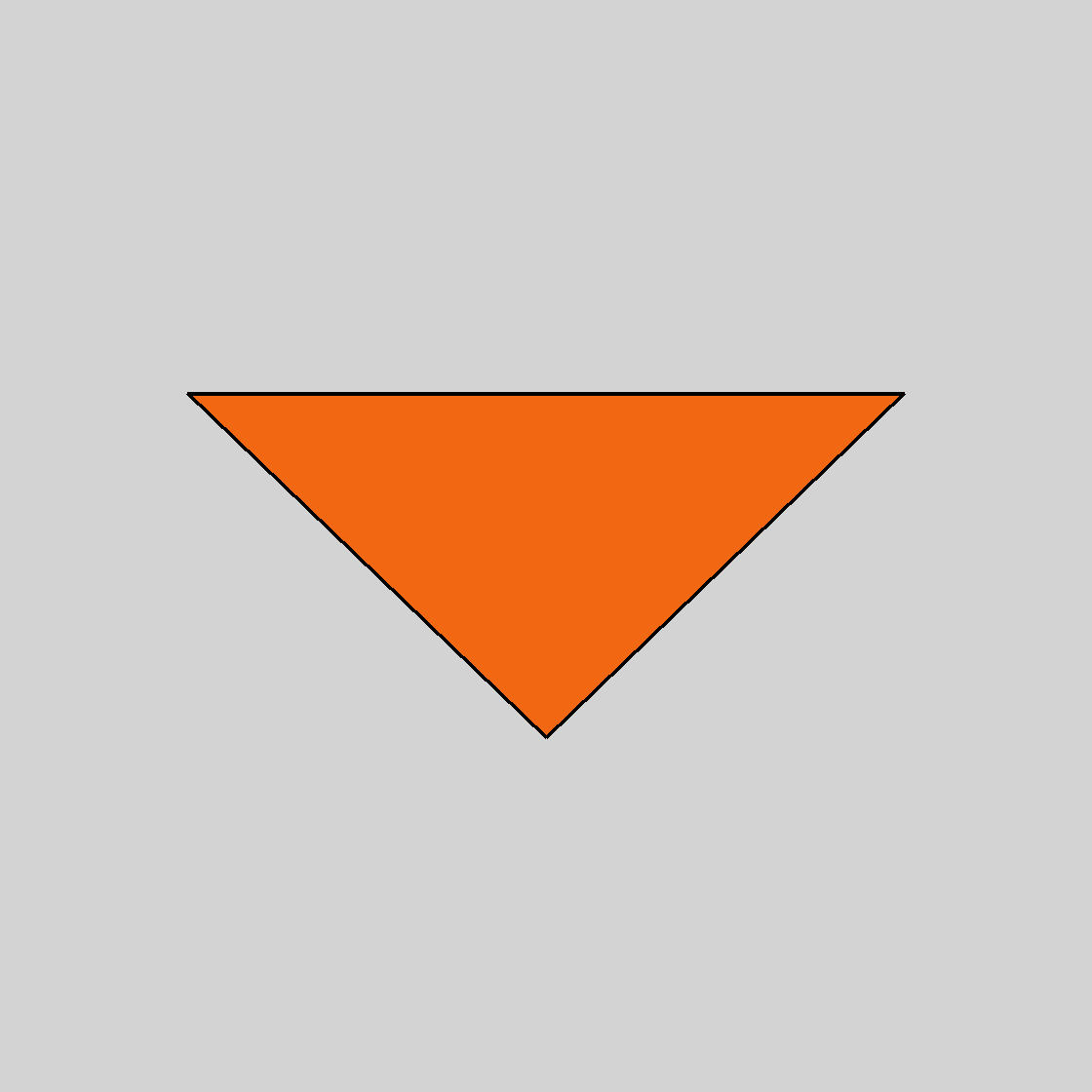}};

\node[img] (b4) at (3\xgap, \ygap)
{\secondrowfig{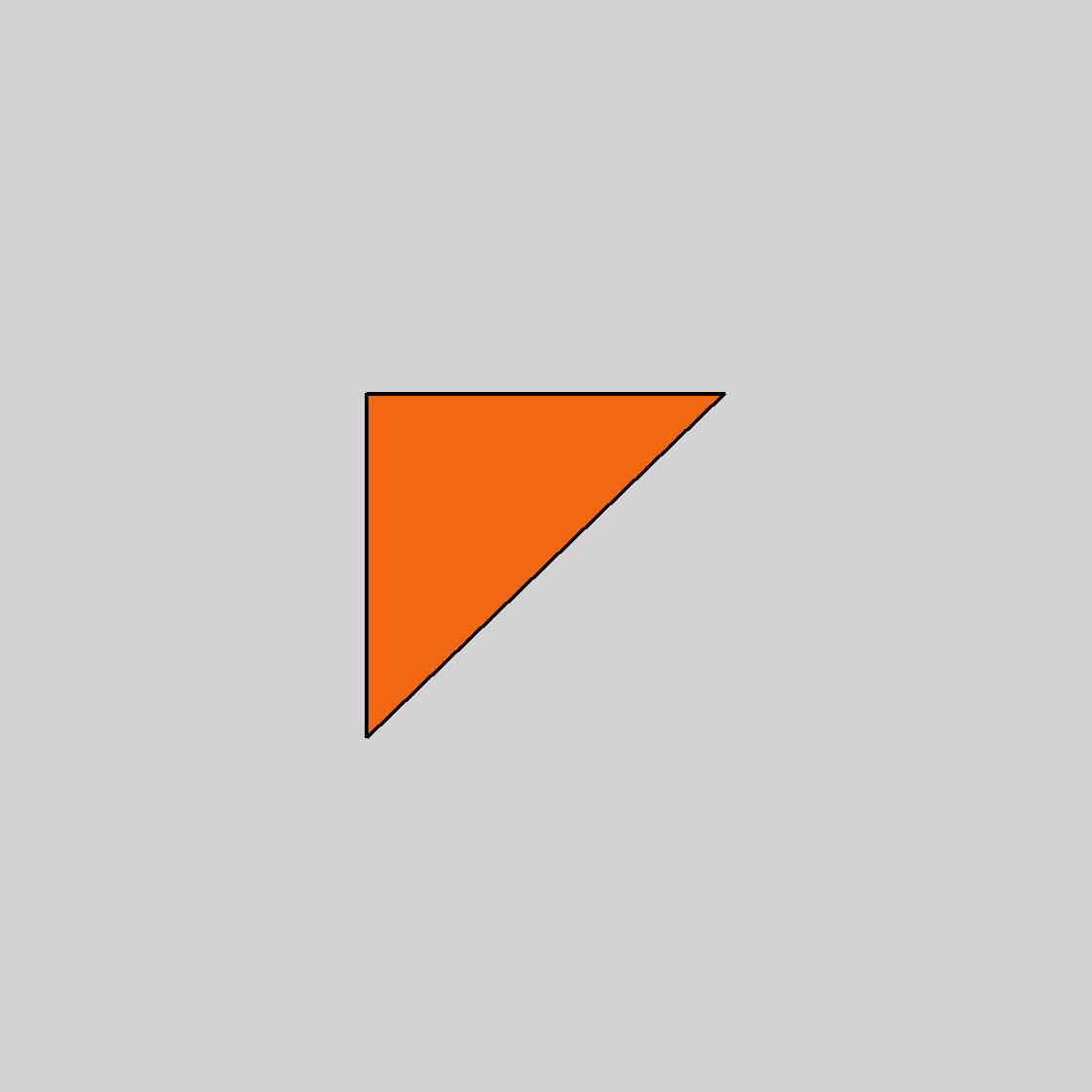}};

\node[img] (b5) at (4\xgap, \ygap)
{\secondrowfig{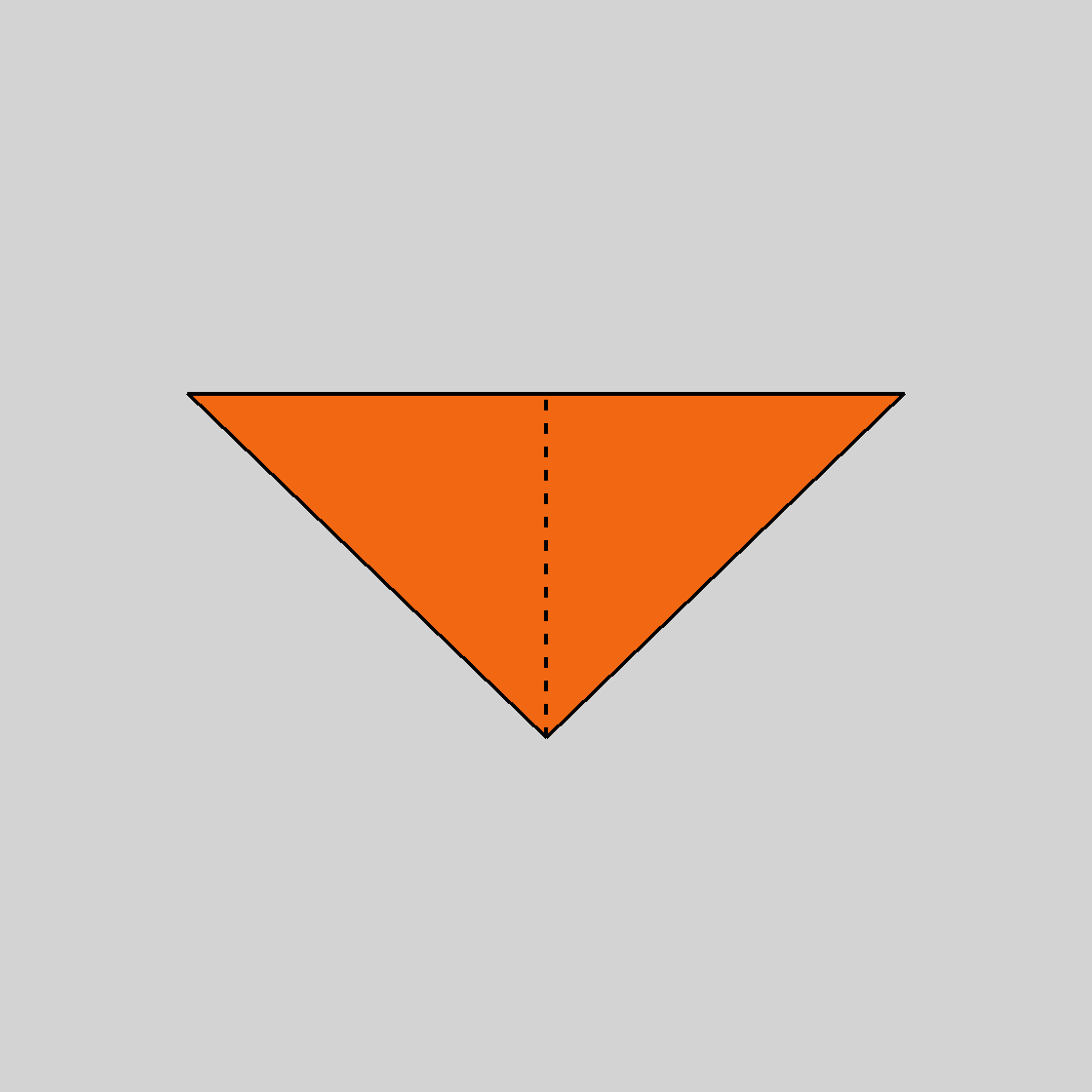}};

\node[img] (b6) at (5\xgap, \ygap)
{\secondrowfig{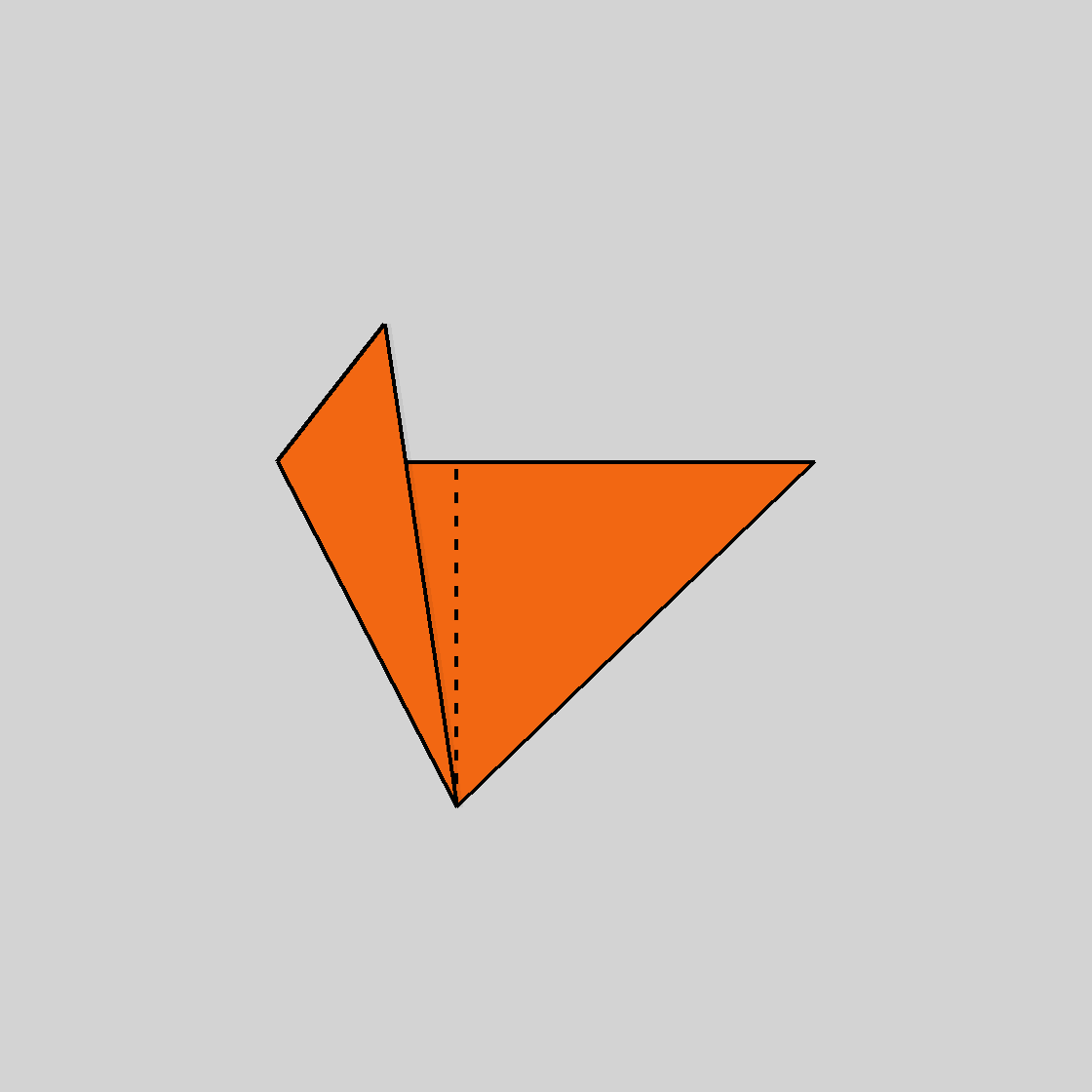}};

\node[img] (b7) at (6\xgap, \ygap)
{\secondrowfig{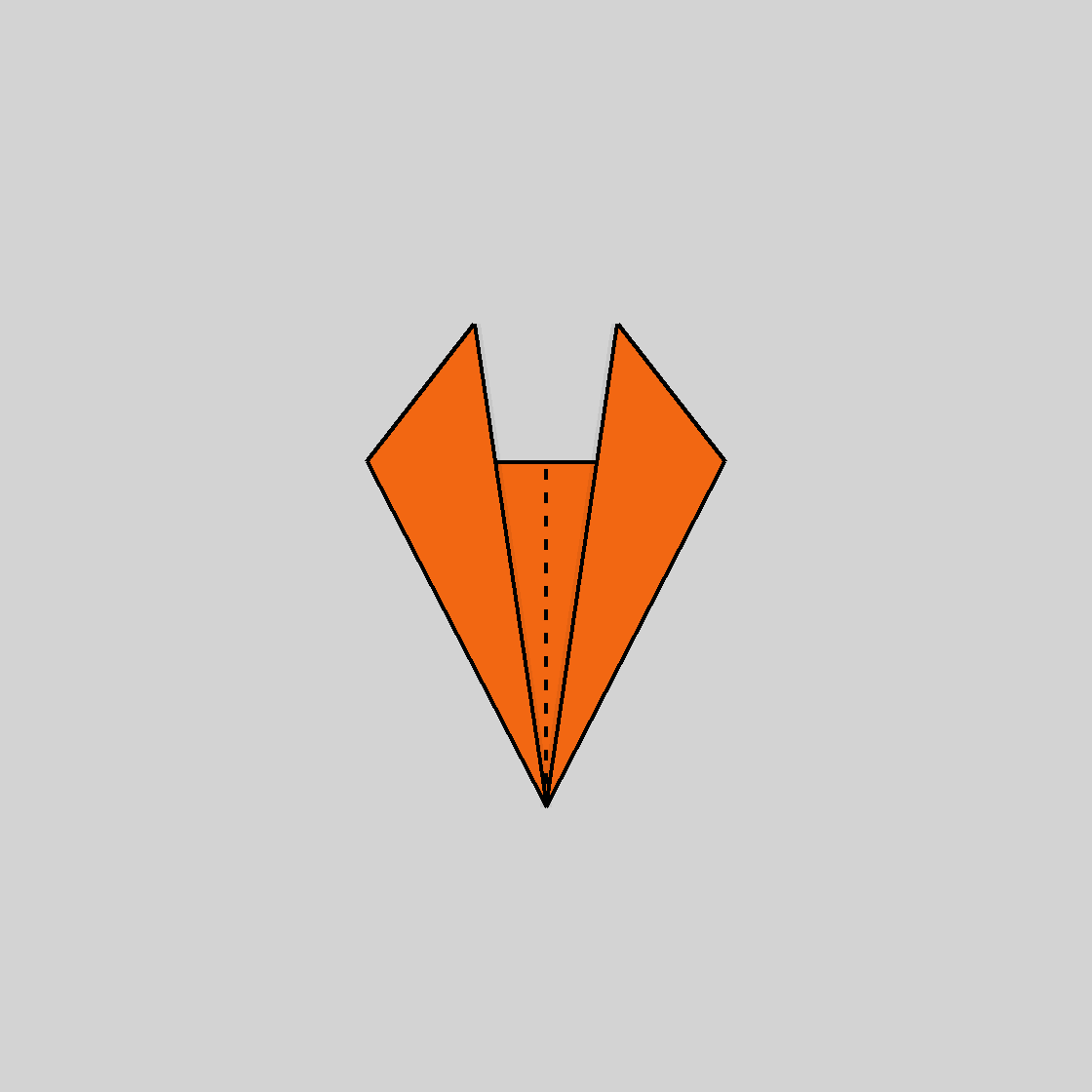}};

\node[img] (b8) at (7\xgap, \ygap)
{\secondrowfig{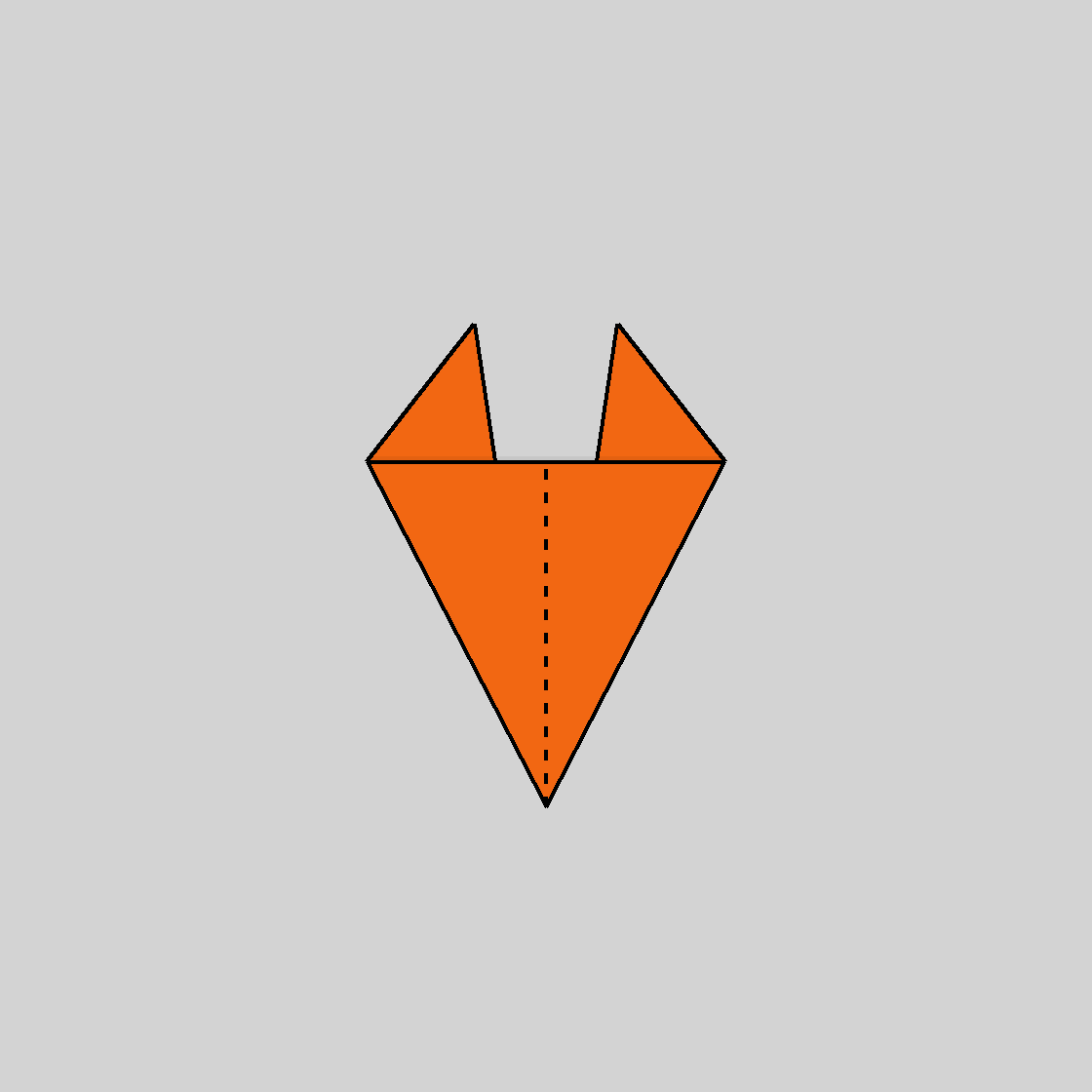}};

\node[cmd] at ($(b1)!0.5!(b2) + (0,-0.77)$) {
rotate(45$^\circ$)
};

\node[cmd] at ($(b2)!0.5!(b3) + (0,-0.77)$) {
fold([4,2],1)
};

\node[cmd] at ($(b3)!0.5!(b4) + (0,-0.65)$) {
add\_v(0.5,[4,2])\\
fold([5,3],-1)
};

\node[cmd] at ($(b4)!0.5!(b5) + (0,-0.77)$) {
unfold()
};

\node[cmd] at ($(b5)!0.5!(b6) + (0,-0.65)$) {
add\_v(0.4,[5,4])\\
fold([6,1],-1)
};

\node[cmd] at ($(b6)!0.5!(b7) + (0,-0.55)$) {
add\_v(0.4,[5,2])\\
fold([7,1],-1)
};

\node[cmd] at ($(b7)!0.5!(b8) + (0,-0.77)$) {
flip(x)
};

\end{tikzpicture}

\vspace{10pt}

\newlength{\catimgw}
\setlength{\catimgw}{0.108\textwidth}

\newlength{\catxgap}
\setlength{\catxgap}{0.1111\textwidth}

\newlength{\catygap}
\setlength{\catygap}{-0.073\textwidth}

\newcommand{\secondrowfigcat}[1]{%
  \includegraphics[
    width=\catimgw,
    trim={0pt 5cm 0pt 5cm},
    clip
  ]{#1}%
}

\begin{tikzpicture}[
    img/.style={
        inner sep=0pt,
        outer sep=0pt
    },
    cmd/.style={
        fill=white,
        fill opacity=0.90,
        text opacity=1,
        draw=black!30,
        rounded corners=1.5pt,
        inner sep=1pt,
        font=\tiny\ttfamily,
        align=left
    },
    frameid/.style={
        fill=white,
        fill opacity=0,
        text opacity=1,
        text=white,
        draw=black!25,
        rounded corners=0pt,
        inner sep=1pt,
        font=\tiny\ttfamily,
        anchor=north west
    }
]

\node[img] (c1) at (0\catxgap, 0)
{\includegraphics[width=\catimgw]{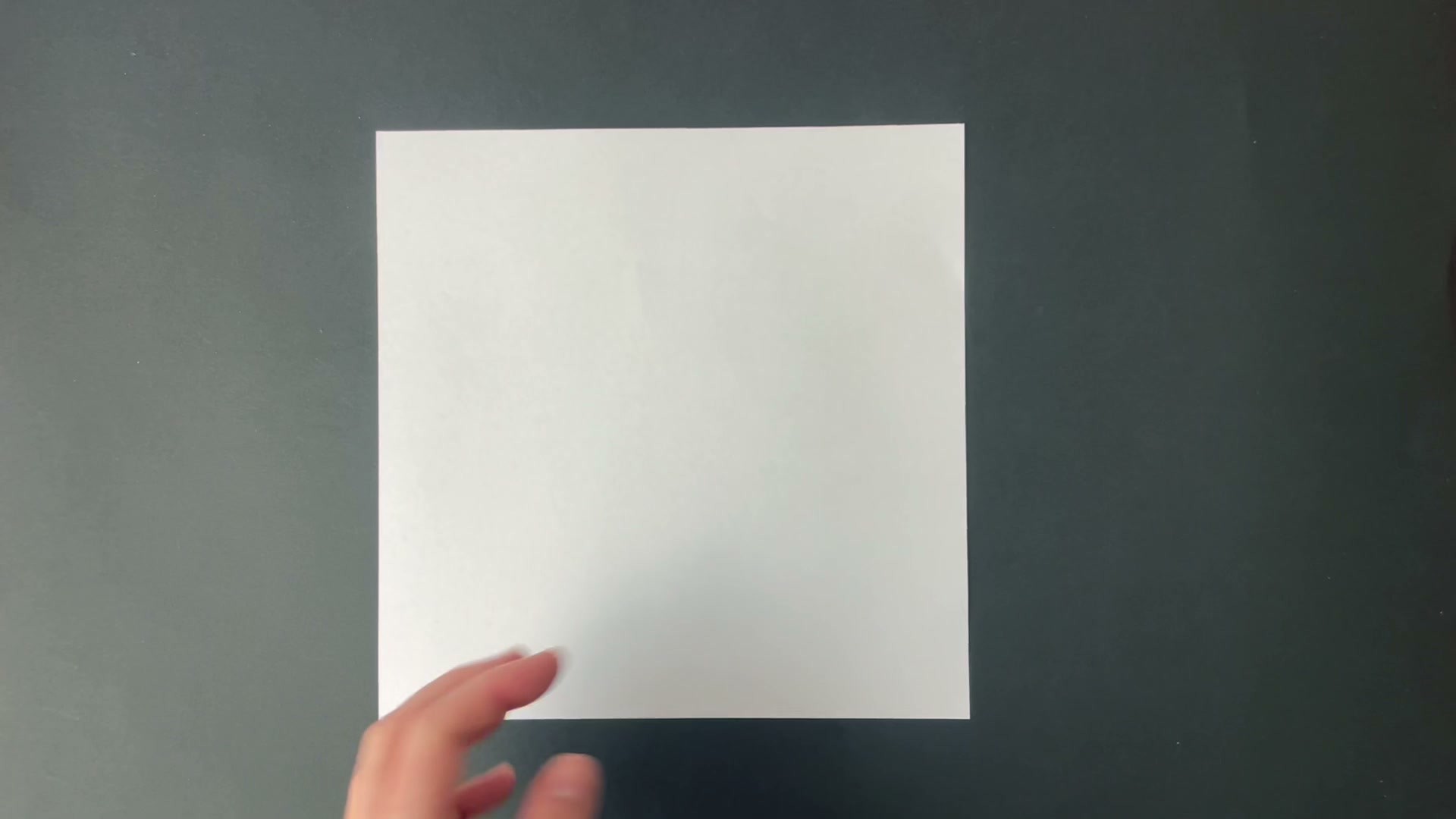}};

\node[img] (c2) at (1\catxgap, 0)
{\includegraphics[width=\catimgw]{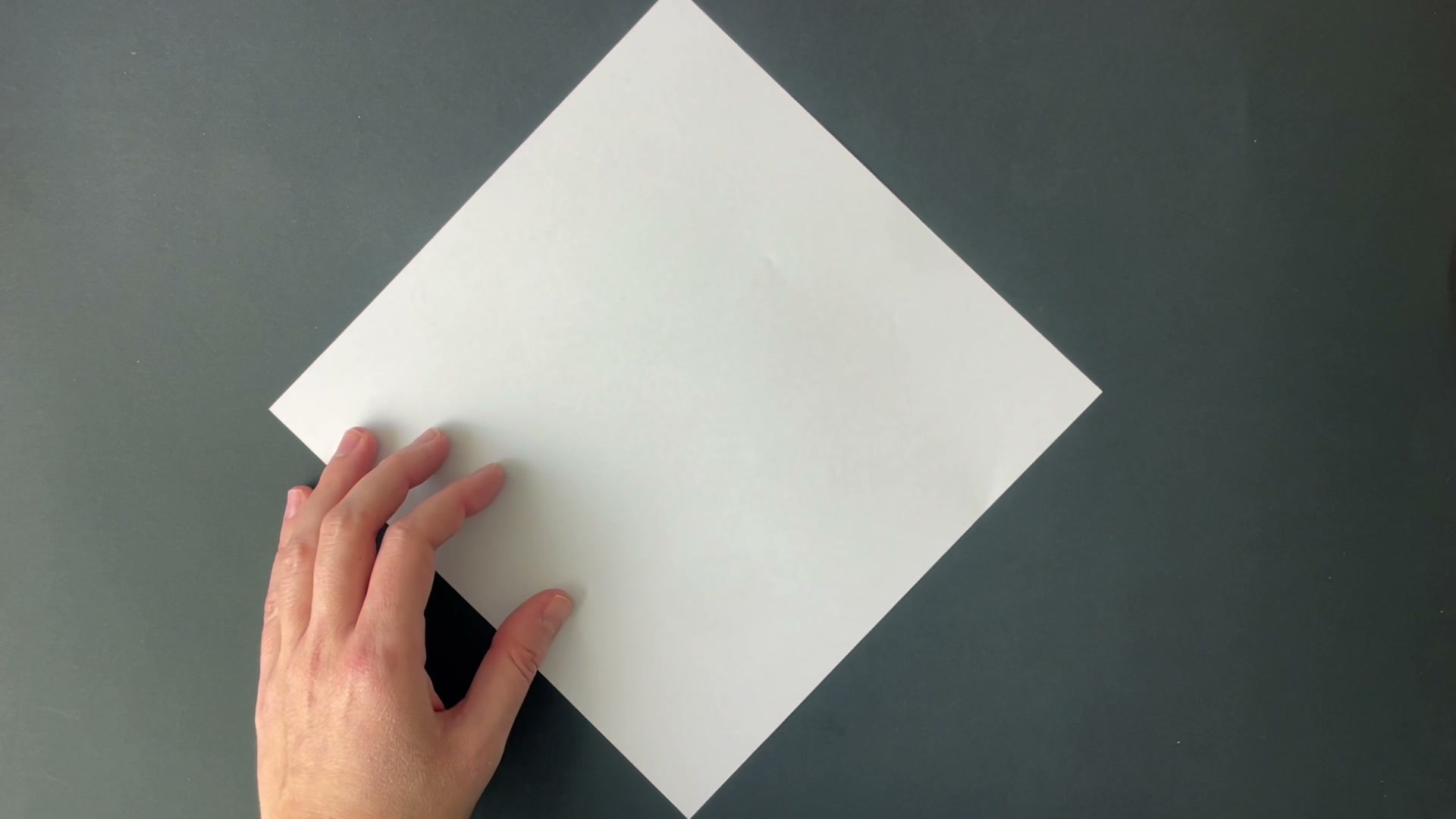}};

\node[img] (c3) at (2\catxgap, 0)
{\includegraphics[width=\catimgw]{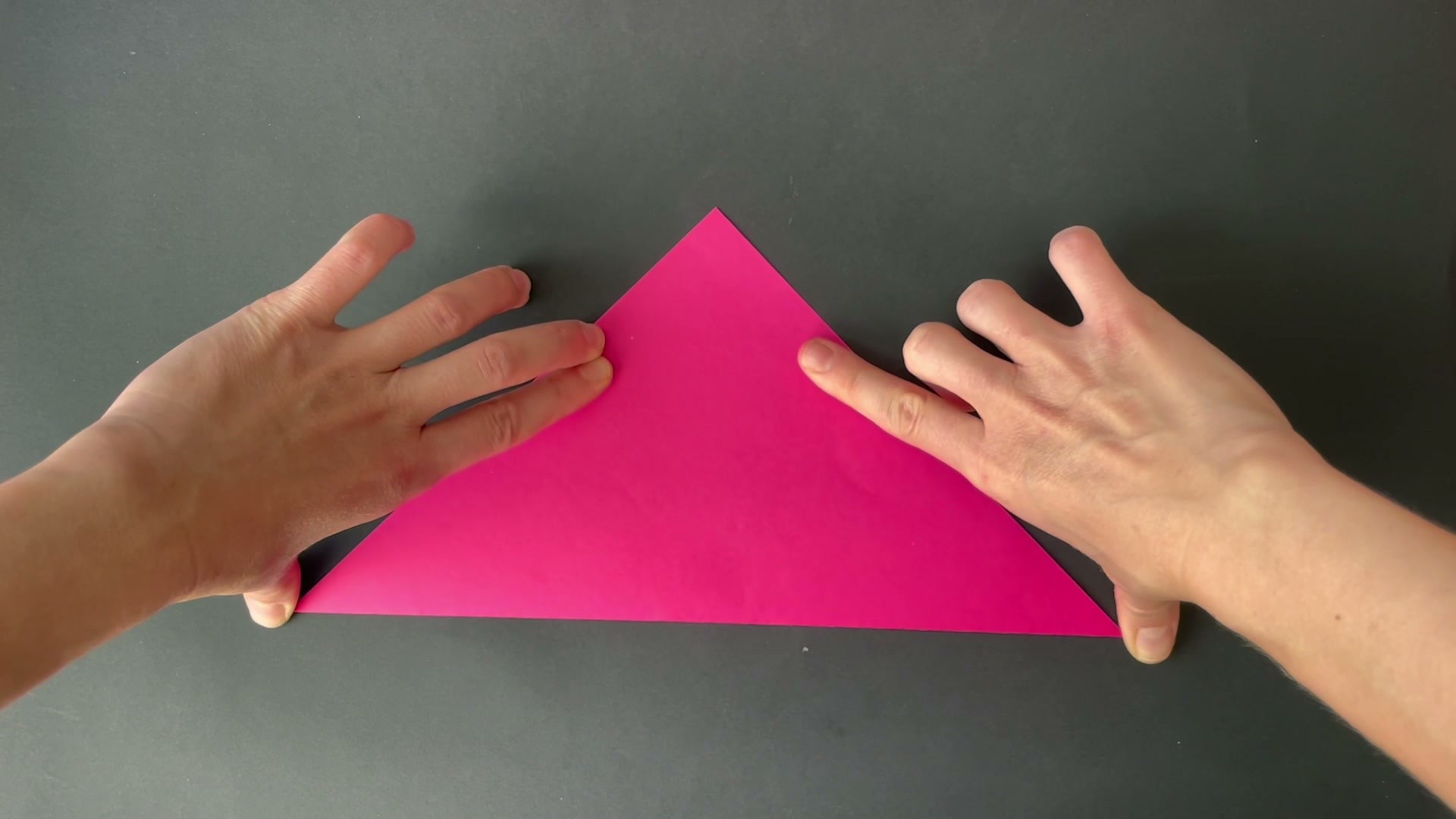}};

\node[img] (c4) at (3\catxgap, 0)
{\includegraphics[width=\catimgw]{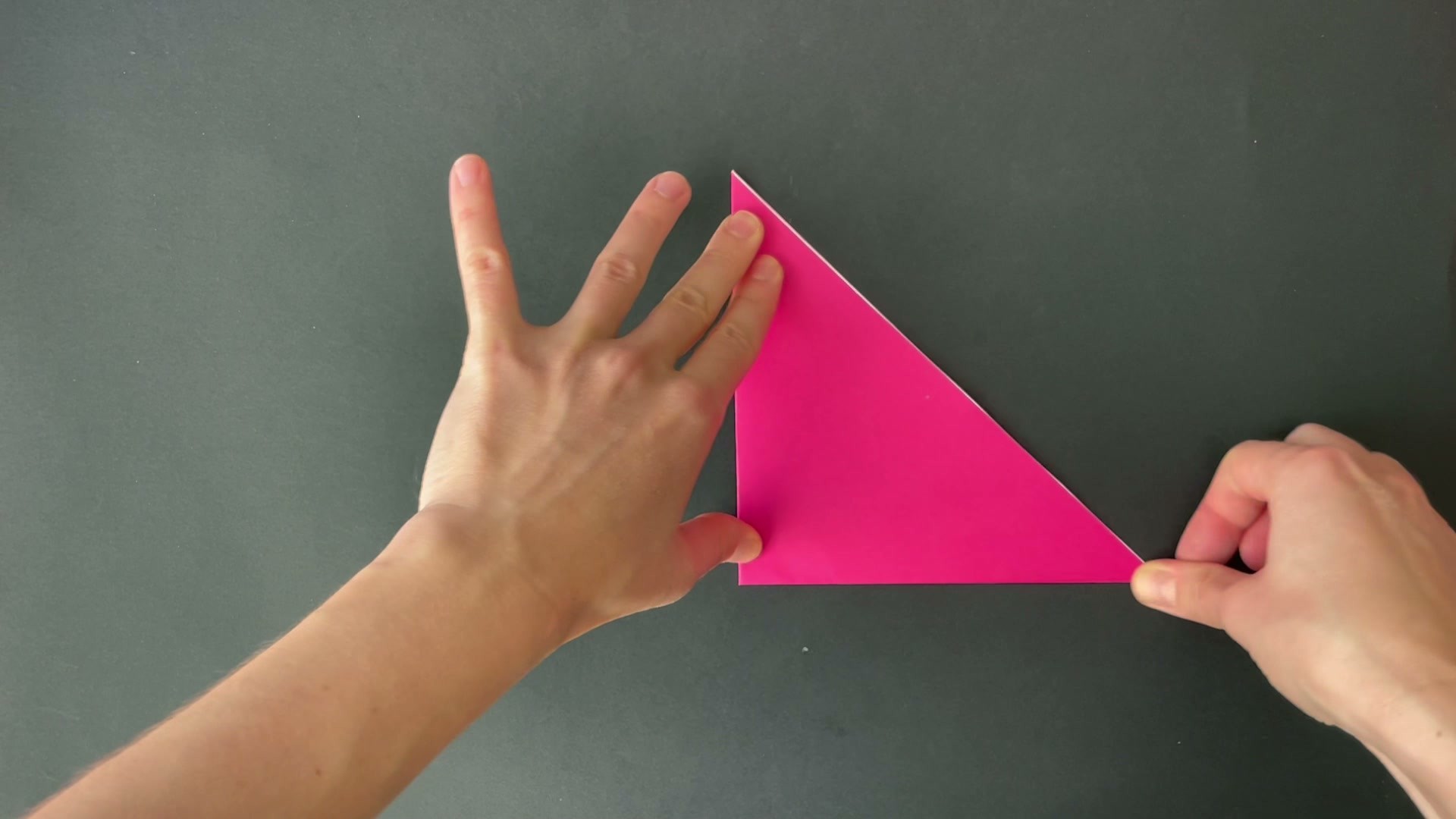}};

\node[img] (c5) at (4\catxgap, 0)
{\includegraphics[width=\catimgw]{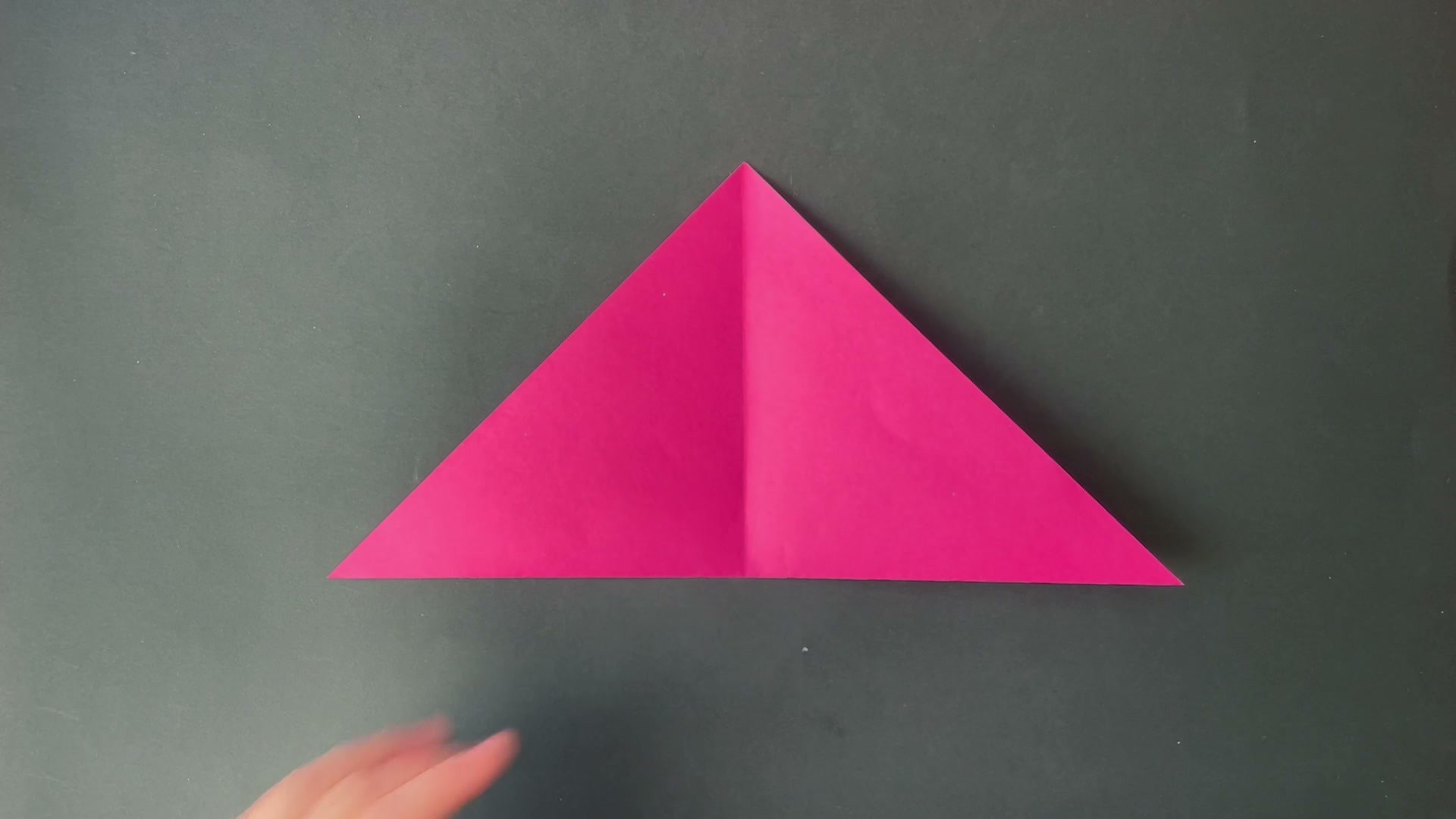}};

\node[img] (c6) at (5\catxgap, 0)
{\includegraphics[width=\catimgw]{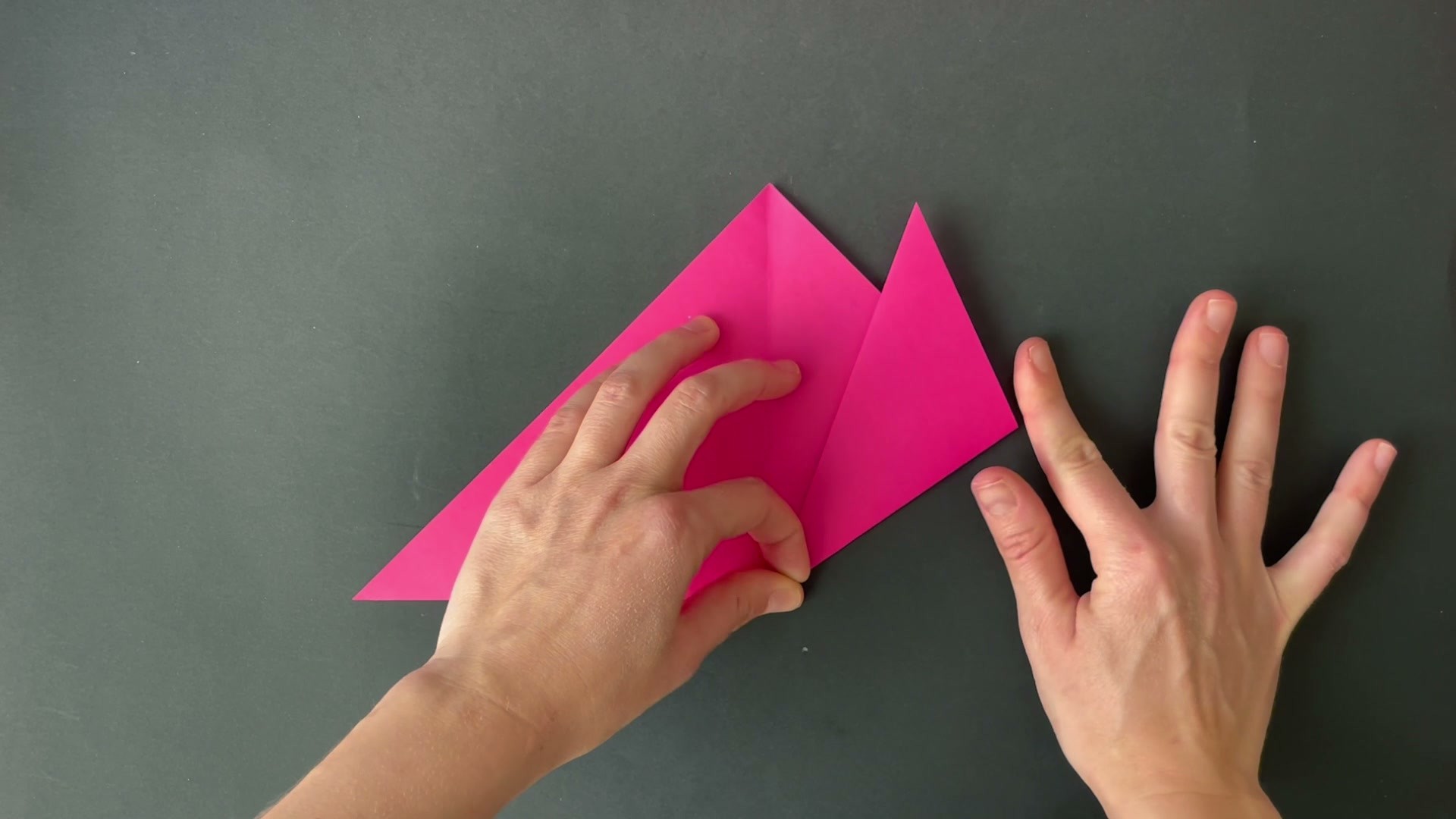}};

\node[img] (c7) at (6\catxgap, 0)
{\includegraphics[width=\catimgw]{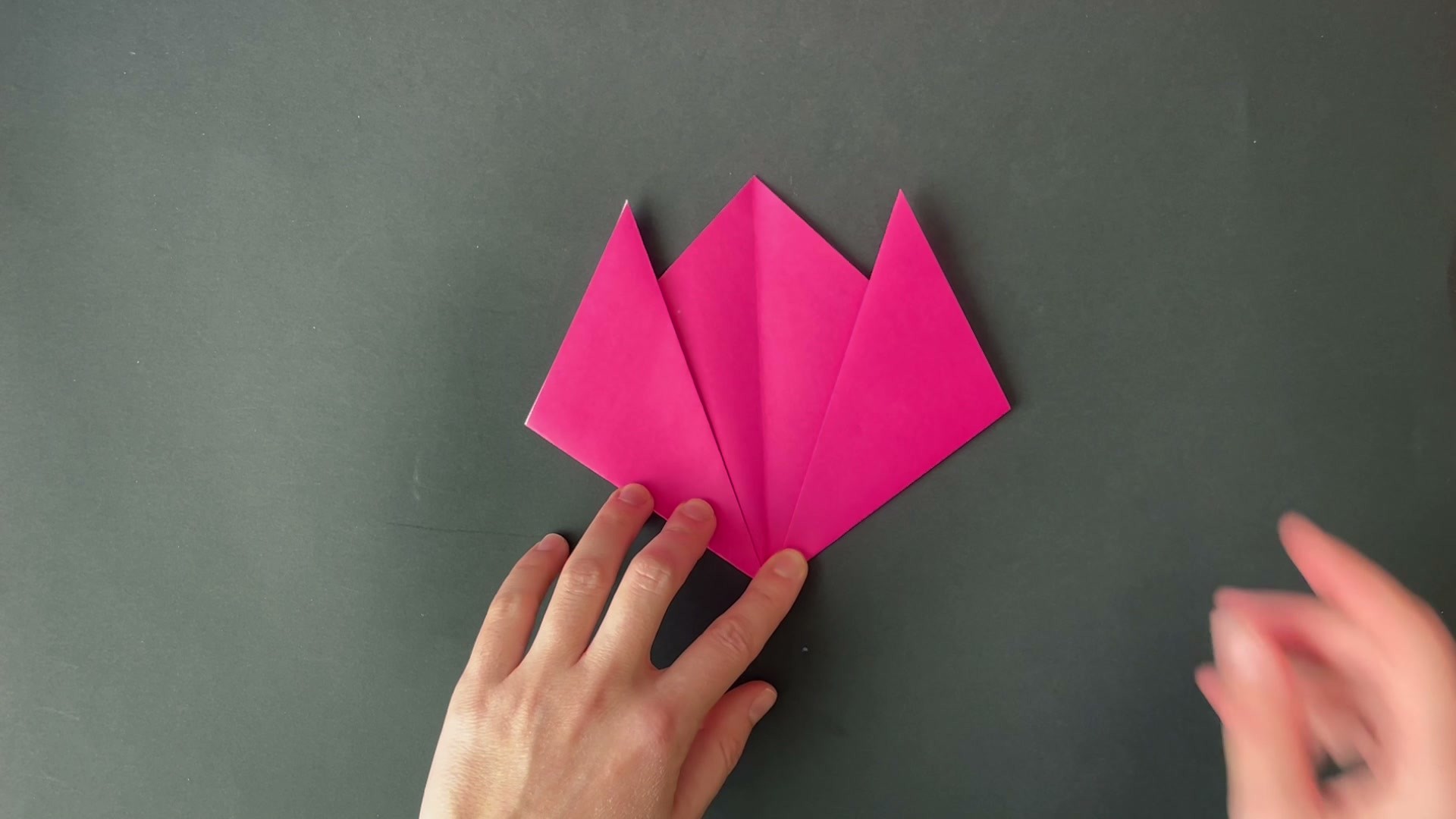}};

\node[img] (c8) at (7\catxgap, 0)
{\includegraphics[width=\catimgw]{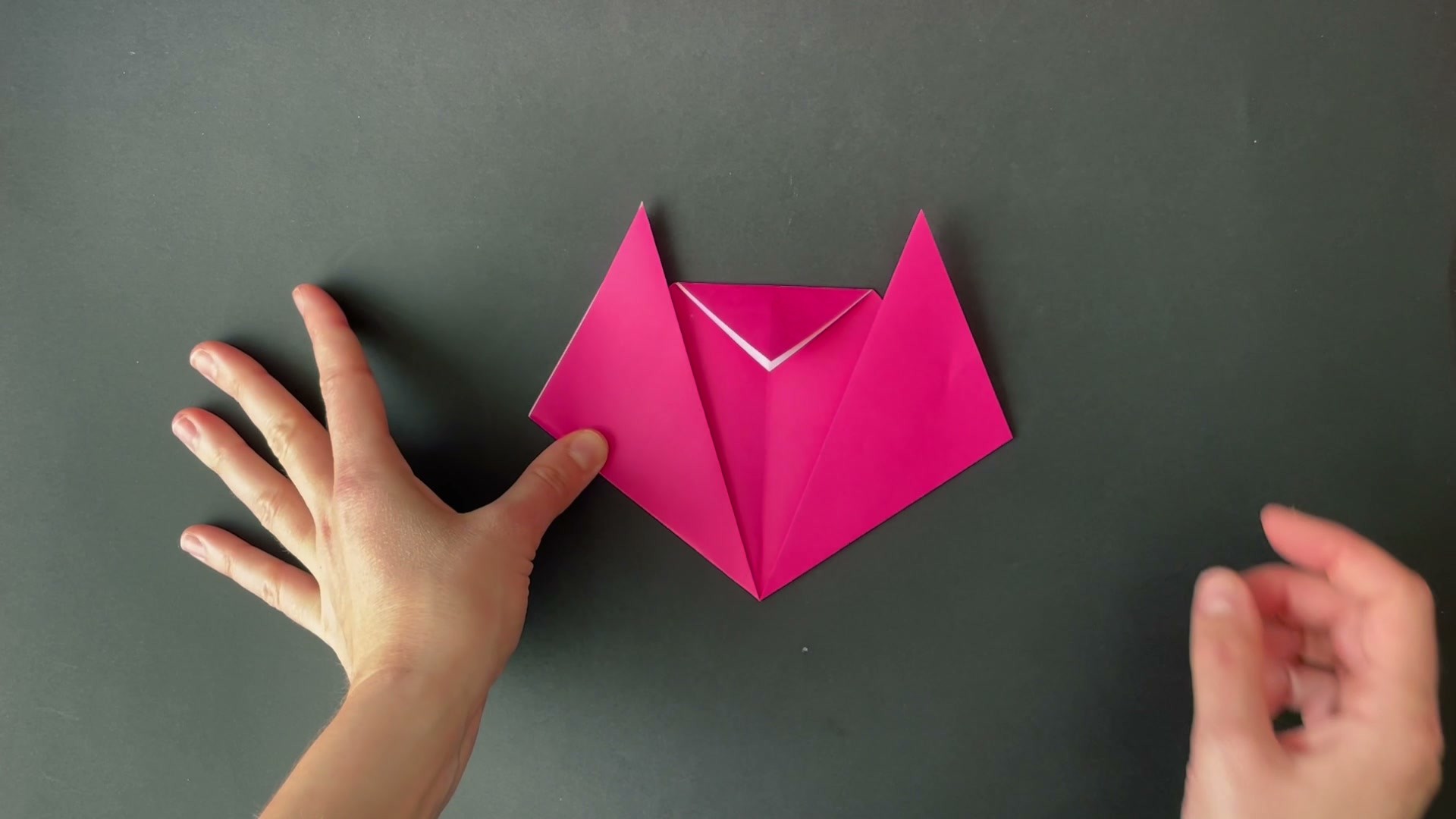}};

\node[img] (c9) at (8\catxgap, 0)
{\includegraphics[width=\catimgw]{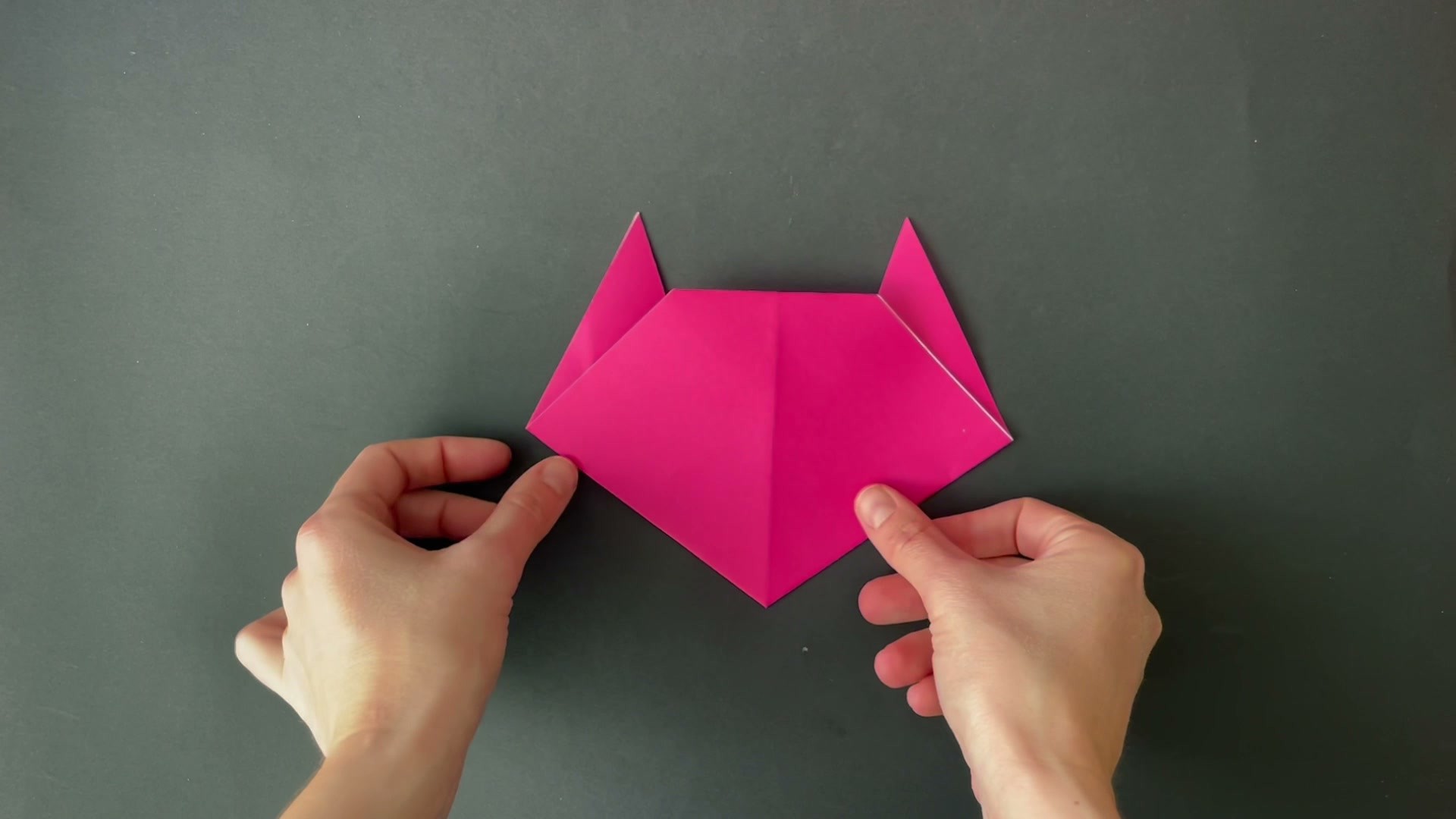}};

\node[frameid] at ([xshift=1pt,yshift=-1pt]c1.north west) {1};
\node[frameid] at ([xshift=1pt,yshift=-1pt]c2.north west) {2};
\node[frameid] at ([xshift=1pt,yshift=-1pt]c3.north west) {3};
\node[frameid] at ([xshift=1pt,yshift=-1pt]c4.north west) {4};
\node[frameid] at ([xshift=1pt,yshift=-1pt]c5.north west) {5};
\node[frameid] at ([xshift=1pt,yshift=-1pt]c6.north west) {6};
\node[frameid] at ([xshift=1pt,yshift=-1pt]c7.north west) {7};
\node[frameid] at ([xshift=1pt,yshift=-1pt]c8.north west) {8};
\node[frameid] at ([xshift=1pt,yshift=-1pt]c9.north west) {9};
\node[img] (d1) at (0\catxgap, \catygap)
{\secondrowfigcat{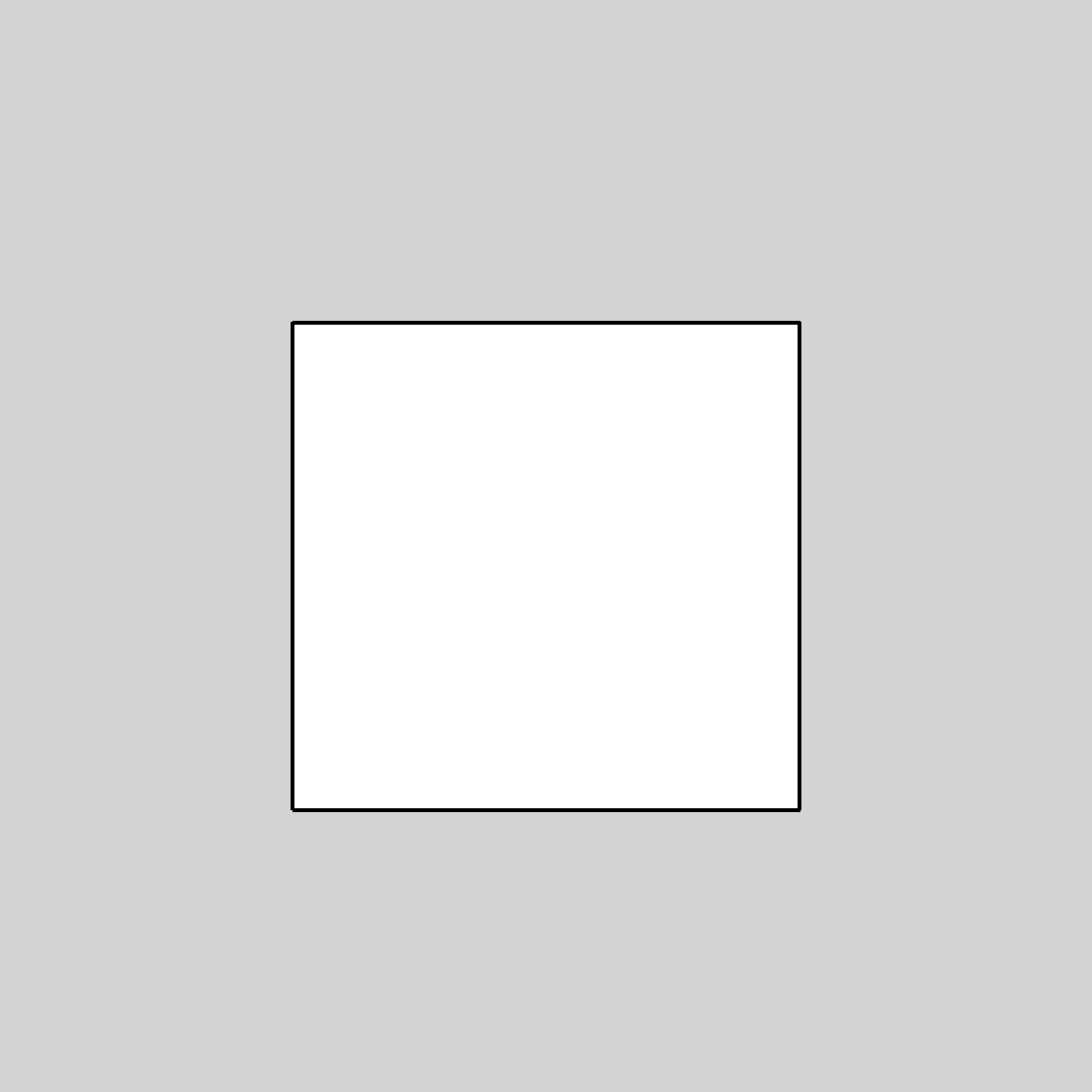}};

\node[img] (d2) at (1\catxgap, \catygap)
{\secondrowfigcat{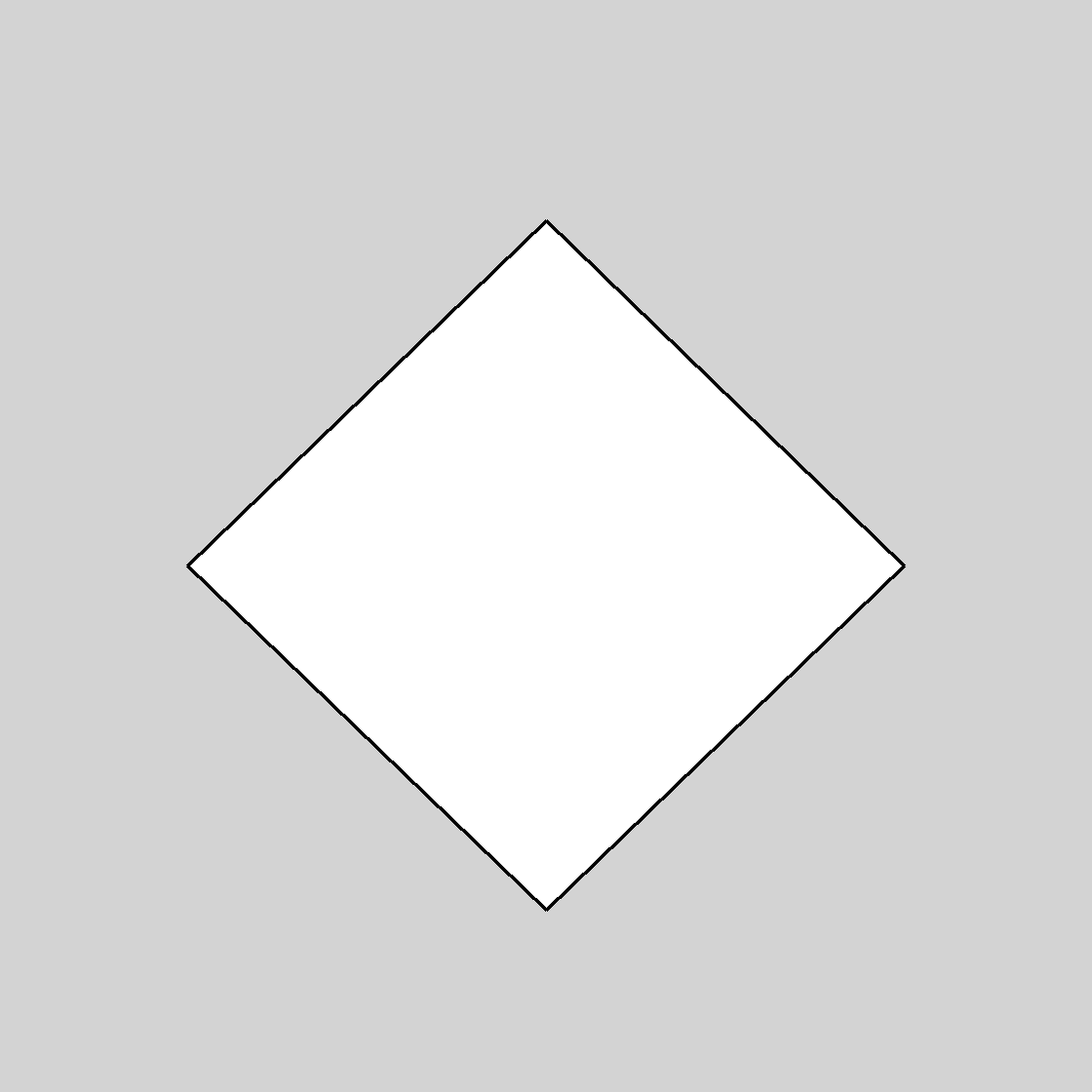}};

\node[img] (d3) at (2\catxgap, \catygap)
{\secondrowfigcat{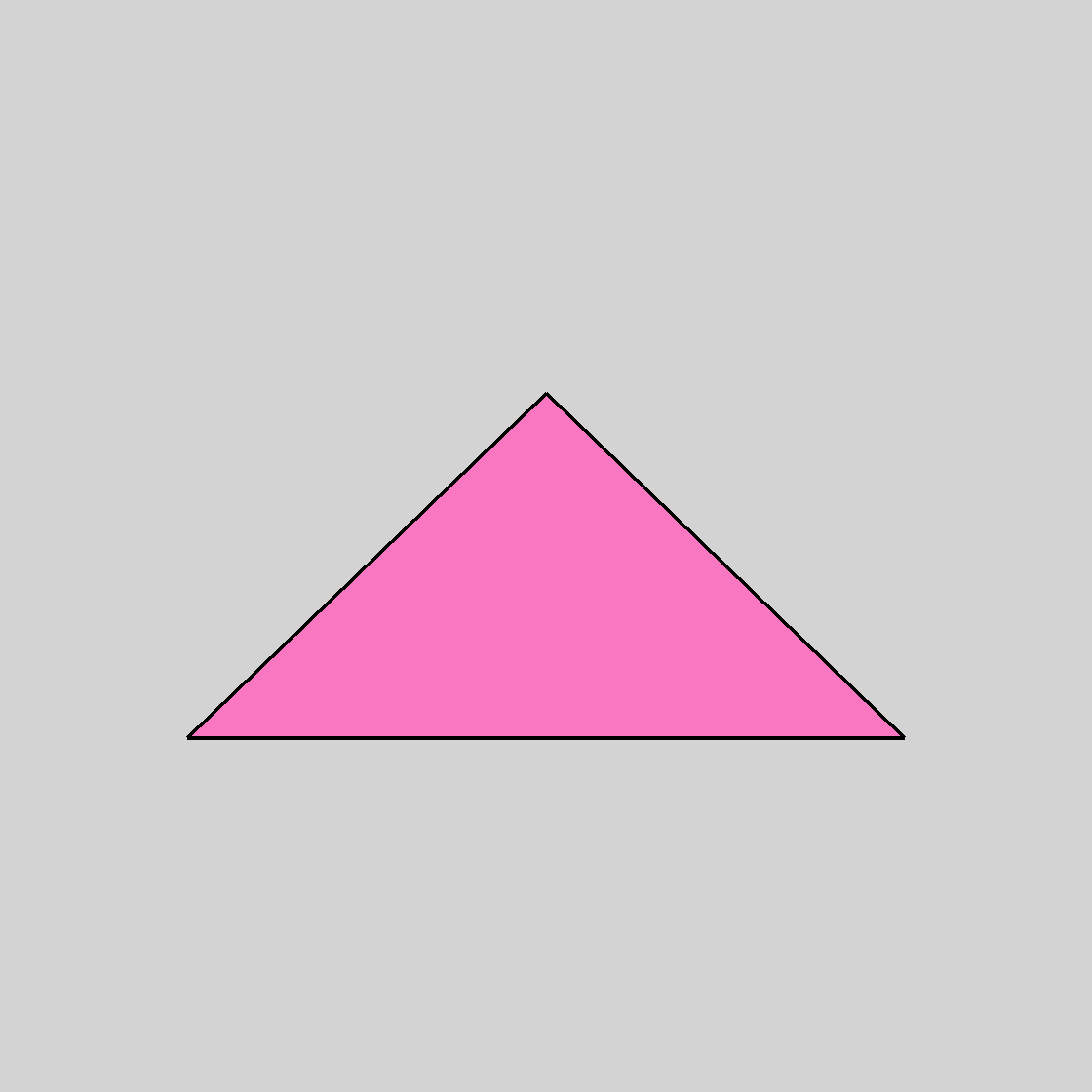}};

\node[img] (d4) at (3\catxgap, \catygap)
{\secondrowfigcat{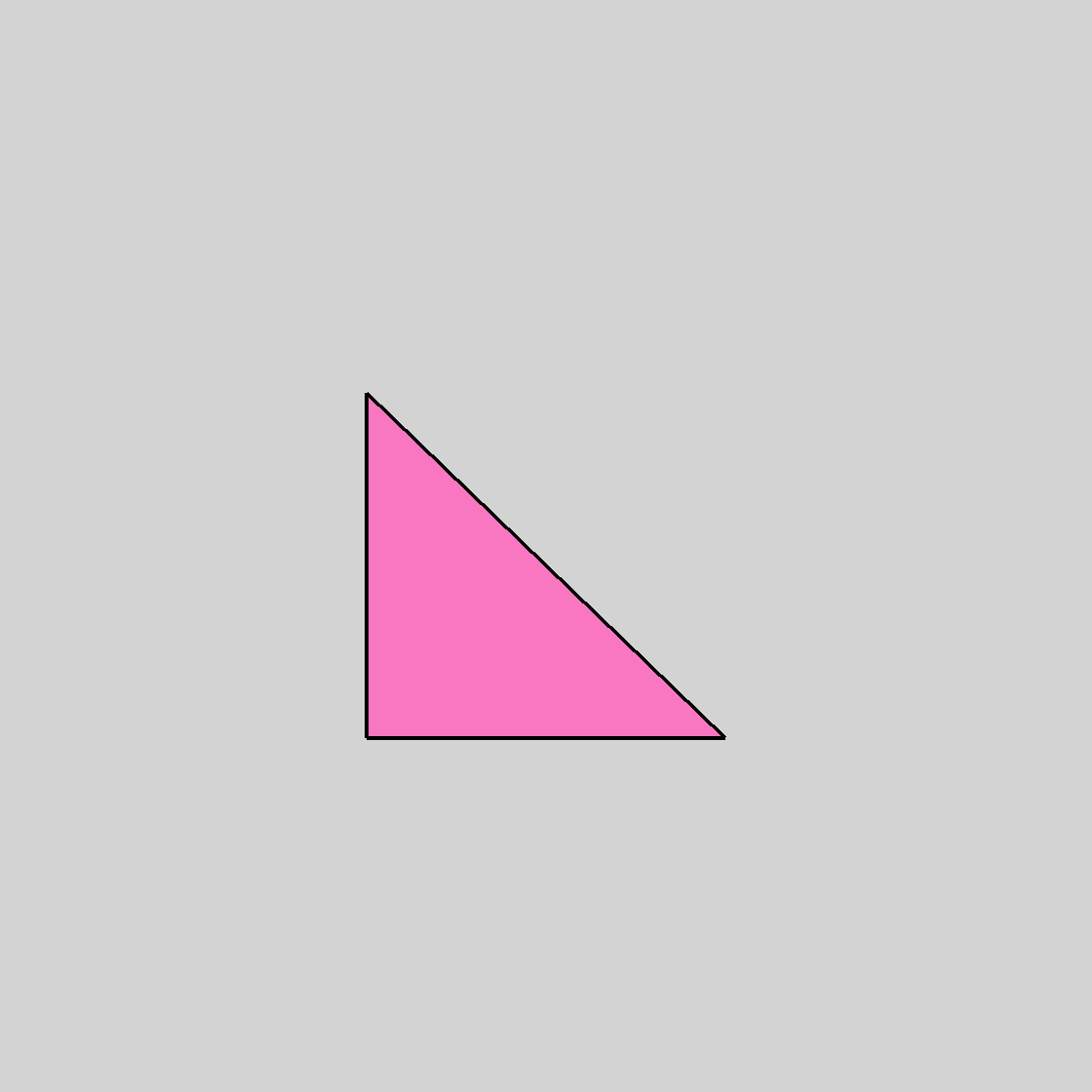}};

\node[img] (d5) at (4\catxgap, \catygap)
{\secondrowfigcat{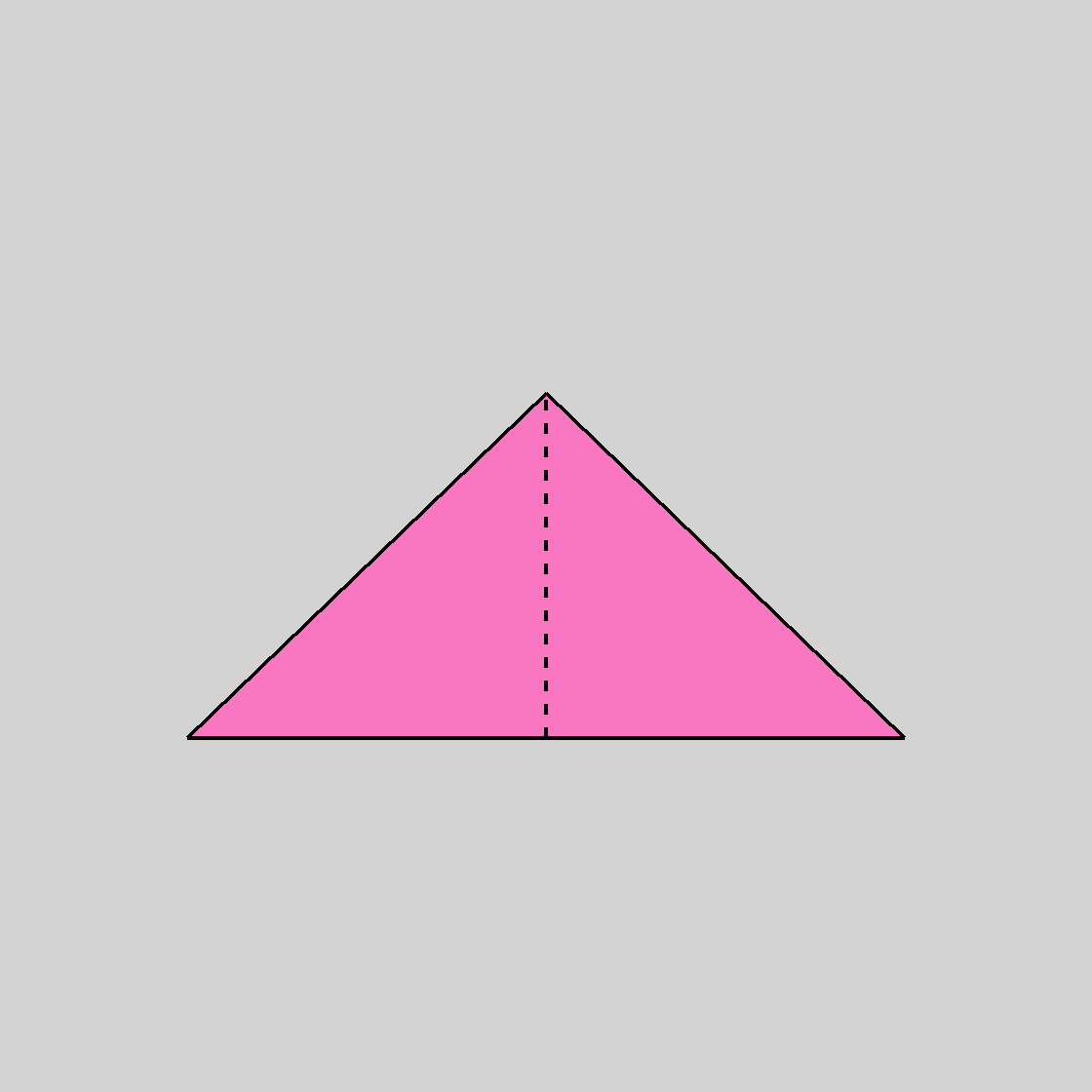}};

\node[img] (d6) at (5\catxgap, \catygap)
{\secondrowfigcat{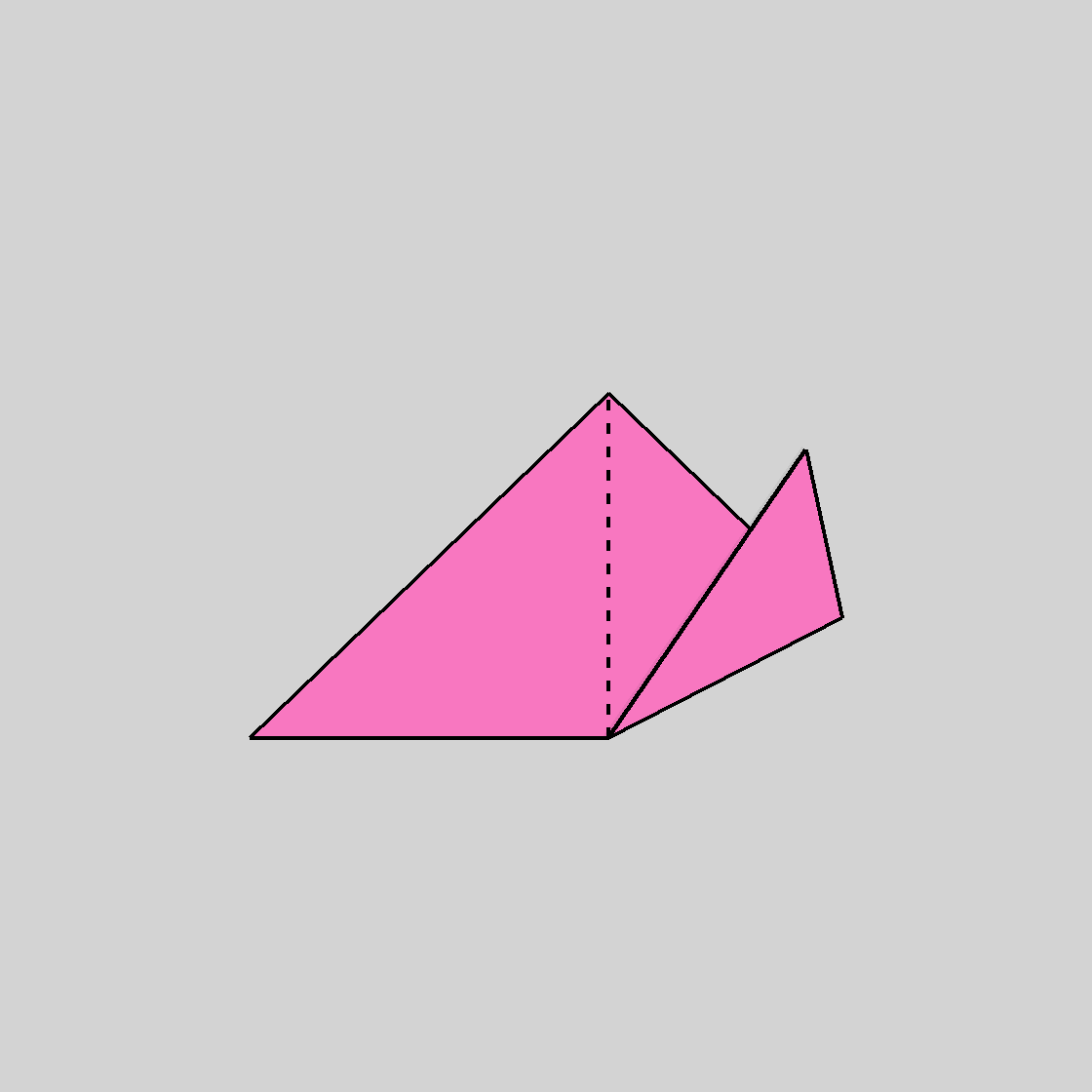}};

\node[img] (d7) at (6\catxgap, \catygap)
{\secondrowfigcat{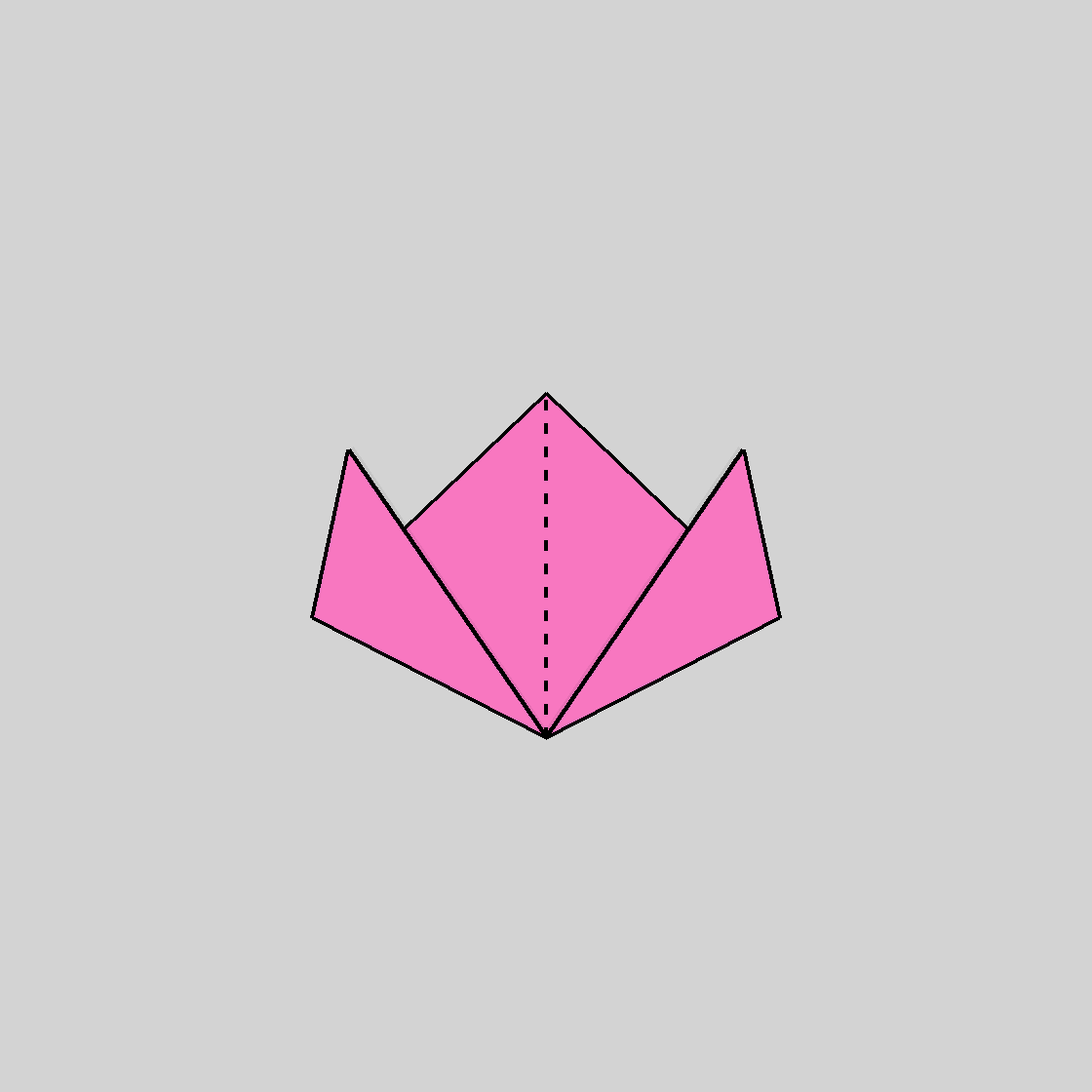}};

\node[img] (d8) at (7\catxgap, \catygap)
{\secondrowfigcat{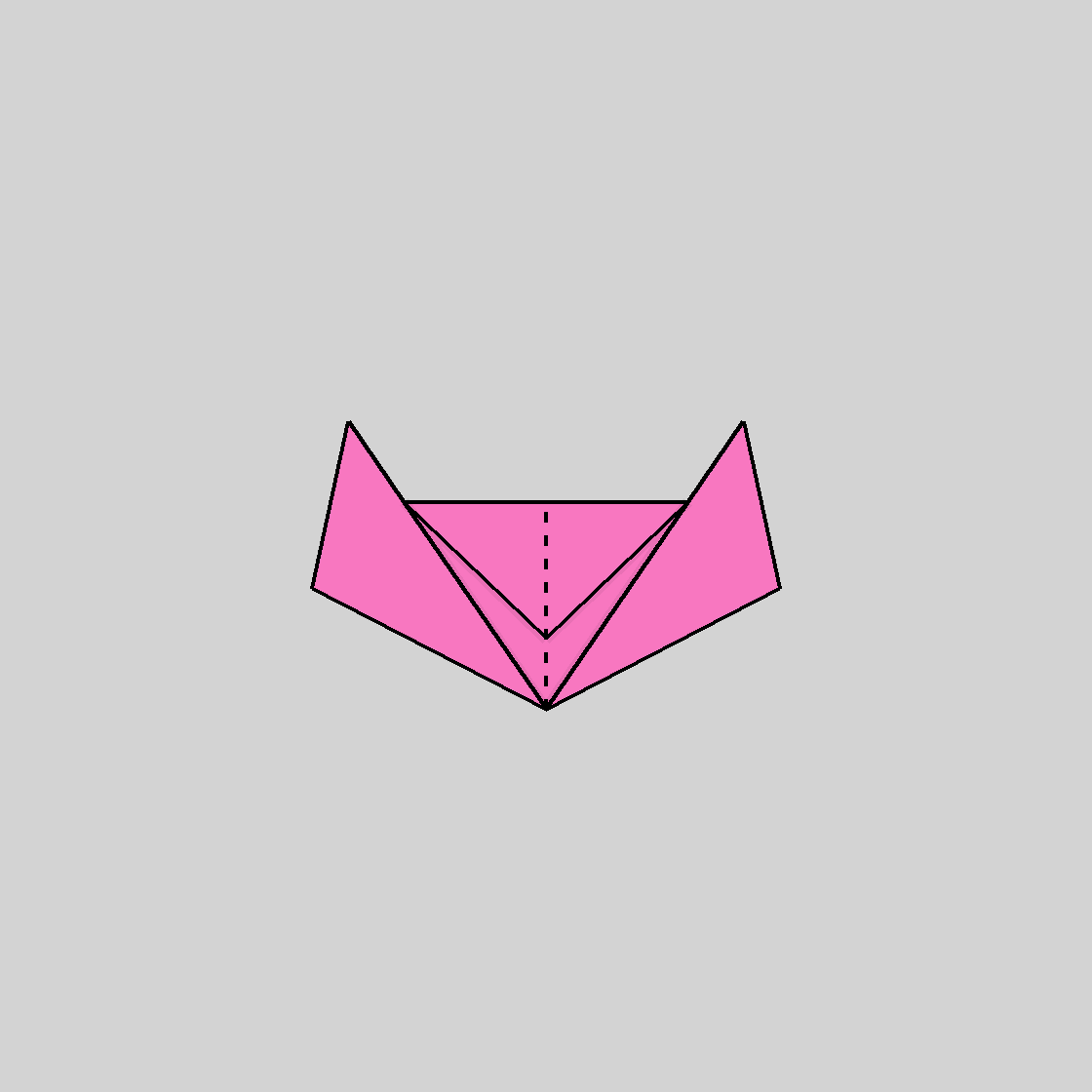}};

\node[img] (d9) at (8\catxgap, \catygap)
{\secondrowfigcat{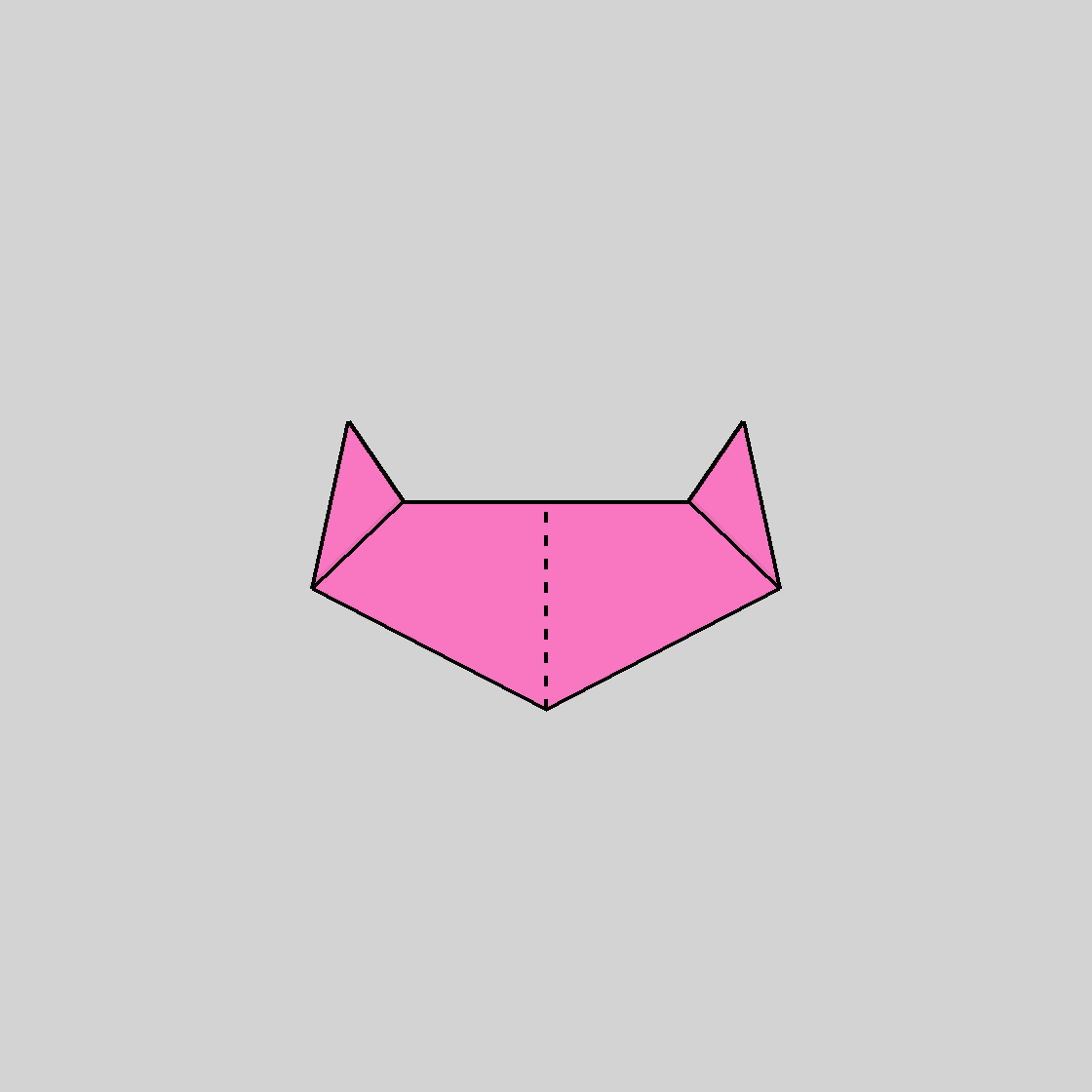}};

\node[cmd] at ($(d1)!0.5!(d2) + (0,-0.77)$) {
rotate(45$^\circ$)
};

\node[cmd] at ($(d2)!0.5!(d3) + (0,-0.77)$) {
fold([4,2],-1)
};

\node[cmd] at ($(d3)!0.5!(d4) + (0,-0.68)$) {
add\_v(0.5, [4,2])\\
fold([5,1], -1)
};

\node[cmd] at ($(d4)!0.5!(d5) + (0,-0.77)$) {
unfold()
};

\node[cmd] at ($(d5)!0.5!(d6) + (0,-0.68)$) {
add\_v(0.65, [1,2])\\
fold([5,6], -1)
};

\node[cmd] at ($(d6)!0.5!(d7) + (0,-0.68)$) {
add\_v(0.35,[4,1])\\
fold([5,8],-1)
};

\node[cmd] at ($(d7)!0.5!(d8) + (0,-0.6)$) {
add\_v(0.61,[1,8])\\
add\_v(0.61,[1,6])\\
fold([10,11],1)
};

\node[cmd] at ($(d8)!0.5!(d9) + (0,-0.77)$) {
flip(x)
};

\end{tikzpicture}

\vspace{10pt}

\setlength{\catimgw}{0.099\textwidth}

\setlength{\catxgap}{0.1\textwidth}

\setlength{\catygap}{-0.07\textwidth}

\begin{tikzpicture}[
    img/.style={
        inner sep=0pt,
        outer sep=0pt
    },
    cmd/.style={
        fill=white,
        fill opacity=0.90,
        text opacity=1,
        draw=black!30,
        rounded corners=1.5pt,
        inner sep=1pt,
        font=\tiny\ttfamily,
        align=left
    },
    frameid/.style={
        fill=white,
        fill opacity=0,
        text opacity=1,
        text=white,
        draw=black!25,
        rounded corners=0pt,
        inner sep=1pt,
        font=\tiny\ttfamily,
        anchor=north west
    }
]

\node[img] (c1) at (0\catxgap, 0)
{\includegraphics[width=\catimgw]{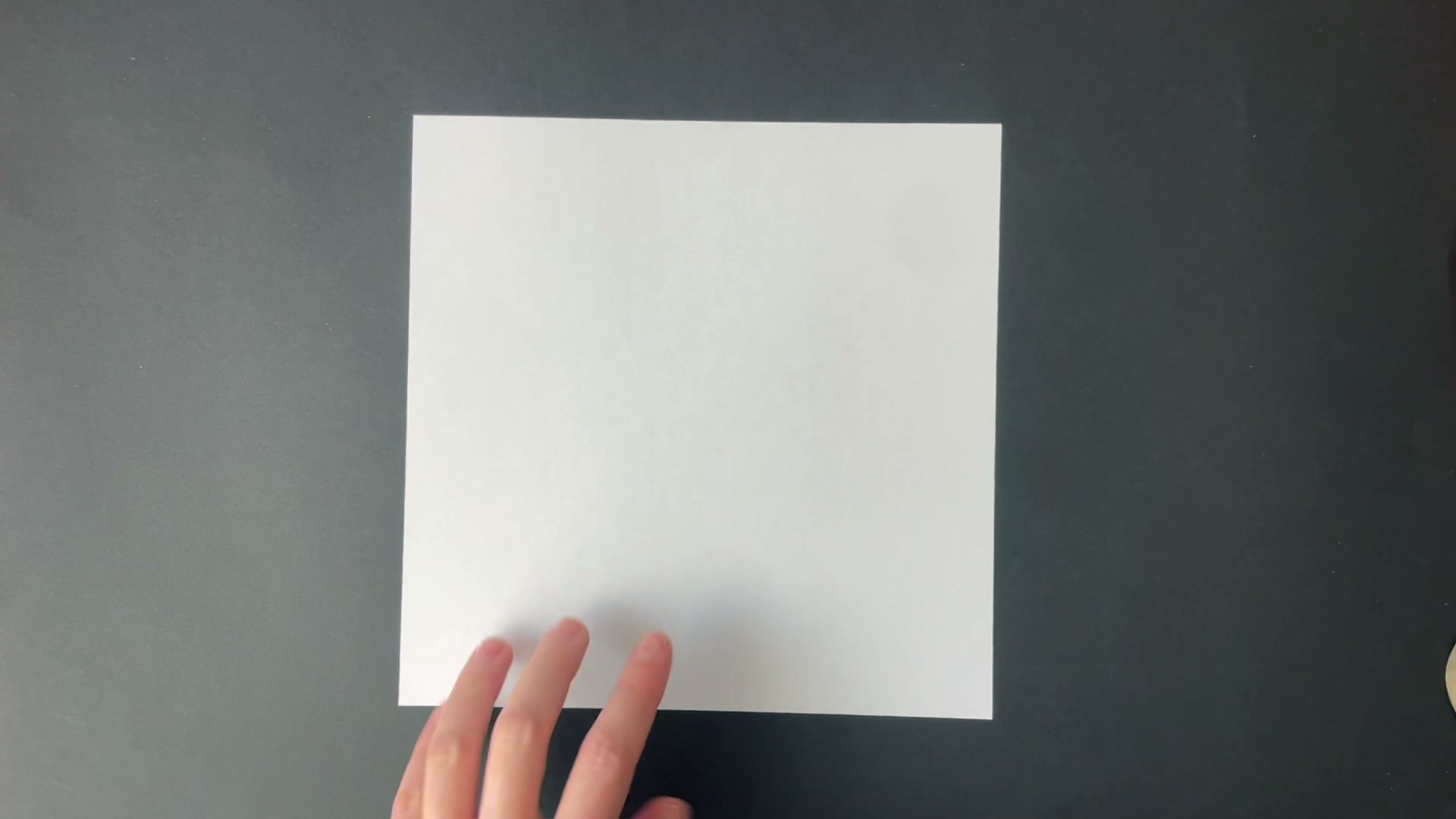}};

\node[img] (c2) at (1\catxgap, 0)
{\includegraphics[width=\catimgw]{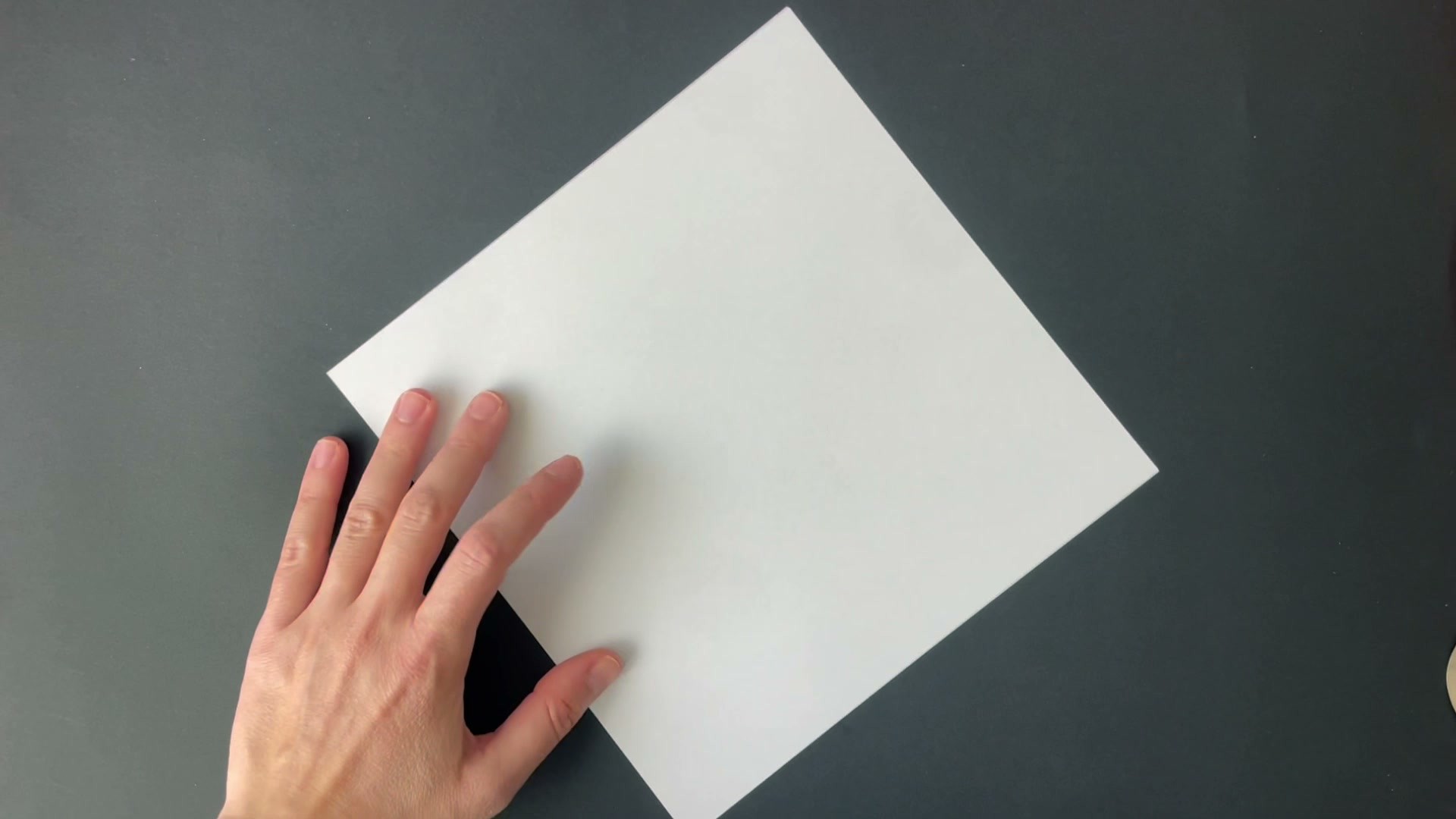}};

\node[img] (c3) at (2\catxgap, 0)
{\includegraphics[width=\catimgw]{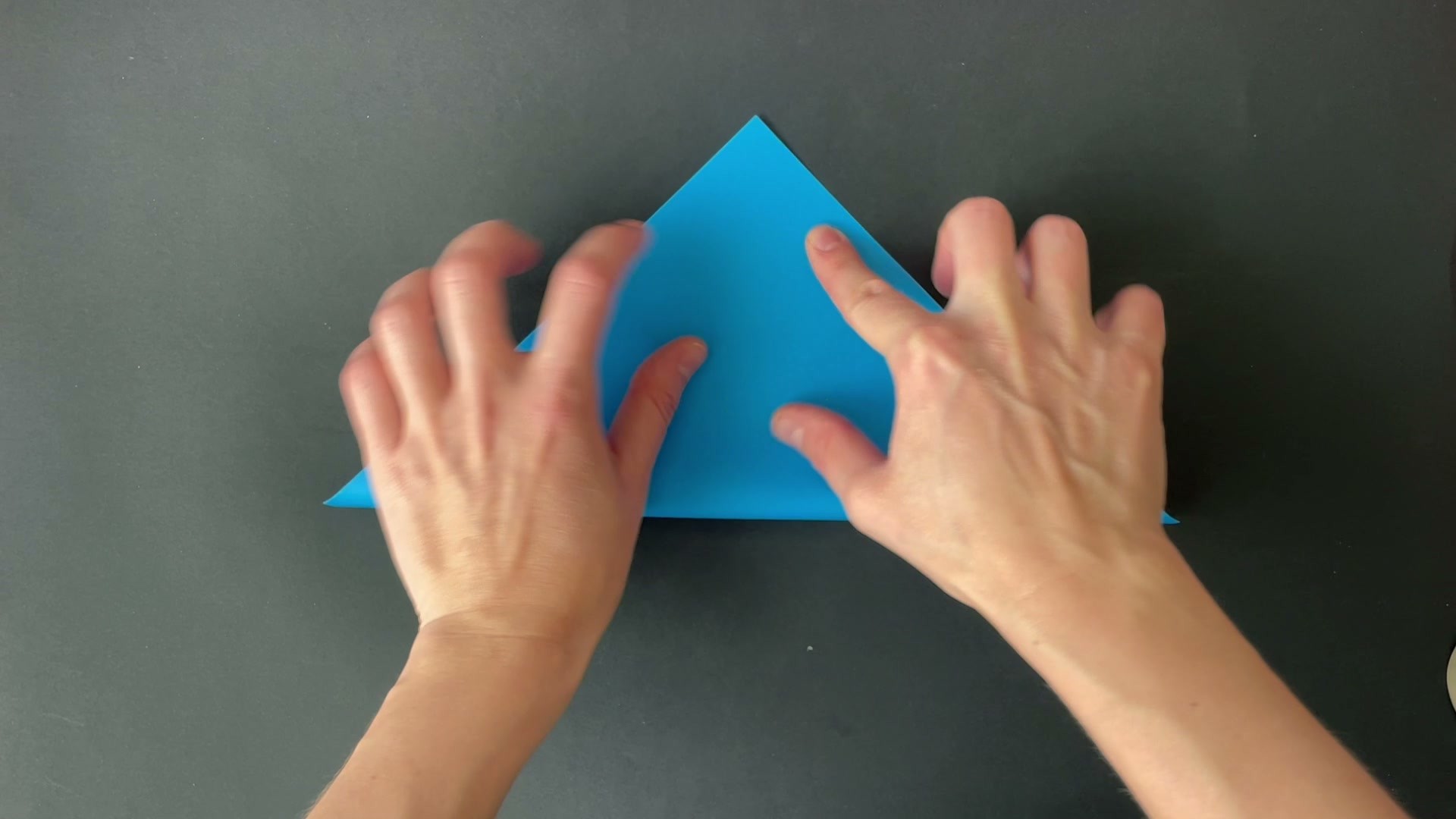}};

\node[img] (c4) at (3\catxgap, 0)
{\includegraphics[width=\catimgw]{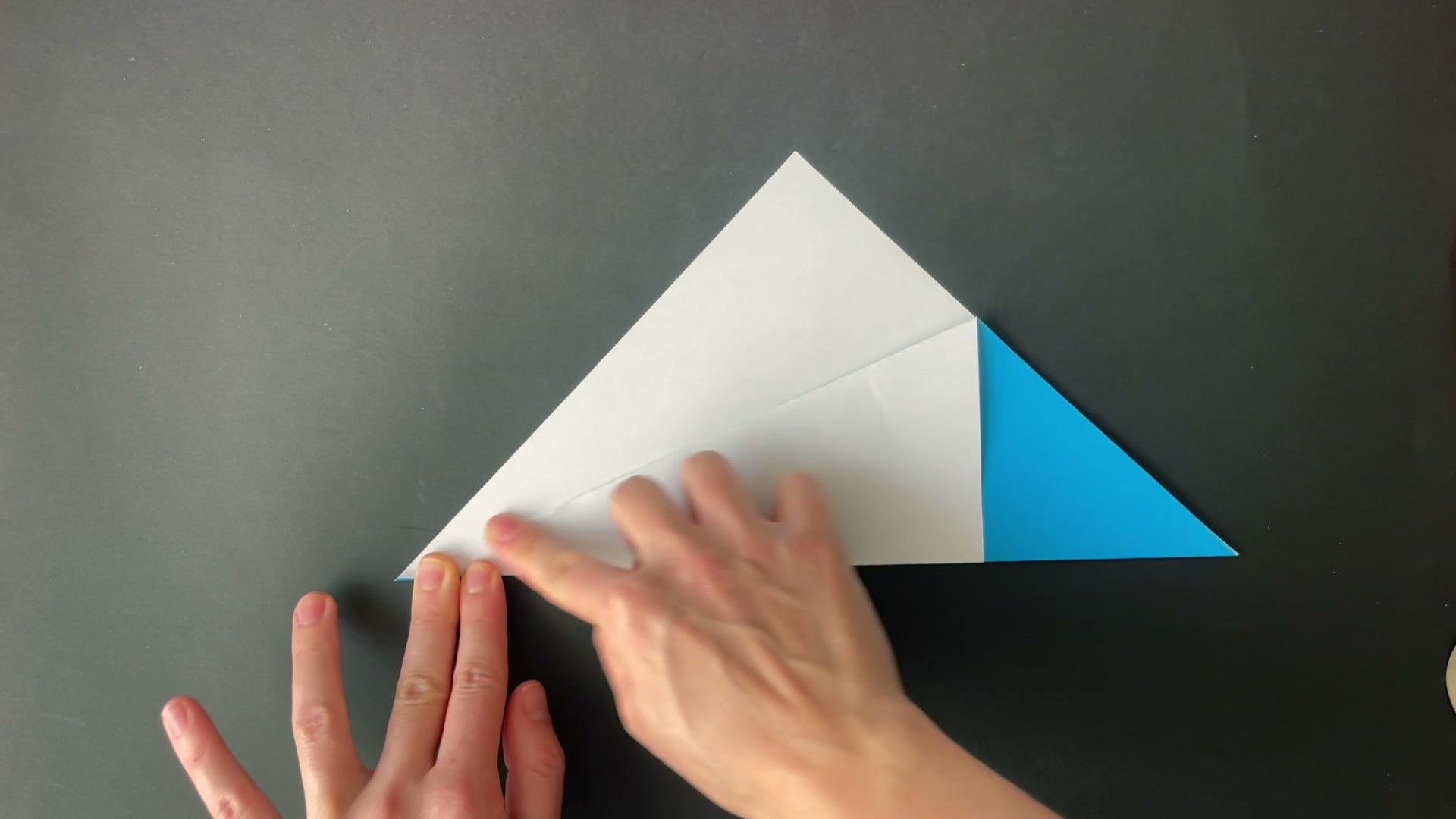}};

\node[img] (c5) at (4\catxgap, 0)
{\includegraphics[width=\catimgw]{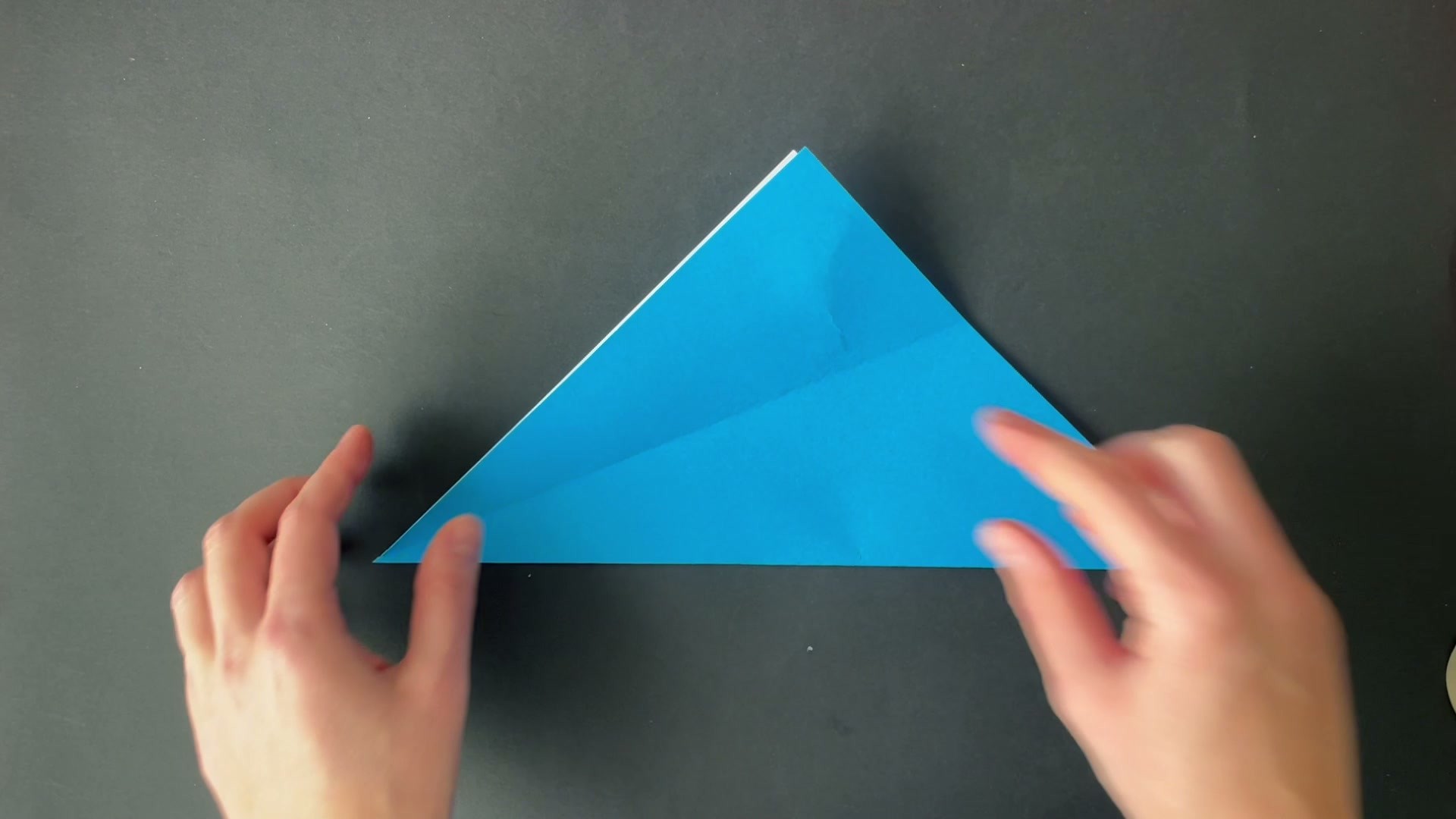}};

\node[img] (c6) at (5\catxgap, 0)
{\includegraphics[width=\catimgw]{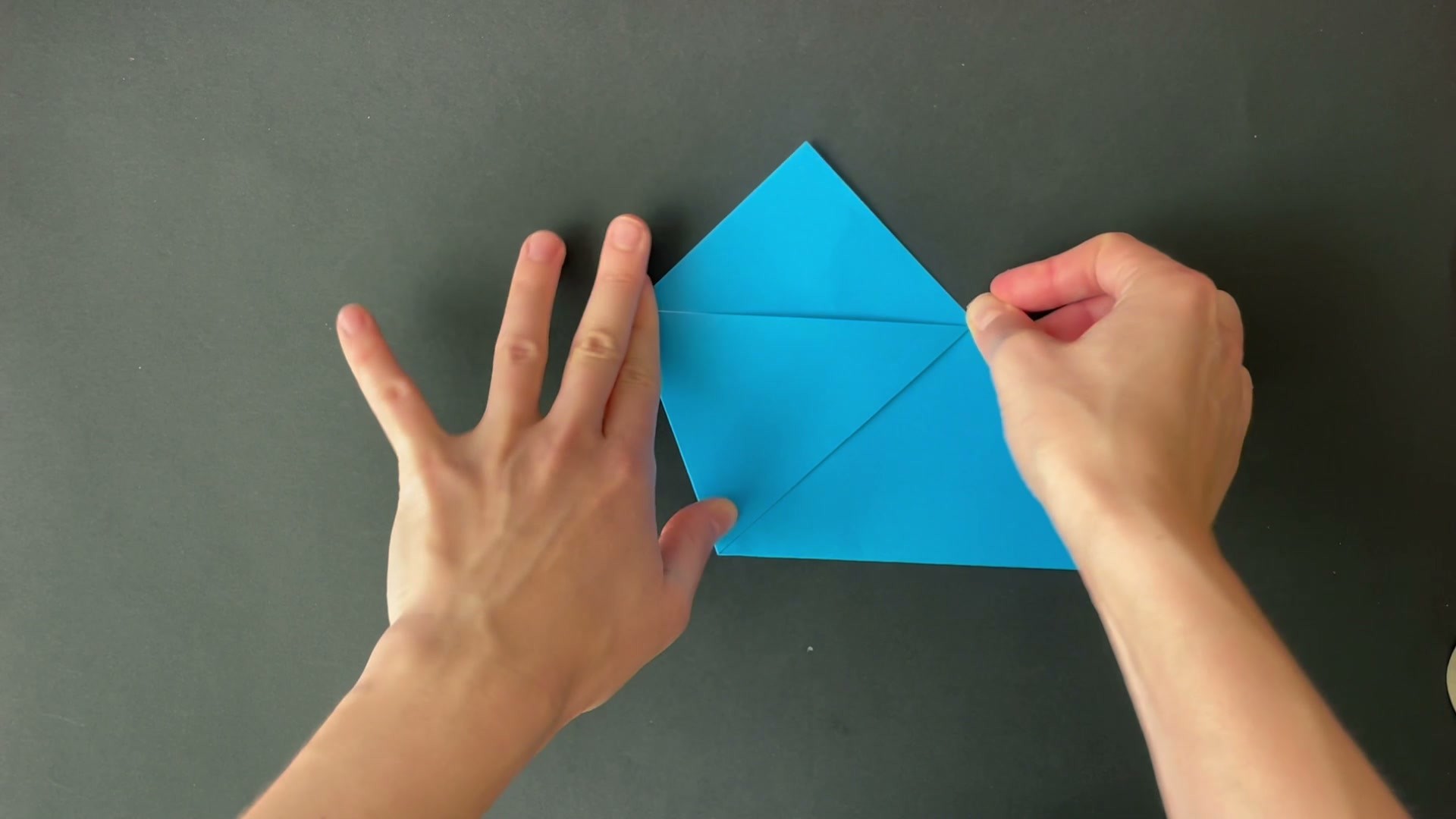}};

\node[img] (c7) at (6\catxgap, 0)
{\includegraphics[width=\catimgw]{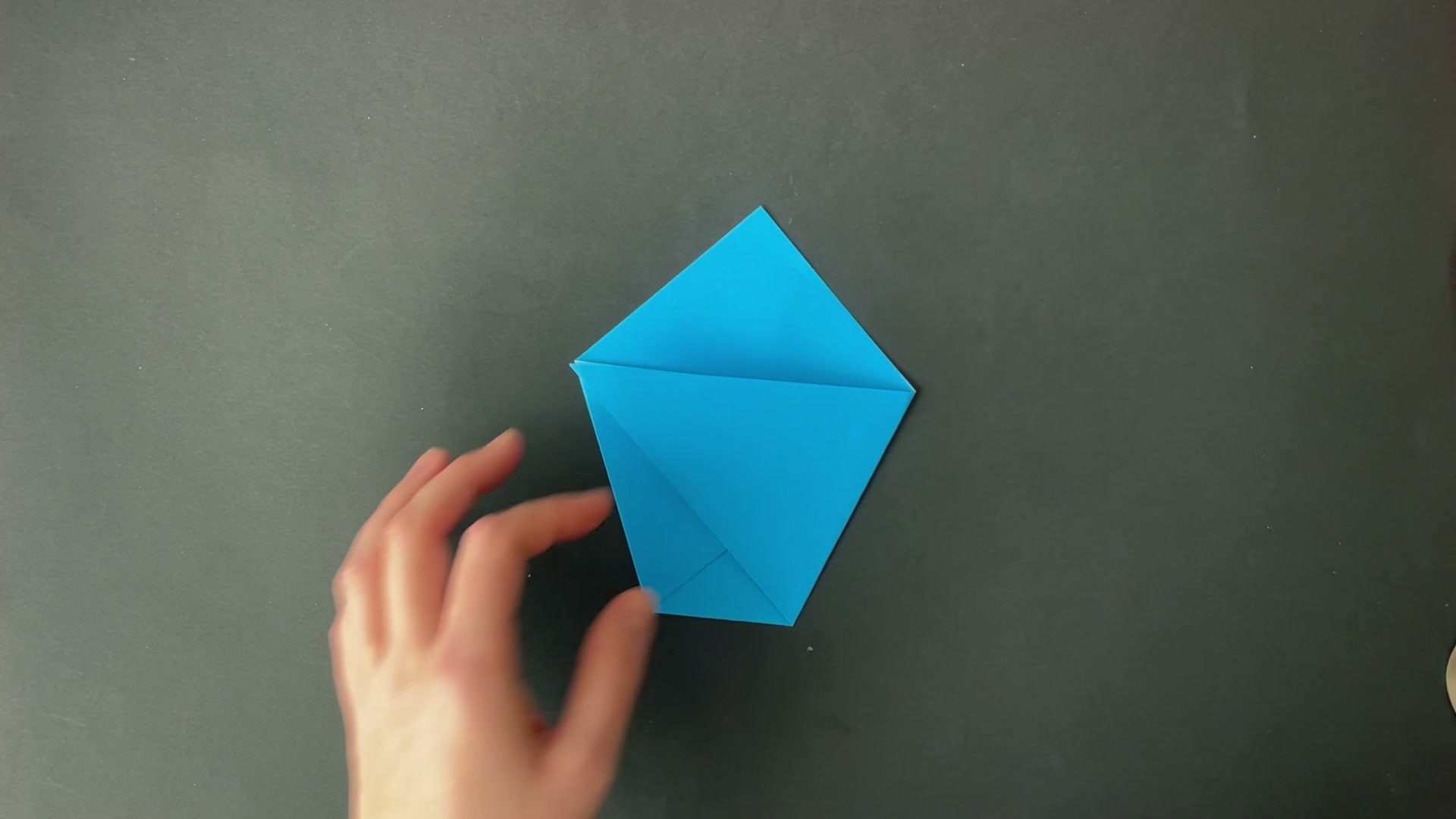}};

\node[img] (c8) at (7\catxgap, 0)
{\includegraphics[width=\catimgw]{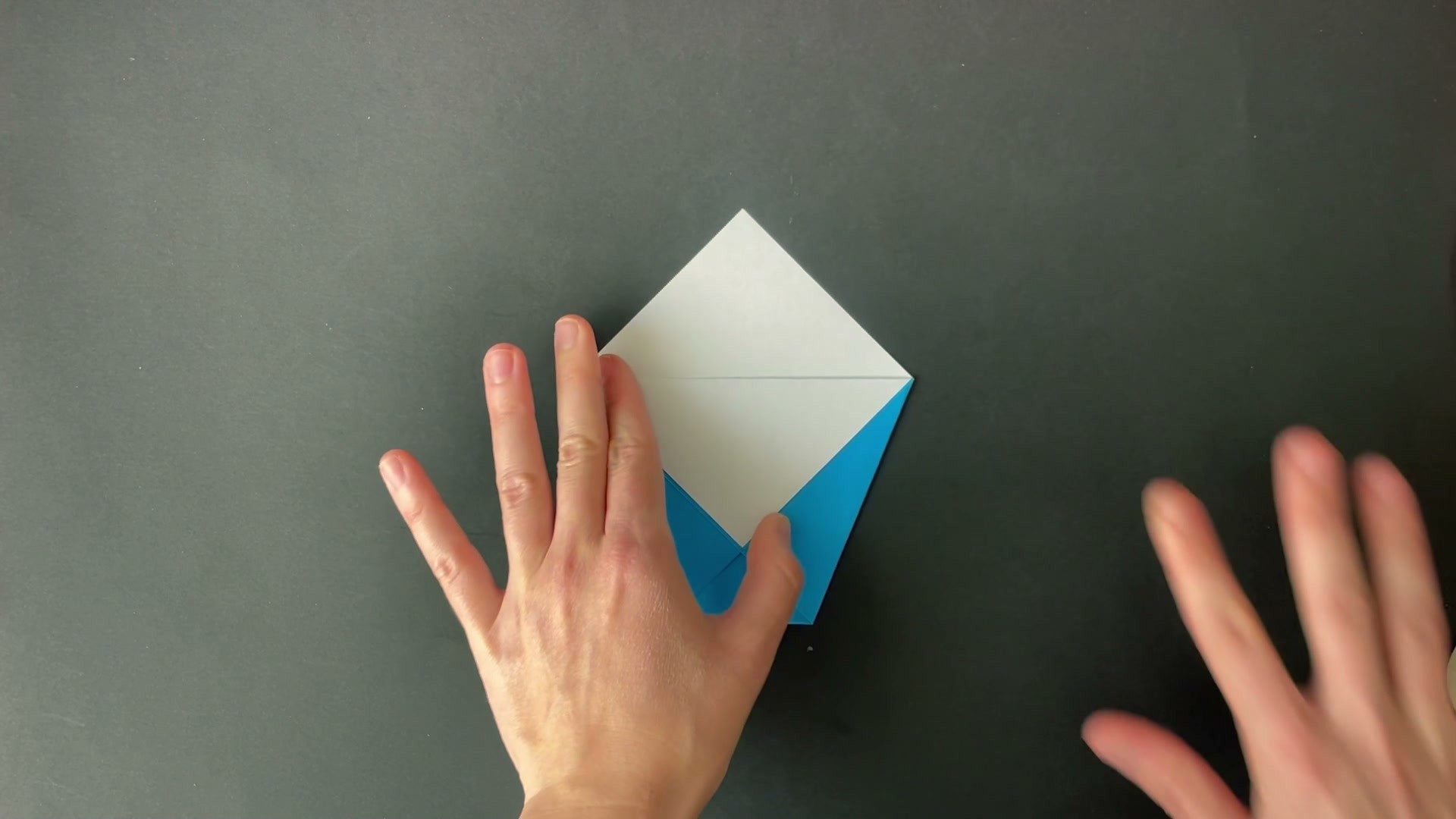}};

\node[img] (c9) at (8\catxgap, 0)
{\includegraphics[width=\catimgw]{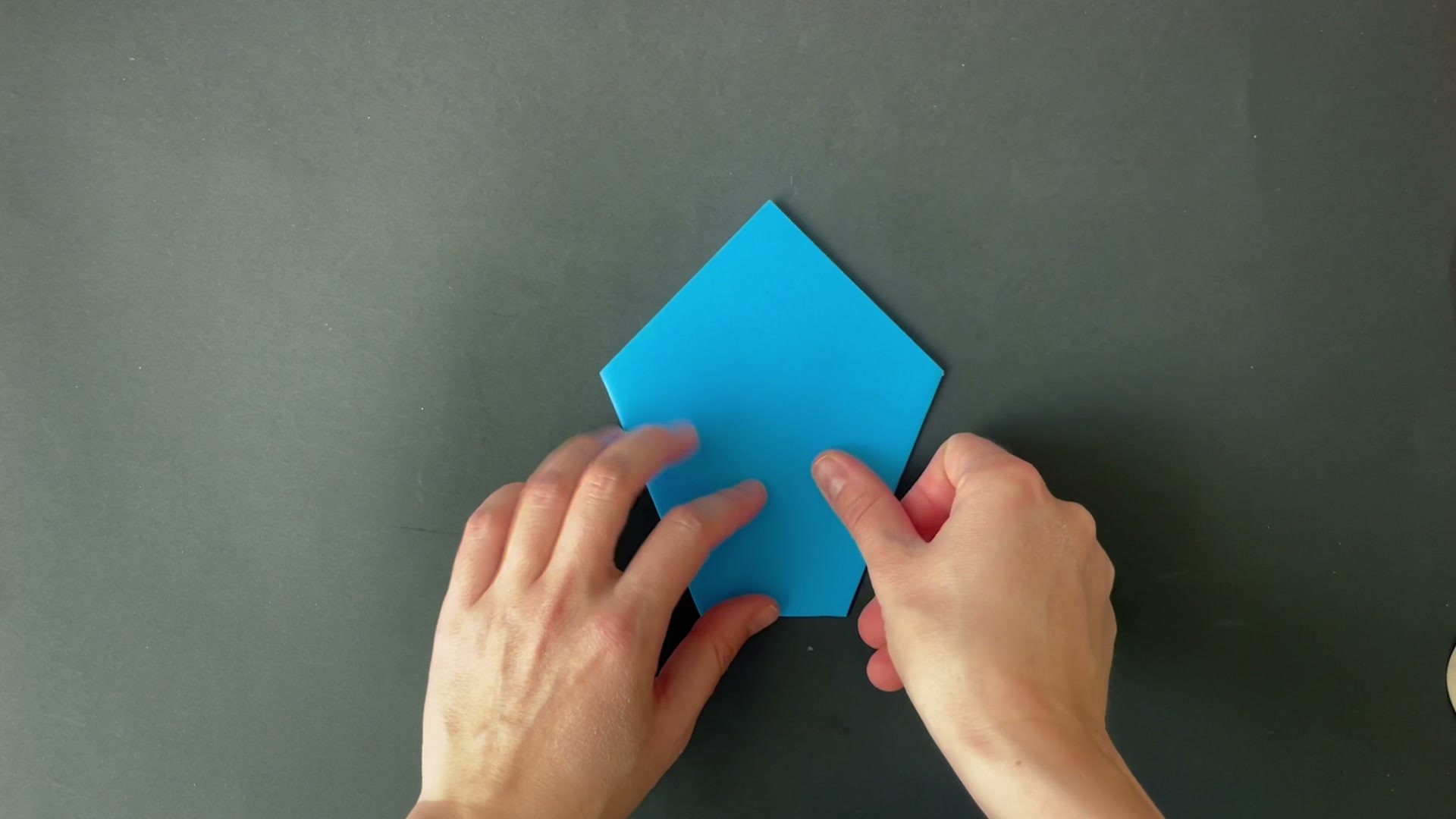}};

\node[img] (c10) at (9\catxgap, 0)
{\includegraphics[width=\catimgw]{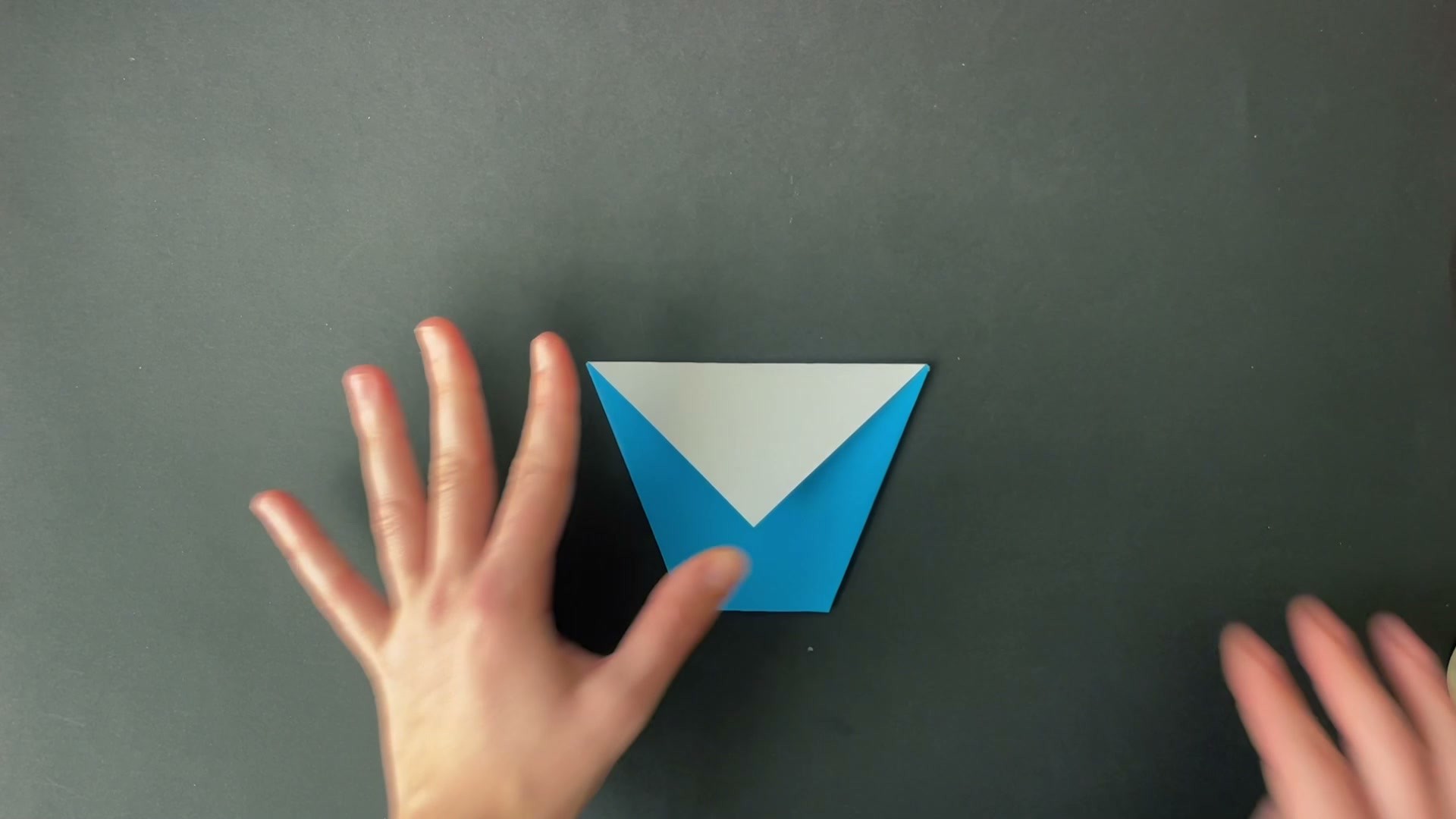}};

\node[frameid] at ([xshift=1pt,yshift=-1pt]c1.north west) {1};
\node[frameid] at ([xshift=1pt,yshift=-1pt]c2.north west) {2};
\node[frameid] at ([xshift=1pt,yshift=-1pt]c3.north west) {3};
\node[frameid] at ([xshift=1pt,yshift=-1pt]c4.north west) {4};
\node[frameid] at ([xshift=1pt,yshift=-1pt]c5.north west) {5};
\node[frameid] at ([xshift=1pt,yshift=-1pt]c6.north west) {6};
\node[frameid] at ([xshift=1pt,yshift=-1pt]c7.north west) {7};
\node[frameid] at ([xshift=1pt,yshift=-1pt]c8.north west) {8};
\node[frameid] at ([xshift=1pt,yshift=-1pt]c9.north west) {9};
\node[img] (d1) at (0\catxgap, \catygap)
{\secondrowfigcat{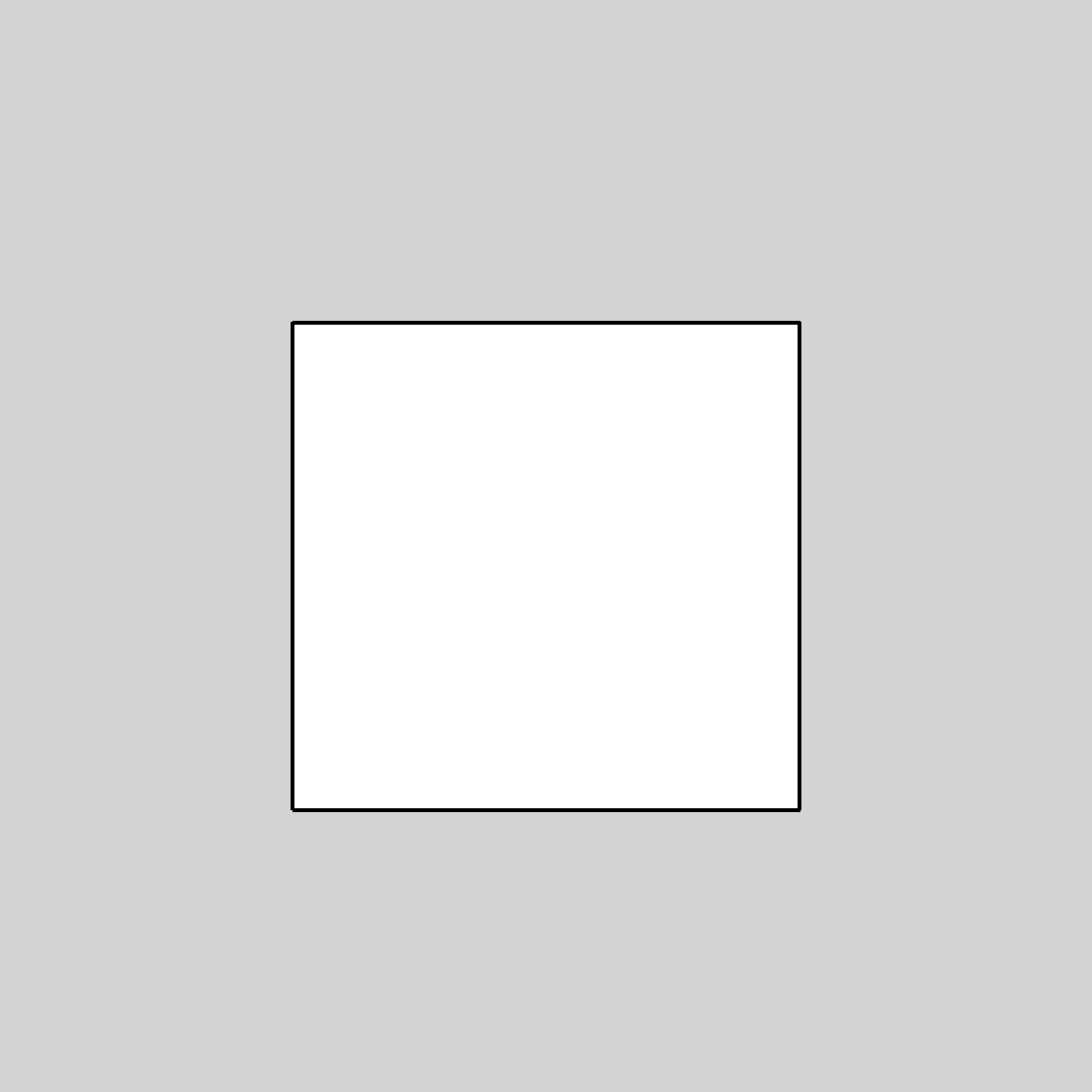}};

\node[img] (d2) at (1\catxgap, \catygap)
{\secondrowfigcat{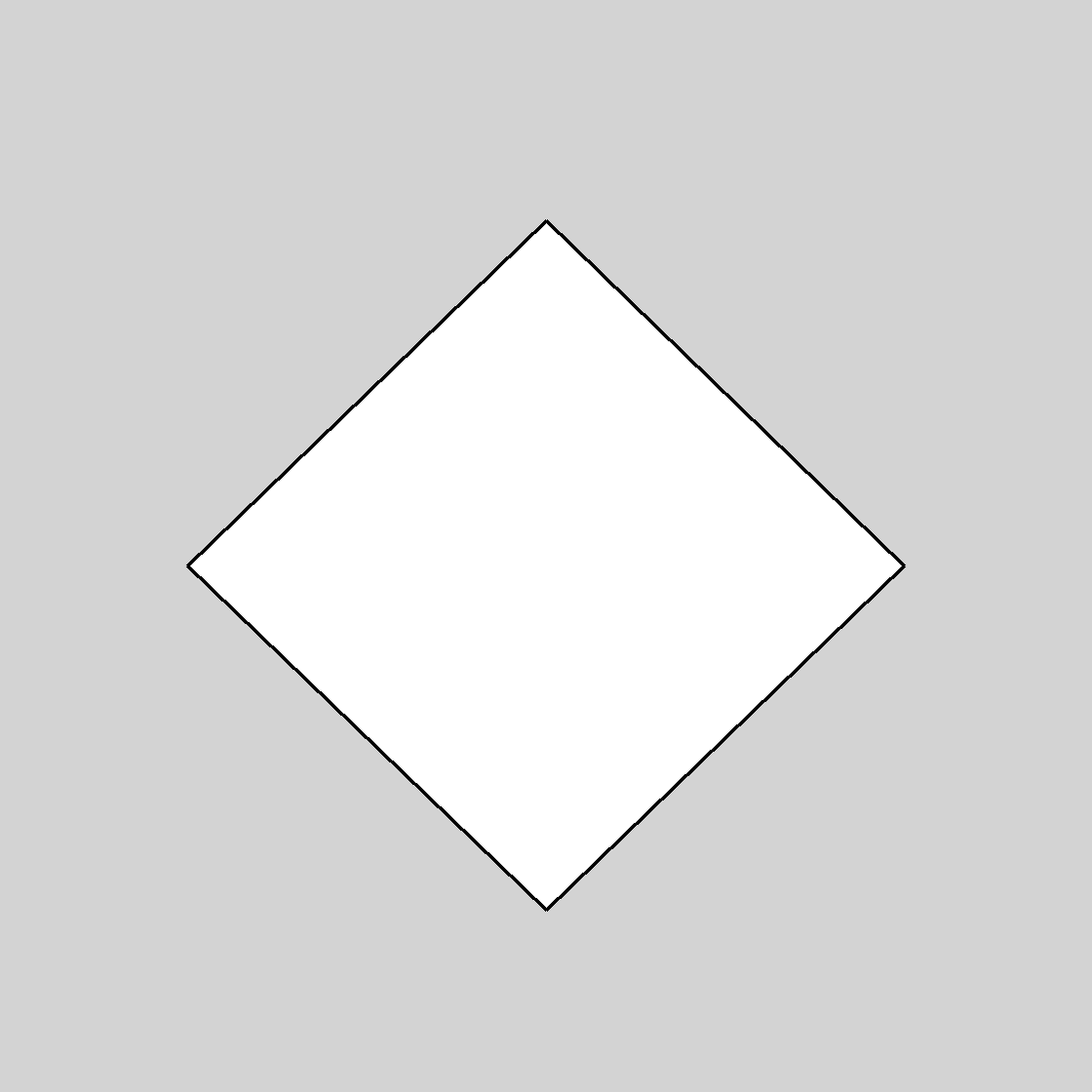}};

\node[img] (d3) at (2\catxgap, \catygap)
{\secondrowfigcat{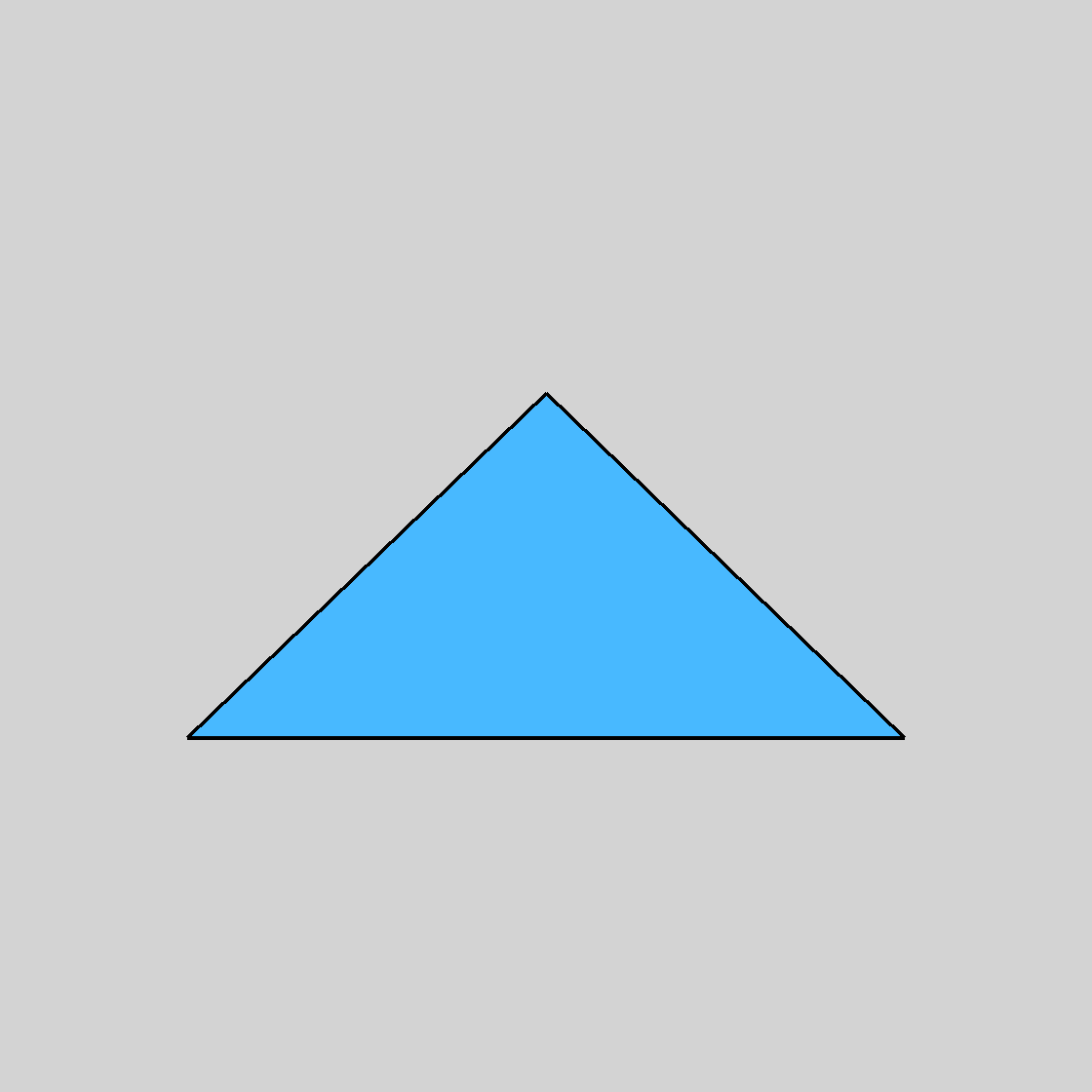}};

\node[img] (d4) at (3\catxgap, \catygap)
{\secondrowfigcat{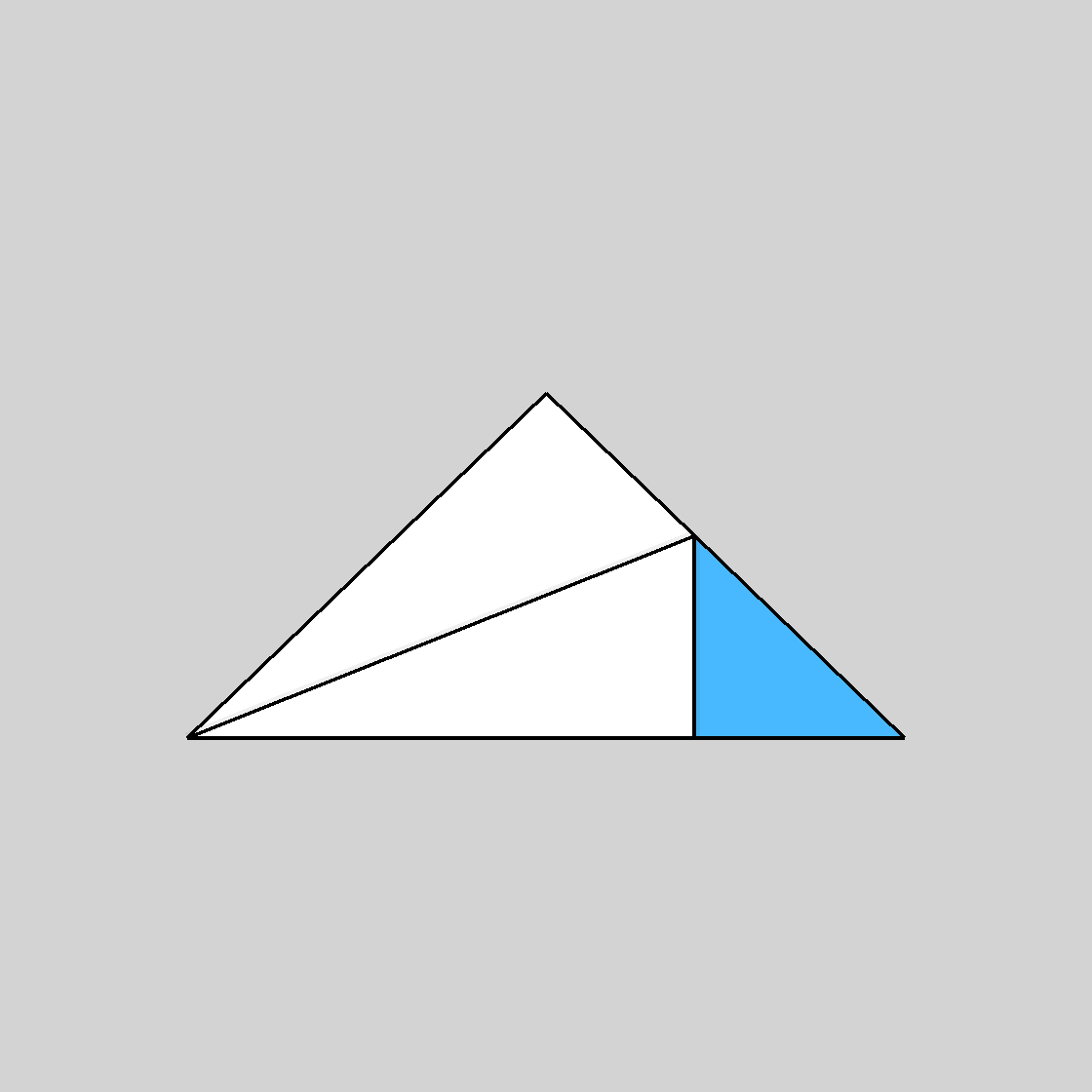}};

\node[img] (d5) at (4\catxgap, \catygap)
{\secondrowfigcat{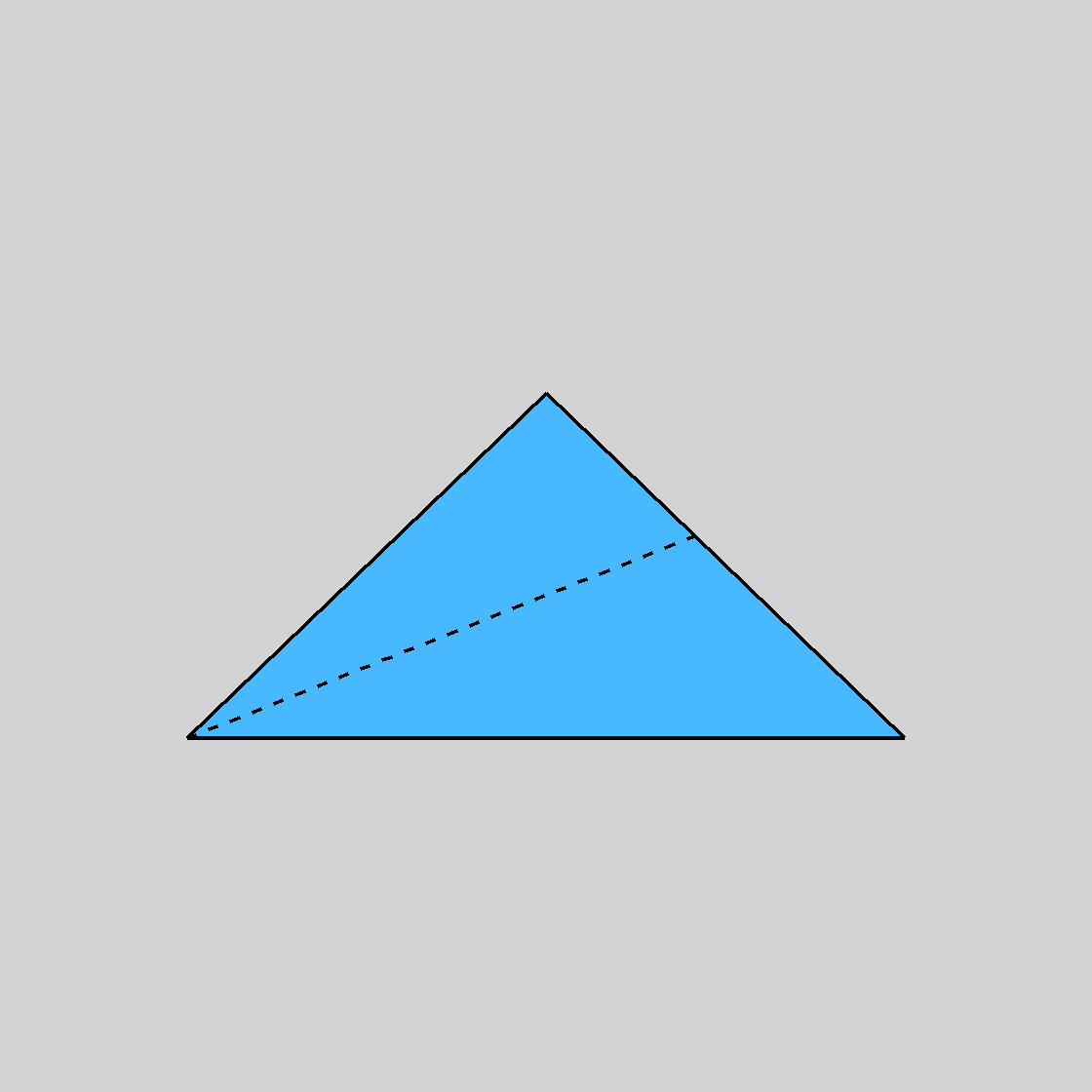}};

\node[img] (d6) at (5\catxgap, \catygap)
{\secondrowfigcat{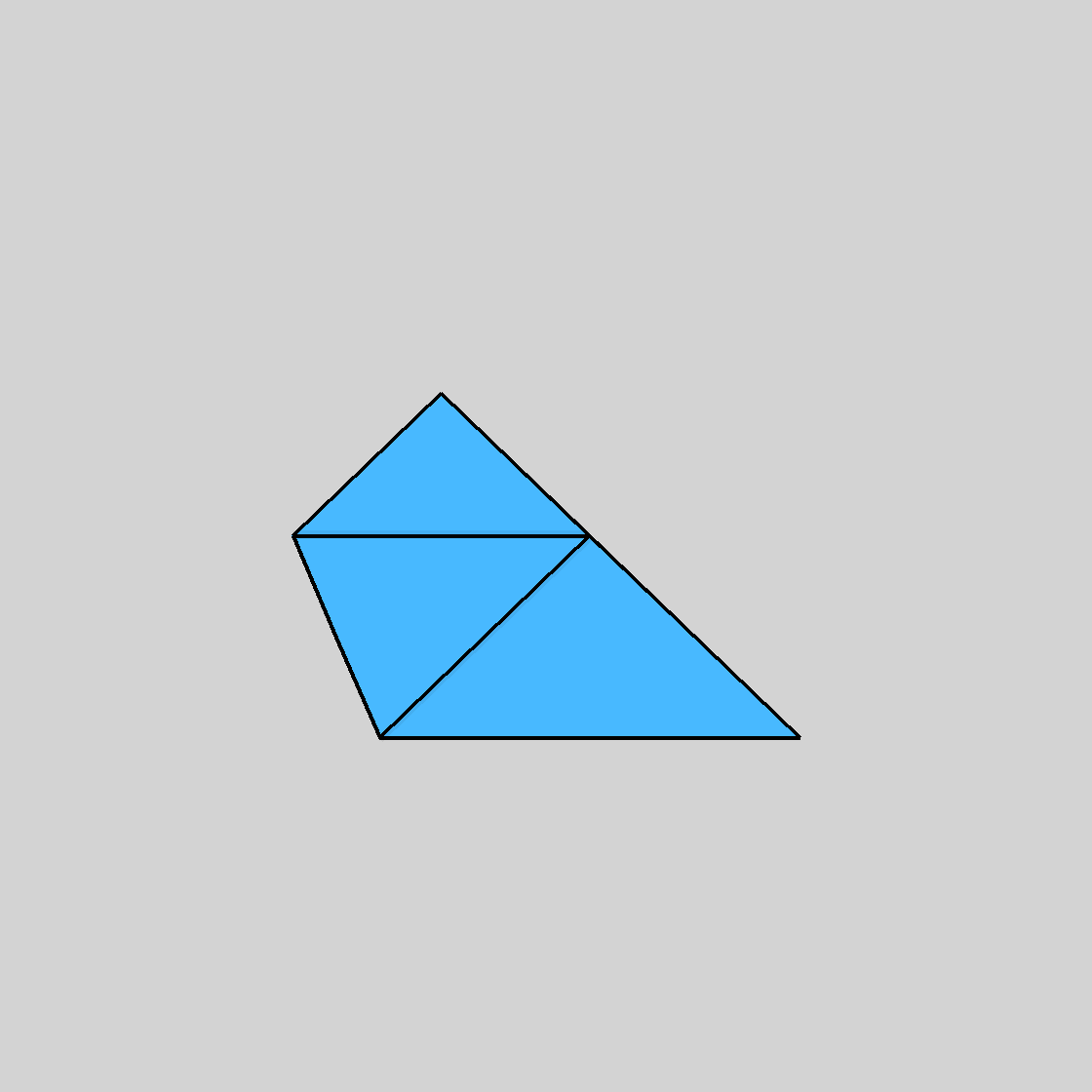}};

\node[img] (d7) at (6\catxgap, \catygap)
{\secondrowfigcat{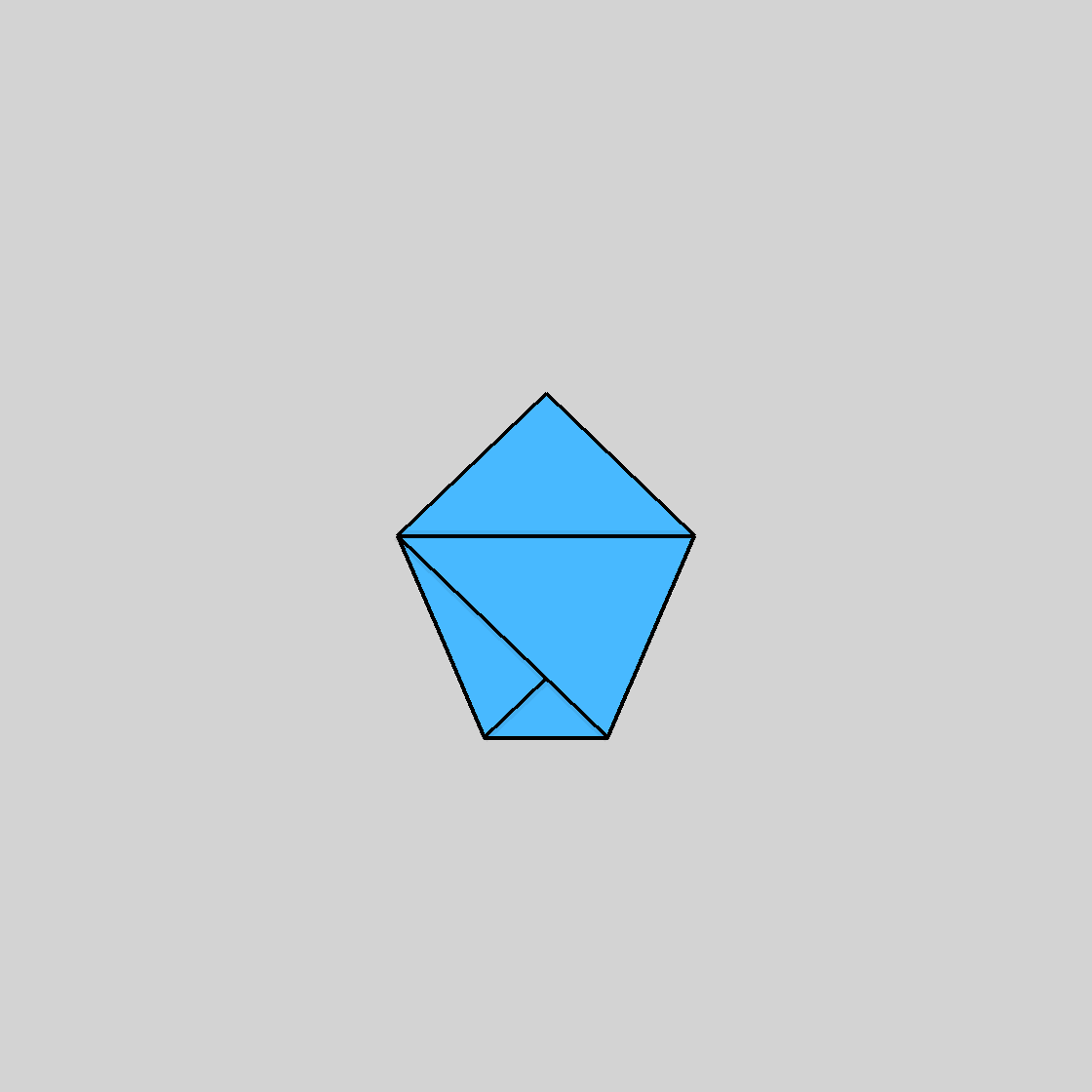}};

\node[img] (d8) at (7\catxgap, \catygap)
{\secondrowfigcat{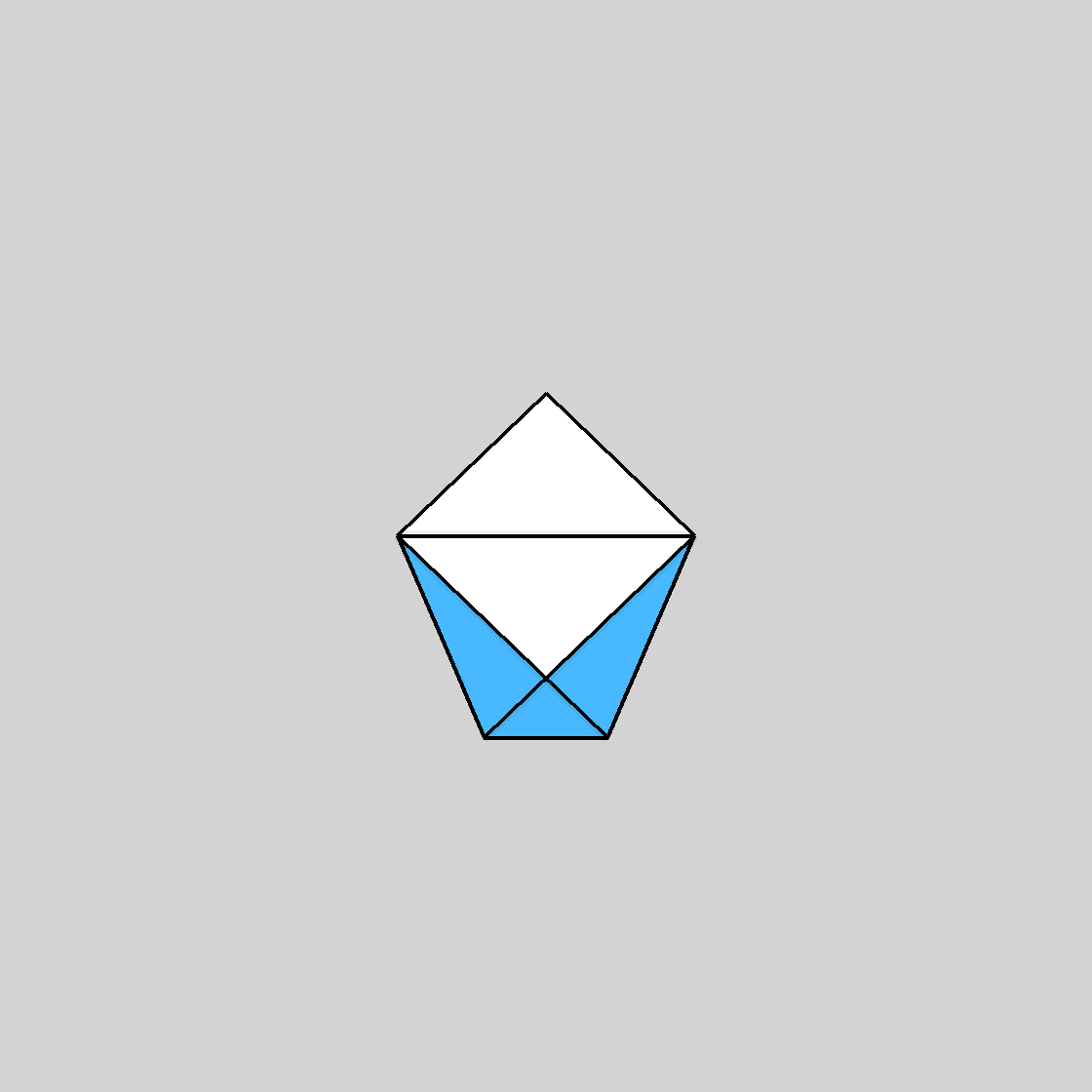}};

\node[img] (d9) at (8\catxgap, \catygap)
{\secondrowfigcat{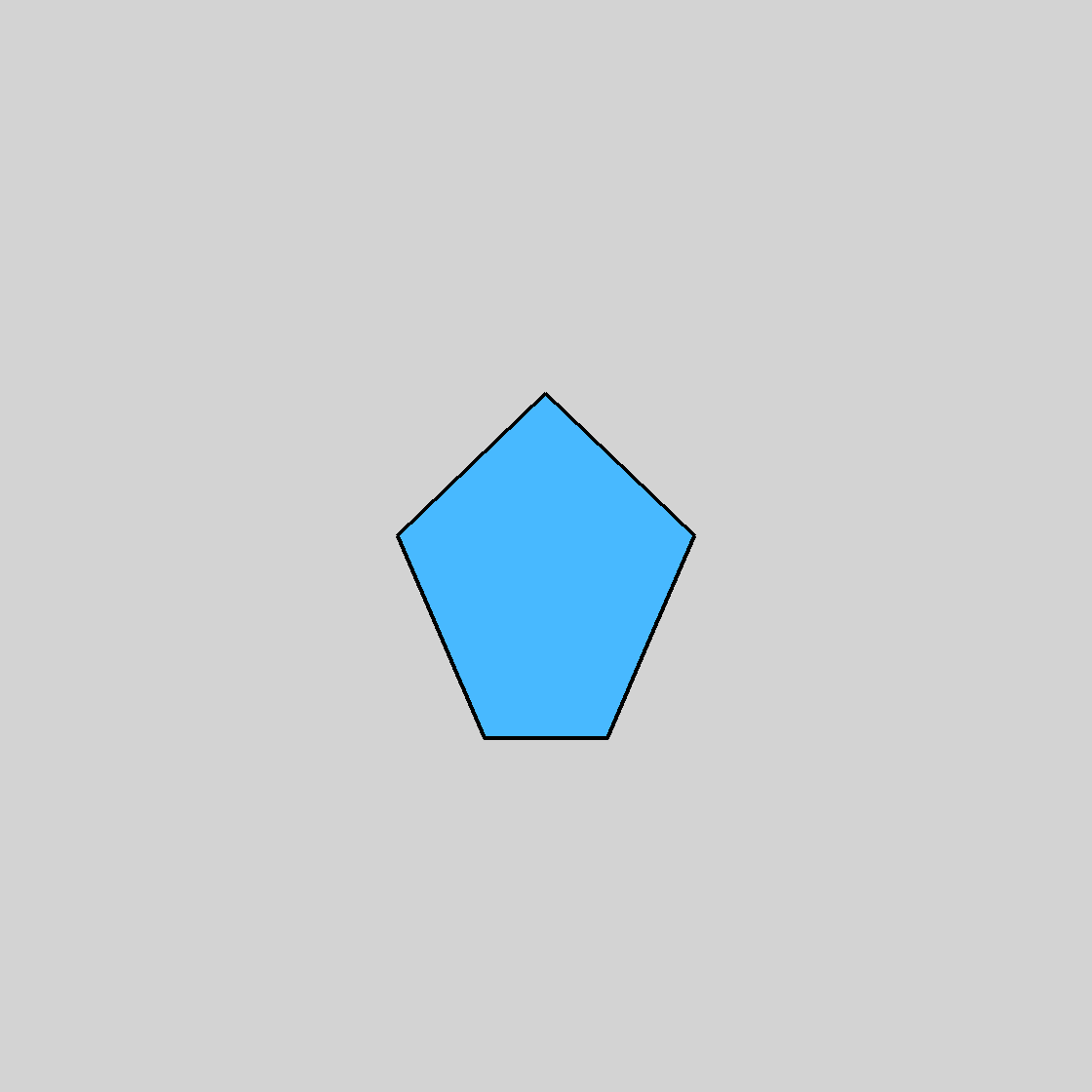}};

\node[img] (d10) at (9\catxgap, \catygap)
{\secondrowfigcat{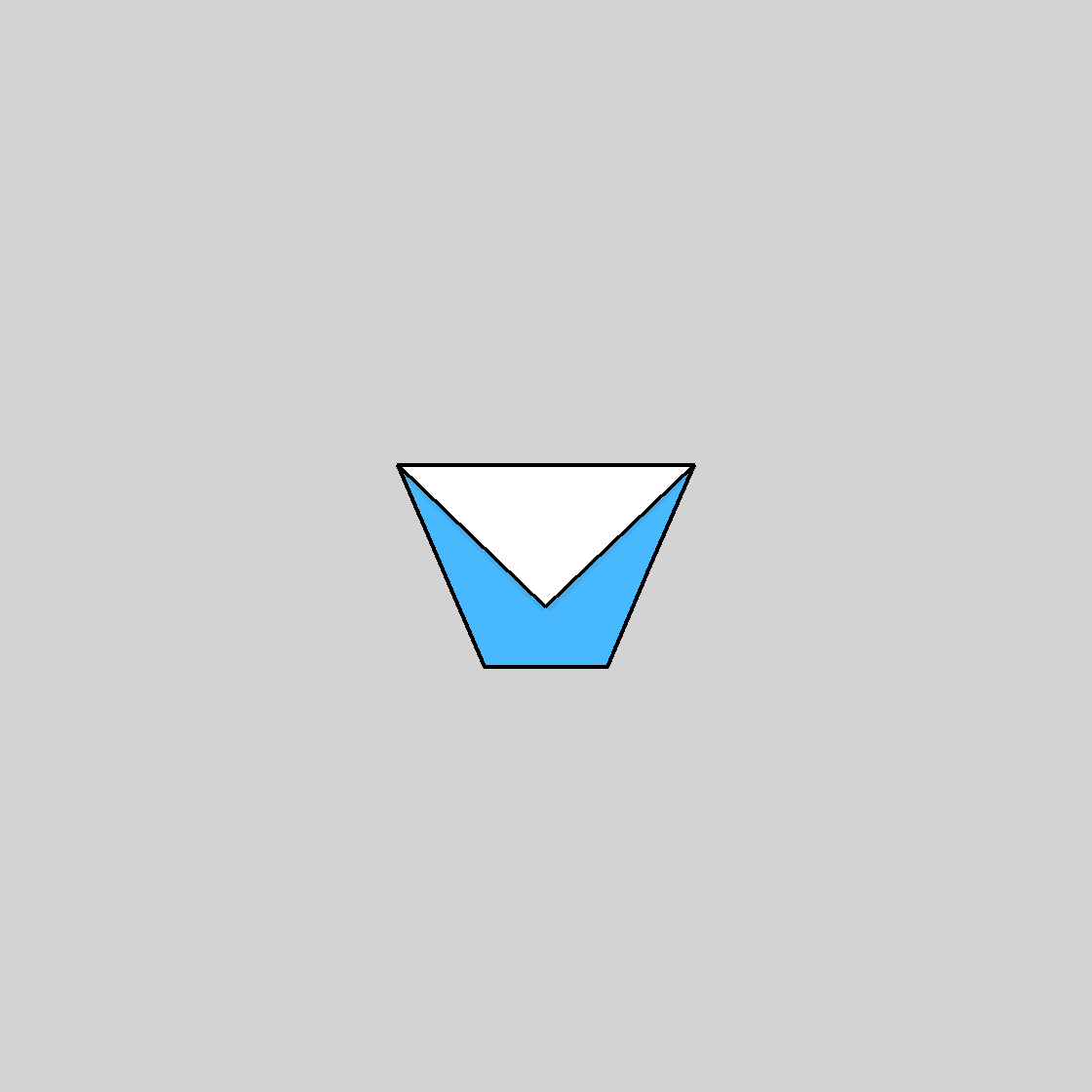}};

\node[cmd] at ($(d1)!0.5!(d2) + (0,-0.77)$) {
rotate(45$^\circ$)
};

\node[cmd] at ($(d2)!0.5!(d3) + (0,-0.77)$) {
fold([4,2],-1)
};

\node[cmd] at ($(d3)!0.5!(d4) + (0,-0.68)$) {
add\_v(0.58,[2,3])\\
fold([4,5],1)
};

\node[cmd] at ($(d4)!0.5!(d5) + (0,-0.77)$) {
unfold()
};

\node[cmd] at ($(d5)!0.5!(d6) + (0,-0.68)$) {
add\_v(0.58,[2,3])\\
add\_v(0.41,[3,4])\\
fold([6,7],-1)
};

\node[cmd] at ($(d6)!0.5!(d7) + (0,-0.68)$) {
add\_v(0.7,[2,6])\\
fold([10,5],-1)
};

\node[cmd] at ($(d7)!0.5!(d8) + (0,-0.77)$) {
fold([5,7],1)
};

\node[cmd] at ($(d8)!0.5!(d9) + (0,-0.77)$) {
flip(x)
};

\node[cmd] at ($(d9)!0.5!(d10) + (0,-0.77)$) {
fold([11,9],1)
};

\end{tikzpicture}
\Description{Qualitative results}
\caption{\emph{Qualitative results.} For each sequence, we show the input keyframes (top) and the rendering of the corresponding folded states predicted by our method (bottom). The actions inferred between consecutive keyframes are shown between the rendered states.} 
\label{fig:fox_results}
\end{figure*}

\subsection{Qualitative Results}
We first present selected qualitative results in \cref{fig:teaser,fig:fox_results,fig:fourteen_images2}, showing rendered folding procedures extracted from input keyframe sequences across diverse concepts. Our method handles challenging occlusions and lighting conditions, as well as complex configurations with accumulating layers.
For example, \cref{fig:fourteen_images2} shows long sequences of 13 and 14 keyframes that include diverse flips and unfolds, as well as vertex definitions requiring precise geometric reasoning, such as in keyframe 9 (top). In addition, the bottom sequence in \cref{fig:fox_results} contains challenging hand occlusions in keyframes 3, 6, 8 and 9, which our method successfully overcomes.

\ifarxiv
    
\begin{figure*}
    \centering
    \includegraphics[width=0.85\linewidth]{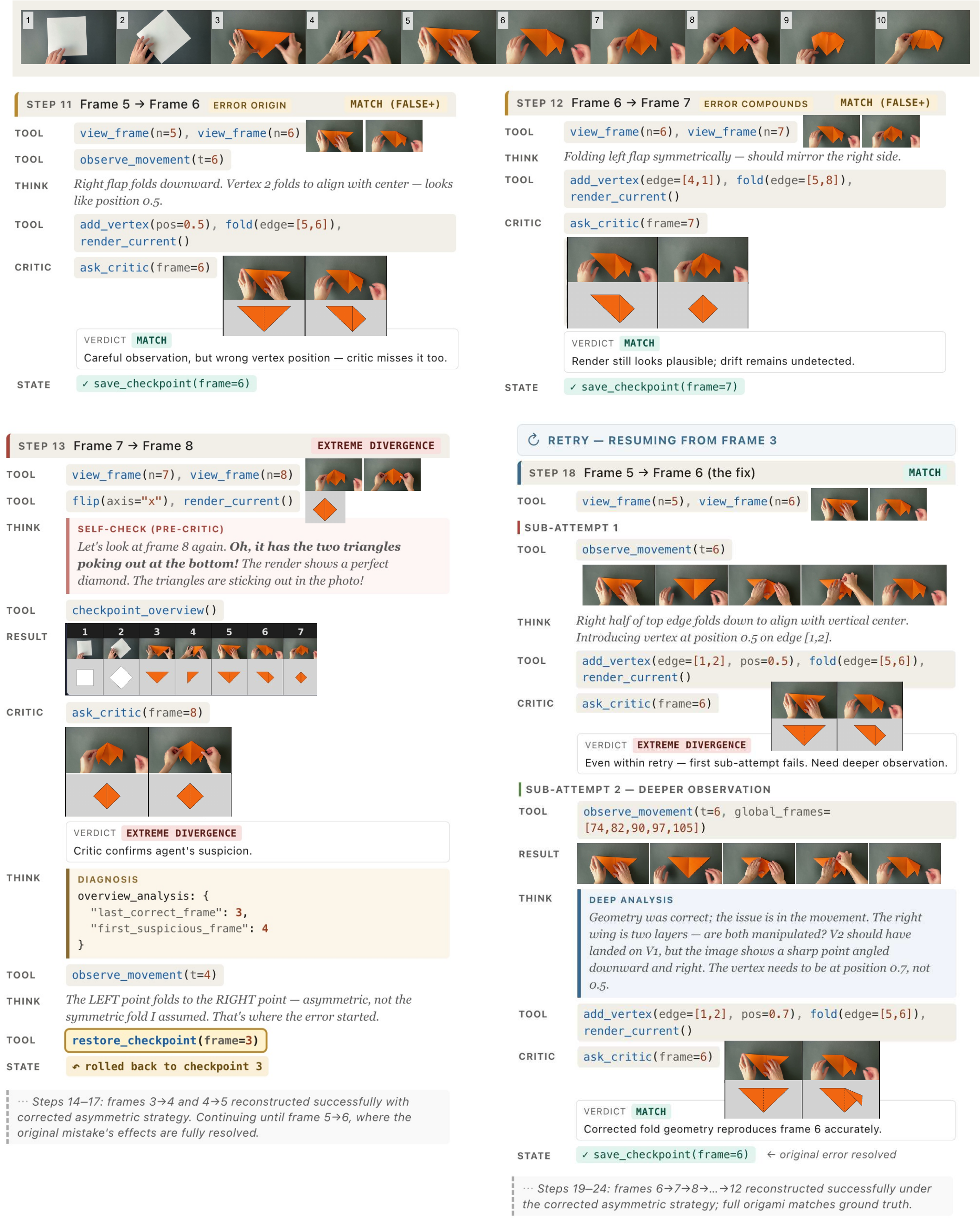}
    \Description{Recovery from critic-approved errors.}
    \caption{\emph{Recovery from critic-approved errors.} The top row shows the full keyframe sequence, while the four panels below illustrate selected reasoning steps. An incorrect fold at frame $5\rightarrow6$ is approved by the critic and propagates to frame 7. The agent detects the divergence at frame $7\rightarrow8$, rolls back to an earlier checkpoint, and reconstructs the sequence using additional tool calls until it finds a corrected fold that remains consistent through the final frame.}
    \label{fig:analysis}
\end{figure*}

\fi

In \cref{fig:analysis}, we illustrate the reasoning process of our agent across selected steps within a single sequence, demonstrating its interaction with the controller, use of different tools, and autonomous decision-making process. Specifically, as seen in the transition from keyframe 5 to 6, the agent first predicts an inaccurate vertex location, which the critic approves, resulting in a wrong fold. This error is then propagated from keyframe 6 to 7, with both steps approved by the critic. Such failures are possible because the critic has an 88\% success rate.
However, the agent can recover from these mistakes through our framework. In the transition from keyframe 7 to 8, the agent recognizes that the action involves a flip, but the resulting state is unsatisfactory. It verifies this concern with the critic and then begins investigating the source of the problem by rolling back to keyframe 3. From there, it successfully reconstructs the actions until it reaches the transition from frame 5 to frame 6 again. This time, the agent makes several attempts and autonomously calls tools, such as observing movement between frames, until it reaches a better decision with a more accurate fold. This improved decision then persists through the final fold.

\subsection{Quantitative Evaluation} \label{sec:benchmark}
\subsubsection{PurelandFold Dataset} \label{sec:data}
Existing origami datasets target either static crease patterns or final folded forms \cite{xu2025origamispace,agarwal2026origamibench,spencer2025gamibench}, but none supports the evaluation of procedural folding states from real-world demonstrations.
We thus curate a new dataset, \textit{PurelandFold}, comprising 27 diverse folding sequences. We sourced the underlying instructions from the \textit{Easy Origami} category on the OrigamiWay website~\shortcite{origamiway}, which provides a clean and diverse collection of 60 origami models, 40 of which are Pureland. We captured all demonstration videos ourselves, and annotated each sequence with
keyframe-level labels of ground-truth folding actions and geometric states (see \cref{sec:actions,sec:representation}), including paper topology, vertex and face geometry, crease assignments, face orientations, and the layer structure. Using curated instructions and self-capture allows us to achieve an unbiased automatic selection of models, without going into filtering and manual inspection of noisy online videos. Our dataset provides a testbed for evaluating the procedural fidelity of our framework. A link to our dataset is available on our project page.

\subsubsection{Metrics}\label{sec:metrics}
A successful prediction should be (1) \emph{valid} -- representing a physically foldable geometry, and (2) \emph{faithful} to the ground truth, both to the annotated geometry and to the raw keyframe. 
To measure (1), we follow OrigamiSpace~\cite{xu2025origamispace} and define Compilation Validity (CV) as whether the predicted geometry can be processed by the Flat-Folder compiler~\cite{ku2022flatfolder}; this requires syntactic validity, consistent topology, and no self-intersections.

Measuring (2) is less straightforward, as faithfulness has several distinct aspects, each of which needs a different metric. We thus follow OrigamiSpace~\cite{xu2025origamispace} and report several complementary metrics: \emph{Topological Structure Similarity (TSS)}, comparing the combinatorial structure of the crease-pattern graph, including vertices, edges, faces, and edge labels. \emph{Geometric Similarity (GS)} evaluates the realized 3D geometry through point-set Hausdorff distance, dihedral angles, and global proportions. \emph{Constraint Satisfaction (CS)} tests whether the prediction satisfies physical folding constraints, including layer ordering and local flat-foldability conditions such as Maekawa's and Kawasaki's theorems. \emph{Final Folded State (FFS)} evaluates the overall shape and the layering order. Finally, \emph{Crease-Pattern Dissimilarity (CPD)}~\cite{oh2015dissimilarity} provides a comparison via explicit bipartite matching between predicted and ground-truth elements (edges and vertices); unlike OrigamiSpace, which compares patterns by aggregated graph statistics, CPD requires per-element correspondence.

\ifarxiv
    
\begin{figure*}
    \centering
    \includegraphics[width=1\linewidth]{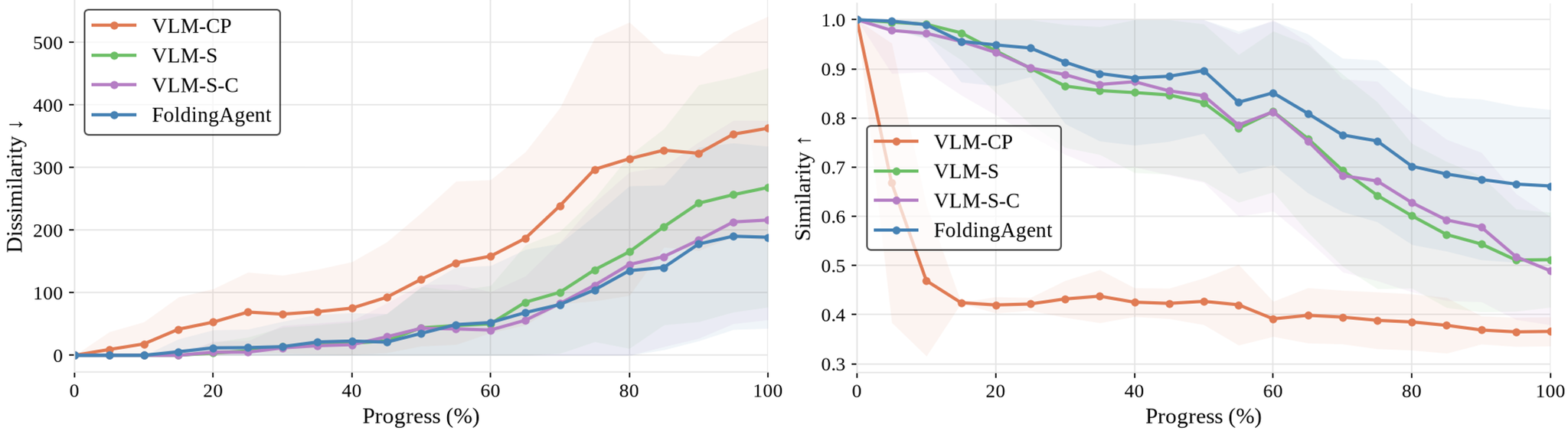}
    \Description{Performance}
    \caption{\emph{Performance of our method and ablation variants as a function of sequence progress.} We report mean Dissimilarity and Similarity scores averaged over completed sequences. As seen, the performance of all methods declines as the sequence progresses, yet our method exhibits a steadier slope with a notable gap to the others, which increases as the complexity grows.}
    \label{fig:graphs}
\end{figure*}

\fi

\subsubsection{Ablations} \label{sec:baselines}
To the best of our knowledge, inferring step-by-step folded states from origami videos has not been addressed in prior work, leaving no method that can be directly applied to our task 
(see the analysis of VIGA~\cite{yin2026viga} in \supp{sec:viga}).
We thus evaluate against a ladder of variants
that progressively add our key components. %
Unless stated otherwise, all configurations use the same VLM (Gemini 3.1 Pro Preview).
\begin{enumerate}[leftmargin=1.5em]
    \item \emph{VLM-CP:} The VLM is given a pair of frames along with the current crease pattern in standard FOLD representation $(I_{t-1}, I_{t}, CP_{t-1})$, and is prompted to predict $CP_{t}$. 
    This variant does not use our representation, tools, simulator, and controller. %
    It serves as a reference for what the underlying VLM can achieve with a conventional crease pattern representation.
    
    \item \emph{VLM-S:} Instead of a crease pattern, the VLM works in our representation and action spaces (\cref{sec:representation,sec:actions}). It receives the current folded state and predicts a symbolic action: $(I_{t-1}, I_{t}, S_{t-1})\rightarrow a_t$. 
    This variant isolates the effect of our state and action representation: rather than directly predicting the next geometry, the model predicts an executable simulator action.
    
    \item \emph{VLM-S + Critic (VLM-S-C):} We add the external critic (\cref{sec:critic}), yet in a \emph{non-agentic} fashion. That is, each step comprises an action prediction followed by critic validation until the critic declares success. If more than 3 attempts are made, the VLM is requested to select the best out of them. This variant allows us to quantify the contribution of richer, more flexible reasoning enabled by our agentic tool-calling procedure. 
\end{enumerate}

\begin{table}[t]
\centering
\caption{\emph{Quantitative comparison.} We report metrics (\cref{sec:metrics}) across all ablation variants (\cref{sec:baselines}) and our full framework (FoldingAgent)\textsuperscript{*}. \#Comp. denotes the percentage of fully completed sequences and CV the compilation success rate on completed steps; our full method achieves the best scores across all similarity and dissimilarity metrics. We further report the percentage of judgments rating our renderings as closer to the keyframes than those of each variant (VLM-CP could not be evaluated since most of its sequences are not compilable).}
\label{tab:quant_res}
\Description[]{}

\newcommand{\std}[1]{$^{\pm#1}$}
\setlength{\tabcolsep}{2pt}
\renewcommand{\arraystretch}{0.95}

\resizebox{\columnwidth}{!}{
\begin{tabular}{@{}lccccccc|c@{}}
\toprule
\multirow{2}{*}{Model} &
\multirow{2}{*}{\#Comp. $\uparrow$} &
\multirow{2}{*}{CV $\uparrow$} &
\multicolumn{4}{c}{Similarity $\uparrow$} &
\multirow{2}{*}{\makecell{CPD $\downarrow$}} &
\multirow{2}{*}{\makecell{User \\ Study}}\\
\cmidrule(lr){4-7}
 & & & TSS & GS & CS & FFS & &\\
\midrule
VLM-CP  & 96\% & 8\%  & 0.76\std{0.1} & 0.20\std{0.0} & 0.20\std{0.0} & 0.30\std{0.0} & 363\std{179}  & - \\
VLM-S   & 85\%          & 96\% & 0.79\std{0.1} & 0.51\std{0.1} & 0.53\std{0.2} & 0.21\std{0.1} & 268\std{191} & 84\%\std{5\%} \\
VLM-S-C & 85\%          & 96\%  & 0.80\std{0.1} & 0.48\std{0.1} & 0.49\std{0.2} & 0.19\std{0.1} & 216\std{159} & 80\%\std{7\%} \\
\algoname  & \textbf{100\%} & 96\% & $\mathbf{0.90}$\std{0.1} & $\mathbf{0.58}$\std{0.1} & $\mathbf{0.77}$\std{0.3} & $\mathbf{0.39}$\std{0.3} & $\mathbf{189}$\std{145} & \\
\bottomrule
\end{tabular}

}

\vspace{2pt}
{\footnotesize\textsuperscript{*}VLM-CP: naive CP prediction; VLM-S: action prediction; VLM-S-C: action + critic.}
\end{table}

We evaluate these ablation variants on our dataset. Quantitative results for the final predicted geometry on completed sequences are reported in \cref{tab:quant_res}, and \cref{fig:graphs} plots performance across all intermediate steps.

As seen in \cref{tab:quant_res}, applying the VLM without our representation, action space, tools, or agentic framework (VLM-CP) yields invalid geometries in the vast majority of cases, with only 8\% of outputs compiling successfully. 
Shifting the task to our state and action representation (VLM-S) substantially improves compilation rates, quantifying the impact of our representation on producing valid crease patterns, alongside improved topology and geometry scores.
Incorporating the critic (VLM-S-C) further improves fine-grained geometry, reflected in a lower dissimilarity score (CPD).

Our full agentic framework improves all metrics significantly, yielding more precise topology, geometry, and crease patterns. This gain stems from the framework's ability to autonomously detect errors, roll back incorrect steps, and iteratively self-correct — capabilities absent in the ablated variants.
This flexibility does, however, come with a tradeoff: reasoning and tool-calling chains can grow long. We therefore use bounded exploration (\cref{sec:agentic}) to limit the number of attempts per keyframe, and cap execution at 300 total tool calls to constrain runtime and cost. %

\begin{figure}[htbp]
\centering

\setlength{\tabcolsep}{1pt}

\renewcommand{\arraystretch}{1.2}

\begin{tabular}{cccc}

Last Frame (GT) &
VLM-S &
VLM-S-C &
Ours \\

\includegraphics[width=0.25\linewidth, height=0.14\linewidth]{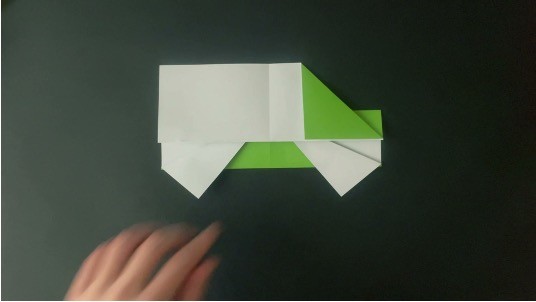} &
\includegraphics[width=0.25\linewidth, height=0.14\linewidth, trim={0pt 1.2cm 0pt 1.2cm}, clip]{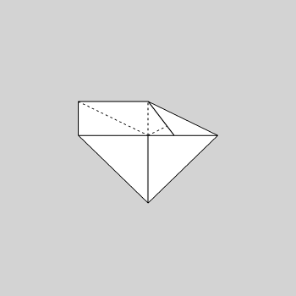} &
\includegraphics[width=0.25\linewidth, height=0.14\linewidth, trim={0pt 1.2cm 0pt 1.2cm}, clip]{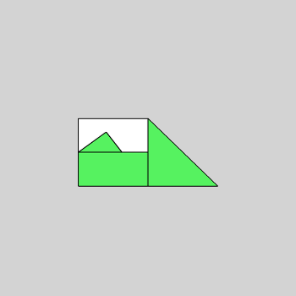} &
\includegraphics[width=0.25\linewidth, height=0.14\linewidth, trim={0pt 1.2cm 0pt 1.2cm}, clip]{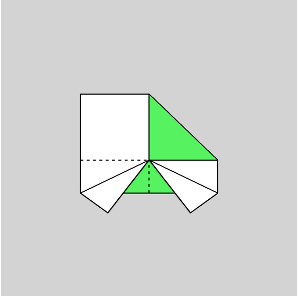}\\

\includegraphics[width=0.25\linewidth]{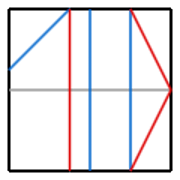} &
\includegraphics[width=0.25\linewidth]{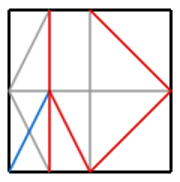} &
\includegraphics[width=0.25\linewidth]{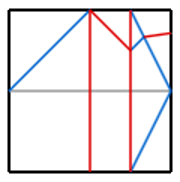} &
\includegraphics[width=0.25\linewidth]{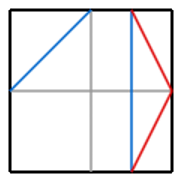}\\

\includegraphics[width=0.25\linewidth, height=0.14\linewidth]{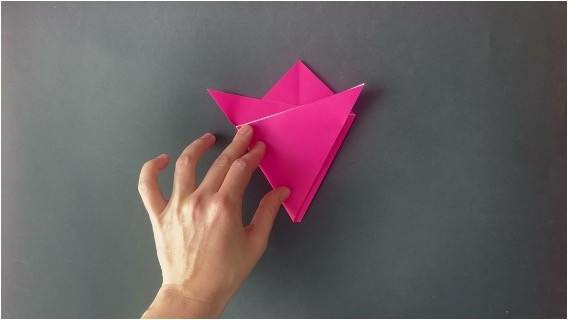} &
\includegraphics[width=0.25\linewidth, height=0.14\linewidth, trim={0pt 1.2cm 0pt 1.2cm}, clip]{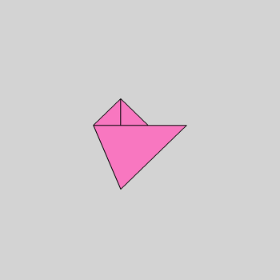} &
\includegraphics[width=0.25\linewidth, height=0.14\linewidth, trim={0pt 1.2cm 0pt 1.2cm}, clip]{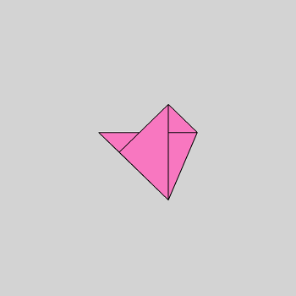} &
\includegraphics[width=0.25\linewidth, height=0.14\linewidth, trim={0pt 1.2cm 0pt 1.2cm}, clip]{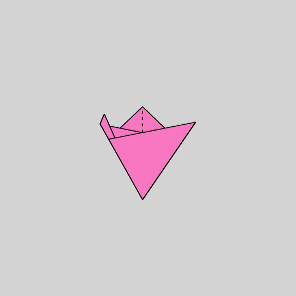}\\

\includegraphics[width=0.25\linewidth]{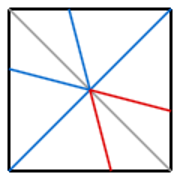} &
\includegraphics[width=0.25\linewidth]{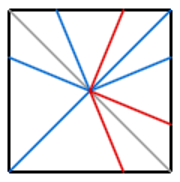} &
\includegraphics[width=0.25\linewidth]{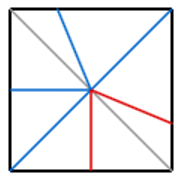} &
\includegraphics[width=0.25\linewidth]{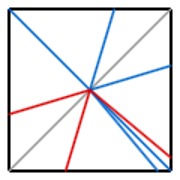}\\

\includegraphics[width=0.25\linewidth, height=0.14\linewidth]{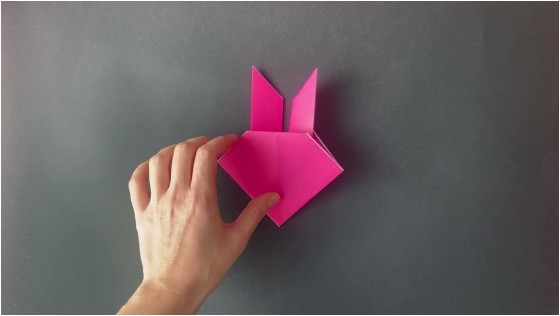} &
\includegraphics[width=0.25\linewidth, height=0.14\linewidth, trim={0pt 1.2cm 0pt 1.2cm}, clip]{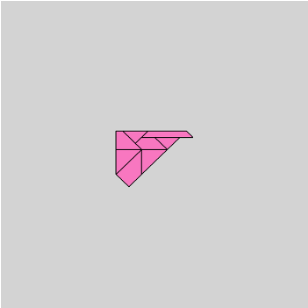} &
\includegraphics[width=0.25\linewidth, height=0.14\linewidth, trim={0pt 1.2cm 0pt 1.2cm}, clip]{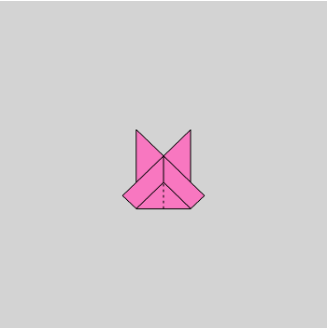} &
\includegraphics[width=0.25\linewidth, height=0.14\linewidth, trim={0pt 1.2cm 0pt 1.2cm}, clip]{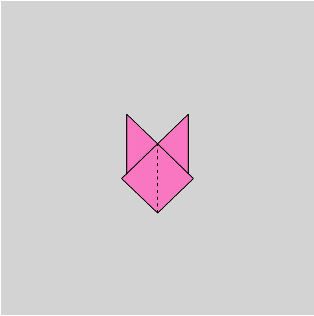}\\

\includegraphics[width=0.25\linewidth]{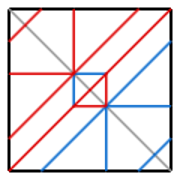} &
\includegraphics[width=0.25\linewidth]{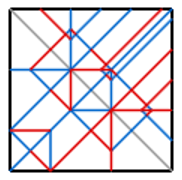} &
\includegraphics[width=0.25\linewidth]{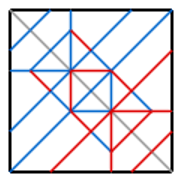} &
\includegraphics[width=0.25\linewidth]{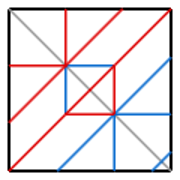}
\end{tabular} \vspace{-0.3cm}
\Description{Qualitative comparison.}
\caption{\emph{Qualitative comparison.} Each row shows a ground-truth keyframe (left) alongside the reconstructed geometry and crease pattern produced by VLM-S, VLM-S-C, and our full framework; mountain and valley creases are marked in red and blue lines, respectively; flat edges are marked in gray.}
\label{fig:comp} %
\end{figure}

These differences are further evidenced visually in \cref{fig:comp}, where our full method produces notably more accurate final states on completed sequences. 
We further validate our results through a perceptual user study,
using human judgments to assess image-to-render similarity.
The participants were shown rendered geometries from two methods (A) and (B), and asked which is more similar to the ground-truth keyframe: A, B, or a tie. 
We collected 600 judgments over a subset of 10 sequences from our data, comprising all sequences jointly completed by all variants within the tool-call budget. As seen in \cref{tab:quant_res}, our method is strongly preferred over both variants (84\% and 80\%).

\ifarxiv
    
    Refer to \supp{sec:in_context,sec:backbones} for an ablation of in-context examples and backbone generalization results.
    
\else
    
    \ifarxiv
        \ifarxiv
    \subsection{Generalization across Backbones}
\else
    \subsubsection{Generalization across Backbones}
\fi
\label{sec:backbones}
We compare a plain agent (VLM-CP, the only variant with none of our design choices) with our FoldingAgent across GPT-5.5, Claude-Opus-4.8, and Gemini-3.1-Pro-Preview
\ifarxiv on a randomly sampled subset of 5 sequences, with metrics computed across all steps.
\else using the same subset of 5 sequences as in the \textit{In-context Examples} experiment. \fi
FoldingAgent substantially improves all backbones, indicating the workflow is not tied to a single VLM.

\begin{table}[h]
\centering
\caption{\textit{Generalization across backbones.}}
\label{tab:vlm-comparison}
\resizebox{\columnwidth}{!}{%
\begin{tabular}{l cc cc cc}
\toprule
\multirow{2}{*}{\textbf{Model}} & \multicolumn{2}{c}{\textbf{\#COMP}} & \multicolumn{2}{c}{\textbf{Similarity (avg) $\uparrow$}} & \multicolumn{2}{c}{\textbf{Dissimilarity $\downarrow$}} \\
\cmidrule(lr){2-3} \cmidrule(lr){4-5} \cmidrule(lr){6-7}
 & VLM-CP & FoldingAgent & VLM-CP & FoldingAgent & VLM-CP & FoldingAgent \\
\midrule
GPT    & 0/5 & 4/5 & 0.37 & 0.73 & 283 & 57  \\
Claude & 0/5 & 5/5 & 0.38 & 0.75 & 252 & 61 \\
Gemini & 1/5 & 5/5 & 0.55 & 0.79 & 170 & 49  \\
\bottomrule
\end{tabular}%
}
\end{table}

        \ifarxiv
    \subsection{Additional Ablation: In Context Examples}
\else
    \paragraph{\textbf{In-context Examples}}
\fi
\label{sec:in_context}
We compare zero-shot prompting against a few-shot setting in which the context includes three transitions of varying difficulty selected from outside the evaluation set. Each example shows the source and target frames alongside the required action $(I_{t-1}, I_t, a_t)$.
\ifarxiv Using the same subset of 5 sequences as in \cref{sec:backbones}, the table below
\else The table below, computed on a randomly sampled subset of 5 sequences, \fi
shows that the zero-shot version outperforms the few-shot one. We attribute this to the breadth of the visual state space and the diversity of valid action sequences: a small, fixed set of examples cannot cover this space and instead biases the VLM toward the specific states and actions it happens to see. Thus, performance becomes sensitive to the particular examples selected, without improving generalization. \ifarxiv \else Metrics are computed across all steps. \fi

\begin{table}[h]

\centering
\caption{\textit{In context examples ablation.}}
\label{tab:examples-ablation}
\setlength{\tabcolsep}{6pt}
\resizebox{\columnwidth}{!}{%
\begin{tabular}{l cccc}
\toprule
 & \textbf{Completed} $\uparrow$ &  \textbf{Similarity (avg) $\uparrow$} & \textbf{Dissimilarity $\downarrow$} \\
\midrule
Few-shot             & 4/5 & 0.72 & 124 \\
Zero-shot (ours)   & \textbf{5/5} & \textbf{0.78} & \textbf{49} \\
\bottomrule
\end{tabular}
}
\end{table}

    \else

    \fi

\fi

\section{Conclusion, Limitations, and Future Work} \label{sec:conclusion}

We introduced FoldingAgent, a novel agentic framework capable of translating unstructured origami demonstrations into executable procedural folding programs. By formalizing a parameterized action space for Pureland origami and coupling a pre-trained Vision-Language Model (VLM) with a suite of specialized simulation, verification, and rollback tools, our framework dramatically enhances the VLM's capabilities in mitigating compounding errors and long-range dependencies inherent in multi-step folding tasks. 

As for limitations, our performance is highly dependent on the VLM's reasoning capabilities. We observed that folds that are heavily occluded -- whether by the demonstrator or by complex, accumulating paper layers -- remain difficult for the model to accurately identify from raw keyframes. Our agent also struggles to identify compound simultaneous actions (e.g., rotating while flipping the paper), as these break the standard sequential assumptions of our parameterized action space.
Finally, our focus in this paper is Pureland origami; extending it to more advanced techniques is conceptually possible but requires extending the action space and may require stronger visual-language reasoning (which may not yet be available). We visualize failure cases in \supp{sec:failure}.

Our work takes a first notable step towards bridging the semantic gap between intuitive human visual demonstrations and structured computational origami representations. We believe that the principles we presented for designing a specialized, flexible agentic framework will trigger more research in this area, opening the door to new applications in robotic manipulation, interactive instructional systems, and automated generative design.

\ifanonymous\else
    \begin{acks}
This research has been generously supported by the Sagol Weizmann-MIT Bridge Program.
\end{acks}

\fi

\bibliographystyle{ACM-Reference-Format}
\bibliography{main}


\begin{thebibliography}{65}


\ifx \showCODEN    \undefined \def \showCODEN     #1{\unskip}     \fi
\ifx \showISBNx    \undefined \def \showISBNx     #1{\unskip}     \fi
\ifx \showISBNxiii \undefined \def \showISBNxiii  #1{\unskip}     \fi
\ifx \showISSN     \undefined \def \showISSN      #1{\unskip}     \fi
\ifx \showLCCN     \undefined \def \showLCCN      #1{\unskip}     \fi
\ifx \shownote     \undefined \def \shownote      #1{#1}          \fi
\ifx \showarticletitle \undefined \def \showarticletitle #1{#1}   \fi
\ifx \showURL      \undefined \def \showURL       {\relax}        \fi
\providecommand\bibfield[2]{#2}
\providecommand\bibinfo[2]{#2}
\providecommand\natexlab[1]{#1}
\providecommand\showeprint[2][]{arXiv:#2}

\bibitem[Ablat and Qattawi(2018)]%
        {ablat2018finite}
\bibfield{author}{\bibinfo{person}{Muhammad~Ali Ablat} {and} \bibinfo{person}{Ala Qattawi}.} \bibinfo{year}{2018}\natexlab{}.
\newblock \showarticletitle{Finite element analysis of origami-based sheet metal folding process}.
\newblock \bibinfo{journal}{\emph{Journal of Engineering Materials and Technology}} \bibinfo{volume}{140}, \bibinfo{number}{3} (\bibinfo{year}{2018}), \bibinfo{pages}{031008}.
\newblock
\href{https://doi.org/10.1115/1.4039505}{doi:\nolinkurl{10.1115/1.4039505}}


\bibitem[Agarwal et~al\mbox{.}(2026)]%
        {agarwal2026origamibench}
\bibfield{author}{\bibinfo{person}{Naaisha Agarwal}, \bibinfo{person}{Yihan Wu}, \bibinfo{person}{Yichang Jian}, \bibinfo{person}{Yikuan Hu}, \bibinfo{person}{Nishad Mansoor}, \bibinfo{person}{Mohan Li}, \bibinfo{person}{Yifei Peng}, \bibinfo{person}{Wang-Zhou Dai}, \bibinfo{person}{Yao-Xiang Ding}, {and} \bibinfo{person}{Emanuele Sansone}.} \bibinfo{year}{2026}\natexlab{}.
\newblock \bibinfo{title}{{OrigamiBench}: An Interactive Environment to Synthesize Flat-Foldable Origamis}.
\newblock
\showeprint[arxiv]{2603.13856}~[cs.LG]
\urldef\tempurl%
\url{https://arxiv.org/abs/2603.13856}
\showURL{%
\tempurl}


\bibitem[Akitaya et~al\mbox{.}(2020)]%
        {akitaya2020rigid}
\bibfield{author}{\bibinfo{person}{Hugo~A. Akitaya}, \bibinfo{person}{Erik~D. Demaine}, \bibinfo{person}{Takashi Horiyama}, \bibinfo{person}{Thomas~C. Hull}, \bibinfo{person}{Jason~S. Ku}, {and} \bibinfo{person}{Tomohiro Tachi}.} \bibinfo{year}{2020}\natexlab{}.
\newblock \showarticletitle{Rigid Foldability is NP-Hard}.
\newblock \bibinfo{journal}{\emph{Journal of Computational Geometry (JoCG)}} \bibinfo{volume}{11}, \bibinfo{number}{1} (\bibinfo{year}{2020}), \bibinfo{pages}{93--124}.
\newblock
\href{https://doi.org/10.20382/jocg.v11i1a4}{doi:\nolinkurl{10.20382/jocg.v11i1a4}}


\bibitem[Akitaya et~al\mbox{.}(2013)]%
        {akitaya2013generating}
\bibfield{author}{\bibinfo{person}{Hugo~A. Akitaya}, \bibinfo{person}{Jun Mitani}, \bibinfo{person}{Yoshihiro Kanamori}, {and} \bibinfo{person}{Yukio Fukui}.} \bibinfo{year}{2013}\natexlab{}.
\newblock \showarticletitle{Generating Folding Sequences from Crease Patterns of Flat-Foldable Origami}. In \bibinfo{booktitle}{\emph{ACM SIGGRAPH 2013 Posters (SIGGRAPH)}}. \bibinfo{publisher}{ACM}, \bibinfo{address}{New York, NY, USA}, \bibinfo{pages}{1--1}.
\newblock
\href{https://doi.org/10.1145/2503385.2503407}{doi:\nolinkurl{10.1145/2503385.2503407}}


\bibitem[Bai et~al\mbox{.}(2025)]%
        {bai2025vlm3drewards}
\bibfield{author}{\bibinfo{person}{Weimin Bai}, \bibinfo{person}{Yubo Li}, \bibinfo{person}{Weijian Luo}, \bibinfo{person}{Wenzheng Chen}, {and} \bibinfo{person}{He Sun}.} \bibinfo{year}{2025}\natexlab{}.
\newblock \bibinfo{title}{Vision-Language Models as Differentiable Semantic and Spatial Rewards for Text-to-3D Generation}.
\newblock
\showeprint[arxiv]{2509.15772}~[cs.CV]
\urldef\tempurl%
\url{https://ai4scientificimaging.org/vlm3d}
\showURL{%
\tempurl}


\bibitem[Balkcom(2004)]%
        {balkcom2004robotic}
\bibfield{author}{\bibinfo{person}{Devin Balkcom}.} \bibinfo{year}{2004}\natexlab{}.
\newblock \emph{\bibinfo{title}{Robotic origami folding}}.
\newblock \bibinfo{thesistype}{Ph.\,D. Dissertation}. \bibinfo{school}{Carnegie Mellon University}, \bibinfo{address}{Pittsburgh, PA}.
\newblock
\urldef\tempurl%
\url{https://www.ri.cmu.edu/publications/robotic-origami-folding/}
\showURL{%
\tempurl}


\bibitem[Balkcom and Mason(2008)]%
        {balkcom2008robotic}
\bibfield{author}{\bibinfo{person}{Devin~J. Balkcom} {and} \bibinfo{person}{Matthew~T. Mason}.} \bibinfo{year}{2008}\natexlab{}.
\newblock \showarticletitle{Robotic origami folding}.
\newblock \bibinfo{journal}{\emph{The International Journal of Robotics Research}} \bibinfo{volume}{27}, \bibinfo{number}{5} (\bibinfo{year}{2008}), \bibinfo{pages}{613--627}.
\newblock
\href{https://doi.org/10.1177/0278364908090235}{doi:\nolinkurl{10.1177/0278364908090235}}


\bibitem[Bern and Hayes(1996)]%
        {bern1996complexity}
\bibfield{author}{\bibinfo{person}{Marshall Bern} {and} \bibinfo{person}{Barry Hayes}.} \bibinfo{year}{1996}\natexlab{}.
\newblock \showarticletitle{The Complexity of Flat Origami}. In \bibinfo{booktitle}{\emph{Proceedings of the Seventh Annual ACM-SIAM Symposium on Discrete Algorithms (SODA)}}. \bibinfo{publisher}{ACM/SIAM}, \bibinfo{address}{New York, NY, USA and Philadelphia, PA, USA}, \bibinfo{pages}{175--183}.
\newblock
\urldef\tempurl%
\url{https://dl.acm.org/doi/10.5555/313852.313918}
\showURL{%
\tempurl}


\bibitem[Bubeck et~al\mbox{.}(2023)]%
        {bubeck2023sparks}
\bibfield{author}{\bibinfo{person}{S{\'e}bastien Bubeck}, \bibinfo{person}{Varun Chandrasekaran}, \bibinfo{person}{Ronen Eldan}, \bibinfo{person}{Johannes Gehrke}, \bibinfo{person}{Eric Horvitz}, \bibinfo{person}{Ece Kamar}, \bibinfo{person}{Peter Lee}, \bibinfo{person}{Yin~Tat Lee}, \bibinfo{person}{Yuanzhi Li}, \bibinfo{person}{Scott Lundberg}, {et~al\mbox{.}}} \bibinfo{year}{2023}\natexlab{}.
\newblock \bibinfo{title}{Sparks of Artificial General Intelligence: Early Experiments with GPT-4}.
\newblock
\showeprint[arxiv]{2303.12712}~[cs.CL]
\urldef\tempurl%
\url{https://arxiv.org/abs/2303.12712}
\showURL{%
\tempurl}


\bibitem[Cai et~al\mbox{.}(2024)]%
        {cai2024delving}
\bibfield{author}{\bibinfo{person}{Mu Cai}, \bibinfo{person}{Zeyi Huang}, \bibinfo{person}{Yuheng Li}, \bibinfo{person}{Haohan Wang}, {and} \bibinfo{person}{Yong~Jae Lee}.} \bibinfo{year}{2024}\natexlab{}.
\newblock \bibinfo{title}{Delving into {LLM}s{\textquoteright} visual understanding ability using {SVG} to bridge image and text}.
\newblock
\urldef\tempurl%
\url{https://openreview.net/forum?id=pwlm6Po61I}
\showURL{%
\tempurl}


\bibitem[Chen et~al\mbox{.}(2023)]%
        {chen2023origamisensei}
\bibfield{author}{\bibinfo{person}{Qiyu Chen}, \bibinfo{person}{Richa Mishra}, \bibinfo{person}{Dina El-Zanfaly}, {and} \bibinfo{person}{Kris Kitani}.} \bibinfo{year}{2023}\natexlab{}.
\newblock \showarticletitle{Origami Sensei: Mixed Reality AI-Assistant for Creative Tasks Using Hands}. In \bibinfo{booktitle}{\emph{Designing Interactive Systems Conference (DIS)}}. \bibinfo{publisher}{ACM}, \bibinfo{address}{New York, NY, USA}, \bibinfo{pages}{147--151}.
\newblock
\href{https://doi.org/10.1145/3563703.3596625}{doi:\nolinkurl{10.1145/3563703.3596625}}


\bibitem[Chen et~al\mbox{.}(2025)]%
        {chen2025origamisensei}
\bibfield{author}{\bibinfo{person}{Qiyu Chen}, \bibinfo{person}{Richa Mishra}, \bibinfo{person}{Lia~Sparingga Purnamasari}, \bibinfo{person}{Dina El-Zanfaly}, {and} \bibinfo{person}{Kris Kitani}.} \bibinfo{year}{2025}\natexlab{}.
\newblock \showarticletitle{Origami Sensei: A Mixed Reality AI-Assistant}. In \bibinfo{booktitle}{\emph{Proceedings of the 2025 CHI Conference on Human Factors in Computing Systems (CHI)}}. \bibinfo{publisher}{ACM}, \bibinfo{address}{New York, NY, USA}, \bibinfo{pages}{1--18}.
\newblock
\href{https://doi.org/10.1145/3706598.3714099}{doi:\nolinkurl{10.1145/3706598.3714099}}


\bibitem[Chen et~al\mbox{.}(2026)]%
        {chen2026know3d}
\bibfield{author}{\bibinfo{person}{Wenyue Chen}, \bibinfo{person}{Wenjue Chen}, \bibinfo{person}{Peng Li}, \bibinfo{person}{Qinghe Wang}, \bibinfo{person}{Xu Jia}, \bibinfo{person}{Heliang Zheng}, \bibinfo{person}{Rongfei Jia}, \bibinfo{person}{Yuan Liu}, {and} \bibinfo{person}{Ronggang Wang}.} \bibinfo{year}{2026}\natexlab{}.
\newblock \bibinfo{title}{{Know3D}: Prompting 3D Generation with Knowledge from Vision-Language Models}.
\newblock
\showeprint[arxiv]{2603.22782}~[cs.CV]
\urldef\tempurl%
\url{https://xishuxishu.github.io/Know3D/}
\showURL{%
\tempurl}


\bibitem[Demaine et~al\mbox{.}(2016)]%
        {FOLD_CGW2016}
\bibfield{author}{\bibinfo{person}{Erik~D. Demaine}, \bibinfo{person}{Jason~S. Ku}, {and} \bibinfo{person}{Robert~J. Lang}.} \bibinfo{year}{2016}\natexlab{}.
\newblock \showarticletitle{A New File Standard to Represent Folded Structures}. \bibinfo{howpublished}{\url{https://github.com/edemaine/fold}}. In \bibinfo{booktitle}{\emph{Abstracts from the 26th Fall Workshop on Computational Geometry}}. \bibinfo{publisher}{CUNY Graduate Center}, \bibinfo{address}{New York, NY, USA}, \bibinfo{pages}{to appear}.
\newblock


\bibitem[Demaine and O'Rourke(2007)]%
        {demaine2007geometric}
\bibfield{author}{\bibinfo{person}{Erik~D. Demaine} {and} \bibinfo{person}{Joseph O'Rourke}.} \bibinfo{year}{2007}\natexlab{}.
\newblock \bibinfo{booktitle}{\emph{Geometric Folding Algorithms: Linkages, Origami, Polyhedra}}.
\newblock \bibinfo{publisher}{Cambridge University Press}, \bibinfo{address}{Cambridge, UK}.
\newblock
\href{https://doi.org/10.1017/CBO9780511735172}{doi:\nolinkurl{10.1017/CBO9780511735172}}


\bibitem[Dudte et~al\mbox{.}(2021)]%
        {dudte2021additive}
\bibfield{author}{\bibinfo{person}{Levi~H. Dudte}, \bibinfo{person}{Gary P.~T. Choi}, {and} \bibinfo{person}{L. Mahadevan}.} \bibinfo{year}{2021}\natexlab{}.
\newblock \showarticletitle{An Additive Algorithm for Origami Design}.
\newblock \bibinfo{journal}{\emph{Proceedings of the National Academy of Sciences (PNAS)}} \bibinfo{volume}{118}, \bibinfo{number}{21} (\bibinfo{year}{2021}), \bibinfo{pages}{e2019241118}.
\newblock
\href{https://doi.org/10.1073/pnas.2019241118}{doi:\nolinkurl{10.1073/pnas.2019241118}}


\bibitem[Eppstein(2025)]%
        {eppstein2025complexities}
\bibfield{author}{\bibinfo{person}{David Eppstein}.} \bibinfo{year}{2025}\natexlab{}.
\newblock \showarticletitle{Computational Complexities of Folding}.
\newblock \bibinfo{journal}{\emph{Journal of Information Processing (JIP)}}  \bibinfo{volume}{33} (\bibinfo{year}{2025}), \bibinfo{pages}{954--973}.
\newblock
\href{https://doi.org/10.2197/ipsjjip.33.954}{doi:\nolinkurl{10.2197/ipsjjip.33.954}}


\bibitem[Felton et~al\mbox{.}(2014)]%
        {felton2014method}
\bibfield{author}{\bibinfo{person}{Samuel~M. Felton}, \bibinfo{person}{Michael~T. Tolley}, \bibinfo{person}{Erik~D. Demaine}, \bibinfo{person}{Daniela Rus}, {and} \bibinfo{person}{Robert~J. Wood}.} \bibinfo{year}{2014}\natexlab{}.
\newblock \showarticletitle{A method for building self-folding machines}.
\newblock \bibinfo{journal}{\emph{Science}} \bibinfo{volume}{345}, \bibinfo{number}{6197} (\bibinfo{year}{2014}), \bibinfo{pages}{644--646}.
\newblock
\href{https://doi.org/10.1126/science.1252610}{doi:\nolinkurl{10.1126/science.1252610}}


\bibitem[Feng et~al\mbox{.}(2020)]%
        {feng2020rigid}
\bibfield{author}{\bibinfo{person}{Huijuan Feng}, \bibinfo{person}{Rui Peng}, \bibinfo{person}{Shixi Zang}, \bibinfo{person}{Jiayao Ma}, {and} \bibinfo{person}{Yan Chen}.} \bibinfo{year}{2020}\natexlab{}.
\newblock \showarticletitle{Rigid Foldability and Mountain-Valley Crease Assignments of Square-Twist Origami Pattern}.
\newblock \bibinfo{journal}{\emph{Mechanism and Machine Theory (MMT)}}  \bibinfo{volume}{152} (\bibinfo{year}{2020}), \bibinfo{pages}{103947}.
\newblock
\href{https://doi.org/10.1016/j.mechmachtheory.2020.103947}{doi:\nolinkurl{10.1016/j.mechmachtheory.2020.103947}}


\bibitem[Filipov et~al\mbox{.}(2017)]%
        {filipov2017bar}
\bibfield{author}{\bibinfo{person}{Evgueni~T. Filipov}, \bibinfo{person}{Ke Liu}, \bibinfo{person}{Tomohiro Tachi}, \bibinfo{person}{Mark Schenk}, {and} \bibinfo{person}{Glaucio~H. Paulino}.} \bibinfo{year}{2017}\natexlab{}.
\newblock \showarticletitle{Bar and hinge models for scalable analysis of origami}.
\newblock \bibinfo{journal}{\emph{International Journal of Solids and Structures}}  \bibinfo{volume}{124} (\bibinfo{year}{2017}), \bibinfo{pages}{26--45}.
\newblock
\href{https://doi.org/10.1016/j.ijsolstr.2017.05.028}{doi:\nolinkurl{10.1016/j.ijsolstr.2017.05.028}}


\bibitem[Ghassaei et~al\mbox{.}(2018)]%
        {ghassaei2018fast}
\bibfield{author}{\bibinfo{person}{Amanda Ghassaei}, \bibinfo{person}{Erik~D. Demaine}, {and} \bibinfo{person}{Neil Gershenfeld}.} \bibinfo{year}{2018}\natexlab{}.
\newblock \showarticletitle{Fast, interactive origami simulation using {GPU} computation}.
\newblock In \bibinfo{booktitle}{\emph{Origami$^7$: Proceedings of the 7th International Meeting on Origami in Science, Mathematics and Education}}. Vol.~\bibinfo{volume}{4}. \bibinfo{publisher}{Tarquin}, \bibinfo{address}{Oxford, England}, \bibinfo{pages}{1151--1166}.
\newblock
\urldef\tempurl%
\url{https://erikdemaine.org/papers/OrigamiSimulator_Origami7/}
\showURL{%
\tempurl}


\bibitem[Han et~al\mbox{.}(2023)]%
        {Han2023ChartLlamaAM}
\bibfield{author}{\bibinfo{person}{Yucheng Han}, \bibinfo{person}{China.~Xiaoyan Zhang}, \bibinfo{person}{Xin Chen}, \bibinfo{person}{Xu Yang}, \bibinfo{person}{Zhibin Wang}, \bibinfo{person}{Gang Yu}, \bibinfo{person}{Bin Fu}, {and} \bibinfo{person}{Hanwang Zhang}.} \bibinfo{year}{2023}\natexlab{}.
\newblock \showarticletitle{ChartLlama: A Multimodal LLM for Chart Understanding and Generation}.
\newblock \bibinfo{journal}{\emph{ArXiv}}  \bibinfo{volume}{abs/2311.16483} (\bibinfo{year}{2023}).
\newblock
\urldef\tempurl%
\url{https://api.semanticscholar.org/CorpusID:265466206}
\showURL{%
\tempurl}


\bibitem[Hawkes et~al\mbox{.}(2010)]%
        {hawkes2010programmable}
\bibfield{author}{\bibinfo{person}{Elliot Hawkes}, \bibinfo{person}{Byoungkwon An}, \bibinfo{person}{Nadia~M. Benbernou}, \bibinfo{person}{Hiroto Tanaka}, \bibinfo{person}{Sangbae Kim}, \bibinfo{person}{Erik~D. Demaine}, \bibinfo{person}{Daniela Rus}, {and} \bibinfo{person}{Robert~J. Wood}.} \bibinfo{year}{2010}\natexlab{}.
\newblock \showarticletitle{Programmable matter by folding}.
\newblock \bibinfo{journal}{\emph{Proceedings of the National Academy of Sciences}} \bibinfo{volume}{107}, \bibinfo{number}{28} (\bibinfo{year}{2010}), \bibinfo{pages}{12441--12445}.
\newblock
\href{https://doi.org/10.1073/pnas.0914069107}{doi:\nolinkurl{10.1073/pnas.0914069107}}


\bibitem[Hu and Liang(2020)]%
        {hu2020folding}
\bibfield{author}{\bibinfo{person}{Yucai Hu} {and} \bibinfo{person}{Haiyi Liang}.} \bibinfo{year}{2020}\natexlab{}.
\newblock \showarticletitle{Folding simulation of rigid origami with Lagrange multiplier method}.
\newblock \bibinfo{journal}{\emph{International Journal of Solids and Structures}}  \bibinfo{volume}{202} (\bibinfo{year}{2020}), \bibinfo{pages}{552--561}.
\newblock
\href{https://doi.org/10.1016/j.ijsolstr.2020.06.016}{doi:\nolinkurl{10.1016/j.ijsolstr.2020.06.016}}


\bibitem[Huang et~al\mbox{.}(2026)]%
        {huang2026learn2fold}
\bibfield{author}{\bibinfo{person}{Yanjia Huang}, \bibinfo{person}{Yunuo Chen}, \bibinfo{person}{Ying Jiang}, \bibinfo{person}{Jinru Han}, \bibinfo{person}{Zhengzhong Tu}, \bibinfo{person}{Yin Yang}, {and} \bibinfo{person}{Chenfanfu Jiang}.} \bibinfo{year}{2026}\natexlab{}.
\newblock \bibinfo{title}{{Learn2Fold}: Structured Origami Generation with World Model Planning}.
\newblock
\showeprint[arxiv]{2603.29585}~[cs.GR]
\urldef\tempurl%
\url{https://arxiv.org/abs/2603.29585}
\showURL{%
\tempurl}


\bibitem[Kanade(1980)]%
        {kanade1980theory}
\bibfield{author}{\bibinfo{person}{Takeo Kanade}.} \bibinfo{year}{1980}\natexlab{}.
\newblock \showarticletitle{A Theory of Origami World}.
\newblock \bibinfo{journal}{\emph{Artificial Intelligence (AIJ)}} \bibinfo{volume}{13}, \bibinfo{number}{1--2} (\bibinfo{year}{1980}), \bibinfo{pages}{279--311}.
\newblock
\urldef\tempurl%
\url{https://www.ri.cmu.edu/publications/a-theory-of-origami-world/}
\showURL{%
\tempurl}


\bibitem[Kato et~al\mbox{.}(2025)]%
        {kato2025origami}
\bibfield{author}{\bibinfo{person}{Hitomi Kato}, \bibinfo{person}{Hirotaka Kato}, \bibinfo{person}{Takatsugu Hirayama}, \bibinfo{person}{Takahiro Komamizu}, {and} \bibinfo{person}{Ichiro Ide}.} \bibinfo{year}{2025}\natexlab{}.
\newblock \bibinfo{booktitle}{\emph{Origami Crease Recognition for Automatic Folding Diagrams Generation}}.
\newblock \bibinfo{publisher}{Springer Nature Singapore}, \bibinfo{address}{Singapore}, \bibinfo{pages}{16--31}.
\newblock
\href{https://doi.org/10.1007/978-981-95-4398-4_2}{doi:\nolinkurl{10.1007/978-981-95-4398-4_2}}


\bibitem[Konjevod and Kupre{\v{s}}anin(2009)]%
        {konjevod2009notation}
\bibfield{author}{\bibinfo{person}{Goran Konjevod} {and} \bibinfo{person}{Ana~Maria Kupre{\v{s}}anin}.} \bibinfo{year}{2009}\natexlab{}.
\newblock \showarticletitle{Notation for a Class of Paperfolded Models}. In \bibinfo{booktitle}{\emph{Proceedings of Bridges 2009: Mathematics, Music, Art, Architecture, Culture}}. \bibinfo{publisher}{Tarquin Group}, \bibinfo{address}{Banff, Alberta, Canada}, \bibinfo{pages}{47--54}.
\newblock
\urldef\tempurl%
\url{https://archive.bridgesmathart.org/2009/bridges2009-47.html}
\showURL{%
\tempurl}


\bibitem[Kraft(2016)]%
        {RabbitEar}
\bibfield{author}{\bibinfo{person}{Robby Kraft}.} \bibinfo{year}{2016}\natexlab{}.
\newblock \bibinfo{title}{Rabbit Ear: Computational Origami Library}.
\newblock \bibinfo{howpublished}{\url{https://github.com/rabbit-ear/rabbit-ear}}.
\newblock
\shownote{Accessed: 2026-05-06}.
\newblock


\bibitem[Ku(2022)]%
        {ku2022flatfolder}
\bibfield{author}{\bibinfo{person}{Jason~S. Ku}.} \bibinfo{year}{2022}\natexlab{}.
\newblock \bibinfo{title}{{Flat-Folder: A Crease Pattern Solver}}.
\newblock \bibinfo{howpublished}{\url{https://github.com/origamimagiro/flat-folder}}.
\newblock
\shownote{GitHub repository}.
\newblock


\bibitem[Kulits et~al\mbox{.}(2024)]%
        {kulits2024rethinking}
\bibfield{author}{\bibinfo{person}{Peter Kulits}, \bibinfo{person}{Haiwen Feng}, \bibinfo{person}{Weiyang Liu}, \bibinfo{person}{Victoria~Fern{\'a}ndez Abrevaya}, {and} \bibinfo{person}{Michael~J. Black}.} \bibinfo{year}{2024}\natexlab{}.
\newblock \showarticletitle{Re-Thinking Inverse Graphics With Large Language Models}.
\newblock \bibinfo{journal}{\emph{Trans. Mach. Learn. Res.}}  \bibinfo{volume}{2024} (\bibinfo{year}{2024}).
\newblock
\urldef\tempurl%
\url{https://api.semanticscholar.org/CorpusID:269302972}
\showURL{%
\tempurl}


\bibitem[Lal et~al\mbox{.}(2023)]%
        {lal2023unsupervised}
\bibfield{author}{\bibinfo{person}{Rohit Lal}, \bibinfo{person}{S. Ruphan}, \bibinfo{person}{C.~A.~O. Sifan}, \bibinfo{person}{Sishen Yuan}, \bibinfo{person}{Lalith}, \bibinfo{person}{Qiu Liang}, {and} \bibinfo{person}{Hongliang Ren}.} \bibinfo{year}{2023}\natexlab{}.
\newblock \showarticletitle{Unsupervised Intelligent Pose Estimation of Origami-Inspired Deployable Robots}.
\newblock In \bibinfo{booktitle}{\emph{Deployable Multimodal Machine Intelligence: Applications in Biomedical Engineering}}. \bibinfo{publisher}{Springer}, \bibinfo{address}{Singapore}, \bibinfo{pages}{569--589}.
\newblock
\href{https://doi.org/10.1007/978-981-19-5932-5_21}{doi:\nolinkurl{10.1007/978-981-19-5932-5_21}}


\bibitem[Lang(1996)]%
        {lang1996computational}
\bibfield{author}{\bibinfo{person}{Robert~J. Lang}.} \bibinfo{year}{1996}\natexlab{}.
\newblock \showarticletitle{A Computational Algorithm for Origami Design}. In \bibinfo{booktitle}{\emph{Proceedings of the Twelfth Annual Symposium on Computational Geometry (SoCG)}}. \bibinfo{publisher}{ACM Press}, \bibinfo{address}{New York, NY, USA}, \bibinfo{pages}{98--105}.
\newblock
\href{https://doi.org/10.1145/237218.237249}{doi:\nolinkurl{10.1145/237218.237249}}


\bibitem[Liu et~al\mbox{.}(2025)]%
        {liu2025ir3dbench}
\bibfield{author}{\bibinfo{person}{Parker Liu}, \bibinfo{person}{Chenxin Li}, \bibinfo{person}{Zhengxin Li}, \bibinfo{person}{Yipeng Wu}, \bibinfo{person}{Wuyang Li}, \bibinfo{person}{Zhiqin Yang}, \bibinfo{person}{Zhenyuan Zhang}, \bibinfo{person}{Yunlong Lin}, \bibinfo{person}{Sirui Han}, {and} \bibinfo{person}{Brandon~Y. Feng}.} \bibinfo{year}{2025}\natexlab{}.
\newblock \bibinfo{title}{{IR3D-Bench}: Evaluating Vision-Language Model Scene Understanding as Agentic Inverse Rendering}.
\newblock
\showeprint[arxiv]{2506.23329}~[cs.CV]
\urldef\tempurl%
\url{https://ir3d-bench.github.io/}
\showURL{%
\tempurl}


\bibitem[Lu et~al\mbox{.}(2025)]%
        {lu2025ll3m}
\bibfield{author}{\bibinfo{person}{Sining Lu}, \bibinfo{person}{Guan Chen}, \bibinfo{person}{Nam~Anh Dinh}, \bibinfo{person}{Itai Lang}, \bibinfo{person}{Ari Holtzman}, {and} \bibinfo{person}{Rana Hanocka}.} \bibinfo{year}{2025}\natexlab{}.
\newblock \bibinfo{title}{{LL3M}: Large Language 3D Modelers}.
\newblock
\showeprint[arxiv]{2508.08228}~[cs.GR]
\urldef\tempurl%
\url{https://threedle.github.io/ll3m/}
\showURL{%
\tempurl}


\bibitem[Maiti et~al\mbox{.}(2025)]%
        {maiti2025gen3deval}
\bibfield{author}{\bibinfo{person}{Shalini Maiti}, \bibinfo{person}{Lourdes Agapito}, {and} \bibinfo{person}{Filippos Kokkinos}.} \bibinfo{year}{2025}\natexlab{}.
\newblock \showarticletitle{{Gen3DEval}: Using {vLLMs} for Automatic Evaluation of Generated 3D Objects}. In \bibinfo{booktitle}{\emph{Proceedings of the IEEE/CVF Conference on Computer Vision and Pattern Recognition (CVPR)}}. \bibinfo{publisher}{IEEE Computer Society}, \bibinfo{address}{Los Alamitos, CA, USA}, \bibinfo{pages}{18552--18562}.
\newblock
\showeprint[arxiv]{2504.08125}~[cs.CV]
\href{https://doi.org/10.1109/CVPR52734.2025.01729}{doi:\nolinkurl{10.1109/CVPR52734.2025.01729}}


\bibitem[Mao et~al\mbox{.}(2015)]%
        {mao2015sequential}
\bibfield{author}{\bibinfo{person}{Yiqi Mao}, \bibinfo{person}{Kai Yu}, \bibinfo{person}{Michael~S. Isakov}, \bibinfo{person}{Jiangtao Wu}, \bibinfo{person}{Martin~L. Dunn}, {and} \bibinfo{person}{H.~Jerry Qi}.} \bibinfo{year}{2015}\natexlab{}.
\newblock \showarticletitle{Sequential self-folding structures by 3D printed digital shape memory polymers}.
\newblock \bibinfo{journal}{\emph{Scientific Reports}}  \bibinfo{volume}{5} (\bibinfo{year}{2015}), \bibinfo{pages}{13616}.
\newblock
\href{https://doi.org/10.1038/srep13616}{doi:\nolinkurl{10.1038/srep13616}}


\bibitem[Mitani(2007)]%
        {mitani2007development}
\bibfield{author}{\bibinfo{person}{Jun Mitani}.} \bibinfo{year}{2007}\natexlab{}.
\newblock \showarticletitle{Development of origami pattern editor ({ORIPA}) and a method for estimating a folded configuration of origami from the crease pattern}.
\newblock \bibinfo{journal}{\emph{IPSJ Journal}} \bibinfo{volume}{48}, \bibinfo{number}{9} (\bibinfo{year}{2007}), \bibinfo{pages}{3309--3317}.
\newblock
\urldef\tempurl%
\url{https://ipsj.ixsq.nii.ac.jp/records/9850}
\showURL{%
\tempurl}


\bibitem[Namiki and Yokosawa(2021)]%
        {namiki2021origami}
\bibfield{author}{\bibinfo{person}{Akio Namiki} {and} \bibinfo{person}{Shuichi Yokosawa}.} \bibinfo{year}{2021}\natexlab{}.
\newblock \showarticletitle{Origami Folding by Multifingered Hands with Motion Primitives}.
\newblock \bibinfo{journal}{\emph{Cyborg and Bionic Systems}}  \bibinfo{volume}{2021} (\bibinfo{year}{2021}).
\newblock
\href{https://doi.org/10.34133/2021/9851834}{doi:\nolinkurl{10.34133/2021/9851834}}


\bibitem[Narumi et~al\mbox{.}(2023)]%
        {narumi2023inkjet}
\bibfield{author}{\bibinfo{person}{Koya Narumi}, \bibinfo{person}{Kazuki Koyama}, \bibinfo{person}{Kai Suto}, \bibinfo{person}{Yuta Noma}, \bibinfo{person}{Hiroki Sato}, \bibinfo{person}{Tomohiro Tachi}, \bibinfo{person}{Masaaki Sugimoto}, \bibinfo{person}{Takeo Igarashi}, {and} \bibinfo{person}{Yoshihiro Kawahara}.} \bibinfo{year}{2023}\natexlab{}.
\newblock \showarticletitle{Inkjet {4D} print: Self-folding tessellated origami objects by inkjet {UV} printing}.
\newblock \bibinfo{journal}{\emph{ACM Transactions on Graphics}} \bibinfo{volume}{42}, \bibinfo{number}{4} (\bibinfo{year}{2023}), \bibinfo{pages}{1--13}.
\newblock
\href{https://doi.org/10.1145/3592409}{doi:\nolinkurl{10.1145/3592409}}


\bibitem[Oh et~al\mbox{.}(2015)]%
        {oh2015dissimilarity}
\bibfield{author}{\bibinfo{person}{Seung~Man Oh}, \bibinfo{person}{Godfried~T. Toussaint}, \bibinfo{person}{Erik~D. Demaine}, {and} \bibinfo{person}{Martin~L. Demaine}.} \bibinfo{year}{2015}\natexlab{}.
\newblock \showarticletitle{A Dissimilarity Measure for Comparing Origami Crease Patterns}. In \bibinfo{booktitle}{\emph{Proceedings of the International Conference on Pattern Recognition Applications and Methods (ICPRAM)}}. \bibinfo{publisher}{SCITEPRESS}, \bibinfo{address}{Setubal, Portugal}, \bibinfo{pages}{386--393}.
\newblock
\href{https://doi.org/10.5220/0005291203860393}{doi:\nolinkurl{10.5220/0005291203860393}}


\bibitem[{Origami Way}(2023)]%
        {origamiway}
\bibfield{author}{\bibinfo{person}{{Origami Way}}.} \bibinfo{year}{2023}\natexlab{}.
\newblock \bibinfo{title}{{Origami Instructions -- Origami Way}}.
\newblock \bibinfo{howpublished}{\url{https://www.origamiway.com/}}.
\newblock


\bibitem[Parodi and Torre(1995)]%
        {parodi1995complexity}
\bibfield{author}{\bibinfo{person}{Pietro Parodi} {and} \bibinfo{person}{Vincent Torre}.} \bibinfo{year}{1995}\natexlab{}.
\newblock \showarticletitle{The Complexity of Understanding Line Drawings of Origami Scenes}.
\newblock \bibinfo{journal}{\emph{International Journal of Computer Vision (IJCV)}}  \bibinfo{volume}{16} (\bibinfo{year}{1995}), \bibinfo{pages}{139--170}.
\newblock
\href{https://doi.org/10.1007/BF00055000}{doi:\nolinkurl{10.1007/BF00055000}}


\bibitem[Qattawi et~al\mbox{.}(2014)]%
        {qattawi2014design}
\bibfield{author}{\bibinfo{person}{Ala Qattawi}, \bibinfo{person}{Mahmoud Abdelhamid}, \bibinfo{person}{Ahmad Mayyas}, {and} \bibinfo{person}{Mohammed Omar}.} \bibinfo{year}{2014}\natexlab{}.
\newblock \showarticletitle{Design analysis for origami-based folded sheet metal parts}.
\newblock \bibinfo{journal}{\emph{SAE International Journal of Materials and Manufacturing}} \bibinfo{volume}{7}, \bibinfo{number}{2} (\bibinfo{year}{2014}), \bibinfo{pages}{488--498}.
\newblock
\href{https://doi.org/10.4271/2014-01-9098}{doi:\nolinkurl{10.4271/2014-01-9098}}


\bibitem[Ray et~al\mbox{.}(2024)]%
        {ray2024origami}
\bibfield{author}{\bibinfo{person}{Lala Ray}, \bibinfo{person}{Daniel Geissler}, \bibinfo{person}{Bo Zhou}, \bibinfo{person}{Paul Lukowicz}, {and} \bibinfo{person}{Berit Greinke}.} \bibinfo{year}{2024}\natexlab{}.
\newblock \showarticletitle{Origami Single-End Capacitive Sensing for Continuous Shape Estimation of Morphing Structures}.
\newblock \bibinfo{journal}{\emph{Scientific Reports (Sci. Rep.)}} \bibinfo{volume}{14}, \bibinfo{number}{1} (\bibinfo{year}{2024}).
\newblock
\href{https://doi.org/10.1038/s41598-024-67149-9}{doi:\nolinkurl{10.1038/s41598-024-67149-9}}


\bibitem[Sabbah(1985)]%
        {sabbah1985computing}
\bibfield{author}{\bibinfo{person}{Daniel Sabbah}.} \bibinfo{year}{1985}\natexlab{}.
\newblock \showarticletitle{Computing with Connections in Visual Recognition of Origami Objects}.
\newblock \bibinfo{journal}{\emph{Cognitive Science (Cogn. Sci.)}} \bibinfo{volume}{9}, \bibinfo{number}{1} (\bibinfo{year}{1985}), \bibinfo{pages}{25--50}.
\newblock
\href{https://doi.org/10.1016/S0364-0213(85)80008-2}{doi:\nolinkurl{10.1016/S0364-0213(85)80008-2}}


\bibitem[Sharma et~al\mbox{.}(2024)]%
        {sharma2024vision}
\bibfield{author}{\bibinfo{person}{Pratyusha Sharma}, \bibinfo{person}{Tamar~Rott Shaham}, \bibinfo{person}{Manel Baradad}, \bibinfo{person}{Stephanie Fu}, \bibinfo{person}{Adrian Rodriguez-Munoz}, \bibinfo{person}{Shivam Duggal}, \bibinfo{person}{Phillip Isola}, {and} \bibinfo{person}{Antonio Torralba}.} \bibinfo{year}{2024}\natexlab{}.
\newblock \showarticletitle{A vision check-up for language models}. In \bibinfo{booktitle}{\emph{Proceedings of the IEEE/CVF Conference on Computer Vision and Pattern Recognition}}. \bibinfo{publisher}{IEEE Computer Society}, \bibinfo{address}{Los Alamitos, CA, USA}, \bibinfo{pages}{14410--14419}.
\newblock


\bibitem[Shimanuki et~al\mbox{.}(2003)]%
        {shimanuki2003recognition}
\bibfield{author}{\bibinfo{person}{Hiroshi Shimanuki}, \bibinfo{person}{Jien Kato}, {and} \bibinfo{person}{Toyohide Watanabe}.} \bibinfo{year}{2003}\natexlab{}.
\newblock \showarticletitle{Recognition of Folding Process from Origami Drill Books}. In \bibinfo{booktitle}{\emph{Proceedings of the Seventh International Conference on Document Analysis and Recognition (ICDAR)}}. \bibinfo{publisher}{IEEE Computer Society}, \bibinfo{address}{Los Alamitos, CA, USA}, \bibinfo{pages}{550--554}.
\newblock
\href{https://doi.org/10.1109/ICDAR.2003.1227725}{doi:\nolinkurl{10.1109/ICDAR.2003.1227725}}


\bibitem[Shimanuki et~al\mbox{.}(2012)]%
        {shimanuki2012folding}
\bibfield{author}{\bibinfo{person}{Hiroshi Shimanuki}, \bibinfo{person}{Toyohide Watanabe}, \bibinfo{person}{Koichi Asakura}, \bibinfo{person}{Hideki Sato}, {and} \bibinfo{person}{Taketoshi Ushiama}.} \bibinfo{year}{2012}\natexlab{}.
\newblock \showarticletitle{Folding Support for Beginners Based on State Estimation of Origami}. In \bibinfo{booktitle}{\emph{2012 IEEE RO-MAN: The 21st IEEE International Symposium on Robot and Human Interactive Communication (RO-MAN)}}. \bibinfo{publisher}{IEEE}, \bibinfo{address}{Piscataway, NJ, USA}, \bibinfo{pages}{789--794}.
\newblock
\href{https://doi.org/10.1109/ROMAN.2012.6343855}{doi:\nolinkurl{10.1109/ROMAN.2012.6343855}}


\bibitem[Smith(1980)]%
        {smith1980pureland}
\bibfield{author}{\bibinfo{person}{John~S. Smith}.} \bibinfo{year}{1980}\natexlab{}.
\newblock \bibinfo{booktitle}{\emph{Pureland Origami}}.
\newblock Number~14 in \bibinfo{series}{BOS Booklet}. \bibinfo{publisher}{British Origami Society}, \bibinfo{address}{UK}. 48 pages.
\newblock
\urldef\tempurl%
\url{https://www.giladorigami.com/origami-database-book/2388/Pureland-Origami-by-John-Smith}
\showURL{%
\tempurl}


\bibitem[Smith(1989)]%
        {smith1989pureland2}
\bibfield{author}{\bibinfo{person}{John~S. Smith}.} \bibinfo{year}{1989}\natexlab{}.
\newblock \bibinfo{booktitle}{\emph{Pureland Origami 2}}.
\newblock Number~29 in \bibinfo{series}{BOS Booklet}. \bibinfo{publisher}{British Origami Society}, \bibinfo{address}{UK}. 72 pages.
\newblock
\urldef\tempurl%
\url{https://www.giladorigami.com/origami-database-book/2389/Pureland-Origami-2-by-John-Smith}
\showURL{%
\tempurl}


\bibitem[Smith(1993)]%
        {smith1993pureland3}
\bibfield{author}{\bibinfo{person}{John~S. Smith}.} \bibinfo{year}{1993}\natexlab{}.
\newblock \bibinfo{booktitle}{\emph{Pureland Origami 3}}.
\newblock Number~43 in \bibinfo{series}{BOS Booklet}. \bibinfo{publisher}{British Origami Society}, \bibinfo{address}{UK}. 52 pages.
\newblock
\urldef\tempurl%
\url{https://www.giladorigami.com/origami-database-book/2390/Pureland-Origami-3-by-John-Smith}
\showURL{%
\tempurl}


\bibitem[Spencer et~al\mbox{.}(2025)]%
        {spencer2025gamibench}
\bibfield{author}{\bibinfo{person}{Ryan Spencer}, \bibinfo{person}{Roey Yaari}, \bibinfo{person}{Ritvik Vemavarapu}, \bibinfo{person}{Joyce Yang}, \bibinfo{person}{Steven Ngo}, {and} \bibinfo{person}{Utkarsh Sharma}.} \bibinfo{year}{2025}\natexlab{}.
\newblock \bibinfo{title}{GamiBench: Evaluating Spatial Reasoning and 2D-to-3D Planning Capabilities of MLLMs with Origami Folding Tasks}.
\newblock
\showeprint[arxiv]{2512.22207}~[cs.AI]
\urldef\tempurl%
\url{https://arxiv.org/abs/2512.22207}
\showURL{%
\tempurl}


\bibitem[Sundaram et~al\mbox{.}(2017)]%
        {sundaram20173d}
\bibfield{author}{\bibinfo{person}{Subramanian Sundaram}, \bibinfo{person}{David~S. Kim}, \bibinfo{person}{Marc~A. Baldo}, \bibinfo{person}{Ryan~C. Hayward}, {and} \bibinfo{person}{Wojciech Matusik}.} \bibinfo{year}{2017}\natexlab{}.
\newblock \showarticletitle{{3D}-printed self-folding electronics}.
\newblock \bibinfo{journal}{\emph{ACS Applied Materials \& Interfaces}} \bibinfo{volume}{9}, \bibinfo{number}{37} (\bibinfo{year}{2017}), \bibinfo{pages}{32290--32298}.
\newblock
\href{https://doi.org/10.1021/acsami.7b10443}{doi:\nolinkurl{10.1021/acsami.7b10443}}


\bibitem[Tachi(2009)]%
        {tachi2009simulation}
\bibfield{author}{\bibinfo{person}{Tomohiro Tachi}.} \bibinfo{year}{2009}\natexlab{}.
\newblock \showarticletitle{Simulation of rigid origami}.
\newblock In \bibinfo{booktitle}{\emph{Origami 4: Fourth International Meeting of Origami Science, Mathematics, and Education}}, \bibfield{editor}{\bibinfo{person}{Robert~J. Lang}} (Ed.). \bibinfo{publisher}{A K Peters}, \bibinfo{address}{Natick, MA}, \bibinfo{pages}{175--187}.
\newblock
\urldef\tempurl%
\url{https://origami.c.u-tokyo.ac.jp/~tachi/cg/SimulationOfRigidOrigami_tachi_4OSME.pdf}
\showURL{%
\tempurl}


\bibitem[Tachi(2010)]%
        {TachiFreeformOrigami2010}
\bibfield{author}{\bibinfo{person}{Tomohiro Tachi}.} \bibinfo{year}{2010}\natexlab{}.
\newblock \showarticletitle{Freeform Variations of Origami}.
\newblock \bibinfo{journal}{\emph{Journal for Geometry and Graphics}} \bibinfo{volume}{14}, \bibinfo{number}{2} (\bibinfo{year}{2010}), \bibinfo{pages}{203--215}.
\newblock
\urldef\tempurl%
\url{https://tsg.ne.jp/TT/cg/TachiFreeformOrigami2010.pdf}
\showURL{%
\tempurl}


\bibitem[Tachi and Hull(2017)]%
        {tachi2017selffoldability}
\bibfield{author}{\bibinfo{person}{Tomohiro Tachi} {and} \bibinfo{person}{Thomas~C. Hull}.} \bibinfo{year}{2017}\natexlab{}.
\newblock \showarticletitle{Self-Foldability of Rigid Origami}.
\newblock \bibinfo{journal}{\emph{Journal of Mechanisms and Robotics (JMR)}} \bibinfo{volume}{9}, \bibinfo{number}{2} (\bibinfo{year}{2017}), \bibinfo{pages}{021008}.
\newblock
\href{https://doi.org/10.1115/1.4035558}{doi:\nolinkurl{10.1115/1.4035558}}


\bibitem[Vinker et~al\mbox{.}(2025)]%
        {vinker2025sketchagent}
\bibfield{author}{\bibinfo{person}{Yael Vinker}, \bibinfo{person}{Tamar Rott~Shaham}, \bibinfo{person}{Kristine Zheng}, \bibinfo{person}{Alex Zhao}, \bibinfo{person}{Judith~E. Fan}, {and} \bibinfo{person}{Antonio Torralba}.} \bibinfo{year}{2025}\natexlab{}.
\newblock \showarticletitle{SketchAgent: Language-Driven Sequential Sketch Generation}. In \bibinfo{booktitle}{\emph{Proceedings of the IEEE/CVF Conference on Computer Vision and Pattern Recognition (CVPR)}}. \bibinfo{publisher}{IEEE Computer Society}, \bibinfo{address}{Los Alamitos, CA, USA}.
\newblock
\showeprint[arxiv]{2411.17673}~[cs.CV]
\href{https://doi.org/10.1109/CVPR52734.2025.02175}{doi:\nolinkurl{10.1109/CVPR52734.2025.02175}}


\bibitem[Wu et~al\mbox{.}(2024)]%
        {wu2024gpt4v}
\bibfield{author}{\bibinfo{person}{Tong Wu}, \bibinfo{person}{Guandao Yang}, \bibinfo{person}{Zhibing Li}, \bibinfo{person}{Kai Zhang}, \bibinfo{person}{Ziwei Liu}, \bibinfo{person}{Leonidas Guibas}, \bibinfo{person}{Dahua Lin}, {and} \bibinfo{person}{Gordon Wetzstein}.} \bibinfo{year}{2024}\natexlab{}.
\newblock \showarticletitle{{GPT-4V(ision)} is a Human-Aligned Evaluator for Text-to-3D Generation}. In \bibinfo{booktitle}{\emph{Proceedings of the IEEE/CVF Conference on Computer Vision and Pattern Recognition (CVPR)}}. \bibinfo{publisher}{IEEE Computer Society}, \bibinfo{address}{Los Alamitos, CA, USA}, \bibinfo{pages}{22227--22238}.
\newblock
\showeprint[arxiv]{2401.04092}~[cs.CV]
\href{https://doi.org/10.1109/CVPR52733.2024.02098}{doi:\nolinkurl{10.1109/CVPR52733.2024.02098}}


\bibitem[Xu et~al\mbox{.}(2025)]%
        {xu2025origamispace}
\bibfield{author}{\bibinfo{person}{Rui Xu}, \bibinfo{person}{Dakuan Lu}, \bibinfo{person}{Zicheng Zhao}, \bibinfo{person}{Xiaoyu Tan}, \bibinfo{person}{Xintao Wang}, \bibinfo{person}{Siyu Yuan}, \bibinfo{person}{Jiangjie Chen}, {and} \bibinfo{person}{Yinghui Xu}.} \bibinfo{year}{2025}\natexlab{}.
\newblock \showarticletitle{{ORIGAMISPACE}: Benchmarking Multimodal {LLMs} in Multi-Step Spatial Reasoning with Mathematical Constraints}. In \bibinfo{booktitle}{\emph{The Thirty-ninth Annual Conference on Neural Information Processing Systems (NeurIPS)}}. \bibinfo{publisher}{Curran Associates, Inc.}, \bibinfo{address}{Red Hook, NY, USA}.
\newblock
\shownote{NeurIPS 2025 spotlight}.
\newblock
\urldef\tempurl%
\url{https://openreview.net/forum?id=y7ahj9RoXQ}
\showURL{%
\tempurl}


\bibitem[Yao et~al\mbox{.}(2023)]%
        {yao2023react}
\bibfield{author}{\bibinfo{person}{Shunyu Yao}, \bibinfo{person}{Jeffrey Zhao}, \bibinfo{person}{Dian Yu}, \bibinfo{person}{Nan Du}, \bibinfo{person}{Izhak Shafran}, \bibinfo{person}{Karthik~R. Narasimhan}, {and} \bibinfo{person}{Yuan Cao}.} \bibinfo{year}{2023}\natexlab{}.
\newblock \showarticletitle{{ReAct}: Synergizing Reasoning and Acting in Language Models}. In \bibinfo{booktitle}{\emph{International Conference on Learning Representations (ICLR)}}. \bibinfo{publisher}{OpenReview.net}, \bibinfo{address}{Amherst, MA, USA}.
\newblock
\showeprint[arxiv]{2210.03629}~[cs.CL]
\urldef\tempurl%
\url{https://openreview.net/forum?id=WE_vluYUL-X}
\showURL{%
\tempurl}


\bibitem[Yasuda et~al\mbox{.}(2020)]%
        {yasuda2020data}
\bibfield{author}{\bibinfo{person}{Hiromi Yasuda}, \bibinfo{person}{Koshiro Yamaguchi}, \bibinfo{person}{Yasuhiro Miyazawa}, \bibinfo{person}{Richard Wiebe}, \bibinfo{person}{Jordan~R. Raney}, {and} \bibinfo{person}{Jinkyu Yang}.} \bibinfo{year}{2020}\natexlab{}.
\newblock \showarticletitle{Data-driven prediction and analysis of chaotic origami dynamics}.
\newblock \bibinfo{journal}{\emph{Communications Physics}}  \bibinfo{volume}{3} (\bibinfo{year}{2020}), \bibinfo{pages}{168}.
\newblock
\href{https://doi.org/10.1038/s42005-020-00431-0}{doi:\nolinkurl{10.1038/s42005-020-00431-0}}


\bibitem[Yin et~al\mbox{.}(2026)]%
        {yin2026viga}
\bibfield{author}{\bibinfo{person}{Shaofeng Yin}, \bibinfo{person}{Jiaxin Ge}, \bibinfo{person}{Zora~Zhiruo Wang}, \bibinfo{person}{Chenyang Wang}, \bibinfo{person}{Xiuyu Li}, \bibinfo{person}{Michael~J. Black}, \bibinfo{person}{Trevor Darrell}, \bibinfo{person}{Angjoo Kanazawa}, {and} \bibinfo{person}{Haiwen Feng}.} \bibinfo{year}{2026}\natexlab{}.
\newblock \bibinfo{title}{Vision-as-Inverse-Graphics Agent via Interleaved Multimodal Reasoning}.
\newblock
\showeprint[arxiv]{2601.11109}~[cs.CV]
\urldef\tempurl%
\url{https://fugtemypt123.github.io/VIGA/}
\showURL{%
\tempurl}


\bibitem[Zhu et~al\mbox{.}(2010)]%
        {zhu2010origami}
\bibfield{author}{\bibinfo{person}{Kening Zhu}, \bibinfo{person}{Owen Noel~Newton Fernando}, \bibinfo{person}{Adrian~David Cheok}, \bibinfo{person}{Mark Fiala}, {and} \bibinfo{person}{Theam~Wei Yang}.} \bibinfo{year}{2010}\natexlab{}.
\newblock \showarticletitle{Origami Recognition System Using Natural Feature Tracking}. In \bibinfo{booktitle}{\emph{2010 9th IEEE International Symposium on Mixed and Augmented Reality (ISMAR)}}. \bibinfo{publisher}{IEEE}, \bibinfo{address}{Piscataway, NJ, USA}, \bibinfo{pages}{289--290}.
\newblock
\href{https://doi.org/10.1109/ISMAR.2010.5643611}{doi:\nolinkurl{10.1109/ISMAR.2010.5643611}}


\bibitem[Zhu and Filipov(2022)]%
        {zhu2022harnessing}
\bibfield{author}{\bibinfo{person}{Yi Zhu} {and} \bibinfo{person}{Evgueni~T. Filipov}.} \bibinfo{year}{2022}\natexlab{}.
\newblock \showarticletitle{Harnessing Interpretable Machine Learning for Holistic Inverse Design of Origami}.
\newblock \bibinfo{journal}{\emph{Scientific Reports (Sci. Rep.)}}  \bibinfo{volume}{12} (\bibinfo{year}{2022}).
\newblock
\href{https://doi.org/10.1038/s41598-022-23875-6}{doi:\nolinkurl{10.1038/s41598-022-23875-6}}


\end{thebibliography}

\ifarxiv \else
    \newpage
    
\begin{figure*}
    \centering
    \includegraphics[width=0.85\linewidth]{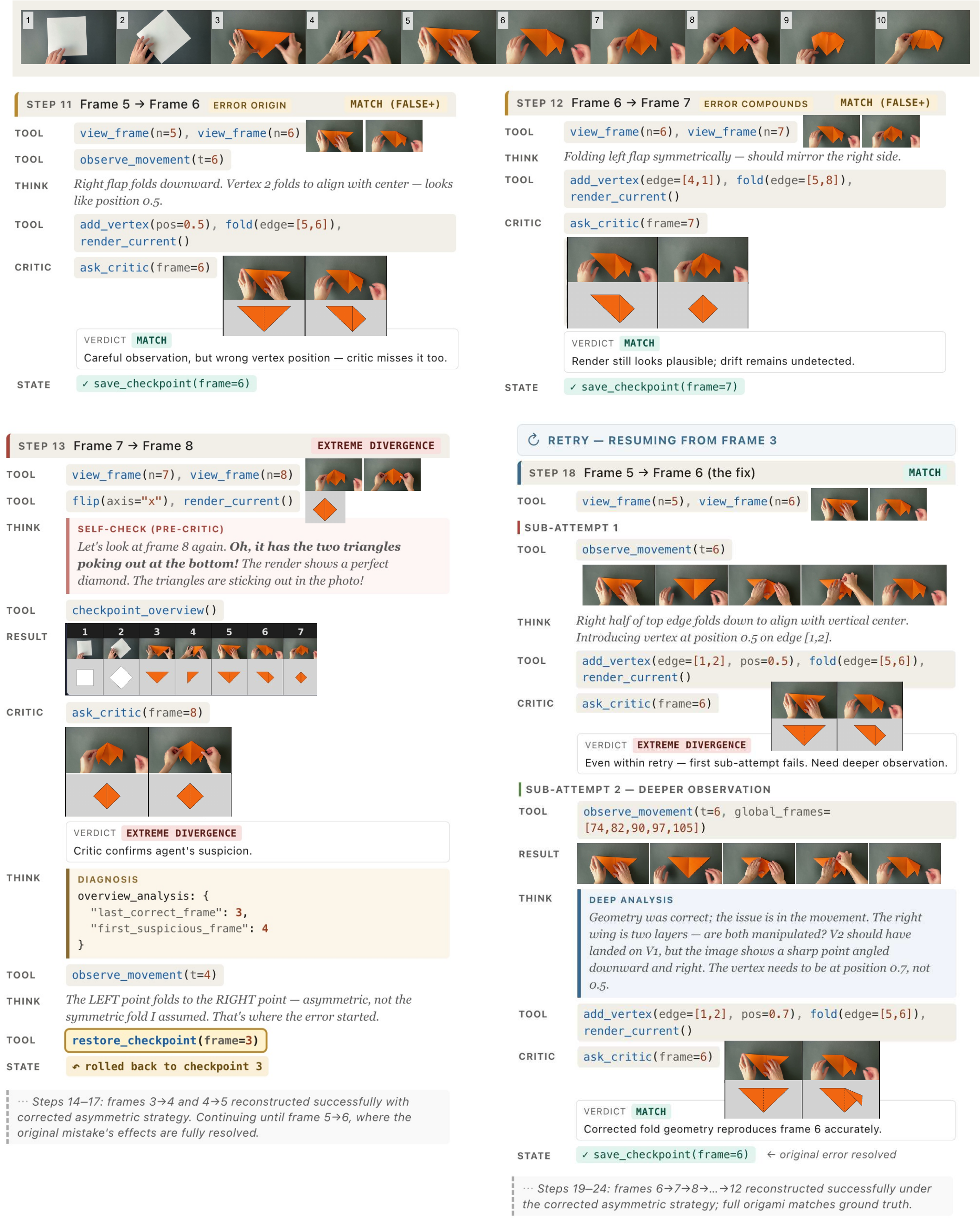}
    \Description{Recovery from critic-approved errors.}
    \caption{\emph{Recovery from critic-approved errors.} The top row shows the full keyframe sequence, while the four panels below illustrate selected reasoning steps. An incorrect fold at frame $5\rightarrow6$ is approved by the critic and propagates to frame 7. The agent detects the divergence at frame $7\rightarrow8$, rolls back to an earlier checkpoint, and reconstructs the sequence using additional tool calls until it finds a corrected fold that remains consistent through the final frame.}
    \label{fig:analysis}
\end{figure*}

\begin{figure*}
    \centering
    \includegraphics[width=1\linewidth]{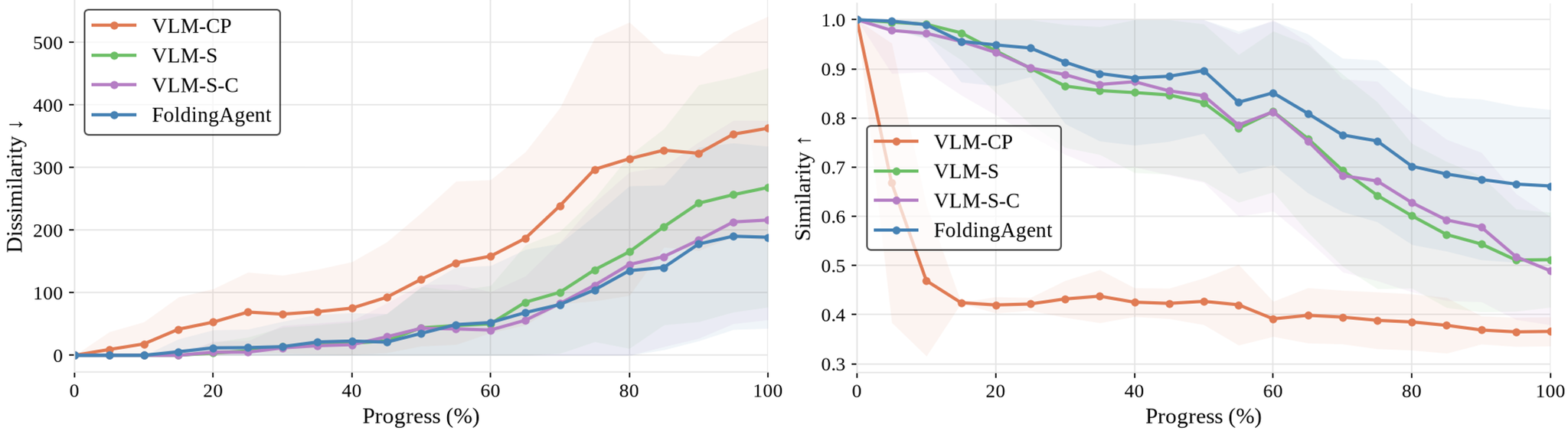}
    \Description{Performance}
    \caption{\emph{Performance of our method and ablation variants as a function of sequence progress.} We report mean Dissimilarity and Similarity scores averaged over completed sequences. As seen, the performance of all methods declines as the sequence progresses, yet our method exhibits a steadier slope with a notable gap to the others, which increases as the complexity grows.}
    \label{fig:graphs}
\end{figure*}

\fi

\ifappendix
    \appendix
    \crefalias{section}{appendix} %
    \crefalias{subsection}{appendix}
    \crefalias{subsubsection}{appendix}    
    \newpage
    \section*{Appendix}  %

This \ifappendix{appendix} \else{supplementary material}\fi provides additional details beyond those in the main paper. While the main paper stands on its own, the details given here may shed more light. 

\ifappendix 
\Cref{sec:full_results} provides access to the full reconstruction results and agent traces. Implementation and dataset construction are detailed in \cref{sec:implementation_details,sec:dataset_details}, and \cref{sec:prompts} presents the complete prompts. Additional quantitative experiments and metric definitions appear in \cref{sec:quantitative_details}. Finally, \cref{sec:qual_res_cont,sec:user_study_screenshot} provide additional qualitative results and a visualization of the user study.
\fi

\section{Full Results} \label{sec:full_results}
\begin{figure*}[t]
    \centering
    \includegraphics[width=\linewidth]{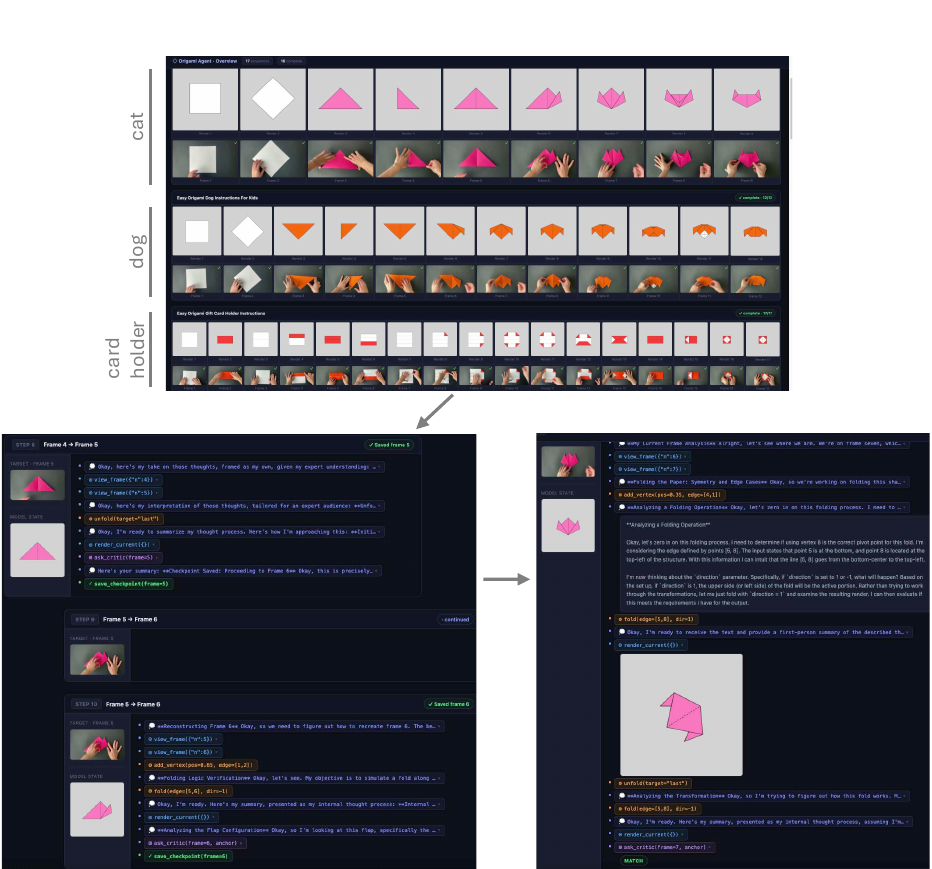}
    \caption{\emph{Results dashboard.} The main screen (top) shows all reconstructed sequences. Each row corresponds to one folding sequence, with dataset keyframes shown below
the rendered geometries. Users can scroll through the sequences to inspect the diversity of examples and the quality of reconstructions. Clicking a sequence opens a detailed trace view (bottom left), which logs all agent interactions. Folding steps are shown with nested indentation, and each interaction element, such as a tool call or response, can be expanded to
show details including agent reasoning, requested images or renders, and critic calls and responses. In the example shown (bottom right), the agent's reasoning and the response to the \texttt{render\_current} tool are expanded. After observing the render, the agent recognizes that the current fold is
insufficient, continues with additional actions, and eventually reaches
the correct fold.
}
    \label{fig:dashboard}
    \Description[]{}  %
\end{figure*}

All results are available 
\ifanonymous
on our supplementary website. To access the website, extract the supplementary ZIP file and open the HTML file.
\else
via the \emph{Results} link on our project page: https://maya-moriya.github.io/origami-page.
\fi
It presents a wide range of folding sequences, along with the full agent interaction for each sample. A visualization of the results page and a description of its usage are shown
in \cref{fig:dashboard}.

\section{Implementation Details} \label{sec:implementation_details}
Both the main agent and the visual critic are powered by Gemini 3.1 Pro
Preview (\texttt{gemini-3.1-pro-preview}). The agent interacts with the
controller exclusively through Gemini's native function-calling API. All
tools, such as \texttt{fold} and \texttt{save\_checkpoint}, are declared
as typed JSON schemas and dispatched by the controller. This allows the agent to issue multiple tool calls within a single iteration and to chain
actions across iterations based on returned results.

\paragraph{\textbf{Token budget.}} 
Each sequence is limited to 300 iterations, with a maximum thinking budget of 8,192 tokens per iteration.

\paragraph{\textbf{Running time.}} Across folding sequences, running times range from 9 to 96 minutes, with an average of 31 and a median of 20 minutes.

\paragraph{\textbf{Thinking-token exhaustion.}} 
Although internal thinking improves our framework's performance, we encountered a developer-reported failure mode in which reasoning tokens can consume the available token budget before a useful response is produced.\footnote{ \url{https://discuss.ai.google.dev/t/thinking-ate-all-the-tokens-and-hit-max-tokens/110971}} In such cases, the request terminates with \texttt{MAX\_TOKENS}, often after producing long, uninformative reasoning traces. We treat excessive thinking-token consumption as a practical failure mode of long agentic runs: when it occurs, we roll back to the beginning of the current folding step and discard the partial actions generated within that step.

\subsection{Controller}

The controller is implemented as a deterministic execution layer around Gemini's function-calling interface. In addition to the high-level role described in the main paper, the controller is responsible for validating each emitted tool call against its JSON schema, converting valid calls into backend-specific function invocations, and serializing all returned outputs into the next agent message. Invalid or malformed calls are rejected before reaching the simulator, and the resulting error message is returned to the agent as part of the interaction history.

\paragraph{\textbf{Context management.}}
Each folding step begins with a fresh context buffer initialized by a text-only summary of all previous steps. Within a step, the controller stores the full interaction history, including agent responses, tool calls, backend outputs, renders, critic verdicts, and checkpoint operations. At each iteration, this accumulated step history is sent back to the agent together with the result of the most recent tool call, if one exists. 

\paragraph{\textbf{State bookkeeping.}}
The controller keeps a synchronized record of the simulator state, checkpoint table, restore counts, and action history. Checkpoint inspection is read-only, whereas checkpoint restoration replaces the current simulator state and increments a persistent restore counter for that checkpoint. These counters are used for logging and diagnosis; the agent receives them as additional context when deciding whether to continue retrying or roll back further.

\paragraph{\textbf{Logging and recovery.}}
All interactions are written to a structured log, including the raw function call arguments, backend return values, rendered image paths, critic outputs, and rollback events. When a request terminates early, fails schema validation, or exhausts its token budget, the controller records the failure and restarts from the beginning of the current folding step, preserving only the previously saved checkpoints.

\subsection{State and Action Representation}

Two details of our representation are specific to reconstruction from visual demonstrations. First, we explicitly model \texttt{rotate} and \texttt{flip}. These operations are usually irrelevant for purely geometric origami analysis, since they do not change the intrinsic crease pattern, but they are necessary here because the demonstrator may reorient or turn over the paper between video frames.

Second, \texttt{fold} is specified by an existing edge rather than by an arbitrary line. The same geometric line can correspond to different folds depending on which faces and layers move. 
Using an edge anchors the command to the current topology, making clear which faces and layers of the current geometry are affected.

\subsection{Simulator}
\label{sec:simulator}
The simulator is a deterministic geometric engine that executes a single action (\ifappendix{\cref{sec:actions}}\else{Sec.~4.2}\fi) on a state $S=(V,F,E,O,L)$ and returns the resulting state $S'$. Recall that $V,F,E,O,L$ denote vertices, faces, edges, orientation, and layering, respectively (\ifappendix{\cref{sec:representation}}\else{Sec.~4.1}\fi). We describe the implementation of \texttt{fold} in detail below; \texttt{unfold} uses similar machinery, and \texttt{rotate} and \texttt{flip} are straightforward. \cref{fig:simulator_algo} demonstrates a simple folding example.

\paragraph{\textbf{Preliminaries and Notation.}}
Let $e$ be the fold edge and $d\in\{-1, +1\}$ the fold direction argument. $e$ is defined by two vertices in $V$, not necessarily in the same face. Let $\ell$ denote the abstract line through $e$ along which the paper is creased. Every edge borders one or two faces; we denote these as the edge's positive-side face (upper/right) and negative-side (lower/left) face. The direction $d$ defines one of these two sides as the moving side of the fold. We define the \emph{moving set} as the set of planes that will move as part of this fold (built below). A \emph{plane} is a set of faces connected by flat edges. A \emph{layer} is a set of planes that are disjoint and share the same topological depth, i.e., lie above the same set of planes and below another shared set of planes. We represent $L$ internally as a directed acyclic graph $G_P=(V_P, E_P)$ over planes, where $V_P$ consists of planes, and a relation $p_i \rightarrow p_j$ denotes that plane $p_i$ lies above plane $p_j$, for $p_i, p_j \in V_P$. This graph can easily be topologically sorted to a list representation $L$, which is more interpretable by the agent.

\paragraph{\textbf{Split faces and planes that straddle $\ell$}}
Each face that straddles $\ell$ is subdivided into a static sub-face and a moving sub-face; each plane containing both moving and static faces is subdivided into a static sub-plane (keeping its static faces) and a moving sub-plane (keeping its moving faces).

\paragraph{\textbf{Moving Set.}}
We use the term \emph{moving set} to denote the subset of planes that gets split in the fold. The \emph{moving set} is initialized by the plane(s) containing the endpoints of $e$. Starting from this set, the simulator repeatedly adds planes to the moving set: any plane that shares an edge lying on the moving side of the fold with a plane already in the moving set; any plane sandwiched between two planes already in the moving set, even if it does not itself share such an edge with them.

\paragraph{\textbf{Mirroring vertices.}}
Each vertex in the \emph{moving set} is updated so that its new location is the mirror image of its original location across $\ell$. See \cref{fig:simulator_algo}(D).

\paragraph{\textbf{Assigning plane relations.}}
Planes in the \emph{moving set} are turned over; existing relations in $G_P$ between two moving planes are reversed. In addition, new relations are added to $G_P$ placing moving sub-planes above their former neighbors on the static side of $\ell$.

\begin{figure}[ht]
    \centering
    \includegraphics[width=\columnwidth]{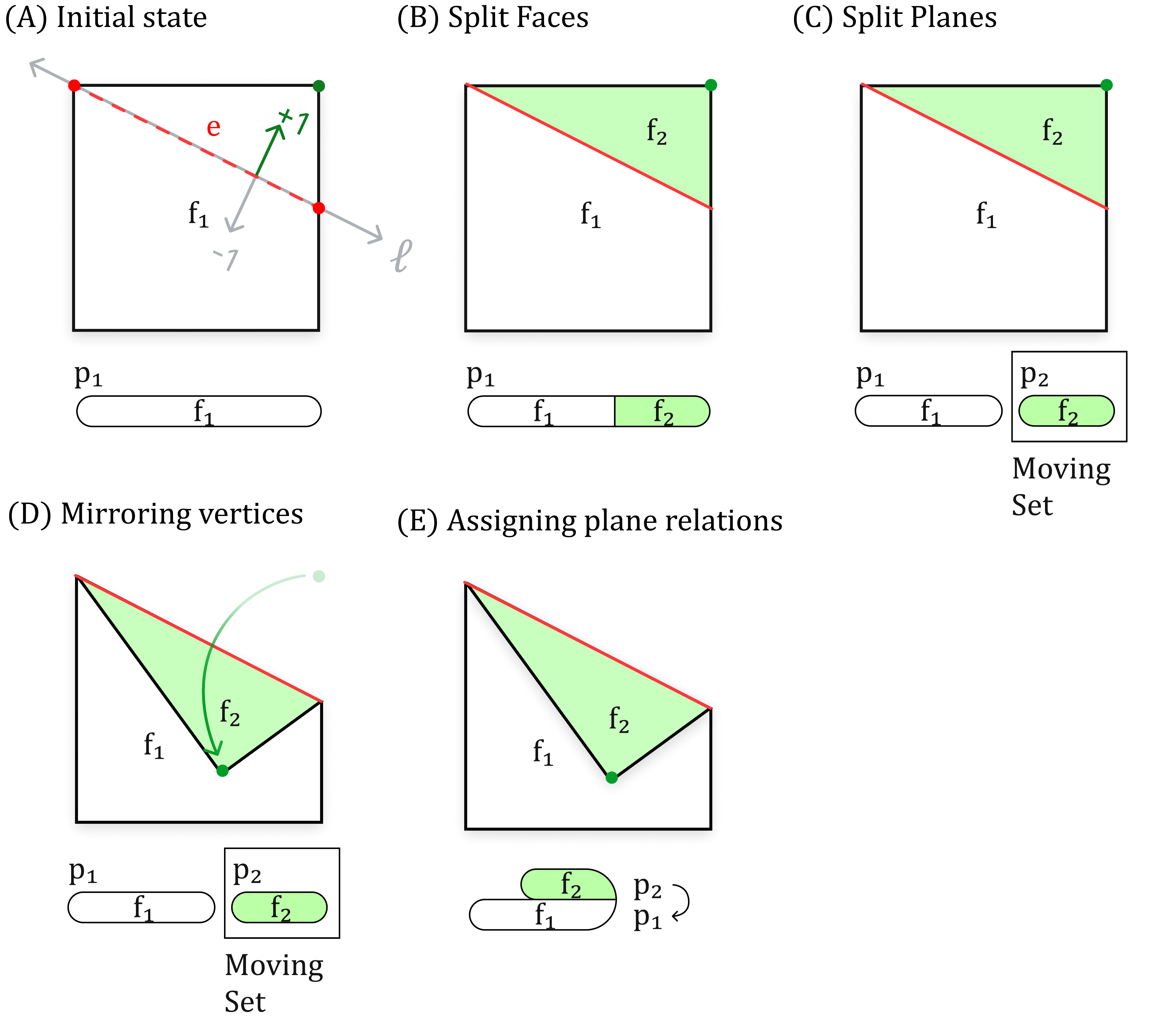}
    \caption{%
    \emph{Detailed run of} \texttt{fold}\emph{$(e, +1)$.}
    (A) \textit{Initial state}: a square face $f_1$ lies in a single plane $p_1$. The fold edge $e$ is depicted by the dashed red line, together with the crease line $\ell$ and its perpendicular directions $+1/-1$. The moving side is colored in green.
    (B) \textit{Split faces}: $f_1$ is subdivided along $\ell$ into a static sub-face $f_1$ and a moving sub-face $f_2$, shown in green.
    (C) \textit{Split planes}: since $f_1$ and $f_2$ originate from the same plane $p_1$, the plane is likewise split: $f_1$ remains on $p_1$, while $f_2$ is assigned to a new plane $p_2$, which constitutes the \emph{moving set} for this fold.
    (D) \textit{Mirroring vertices}: all vertices belonging to the moving sub-plane $p_2$ are reflected across $\ell$.
    (E) \textit{Assigning plane relations}: the layering graph $G_P$ is updated so that $p_2$ (containing the folded face $f_2$) is recorded as lying above $p_1$.
    }
    \label{fig:simulator_algo}
    \Description[]{}  %
\end{figure}

\subsection{Critic}

Conceptually, the critic performs a relational comparison: it checks whether the simulated transition from source render to candidate render matches the observed transition from source photo to target photo. Thus, the critic does not judge the candidate render in isolation, but evaluates whether the same geometric change is expressed in the photo pair and in the simulator pair. Visual illustrations of such relational comparisons, as well as examples for critic verification, are provided in the main paper.

\paragraph{\textbf{Interaction with Agent}}
The agent may invoke \texttt{ask\_critic} at any point during an iteration, and is explicitly instructed to do so once it believes the current action sequence matches the target frame. The intended control flow is to save a checkpoint after a positive critic verdict and to retry or roll back after a negative one. In practice, however, the agent retains discretion: it sometimes saves a checkpoint without an additional critic call, or decides whether to follow a critic verdict based on the accompanying analysis and its own visual reasoning. This is useful because the critic, while generally reliable, can occasionally produce both false positives and false negatives.

Each call to the critic uses the following inputs.
\begin{description}[leftmargin=0pt,labelindent=0pt,itemindent=5pt]
    \item[\textbf{Images.}] The critic receives four images: the source photo, target photo, source render, and candidate render. In practice, these images are passed separately; the controller assembles the grid only when returning the critic's response to the agent.
    \item[\textbf{Action classes.}] The predicted transition type, such as \texttt{fold}, \texttt{unfold}, \texttt{rotate}, or \texttt{flip}.
    \item[\textbf{Geometry JSON.}] The source and candidate geometry states.
    \item[\textbf{Fold anchor.}] An optional local relation for fold actions, specified by a moving element, a static reference, and their expected relationship. For example, a fold anchor may specify that the upper-left triangular flap is the moving element, the right vertical edge of the base layer is the static reference, and the flap's free edge should align with that edge while preserving a small visible gap near the top corner.

\end{description}

The critic responds with these details:
\begin{description}[leftmargin=0pt,labelindent=0pt,itemindent=5pt]
    \item[\textbf{Verdict.}] One of \textsc{Match}, \textsc{Mismatch}, or \textsc{Extreme Divergence}.
    \item[\textbf{Analysis.}] A short textual comparison focused on geometric agreement, including silhouette, flap proportions, fold-line positions, contact topology, gaps, offsets, and layer order.
    \item[\textbf{Anchor check.}] A sentence indicating whether the fold anchor is satisfied, or that no anchor was provided.
    \item[\textbf{Discrepancies.}] A list of concrete geometric issues used by the agent to revise the next attempt.
\end{description}

\subsection{Selector}
\label{sec:selector}
The selector is a lightweight VLM-based comparator, assisting the bounded exploration scheme described in the main paper. The selector is shown a target video keyframe, denoted A, together with several candidate diagrams, denoted B1, B2, \ldots, and selects the candidate whose shape, folds, and proportions best match the target. Its response is constrained to the label of a single candidate, with no additional text or reasoning. The complete system and user prompts used by the selector are provided at the end.

\section{Dataset Construction Details} \label{sec:dataset_details}
We captured the videos for the \benchname dataset using a front-facing LED ring light with a phone
holder, together with ambient daylight from a nearby window. 
Filming was done with an ordinary iPhone 12 device.
Folding actions were performed on black Bristol board to provide a high-contrast background.
Keyframes representing geometric transitions were manually extracted from each video, and the corresponding actions were annotated. Action classification was usually straightforward, whereas determining numerical action parameters, such as vertex locations, required an iterative trial-and-error process.
Altogether, our dataset comprises 27 sequences, with an average length of 12 frames, ranging from 5 to 21 frames.

\section{Prompts} \label{sec:prompts}
All prompts are provided in full at the end.
\subsection{Agent Prompts}
The system prompt provides a comprehensive specification of the agent's role, environment, available actions, and expected reconstruction behavior. It consists of:
\begin{description}[leftmargin=0pt,labelindent=0pt,itemindent=5pt]
    \item[\textbf{Task definition.}] Defines the overall goal: reconstructing an origami folding sequence from real-world photos by producing matching simulator states and saving them as checkpoints.
    \item[\textbf{State schema.}] Specifies the simulator representation exposed to the agent, including vertices, faces, face orientations, edge labels, and layer ordering.
    \item[\textbf{Tool library.}] Lists the available tools for viewing frames, observing motion, querying and rendering states, applying folding actions, saving and restoring checkpoints, and invoking the critic.
    \item[\textbf{Critic context.}] Explains the critic's role, the image grid used for verification, the meaning of critic verdicts, and how the agent should use critic feedback.
    \item[\textbf{Operational guidelines.}] Provides practical instructions for reconstruction, including using renders for visual feedback, preferring precise vertex placement over idealized symmetry, and reconsidering assumptions after repeated mismatches.
\end{description}

The user prompts provide step-specific execution instructions. The first-step prompt initializes the run: it tells the agent that frame 1 is the initial flat square and asks it to save the initial checkpoint and then process the transition from frame 1 to 2. Later-step prompts are used after a checkpoint has already been saved; they state the next target frame, report how many frames remain, and provide a compact history log after context trimming.

\subsection{Critic Prompts}
The critic system prompt defines the critic's role as an origami-specific visual verifier. It specifies the four-image comparison setup, the diagram visual language, the geometric criteria to inspect, the fold anchor mechanism, the allowed verdicts, and the required JSON output format. 

The critic user prompt provides the instance-specific information for a particular query: the source and candidate geometry JSONs, the optional fold anchor to verify, and the instruction to compare the candidate render against the target photo and return a structured verdict.

\ifarxiv
    \section{More Quantitative Results and Details} \label{sec:quantitative_details}

    \subsection{Detailed Metrics}

\else
    \section{More Quantitative Details} \label{sec:quantitative_details}
    \subsection{Metrics}
\fi

We adopt metrics from OrigamiSpace \cite{xu2025origamispace} and \citet{oh2015dissimilarity}. The former presents four metrics to measure origami similarity, and the latter measures origami dissimilarity. Neither work provides code; therefore, we use our own implementation based on their description.

Let $N$ denote the number of folding sequences in the benchmark, and let $K_n$ denote the number of folding steps in sequence $n$.
Let $S_t^n, S_t^{n\prime}$ denote the predicted and ground-truth geometric descriptions of state $t$ in sequence $n$, respectively, where $t \in \{1, \ldots, K_n\}$ and $n\in\{1, \ldots, N\}$.
In the following, $\langle\mathrm{FUNC}\rangle_1(\cdot)$ denotes a generic per-state metric, while $\langle\mathrm{FUNC}\rangle(\cdot)$ denotes its average over all completed sequences.

\begin{description}[leftmargin=0pt,labelindent=0pt,itemindent=5pt]
    \item 
    \textbf{Compilation Validity (CV).}
    Verify whether the generated geometric representation can be processed by the origami compiler. 
    Successful compilation requires the geometric description to be syntactically valid, geometrically foldable, and free of self-intersections and to produce a definite folded state.
    We compute a binary score: $\text{CV}_1(S_t^n) \in \{0,1\}$, and report the average score on all data.
    \begin{equation}
        \mathrm{CV}
        = \frac{1}{N} \sum_{n=1}^{N}
          \frac{1}{K_n} \sum_{t=1}^{K_n}
          \mathrm{CV_1}(S_t^n).
    \end{equation}
    For consistency with OrigamiSpace, we use the Flat-Folder compiler~\cite{ku2022flatfolder} to measure compilability.
\end{description}

For the OrigamiSpace similarity metrics, we provide only the high-level equations and refer readers to the original formulation for complete details.
When compilation fails, the subsequent GS, CS, and FFS similarity metrics default to penalty values of $0.2$, $0.2$, and $0.3$, respectively, enabling evaluation across all completed sequences. 
\begin{description}[leftmargin=0pt,labelindent=0pt,itemindent=5pt]
    \item
    \textbf{Topological Structure Similarity (TSS).}
    Compare the topological structure of the generated and ground-truth patterns in terms of:
    (a) number of vertices (NV);
    (b) edge connectivity (EC), measured by the similarity of the vertex-degree distribution and the number of connected components; 
    (c) face relationships (FR), \eg, number of faces, distribution of face sizes; and
    (d) distributional similarity of crease types (DSC), where types are mountain, valley, etc.
    \begin{equation}
    \begin{aligned}
    \mathrm{TSS}_1(S_t^n, S_t^{n\prime})
    = {}& 0.2 \cdot \mathrm{NV}(S_t^n, S_t^{n\prime})
       + 0.3 \cdot \mathrm{EG}(S_t^n, S_t^{n\prime}) \\
      &+ 0.3 \cdot \mathrm{FR}(S_t^n, S_t^{n\prime})
       + 0.2 \cdot \mathrm{DSC}(S_t^n, S_t^{n\prime}),\\
    \mathrm{TSS}
    = {}& \frac{1}{N} \sum_{n=1}^{N}
          \mathrm{TSS}_1(S_{K_n}^n, S_{K_n}^{n\prime}).
    \end{aligned}
    \end{equation}
    \item
    \textbf{Geometric Similarity (GS).}
    Compare the spatial characteristics of the generated and ground-truth origami states, in terms of:
    (a) point position similarity (PPS), measured by calculating the bidirectional Hausdorff distance between their normalized 3D point sets;
    (b) angular similarity (AS), by comparing the distribution of dihedral angles at the creases; and
    (c) size and proportion similarity (SPS), by comparing the aspect ratios of the overall bounding boxes of the models.
    \begin{equation}
    \begin{aligned}
    \mathrm{GS}_1(S_t^n, S_t^{n\prime})
    = {}& 0.4 \cdot \mathrm{PPS}(S_t^n, S_t^{n\prime})
       + 0.3 \cdot \mathrm{AS}(S_t^n, S_t^{n\prime}) \\
      &+ 0.3 \cdot \mathrm{SPS}(S_t^n, S_t^{n\prime}),\\
    \mathrm{GS}
    = {}& \frac{1}{N} \sum_{n=1}^{N}
          \mathrm{GS}_1(S_{K_n}^n, S_{K_n}^{n\prime}).
    \end{aligned}
    \end{equation}
    \item
    \textbf{Constraint Satisfaction (CS).}
    Evaluate whether the generated geometry satisfies key mathematical origami constraints:
    (a) presence and matching degree of critical constraint types, such as Taco-Taco (TT), Taco-Tortilla (TTo), and transitivity constraints (Trans); and
    (b) satisfaction of fundamental local flat-foldability theorems (FF), such as Maekawa's theorem and Kawasaki's theorem.
    \begin{equation}
    \begin{aligned}
    \mathrm{CS}_1(S_t^n, S_t^{n\prime})
    = {}& 0.3 \cdot \mathrm{TT}(S_t^n, S_t^{n\prime})
       + 0.3 \cdot \mathrm{TTo}(S_t^n, S_t^{n\prime}) \\
      &+ 0.2 \cdot \mathrm{Trans}(S_t^n, S_t^{n\prime})
       + 0.2 \cdot \mathrm{FF}(S_t^n, S_t^{n\prime}) \\
    \mathrm{CS}
    = {}& \frac{1}{N} \sum_{n=1}^{N}
          \mathrm{CS}_1(S_{K_n}^n, S_{K_n}^{n\prime}).
    \end{aligned}
    \end{equation}
    \item
    \textbf{Final Folded State (FFS).}
    Compare the final folded state of the generated and reference geometries in terms of:
    (a) Hausdorff distance (HD) between their point sets; and
    (b) layering consistency relationships (LCR) between faces.
    \begin{equation}
    \begin{aligned}
    \mathrm{FFS}_1(S_t^n, S_t^{n\prime})
    = {}& 0.7 \cdot \mathrm{HD}(S_t^n, S_t^{n\prime})
       + 0.3 \cdot \mathrm{LCR}(S_t^n, S_t^{n\prime}) \\
    \mathrm{FFS}
    = {}& \frac{1}{N} \sum_{n=1}^{N}
          \mathrm{FFS}_1(S_{K_n}^n, S_{K_n}^{n\prime}).
    \end{aligned}
    \end{equation}
\end{description}

Finally, for dissimilarity, we use the metric of \citet{oh2015dissimilarity}.
\begin{description}[leftmargin=0pt,labelindent=0pt,itemindent=5pt]
    \item 
    \textbf{Crease-Pattern Dissimilarity (CPD).} 
    Measure dissimilarity between the crease-patterns (CP) extracted from the generated and reference geometries.
    We follow the geometric graph dissimilarity measure of \citet{oh2015dissimilarity}, which compares two CPs by constructing separate bipartite graphs for edges and vertices and computing the corresponding minimum-weight matchings:
    (a) an edge-matching cost (EC), based on differences in crease length, orientation, spatial distance, and crease type; and
    (b) a vertex-matching cost (VC), based on differences in vertex degree and position.
    \begin{equation}
    \begin{aligned}
    \mathrm{CPD}_1(S_t^n, S_t^{n\prime})
    = {}& EC(S_t^n, S_t^{n\prime}) + VC(S_t^n, S_t^{n\prime}), \\
    \mathrm{CPD}
    = {}& \frac{1}{N} \sum_{n=1}^{N}
          \mathrm{CPD}_1(S_{K_n}^n, S_{K_n}^{n\prime}),
    \end{aligned}
    \end{equation}
    \end{description}

\section{More Qualitative Results} \label{sec:qual_res_cont}

\subsection{Failure Cases} \label{sec:failure}
\begin{figure}[t]
    \centering
    \includegraphics[width=\columnwidth]{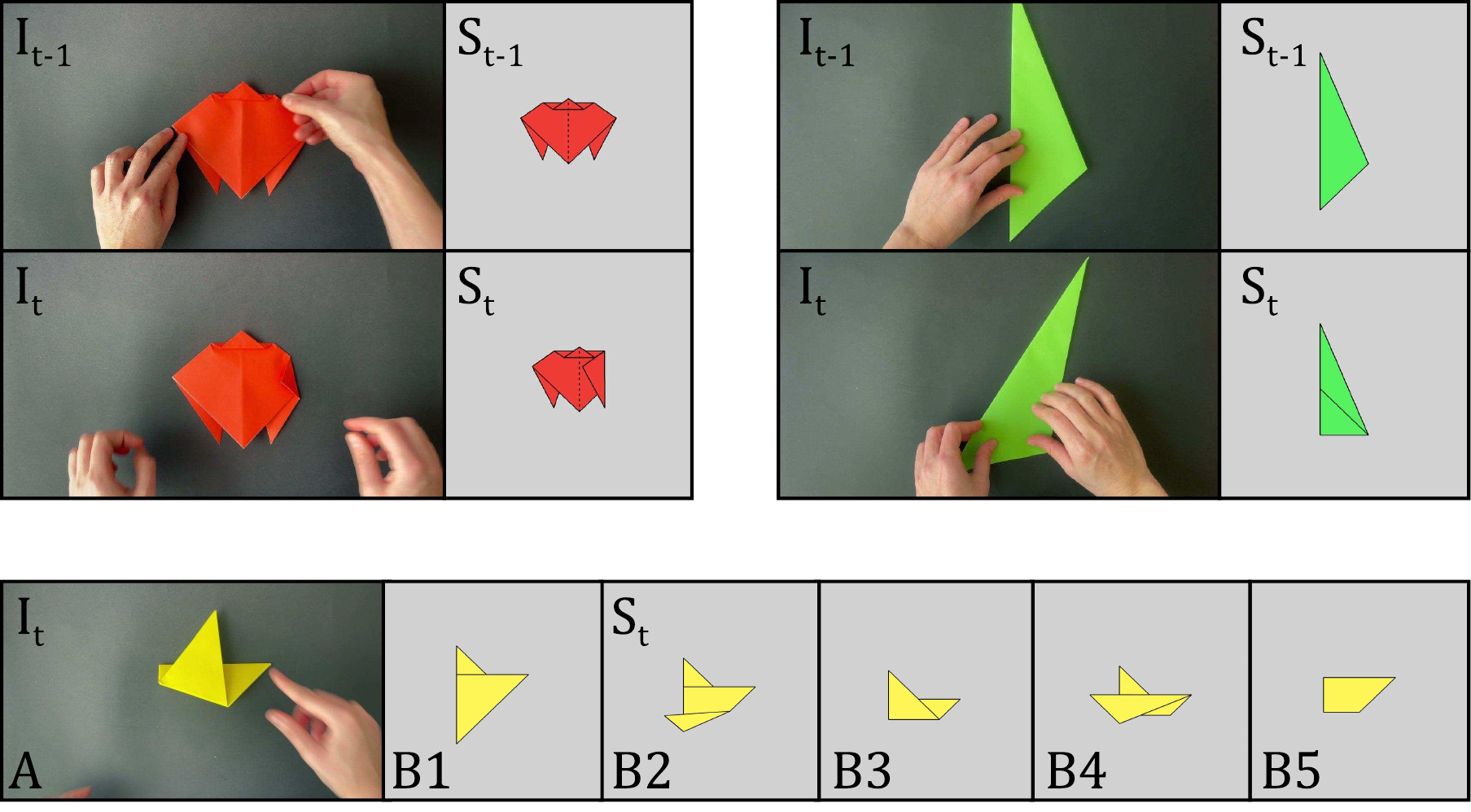}
    \caption{\emph{Failure cases.} 
    The top two rows show the source and target keyframes alongside their corresponding source and target geometries. The bottom row shows the target frame alongside $C=5$ candidates. Top left: the agent reuses an existing vertex instead of adding a new one, leading to incorrect crease placement. Top right: hand occlusion makes the target fold ambiguous, causing an incorrect reconstruction. Bottom: the selector must choose from $C$ bad candidates, resulting in a poor source for the next step.
    }
    \label{fig:failure_cases}
    \Description[]{}  %
\end{figure}

\cref{fig:failure_cases} shows representative failure cases. In addition to the limitations discussed in the main paper, we observe several recurring agent-level errors. First, the agent sometimes prefers to reuse existing vertices even when the target fold clearly terminates along the interior of an edge. This avoids an \texttt{add\_vertex} call and produces incorrect flap proportions or crease placement. 
Second, when bounded exploration produces only poor candidates, the selector must still choose among them. The selected state then becomes the source for the next transition, potentially propagating the error to subsequent steps.

\begin{figure}[t]
    \centering
    \includegraphics[width=\columnwidth]{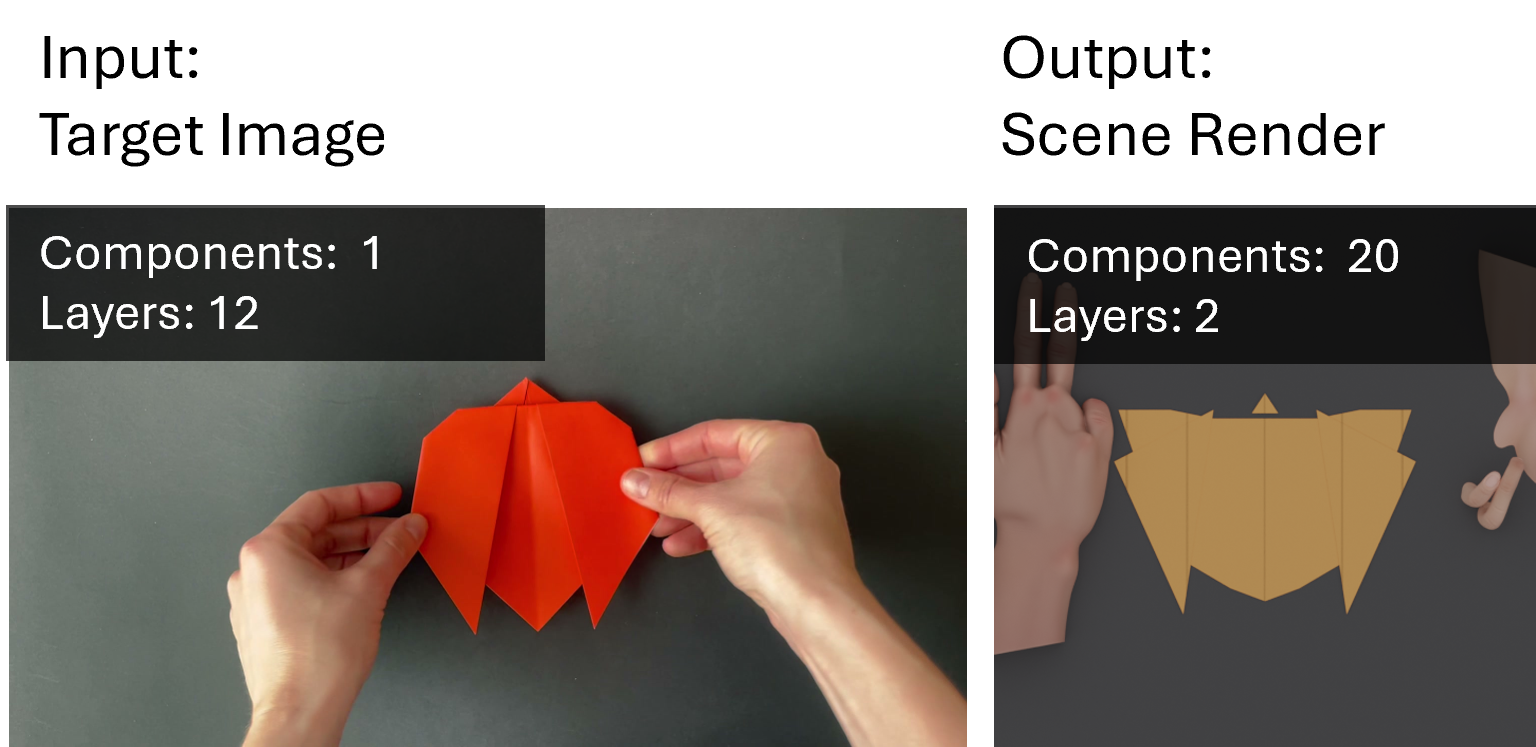}
    \caption{
    \emph{VIGA -- reconstruction from a single final keyframe.} 
    Left: input target keyframe with ground-truth component/layer counts. Right: VIGA's final scene render. Despite 96 iterations, VIGA fragments the connected sheet into 20 disconnected components with only 2 overlapping layers, failing to recover the paper's single-sheet, multi-layer structure.
    }
    \label{fig:viga:scene_recon}
    \Description[]{}  %
\end{figure}

\begin{figure}[t]
    \centering
    \includegraphics[width=\columnwidth]{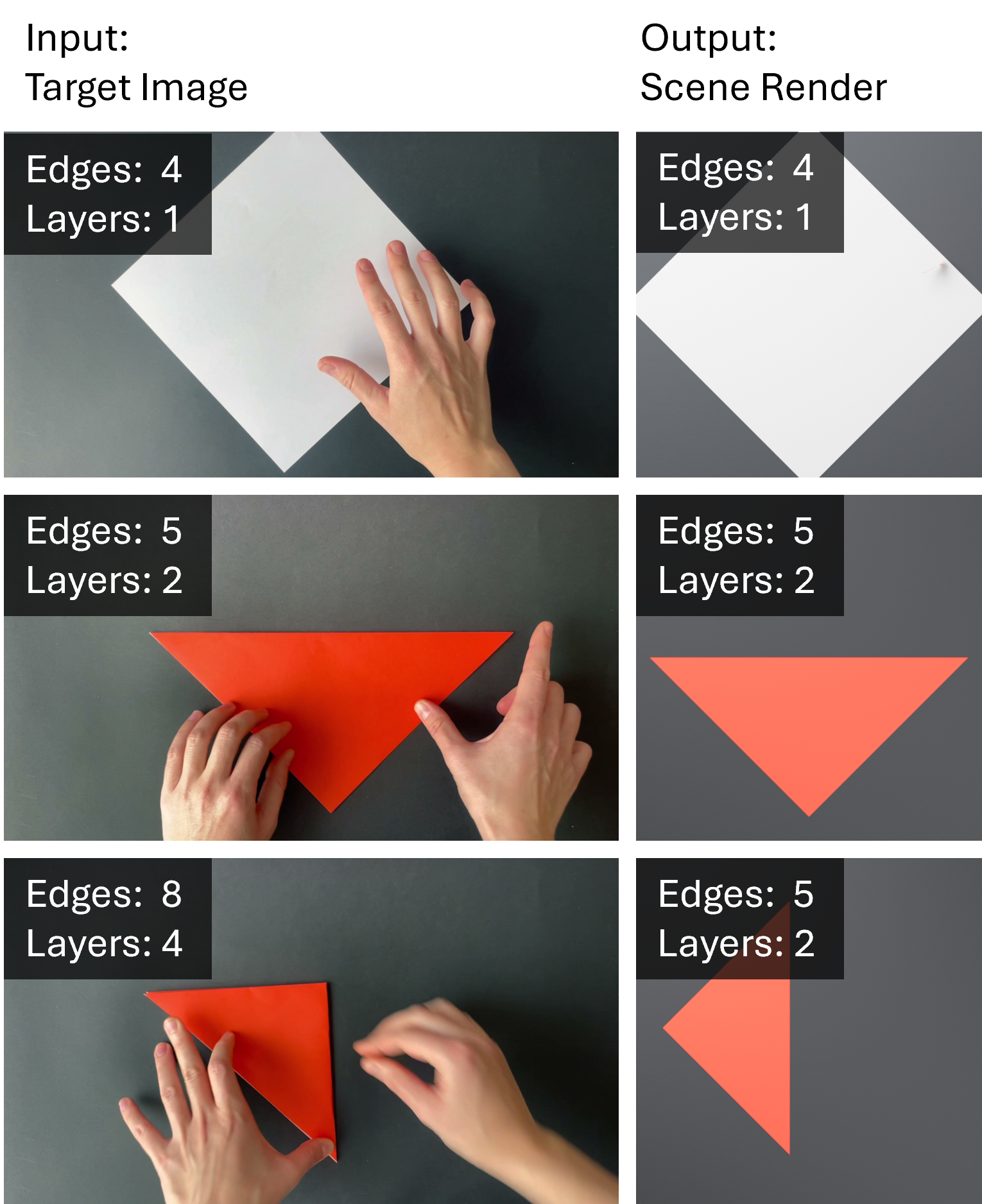}
    \caption{
    \emph{VIGA -- scene edit results.} 
    Each row shows the input target frame (left) with ground-truth edge/layer counts, alongside VIGA's rendered result (right). 
    Top: reconstruction of the flat sheet (10 iterations) is an exact match. Middle: the first fold (16 iterations) matches geometry and topology. 
    Bottom: as early as the second fold, the reconstruction stalls. The agent halts after 3 iterations with unchanged edge counts, unable to recover the additional crease and deeper layering required by the target.
    }
    \label{fig:viga:scene_edit}
    \Description[]{}  %
\end{figure}

\subsection{Comparison with VIGA}
\label{sec:viga}

VIGA \cite{yin2026viga} generates general 3D scenes in Blender through iterative code generation and visual feedback. We use it to assess the extent to which paper geometry can be recovered without an origami-specific representation. 

Since VIGA is not directly designed for our task, we evaluate it in two settings: (i) scene reconstruction from the final keyframe, and (ii) sequential scene editing across keyframes. The first evaluates the recovery of paper geometry from a single image; the second evaluates whether the geometry and accumulated layer structure can be maintained throughout a folding sequence.

Since VIGA's output comprises geometric entities unrelated to folding, we prompted it to output meshes only, allowing us to interpret the resulting geometry as folded paper. In both settings, however, the outputs violate key origami requirements, such as a single square sheet, straight edges and no face intersections.
Both experiments use VIGA's default configuration: 100 iterations, GPT-5, and a memory window of 12. The VIGA prompts used in our experiments are given at the end.

\textbf{Scene reconstruction.} 
First, we applied VIGA to the final folded keyframe of a sequence. The generated result resembles the observed outer shape but lacks internal structure.
VIGA's scene representation, built for composing independent rigid assets, has no notion of a single deformable sheet that must stay connected through creases and stacked layers, as demonstrated in \cref{fig:viga:scene_recon}. The agent stopped early after 96 iterations without converging. 

\textbf{Scene editing.} 
Next, we used VIGA's editing mode, reconstructing the first keyframe and sequentially editing the reconstruction to match each subsequent keyframe, mirroring our own protocol. However, after a few steps VIGA discarded the internal structure and flattened it, as shown in \cref{fig:viga:scene_edit}.

\section{User Study - Screenshot Visualization} \label{sec:user_study_screenshot}
\begin{figure}[t]
    \centering
    \includegraphics[width=.9\columnwidth]{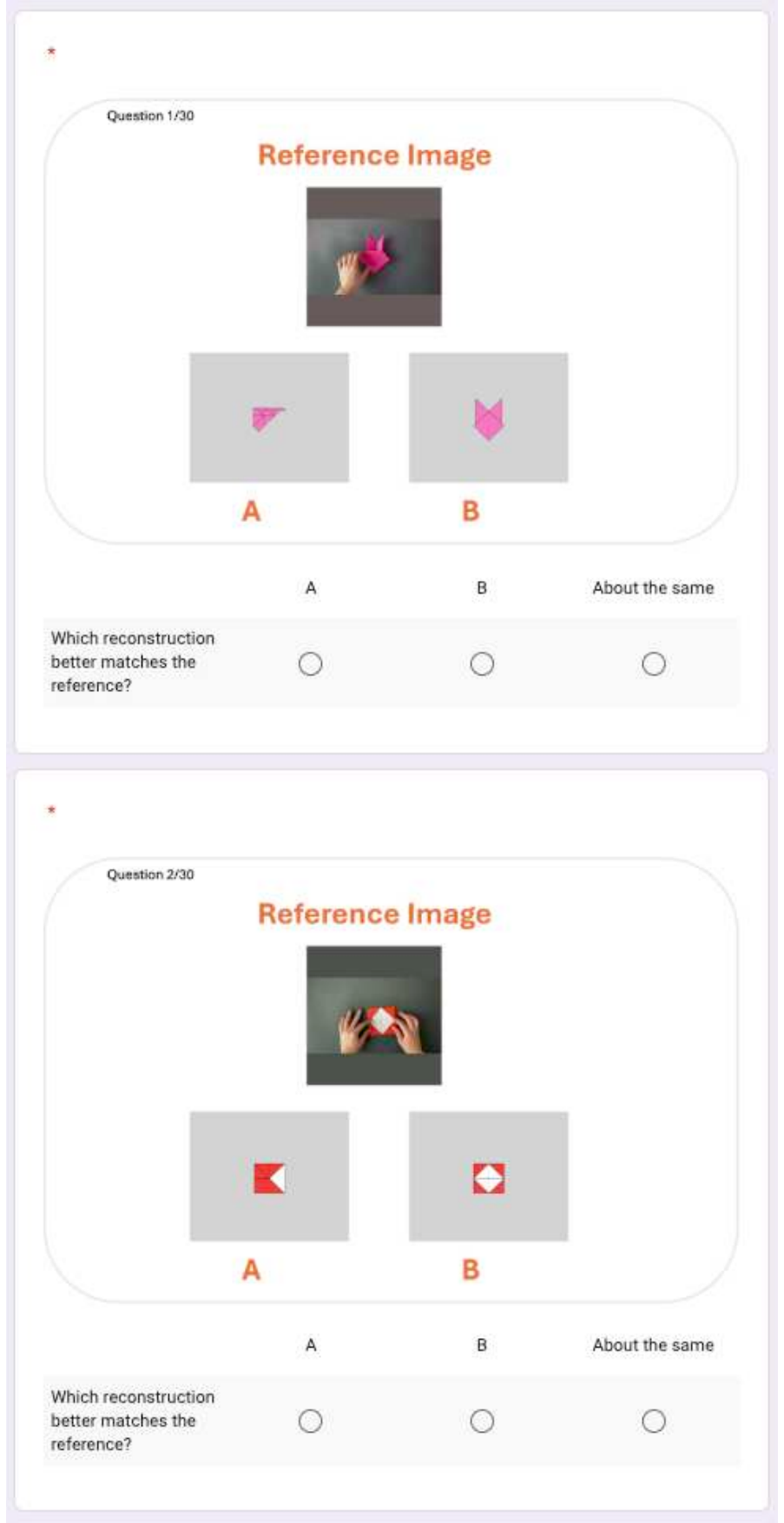}
    \caption{\emph{A screenshot of our user study.} Participants are presented with two geometric representations and asked to select the one that more closely matches a given video keyframe.}
    \label{fig:user_study_screenshot}
    \Description[]{}  %
\end{figure}

Our user study presents a keyframe from an instructional origami-folding video alongside a pair of rendered geometries and asks the user to select the geometry that more closely matches the video keyframe. A representative screenshot is shown in \cref{fig:user_study_screenshot}.

\begin{promptbox}
[FoldingAgent -- System Prompt]
=== YOUR GOAL ===

You are an origami reconstruction agent. You have access to a sequence of real-world photos showing an origami model being folded step-by-step. Your goal is to reproduce this exact folding sequence in a physics-based simulator.

Success means: for every photo in the sequence, you create a matching simulator state and save it as a checkpoint.


=== WHAT YOU HAVE ===

- A collection of real-world photos (frames) numbered 1 to N
- A simulator that starts with a flat square sheet of paper
- Tools to view photos, manipulate the simulator, render diagrams, and verify results


=== SIMULATOR STATE ===

The simulator represents origami as a mesh with these fields:

**vertices** : `{ "<id>": [x, y], ... }`
    2D coordinates (x rightward, y upward, origin at bottom-left)

**faces** : `{ "<id>": [v1, v2, v3, ...], ... }`
    Each face is a counter-clockwise ordered list of vertex IDs

**faces_orientations** : `{ "<id>": 0 | 1 }`
    0 = front side up, 1 = back side up

**edges** : `[ [u, v, kind], ... ]`
    kind in {"B" boundary, "F" flat, "M" mountain, "V" valley}

**layers** : `[ [ [face_id, ...], [face_id, ...] ], ... ]`
    Stack from bottom to top. Each layer contains one or more
    planes (sets of faces joined by flat edges, representing
    continuous sheets at the same altitude).

Coordinate system: Cartesian plane, not image pixels. X increases left-to-right, Y increases bottom-to-top.


=== AVAILABLE TOOLS ===

----------
[VIEWING]
----------

view_frame(n)
  Returns the real-world photo for frame n.

observe_movement(t)
  For step t >= 2, returns a PNG filmstrip showing the motion 
  between selected frames t-1 and t. Use this when the action 
  between those frames is ambiguous.

get_current_state()
  Returns the current simulator state JSON (vertices, faces, 
  edges, etc.)

get_checkpoint_state(n)
  Returns the saved state for frame n without changing current 
  state.

render_current()
  Returns a diagram of the current simulator state.
  Visual language:
    - Gray background
    - {front_color} = front side facing up
    - {back_color} = back side facing up
    - Solid lines = active folds and boundaries
    - Dashed lines = crease marks (unfolded creases)

---------------------
[SIMULATION ACTIONS]
---------------------

add_vertex(edge, position)
  Subdivides edge [u,v] at fractional position (0.0 to 1.0).
  Returns new_vertex_id for use in subsequent folds.

fold(edge, direction)
  Folds along edge [v1, v2].
  direction = +1 or -1 (which side of the crease moves toward 
  viewer)
    Non-vertical: +1 = upper side moves, -1 = lower side moves
    Vertical: +1 = right side moves, -1 = left side moves

unfold(edge) / unfold(target='last')
  Reverses a fold, leaving a crease mark.

rotate(angle)
  Rotates entire model (degrees, positive = counter-clockwise).

flip(axis)
  Flips model about axis in {'x', 'y', 'y=x', 'y=-x'}.

--------------
[CHECKPOINTS]
--------------

save_checkpoint(n)
  Saves current simulator state as frame n.

restore_checkpoint(n)
  Restores simulator to the saved state for frame n.
  Returns next_attempt_index for tracking retries.

---------------
[VERIFICATION]
---------------

ask_critic(frame_n, action_classes, fold_anchor=None)
  Sends a 4-image grid to a visual critic:
    A = real photo of frame (n-1)   B = real photo of frame n
    C = diagram of frame (n-1)      D = diagram of current state

  action_classes: list of transition types, e.g. ["fold"], 
  ["rotate"], ["rotate", "flip"]

  fold_anchor (optional, required if "fold" in action_classes):
  {
  "moving_element": "<what moves>",
  "static_reference": "<what stays fixed>",
  "relationship": "<measurable relationship in target photo>"
  }

  Returns:
    {
      "critic_verdict": "EXACT_MATCH" | "MISMATCH" | "EXTREME 
      DIVERGENCE",
      "critic_analysis": "<prose comparison of D vs B>",
      "critic_anchor_check": "<whether anchor is satisfied>",
      "critic_discrepancies": [<list of geometric issues>],
      "transition_attempt": <retry count for this transition>
    }

  Verdicts:
    EXACT_MATCH - D matches B well enough to save checkpoint
    MISMATCH - retry with different parameters
    EXTREME DIVERGENCE - false assumption, consider rollback

generate_checkpoint_overview()
  Creates a 2-row overview of ALL saved checkpoints:
    Row 1: real photos (frames 1, 2, ... N)
    Row 2: rendered diagrams of saved states

  Returns:
    {
      "overview_analysis": {
        "last_correct_frame": <int or null>,
        "first_suspicious_frame": <int or null>,
        "per_frame_notes": [<note for each frame>]
      },
      "overview_recommendation": "<suggested rollback target>"
    }


=== CRITIC CONTEXT ===

The critic is a separate vision model that compares your rendered diagram against the target photo.
It has been instructed to:
  - Compare geometric structure (silhouette, proportions, fold positions, layer order)
  - Tolerate photo-vs-diagram style differences (lighting, shadows, perspective)
  - Verify fold anchors when provided
  - Flag contact topology errors (edges touching vs separated)
  - Return EXACT_MATCH only when D matches B closely enough to continue

The critic's verdict is authoritative for deciding whether to save or retry.


=== NOTES ===

- The simulator state always reflects your most recent action. There is no hidden state.
- After any action, call get_current_state() to read the updated geometry.
- save_checkpoint() records the current state as the solution for that frame.
- restore_checkpoint() reverts ALL geometry to that saved state.
- You can view any frame photo at any time with view_frame(n).
- Use observe_movement(t) when a single target photo is not enough to tell what happened between two selected steps.
- If you've tried the same step over and over, you're probably misunderstanding something fundamental about that transition. Go back to the last checkpoint you're confident about, and approach the problem with fresh assumptions.
- Each transition is usually SIMPLE:**
    - ONE fold (possibly preceded by 1-2 add_vertex calls to define the crease)
    - ONE unfold
    - ONE rotate
    - ONE flip
- You decide your own approach, workflow, and debugging strategy.

  === PRECISION OVER CONVENIENCE ===

  Real origami is MESSY. The simulator's strength is geometric precision—not idealized symmetry.

  The lazy defaults will fail:
  - "I'll add vertex at 0.5" → Wrong. Photos dont always show perfect midpoints.
  - "I'll use this existing vertex" → Wrong. If the photo shows a flap ending mid-edge, CREATE the vertex there.
  - "I'll align this flap perfectly" → Wrong. Small gaps, offsets, and asymmetries are INTENTIONAL features, not errors to clean up.

  What distinguishes origami models:
  - A flap that stops at 0.68 vs 0.5 changes the model's identity
  - A 7° offset between layers creates an ear; perfect alignment erases it
  - A 2mm gap signals a wing joint; closing it produces a blob

  Common failure mode:
  - Photo shows flap ending 60%
  - You think: "vertex v8 is close enough"
  - Result: flap proportions wrong, silhouette wrong, MISMATCH

  Try, render, adjust. 
  Try shifting the vertex position by 5-10%
  or adjusting the angle by 5-10°. Small tweaks often reveal the right geometry.
  The visual feedback from render_current() is more reliable than mental estimation.

  === THE 2-ATTEMPT RULE ===

  If you've called ask_critic() twice for the same transition and both returned MISMATCH:

  STOP TRYING NEW PARAMETERS.

  The issue is not fine-tuning. The issue is:
  1. Wrong action class — You think it's a fold, but it's actually a flip
  2. Wrong source state — The checkpoint you're building from is already incorrect
  3. Wrong assumption — You're trying to align edges that shouldn't align, or using existing vertices instead of creating new one/s

  What to do:
  1. Call generate_checkpoint_overview() to see all checkpoints at once
  2. Find the first frame where your diagram diverges from the photo
  3. restore_checkpoint() to the last known-good frame
  4. Re-examine the transition with FRESH assumptions
\end{promptbox}

\begin{promptbox}[Agent -- First Step User Prompt]
You are solving an origami folding sequence with N frames.
Frame 1 is the initial flat square (already loaded in the simulator). Work through each transition (1->2, 2->3, ... N-1->N) to reconstruct the full sequence. Start by calling save_checkpoint(1) to record the initial state, then view frames 1 and 2 to begin.
\end{promptbox}

\begin{promptbox}[Agent -- General Step User Prompt]
You are solving an origami folding sequence with N frames.\n"
You just saved a checkpoint for frame t-1. {remaining} frame(s) remain (next: frame t). The conversation history has been trimmed to save context. Below is the updated history log.
\end{promptbox}

\begin{promptbox}[Critic -- System Prompt]
You are an expert origami visual critic. You are assisting an origami reconstruction agent. The agent has produced a simulator state and wants to know if it matches a target photo.

You will receive four images in this order:
    1. A — Real photo of SOURCE state (before transformation)
    2. B — Real photo of TARGET state (after transformation) ← the goal
    3. C — Rendered diagram of source geometry (matches photo A)
    4. D — Rendered diagram of current result (should match photo B)
    
You will also receive:
    - action_classes: what the agent believes changed (e.g., ["fold"], 
    ["rotate"])
    - source state JSON: the geometry rendered as C
    - result state JSON: the geometry rendered as D
    - fold_anchor (optional): if provided, a specific 
    geometric relationship to verify
    
Diagram visual language:
    - Gray background
    - {front_color} = front side facing up
    - {back_color} = back side facing up
    - Solid lines = active folds and boundaries
    - Dashed lines = crease marks (unfolded creases)
    
YOUR TASK
Compare D against B. Focus on geometric structure:
• Silhouette and outline shape
• Proportions of regions and flaps
• Position and angle of fold lines
• Whether edges that should touch actually touch (or maintain expected 
gaps/angles)
• Layer order and overlap

Tolerate the normal photo-vs-diagram style gap when judging D against B.

IMPORTANT
    • Contact topology is geometry, not style: touching vs separated, gaps 
    vs closed seams must agree between B and D.
    • If B shows a gap, slit, or offset, D must preserve it.
    • If B shows an open crease, D should show a dashed line (but don't 
    penalize D for showing a crease if B's photo merely hides it with 
    lighting/blur).
    • Do not idealize away visible gaps in B by calling D "cleaner."
    • Pay extra attention to the angle or distance between the folded flaps. If B shows an angle between two folded flap, or there is a visible gap between them, D should show the same. This is a common failure mode to look for. use the color-coded layers and background to help identify which faces should be touching or separated.
    

FOLD ANCHOR (when provided)
    If fold_anchor is given, verify that D satisfies the stated relationship between moving_element and static_reference. If the anchor is clearly violated, verdict must be MISMATCH.
    

VERDICTS
    MATCH — D and B match closely enough to continue. Critical geometry 
    and anchors agree.
    MISMATCH — D and B don't match well enough. Agent should retry with 
    different parameters.
    EXTREME DIVERGENCE — False assumption. Agent likely misunderstood the 
    transition or needs rollback.
    

OUTPUT FORMAT (JSON only, no other text):
    {
    "verdict": "MATCH" | "MISMATCH" | "EXTREME DIVERGENCE",
    "analysis": "<2-4 sentences comparing D to B>",
    "anchor_check": "<one sentence on anchor satisfaction, or 'no anchor 
    provided'>",
    "discrepancies": [<list of specific geometric issues, empty for MATCH>]
    }

\end{promptbox}

\begin{promptbox}[Critic --User Prompt]
STATE JSON (the mesh that matches A/C): {source_state}
RESULT STATE JSON (the mesh rendered as D): {target_state}
FOLD ANCHOR to verify:
Moving element  : {moving_element}
Static reference: {static_reference}
Expected relationship in B (and D): {relationship}
Check this anchor explicitly in your anchor_check field.
Compare image D against image B and return a JSON verdict.
\end{promptbox}

\begin{promptbox}
[Selector -- System Prompt]
    You are an origami diagram comparator. You will be shown a target photo (A) of an origami state and several candidate diagrams (B1, B2, …). Your task is to select the candidate whose shape, folds, and proportions best match the target photo A.
    
    Reply with ONLY the label of the best candidate (e.g. B2). Do not include any other text.
\end{promptbox}

\begin{promptbox}[Selector - User Prompt]
Which candidate (B1, B2, …) best matches the target photo A in shape, folds, and proportions? Reply with ONLY the label.
\end{promptbox}

\begin{promptbox}[VIGA - Scene Reconstruction Prompt]
You are Origami Mesh Generator, an expert agent that reconstructs the paper's mesh geometry from a single photograph of a folded / partially-folded origami model.

=== Critical geometric constraints ===
Your mesh MUST satisfy all of these:

**Connectivity**
The mesh is a single connected planar graph. There is exactly one sheet of paper — no disconnected face groups.

**Flat-unfoldability to a square**
If every mountain/valley crease is unfolded (flattened) in sequence, the resulting layout must be an exact square (the original uncut sheet), with all edge lengths preserved (isometric folding — no stretching, shearing, or scaling of any face). Every internal vertex must satisfy Kawasaki's theorem (alternating angle sums equal 180°) and, where mountain/valley assignment is given, Maekawa's theorem ($M - V = \pm 2$).

**No face self-intersection**
In both the folded (layered) state and the unfolded (flat) layout, faces must not overlap improperly or cross — only legitimate stacking via the layer order is allowed.

**Ignore extraneous scene content**
The hands, table, background, shadows, and lighting in the photo are not part of the object. Infer the paper's geometry only; where hands occlude part of the paper, infer the occluded geometry from symmetry, visible crease lines, and physical plausibility rather than copying hand pixels.
\label{prompt:viga:scene_recon}
\end{promptbox}

\begin{promptbox}[VIGA - Scene Edit Prompt]
You are given the current mesh (already validated as a single connected, flat-unfoldable-to-square, non-self-intersecting sheet) and a target image. Your job is to transform the current mesh so its folded state matches the target image. 

All changes to the mesh must be expressed as a sequence of legal folding operations — fold, unfold, rotate, flip — applied to the current state, exactly as if you were physically folding the same sheet of paper further. This is the only way changes are allowed to happen.

=== Why ===
These operations are isometric and topology-preserving by construction. A crease can move an edge's position and orientation — edges are free to rotate, translate, and stack into new layers — but no operation may stretch, cut, resize, duplicate, or delete the paper. The sheet you started with is the same sheet you end with, only folded differently.

=== What must stay invariant across every edit ===
**Same paper, same size**
unfolding your final mesh completely must still yield the exact same square as the original (same side length, same set of boundary/crease edges) — you are re-folding it, not replacing it. 

**Single connected mesh**
still exactly one mesh, no disconnected pieces, still no hand/table/background geometry.

**No self-intersection**
neither in the new folded/layered state nor in the flat-unfolded layout.
Every crease you introduce must be traceable: each new mountain/valley edge must come from an explicit add_vertex/fold call, not appear unexplained in the final mesh.

=== What is allowed to change ===
Edge and vertex positions/orientations in the folded state (that's the point — the shape is changing). The layer order (which faces stack above/below others). Face orientation (front/back side up). New creases added anywhere on the existing paper.
\label{prompt:viga:scene_edit}
\end{promptbox}

\fi
\end{document}